%% file: main.tex
\documentclass[12pt]{article}
\usepackage[utf8]{inputenc}
\usepackage[english]{babel}
\usepackage[letterpaper,top=2cm,bottom=2cm,left=2.5cm,right=2.5cm,marginparwidth=1.75cm]{geometry}
\usepackage{amsmath}
\usepackage{amssymb}
\usepackage{bm}
\usepackage{graphicx}
\usepackage{color}
\usepackage{upgreek}
\usepackage{placeins}
\usepackage{pgfplots}
\pgfplotsset{compat=newest}
\usepackage{tikz}
\usetikzlibrary{chains,shapes.arrows, arrows, positioning}
\usepackage{subcaption}
\usepackage{hyperref}
\usepackage{multirow}
\pgfkeys{/pgf/number format/.cd, 1000 sep={\,}}
\usepackage{comment}
\tikzstyle{mybox} = [rectangle, rounded corners, minimum width=3cm, minimum height=1cm, text centered, draw=black]
\graphicspath{{Figures/Modes/}{Figures/Modes_k50/}{Figures/Modes_k1000/}{Figures/}{Figures/Dots}{Figures/Mixed}{Figures/Fingers}{Figures/Dots_few}{Figures/Mixed_few}{Figures/Fingers_few}}
\usepackage{authblk}
\usepackage{mathptmx}
\usepackage[backend=biber, style=ieee]{biblatex}
\usepackage{csquotes}
\usepackage[output-decimal-marker={.}, locale = US, group-separator={\,},   group-minimum-digits=4, group-digits=integer]{siunitx}
\DeclareSIUnit\px{px}
\DeclareSIUnit\dpi{dpi}
\DeclareSIUnit\bit{bit}
\DeclareSIUnit\lines{lines}

\newcommand*\circled[1]{\tikz[baseline=(char.base)]{
            \node[shape=circle,draw,inner sep=2pt] (char) {\textbf{#1}};}}
\newcommand*\circledFill[1]{\tikz[baseline=(char.base)]{
            \node[shape=circle,draw,fill=white,inner sep=2pt] (char) {\textbf{#1}};}}

\title{\textbf{Reduced-order modeling and classification of hydrodynamic pattern formation in gravure printing}}
\author[1]{Pauline Rothmann-Brumm}
\author[2]{Steven L. Brunton}
\author[3]{Isabel Scherl}
\affil[1]{Technical University of Darmstadt, Department of Mechanical Engineering, Institute of Printing Science and Technology, 64289 Darmstadt, Germany}
\affil[2]{University of Washington, Department of Mechanical Engineering, Seattle, WA 98195, USA}
\affil[3]{California Institute of Technology, Department of Mechanical and Civil Engineering, Pasadena, CA 91125, USA}
\date{\today}

\begin{document}

\maketitle

\begin{abstract}
Hydrodynamic pattern formation phenomena in printing and coating processes are still not fully understood. However, fundamental understanding is essential to achieve high-quality printed products and to tune printed patterns according to the needs of a specific application like printed electronics, graphical printing, or biomedical printing. The aim of the paper is to develop an automated pattern classification algorithm based on methods from supervised machine learning and reduced-order modeling. We use the HYPA-p dataset, a large image dataset of gravure-printed images, which shows various types of hydrodynamic pattern formation phenomena. It enables the correlation of printing process parameters and resulting printed patterns for the first time. \num{26880} images of the HYPA-p dataset have been labeled by a human observer as dot patterns, mixed patterns, or finger patterns; \num{864000} images (\SI{97}{\percent}) are unlabeled. A singular value decomposition (SVD) is used to find the modes of the labeled images and to reduce the dimensionality of the full dataset by truncation and projection. Selected machine learning classification techniques are trained on the reduced-order data. We investigate the effect of several factors, including classifier choice, whether or not fast Fourier transform (FFT) is used to preprocess the labeled images, data balancing, and data normalization. The best performing model is a k-nearest neighbor (kNN) classifier trained on unbalanced, FFT-transformed data with a test error of \SI{3}{\percent}, which outperforms a human observer by \SI{7}{\percent}. Data balancing slightly increases the test error of the kNN-model to \SI{5}{\percent}, but also increases the recall of the mixed class from \SI{90}{\percent} to \SI{94}{\percent}. Finally, we demonstrate how the trained models can be used to predict the pattern class of unlabeled images and how the predictions can be correlated to the printing process parameters, in the form of regime maps.
\end{abstract}


\input{1_Introduction}
\input{2_Methods}
\input{3_Results}
\input{4_Conclusion}
\input{Acknowledgements}

\FloatBarrier
\printbibliography

\newpage
\appendix
\input{Appendix}

\end{document}

%% file: 1_Introduction.tex
\section{Introduction}

Pattern-forming systems are of wide interest to the scientific community and patterns are ubiquitous in nature~\cite{cross1993pattern, ball1999self}. Well-known examples in fluids include coherent patterns in turbulent flows~\cite{holmes2012turbulence}, viscous fingering patterns in a lifted Hele-Shaw cell~\cite{brulin2020fingering}, and biological pattern formation on animal skin~\cite{kondo2002reaction}. For some pattern-forming systems, the underlying principles are known, however, in others the principles remain unknown. In these cases, machine learning offers an opportunity to study these fluid systems~\cite{brunton2020machine} and understand pattern formation~\cite{schmekel2022predicting, Brumm2022Deep}. In this work, we investigate hydrodynamic pattern formation phenomena using images generated from gravure printing. Hydrodynamic pattern formation in gravure printing is not fully understood due to the complex surfaces and fluids involved~\cite{roisman2023forced, Kumar.2015}.

\begin{figure}[t]
\def\svgwidth{\textwidth}
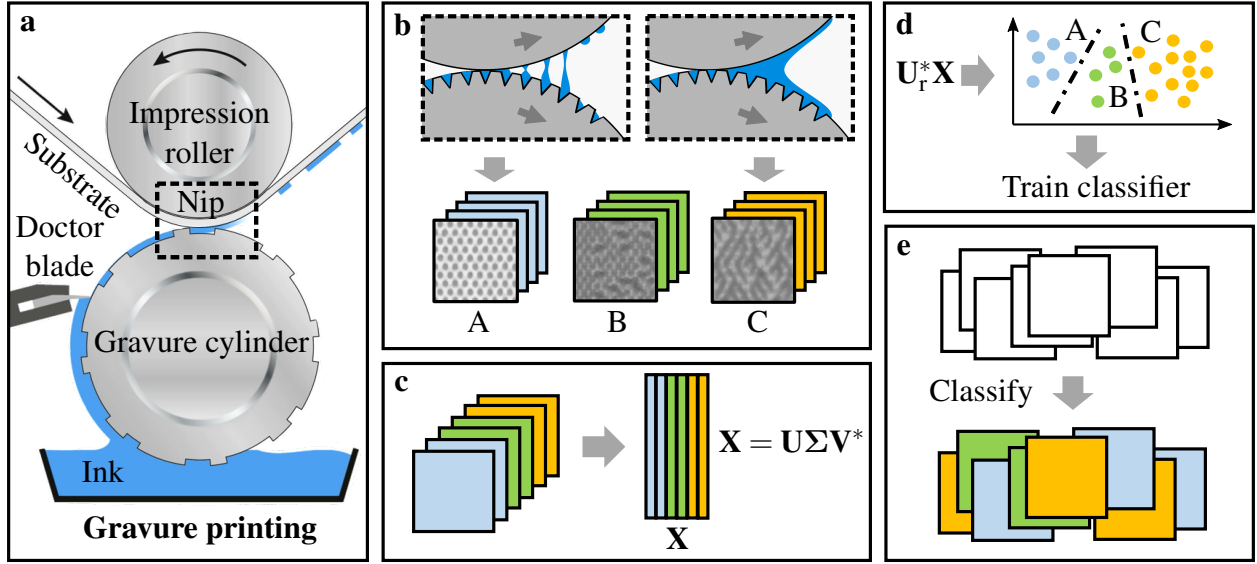
\caption{We investigate hydrodynamic pattern formation phenomena during gravure printing (a). The patterns are mainly formed during the sub-process of fluid transfer (dashed box) and can be assigned to one of three classes (b): point splitting, transition regime, and lamella splitting. The three classes are represented in the printed images as dot patterns (A), mixed patterns (B), and finger patterns (C). We use a large image dataset of gravure-printed patterns, including \num{26880} labeled images. The labeled images are randomly divided into train and test dataset and the train dataset is rearranged to a data matrix $\mathbf{X}$ and a singular value decomposition (SVD) is performed (c). For dimensionality reduction, the matrix $\mathbf{X}$ is projected onto the truncated SVD modes $\mathbf{U_{\text{r}}}$. This yields reduced-order data for training of several classifiers (d). The trained classifiers can be used to classify the unlabeled images of the dataset and to enable application-oriented scientific analysis on hydrodynamic pattern formation in gravure printing (e).}
\label{fig:Teaser}
\end{figure}

Gravure printing, also known as rotogravure, belongs to the group of printing processes in which the image-forming elements are recessed \cite{Gray.2003, kipphan2001handbook}. The main component of this printing process is the gravure cylinder. Small, micrometer-scale cells are engraved or etched into its surface. During printing, the cells are filled with ink from the ink reservoir and the excess ink is wiped away by a doctor blade (\autoref{fig:Teaser}a). Under high pressure, the ink from the cells is deposited onto the substrate. However, only a portion of the ink is transferred due to fluid splitting (\autoref{fig:Teaser}b). Pressure during fluid transfer is created by the mechanical contact between the gravure cylinder and the impression roller. During the printing process, the substrate is fed through the nip between the gravure cylinder and the impression roller. Electrostatic printing assist (i.\,e. electrostatic charges) can help to improve ink deposition from the printing form to the substrate.

The patterns that are generated in gravure printing are only partially pre-determined by the chosen printing image and the engraving parameters. This is due to a fingering instability of the ink-air interface during fluid splitting. For example, when using a relatively large ink volume per area, spontaneous finger patterns are created on the substrate. The fingering instability and Saffman-Taylor instability~\cite{Saffman.1958} are similar since they are both driven by pressure gradients at the interface. However, due to the complex surfaces and non-Newtonian fluids used in gravure printing, the finger frequency scales differently than in the Saffman-Taylor theory \cite{schafer2019millisecond, rothmann2024gravure}.

H{\"u}bner~\cite{Hubner.1991} distinguishes two classes of fluid splitting during fluid transfer: point splitting and lamella splitting. Brumm et al.~\cite{Brumm2021Ink} extend this view by adding a third class, the transition regime between point and lamella splitting. Depending on the class of fluid splitting, different hydrodynamic pattern formation phenomena occur in the gravure printing process. In the lamella splitting regime, the volume of the cells per surface area is relatively large causing a continuous meniscus in the nip due to capillary forces. The ink-air interface may become unstable leading to stochastic finger patterns printed (or viscous fingering). When there is only a weak amount of viscous fingering, the lamella splitting regime creates a rather closed homogeneous layer of ink. The point splitting regime is characterized by dot patterns on the printed product originating from the raster pattern of the cells on the printing cylinder. Each cell deposits a portion of ink onto the substrate independently. In the transition regime, both dot and finger patterns appear on the printed product. This is where the system transitions from point splitting to lamella splitting or vice versa. The transition regime is especially challenging to characterize~\cite{Brumm2021Ink}.

There are many industrial applications for printed patterns besides graphical printing, including printed electronics, biomedical printing, security printing, or functional coatings. Printed electronics typically require defect-free closed layers of constant thickness, which can be achieved in the regime of lamella splitting. However, this regime comes with the pattern formation phenomenon of viscous fingering, which leads to surface undulations that can cause defects in the printed, multi-layered electronic devices. In many cases, a common goal is to suppress viscous fingering. In contrast to printed electronics, graphical printing often operates in the point splitting regime, which produces dot-like raster patterns that are needed to form high-resolution graphical images and text. In the point splitting regime, printing defects like missing dots may spoil the printing quality. In biomedical printing, the pattern formation phenomenon of viscous fingering can be used to create tailored vascular network structures for organs-on-a-chip \cite{brumm2022fabrication}. The distance, height, and width of the printed vascular structures must be adjusted properly in order to have biomimetic properties. Each application requires special properties of the printed patterns. Consequently, a thorough understanding of pattern formation in printing processes is essential. The ultimate goal is to tune the printed patterns according to the needs of the specific application. 

Manual classification by a human observer, as in~\cite{Brumm2021Ink}, is one way to classify printed patterns. However, it is time-consuming and prone to errors and subjective judgment. Consequently, manual classification is not suitable for very large datasets, which are needed to provide deep insights into fluid dynamics. To print with the necessary characteristics for each application, we first need a method to efficiently classify printed pattern samples. Then, we can apply the method to large datasets and correlate the pattern classes to the printing process parameters. Additionally, these correlations illuminate the fluid dynamics behind pattern formation phenomena. The aim of this research is therefore to establish an automated pattern classification approach that can automatically distinguish printed patterns with high accuracy.

In this paper we develop an automated pattern classification algorithm using methods from supervised machine learning and reduced-order modeling. Printed pattern classification is challenging since the variety of patterns is vast. Even for an experienced scientist, it is difficult to categorize some printed patterns into predefined classes. Previous studies show that the distinction between the transition regime and other regimes is the most challenging~\cite{Brumm2021Ink}. In contrast to studies that use deep learning for classification \cite{Brumm2022Deep, Brumm2021A}, we establish an accurate and efficient classification algorithm that makes use of reduced-order modeling and supervised machine learning. Classic machine learning models are more interpretable than deep learning models allowing us to easily detect biases and tune model parameters. 

The scope of this paper is shown in \autoref{fig:Teaser}. We use a large image dataset that was created using gravure printing (\autoref{fig:Teaser}a). The images show hydrodynamic pattern formation phenomena in form of dot patterns (A), mixed patterns (B), and finger patterns (C) (\autoref{fig:Teaser}b), which correspond to the fluid splitting class of point splitting, transition regime, and lamella splitting, respectively; see \autoref{tab:Wording}. However, we mainly use the terms dot, mixed, and finger pattern throughout this paper, since they are more descriptive. We randomly divide the labeled data into train and test data. The training dataset is rearranged to a data matrix $\mathbf{X}$. To obtain a reduced-order model, an SVD is performed on $\mathbf{X}$ (\autoref{fig:Teaser}c). The SVD yields the modes $\mathbf{U}$, which serve as a new coordinate system for our pattern dataset. Similarly, Sirovich and Kirby~\cite{sirovich.1987} created eigenpictures to serve as a coordinate system for the representation of facial images. Inspired by their term, our SVD modes could be called \emph{eigenpatterns}. Typically, the SVD is used for dimensionality reduction. The aim is to capture as much energy as possible using as few modes as possible, resulting in a low rank representation~\cite{brunton2022data}. To extract the dominant, low-dimensional patterns, the SVD is truncated, i.\,e., only the first $r$ SVD modes are kept. This leads to $\mathbf{U_{\text{r}}}$. The data matrix $\mathbf{X}$ is projected onto $\mathbf{U_{\text{r}}}$ and the resulting reduced-order data is fed to several machine learning classifiers. The trained classifiers are evaluated on the test dataset and can be used for automated classification of the unlabeled images of the dataset. Another approach uses the magnitude of the FFT-transformed data matrix $\mathbf{X}$ because the data contains spatially repeating patterns as well as stochastic patterns with a dominant wavelength. We increase the performance of our developed workflow by using FFT; see \autoref{fig:Schematic_workflow}. Additionally, we investigate the effect of classifier choice, data balancing, and data normalization.

\begin{table}[t]
    \centering
    \caption{Pattern formation phenomena and their corresponding fluid splitting class.}
	\label{tab:Wording}
\begin{tabular}{|l|l|l|}
    \hline
        & \textbf{Pattern formation phenomenon} & \textbf{Fluid splitting class} \\
     \hline
     A & Dot patterns & Point splitting\\
     B & Mixed patterns & Transition regime\\
     C & Finger patterns & Lamella splitting\\
     \hline
\end{tabular}
\end{table}

A description of the image dataset and the developed workflow for automated image classification is presented in Section~\ref{sec:Methods}. The results of the SVD and the classification as well as exemplary regime maps are shown in Section~\ref{sec:Results}. Conclusions and discussion can be found in Section~\ref{sec:Conclusions}. 

%% file: Figures/Teaser_V8_with_nip.pdf_tex
%% Creator: Inkscape inkscape 0.92.5, www.inkscape.org
%% Accompanies image file 'Teaser_V8_with_nip.pdf' (pdf, eps, ps)
\begingroup%
  \makeatletter%
  \providecommand\color[2][]{%
    \errmessage{(Inkscape) Color is used for the text in Inkscape, but the package 'color.sty' is not loaded}%
    \renewcommand\color[2][]{}%
  }%
  \providecommand\transparent[1]{%
    \errmessage{(Inkscape) Transparency is used (non-zero) for the text in Inkscape, but the package 'transparent.sty' is not loaded}%
    \renewcommand\transparent[1]{}%
  }%
  \providecommand\rotatebox[2]{#2}%
  \newcommand*\fsize{\dimexpr\f@size pt\relax}%
  \newcommand*\lineheight[1]{\fontsize{\fsize}{#1\fsize}\selectfont}%
  \ifx\svgwidth\undefined%
    \setlength{\unitlength}{894.87293153bp}%
    \ifx\svgscale\undefined%
      \relax%
    \else%
      \setlength{\unitlength}{\unitlength * \real{\svgscale}}%
    \fi%
  \else%
    \setlength{\unitlength}{\svgwidth}%
  \fi%
  \global\let\svgwidth\undefined%
  \global\let\svgscale\undefined%
  \makeatother%
  \begin{picture}(1,0.44967177)%
    \lineheight{1}%
    \setlength\tabcolsep{0pt}%
    \put(0,0){\includegraphics[width=\unitlength,page=1]{Teaser_V8_with_nip.pdf}}%
    \put(0.06990289,0.16855762){\color[rgb]{0,0,0}\makebox(0,0)[lt]{\lineheight{1.25}\smash{\begin{tabular}[t]{l}Gravure cylinder\end{tabular}}}}%
    \put(0.154251,0.35204129){\color[rgb]{0,0,0}\makebox(0,0)[t]{\lineheight{1.25}\smash{\begin{tabular}[t]{c}Impression\\roller \end{tabular}}}}%
    \put(0.05998737,0.06449153){\color[rgb]{0,0,0}\makebox(0,0)[lt]{\lineheight{1.25}\smash{\begin{tabular}[t]{l}Ink\end{tabular}}}}%
    \put(0.06088843,0.01898835){\color[rgb]{0,0,0}\makebox(0,0)[lt]{\lineheight{1.25}\smash{\begin{tabular}[t]{l}\textbf{Gravure printing}\end{tabular}}}}%
    \put(0,0){\includegraphics[width=\unitlength,page=2]{Teaser_V8_with_nip.pdf}}%
    \put(0.310017,0.4229079){\color[rgb]{0,0,0}\makebox(0,0)[lt]{\lineheight{1.25}\smash{\begin{tabular}[t]{l}\textbf{b}\end{tabular}}}}%
    \put(0.01045922,0.42564823){\color[rgb]{0,0,0}\makebox(0,0)[lt]{\lineheight{1.25}\smash{\begin{tabular}[t]{l}\textbf{a}\end{tabular}}}}%
    \put(0.31052506,0.13683491){\color[rgb]{0,0,0}\makebox(0,0)[lt]{\lineheight{1.25}\smash{\begin{tabular}[t]{l}\textbf{c}\end{tabular}}}}%
    \put(0,0){\includegraphics[width=\unitlength,page=3]{Teaser_V8_with_nip.pdf}}%
    \put(0.71239267,0.42320917){\color[rgb]{0,0,0}\makebox(0,0)[lt]{\lineheight{1.25}\smash{\begin{tabular}[t]{l}\textbf{d}\end{tabular}}}}%
    \put(0,0){\includegraphics[width=\unitlength,page=4]{Teaser_V8_with_nip.pdf}}%
    \put(0.71309778,0.24663102){\color[rgb]{0,0,0}\makebox(0,0)[lt]{\lineheight{1.25}\smash{\begin{tabular}[t]{l}\textbf{e}\end{tabular}}}}%
    \put(0,0){\includegraphics[width=\unitlength,page=5]{Teaser_V8_with_nip.pdf}}%
    \put(0.36870611,0.184219){\color[rgb]{0,0,0}\makebox(0,0)[lt]{\lineheight{1.25}\smash{\begin{tabular}[t]{l}A\end{tabular}}}}%
    \put(0.48103747,0.18414886){\color[rgb]{0,0,0}\makebox(0,0)[lt]{\lineheight{1.25}\smash{\begin{tabular}[t]{l}B\end{tabular}}}}%
    \put(0.59296589,0.18429683){\color[rgb]{0,0,0}\makebox(0,0)[lt]{\lineheight{1.25}\smash{\begin{tabular}[t]{l}C\end{tabular}}}}%
    \put(0.73910945,0.12948353){\color[rgb]{0,0,0}\makebox(0,0)[lt]{\lineheight{1.25}\smash{\begin{tabular}[t]{l}Classify\end{tabular}}}}%
    \put(0.7973539,0.29267535){\color[rgb]{0,0,0}\makebox(0,0)[lt]{\lineheight{1.25}\smash{\begin{tabular}[t]{l}Train classifier\end{tabular}}}}%
    \put(0,0){\includegraphics[width=\unitlength,page=6]{Teaser_V8_with_nip.pdf}}%
    \put(0.01248628,0.33396214){\color[rgb]{0,0,0}\rotatebox{-40}{\makebox(0,0)[lt]{\lineheight{1.25}\smash{\begin{tabular}[t]{l}Substrate\end{tabular}}}}}%
    \put(0.71246618,0.38397983){\color[rgb]{0,0,0}\makebox(0,0)[lt]{\lineheight{1.25}\smash{\begin{tabular}[t]{l}$\mathbf{U_{\text{r}}^*X}$\end{tabular}}}}%
    \put(0.5703623,0.08863767){\color[rgb]{0,0,0}\makebox(0,0)[lt]{\lineheight{1.25}\smash{\begin{tabular}[t]{l}$\mathbf{X}=\mathbf{U} \bm{\Sigma} \mathbf{V^*}$\end{tabular}}}}%
    \put(0.52820917,0.01010453){\color[rgb]{0,0,0}\makebox(0,0)[lt]{\lineheight{1.25}\smash{\begin{tabular}[t]{l}$\mathbf{X}$\end{tabular}}}}%
    \put(0,0){\includegraphics[width=\unitlength,page=7]{Teaser_V8_with_nip.pdf}}%
    \put(0.04237028,0.26076775){\color[rgb]{0,0,0}\makebox(0,0)[t]{\lineheight{1.25}\smash{\begin{tabular}[t]{c}Doctor\\blade \end{tabular}}}}%
    \put(0.84760457,0.41851439){\color[rgb]{0,0,0}\makebox(0,0)[lt]{\lineheight{1.25}\smash{\begin{tabular}[t]{l}A\end{tabular}}}}%
    \put(0.88255822,0.36514201){\color[rgb]{0,0,0}\makebox(0,0)[lt]{\lineheight{1.25}\smash{\begin{tabular}[t]{l}B\end{tabular}}}}%
    \put(0.90831009,0.41717304){\color[rgb]{0,0,0}\makebox(0,0)[lt]{\lineheight{1.25}\smash{\begin{tabular}[t]{l}C\end{tabular}}}}%
    \put(0.1553184,0.2809871){\color[rgb]{0,0,0}\makebox(0,0)[t]{\lineheight{1.25}\smash{\begin{tabular}[t]{c}Nip \end{tabular}}}}%
  \end{picture}%
\endgroup%

%% file: 2_Methods.tex
\section{Methods}
\label{sec:Methods}

In this section, we first provide a description of the used HYPA-p dataset~\cite{HYPA-p} and explain how it was created using gravure printing and why it is suitable for analysis of hydrodynamic pattern formation (Section~\ref{subsec:Dataset}). We then present our algorithm for analysis of the dataset (Section~\ref{subsec:Data_set_analysis}). It uses the singular value decomposition (SVD) as well as machine learning classification algorithms for pattern identification. We also outline our chosen performance metrics. Our workflow is implemented in Matlab and the code is publicly available (Section~\ref{subsubsec:Code availability}).

\subsection{Dataset}
\label{subsec:Dataset}

For our investigation, we use the HYPA-p dataset~\cite{HYPA-p}. HYPA-p stands for \glq{}\textbf{hy}drodynamic \textbf{pa}tterns dataset - \textbf{p}rocessed\grq. To the best of our knowledge, this dataset is the only open-access dataset of its kind. It is a large image dataset of \num{890880} printed patterns, from which \num{26880} (\SI{3}{\percent}) are labeled. It enables the correlation of printing process parameters and resulting printed patterns for the first time. The labeled images were selected with the aim to represent the diversity of the dataset. Labeling was performed by a human observer according to the classification scheme of~\cite{Brumm2021Ink} and took around 15 to 22 hours of active identification~\cite{thesis_Pauline}. Thus, labeling of all images from the dataset would theoretically require 500 to 733 hours. This time-consuming labeling process emphasizes the need for an automated classification workflow. The dataset consists of \SI{34.8}{\percent} of the labeled dot patterns images, \SI{13.9}{\percent} mixed pattern images, and \SI{51.3}{\percent} finger pattern images. The labels correspond to the fluid splitting classes of point splitting, transition regime, and lamella splitting, respectively (\autoref{tab:Wording}).

\begin{figure}[t]
\centering
\begin{subfigure}[b]{0.325\textwidth}
\frame{\includegraphics[width=1\textwidth]{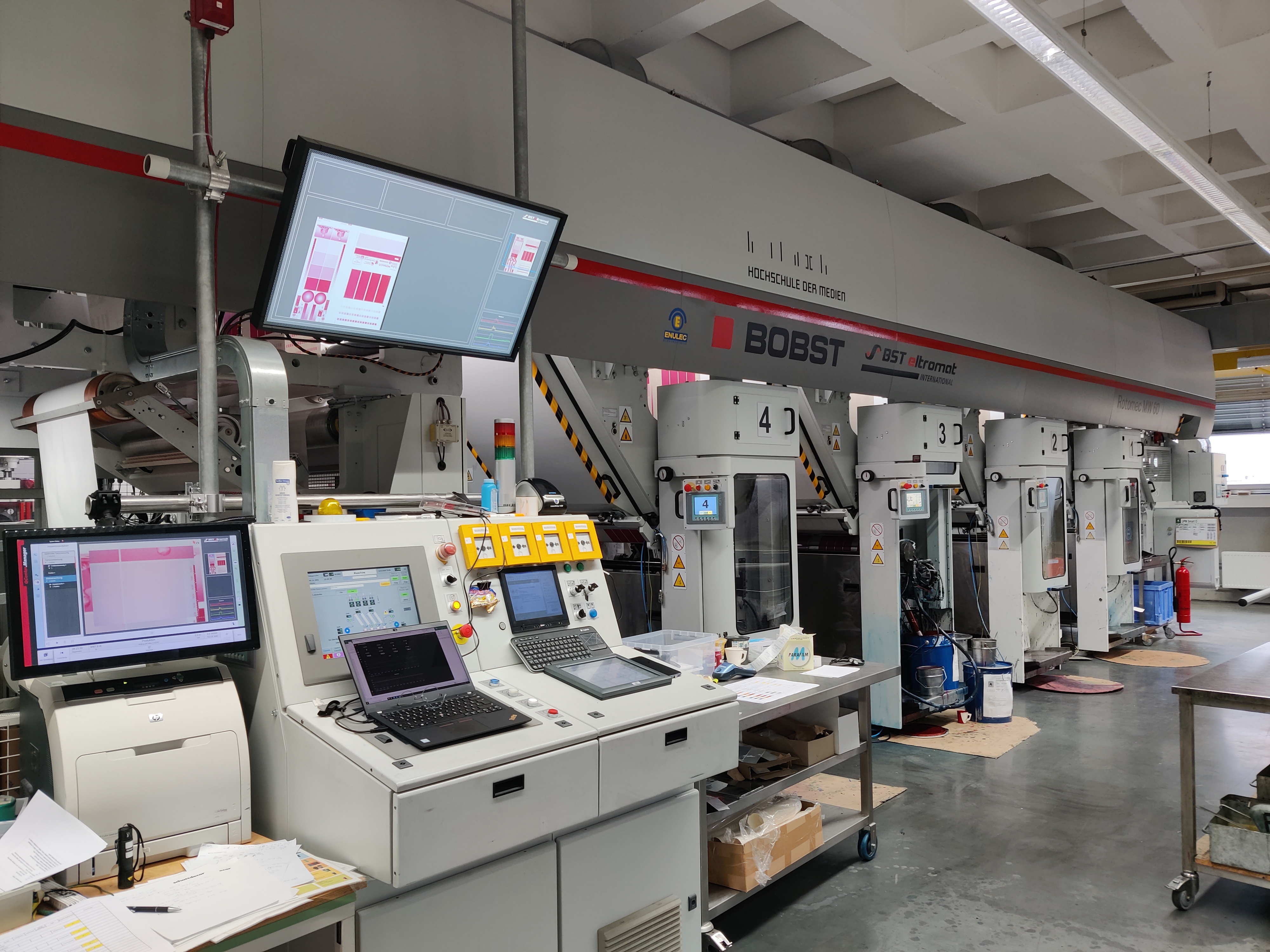}}
\caption{{\footnotesize Bobst machine}}
\end{subfigure}
\hfill
\begin{subfigure}[b]{0.325\textwidth}
\frame{\includegraphics[width=1\textwidth]{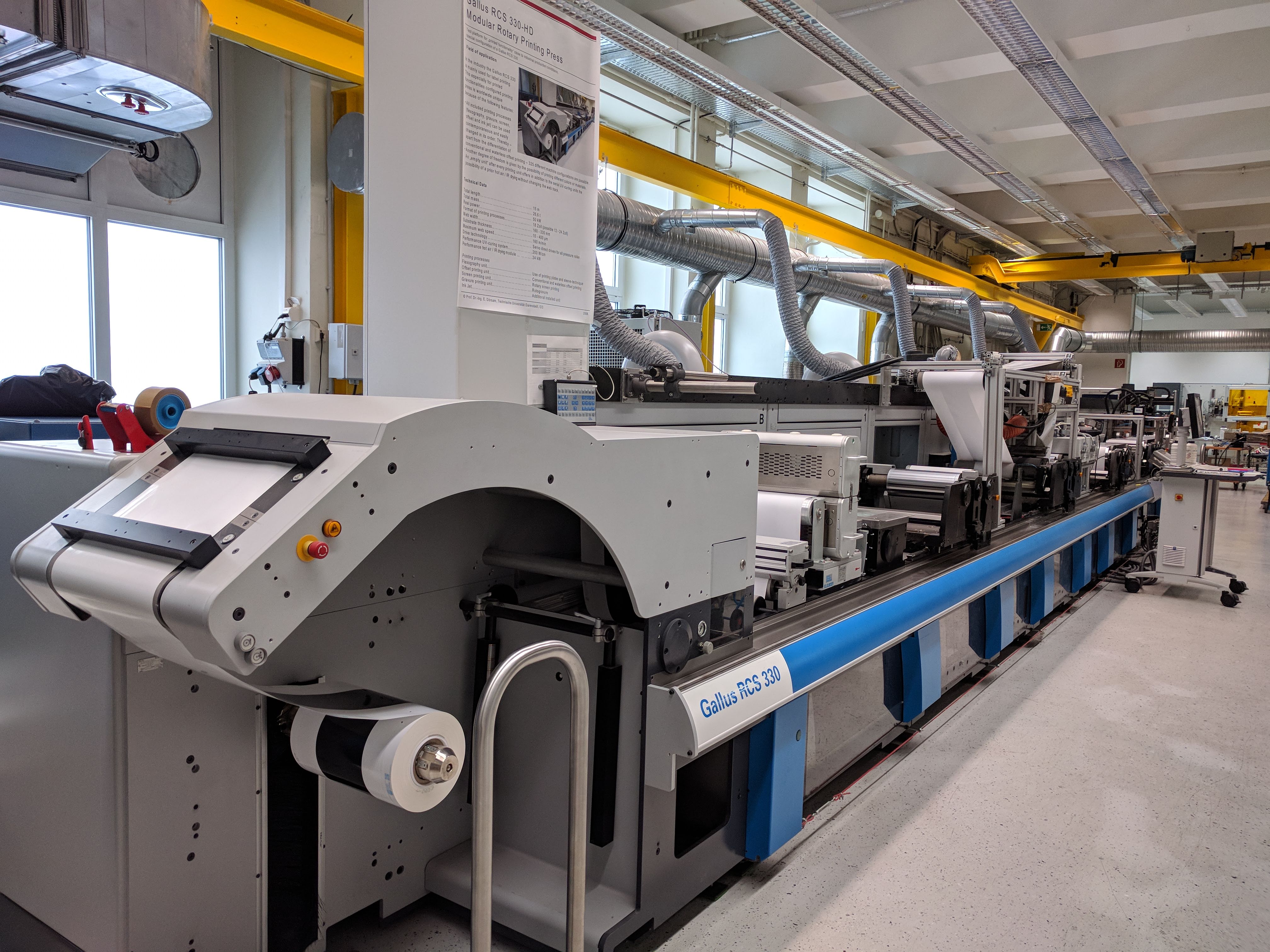}}
\caption{{\footnotesize Gallus machine}}
\end{subfigure}
\hfill
\begin{subfigure}[b]{0.325\textwidth}
\frame{\includegraphics[width=1\textwidth]{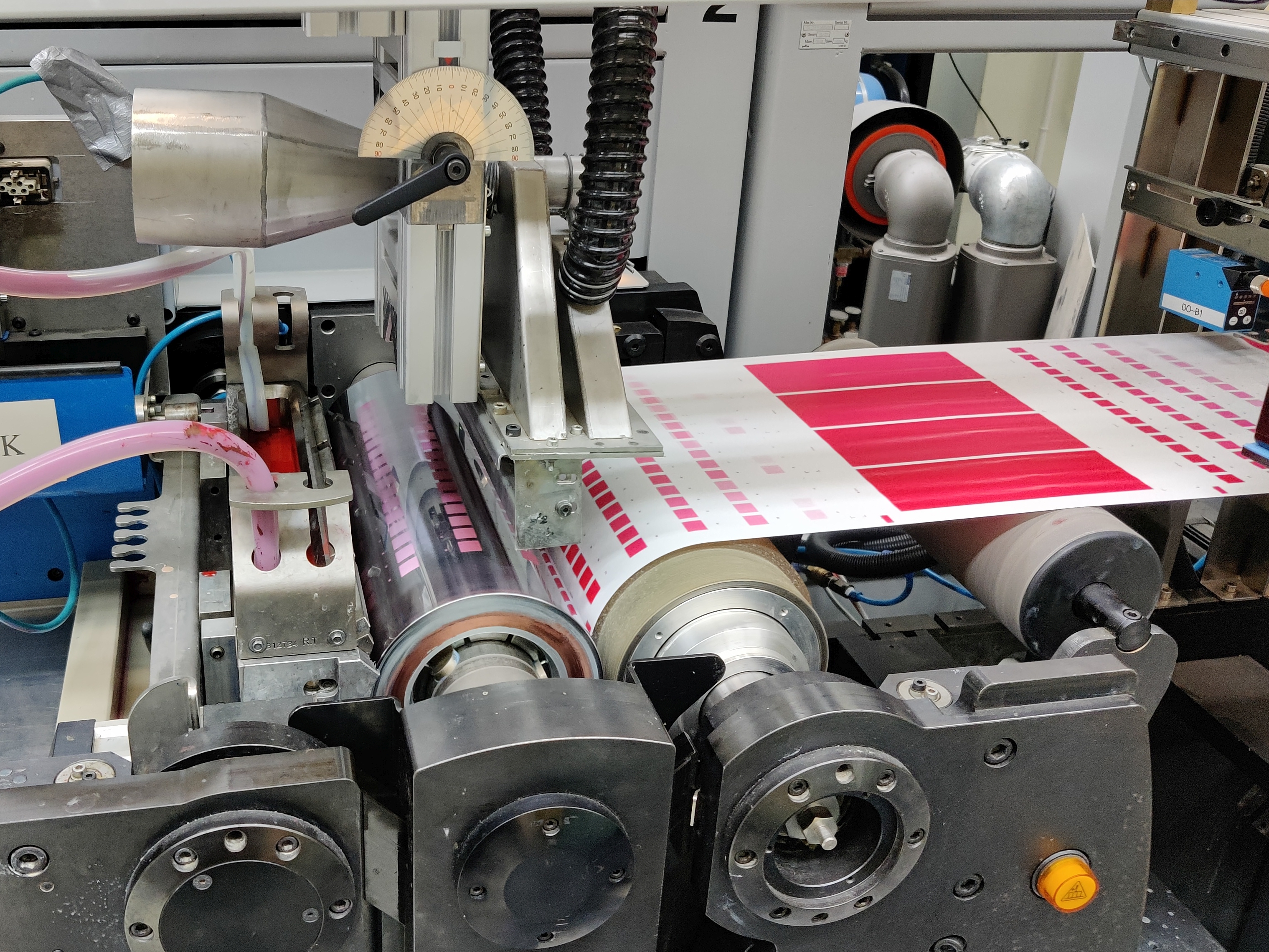}}
\caption{{\footnotesize Gallus gravure unit}}
\end{subfigure}
\par \smallskip
\begin{subfigure}[b]{0.325\textwidth}
\centering
\resizebox{\textwidth}{!}{\input{Figures/Dots_few/dots_few}}
\caption{{\footnotesize Dot patterns}}
\end{subfigure}
\hfill
\begin{subfigure}[b]{0.325\textwidth}
\centering
\resizebox{\textwidth}{!}{\input{Figures/Mixed_few/mixed_few}}
\caption{{\footnotesize Mixed patterns}}
\end{subfigure}
\hfill
\begin{subfigure}[b]{0.325\textwidth}
\centering
\resizebox{\textwidth}{!}{\input{Figures/Fingers_few/fingers_few}}
\caption{{\footnotesize Finger patterns}}
\end{subfigure}
\caption{Industrial gravure printing machines used for creation of the HYPA-p dataset. Bobst Rotomec MW 60-600/250 (Bobst, Mex, Switzerland) (a) and Gallus RCS 330-HD (Gallus Ferd. Rüesch AG, St. Gallen, Switzerland) (b). The gravure printing unit of the latter is shown in (c). Examples for dot pattern images (d), mixed pattern images (e), and finger pattern images (f) from the labeled dataset. Both machines can create all types of patterns. Each printed example has a size of \SI{260}{\px}~x~\SI{260}{\px} (\SI{2.75}{\mm}~x~\SI{2.75}{\mm}). More examples of printed patterns can be found in \autoref{fig:patterns_many} in the Appendix.}
\label{fig:patterns}
\end{figure}

The HYPA-p dataset was created by large-scale printing experiments on two industrial roll-to-roll gravure printing machines: Bobst Rotomec MW 60-600/250 (Bobst, Mex, Switzerland) and Gallus RCS 330-HD (Gallus Ferd. Rüesch AG, St. Gallen, Switzerland). Both machines are depicted in \autoref{fig:patterns}a-b. \autoref{fig:patterns}c provides a closer look at the Gallus gravure unit. \autoref{fig:patterns}d-f show printed pattern examples. Each printed example is \SI{260}{\px}~x~\SI{260}{\px} or \SI{2.75}{\mm}~x~\SI{2.75}{\mm}. For more examples of printed patterns; see \autoref{fig:patterns_many} in the Appendix.

In the following paragraphs, we briefly describe the experimental setup and the design of experiments that was used to create the HYPA-p dataset. More details can be found in ~\cite{HYPA-p, thesis_Pauline}.

The HYPA-p dataset was created for the analysis of hydrodynamic pattern formation in gravure printing. The hydrodynamics are strongly influenced by the choice of printing parameters such as ink viscosity, printing speed, or transfer volume of the printing form. Therefore, a large range of printing parameters were tested to observe a variety of printed patterns. The large-scale printing trial is summarized in \autoref{tab:DoE}. The adjustable parameters are: printing machine, type of fluid, fluid viscosity, substrate, doctor blade angle, printing velocity, and the use of electrostatic printing assist. An electromechanically engraved printing form was used, which exhibited four different raster frequencies (60, 70, 80 and 100 lines/cm) and twenty different tonal values (\SI{5}{\percent} to \SI{100}{\percent}). The raster frequency determines the spacing of the engraved raster cells, whereas the tonal value corresponds to the volume of the cells and thus the amount of ink transferred to the substrate. The raster angle was kept constant.

\begin{table}[t]
	\caption{Design of experiments for the large-scale printing trial, which was conducted by \cite{thesis_Pauline} to create the HYPA-p dataset \cite{HYPA-p}.}
	\label{tab:DoE}
	\centering
	\vspace{0.5\baselineskip}
		\begin{tabular}{|p{1.5cm} |p{2cm} |p{2cm} |p{2.5cm} |p{1.2cm}| p{2cm}| p{2.3cm}|}
		\hline
		\textbf{Name} & \textbf{Fluid} &  \textbf{Viscosity} & \textbf{Substrate} &  \textbf{Doctor blade angle} & \textbf{Printing velocity} & \textbf{Electrostatic printing assist} \\
		\hline
        B1-01 & Fluid \#1 & High & Substrate \#1 & 55° &  \multirow{17}{2cm}{[15, 30, 60, 90, 120, 180, 240]~m/min} & \multirow{17}{*}{On/off}\\
		B1-02 & Fluid \#1 & Medium & Substrate \#1 & 55° &  &\\
		B1-03 & Fluid \#1 & Low & Substrate \#1 & 55° & & \\
		B1-04 & Fluid \#1 & Low & Substrate \#1 & 60° && \\
		B1-05 & Fluid \#1 & Low & Substrate \#1 & 48° &  & \\
        \cline{1-5}
		B2-01 & Fluid \#2 & High & Substrate \#2 & 55° & & \\
		B2-02 & Fluid \#2 & High & Substrate \#3 & 55° & & \\
		B2-03 & Fluid \#2 & Medium & Substrate \#3 & 55° &  & \\
		B2-04 & Fluid \#2 & Medium & Substrate \#2 & 55° &  &\\
		B2-05 & Fluid \#2 & Low & Substrate \#2 & 55° & & \\
		B2-09 & Fluid \#2 & Low & Substrate \#3 & 55° && \\
		B2-10 & Fluid \#2 & Low & Substrate \#3 & 60° & & \\
		B2-11 & Fluid \#2 & Low & Substrate \#3 & 48° & & \\
        \cline{1-5}
		B3-01 & Fluid \#3 & Very high & Substrate \#3 & 55° & & \\
		B3-02 & Fluid \#3 & High & Substrate \#3 & 55° &  & \\
		B3-04 & Fluid \#3 & Medium & Substrate \#3 & 55° & & \\
		B3-05 & Fluid \#3 & Low & Substrate \#3 & 55° & & \\
		\hline
        G1-01 & Fluid \#3 & Very high & Substrate \#3 & 37\textdegree &\multirow{4}{2cm}{[5, 10, 15, 30, 60, 90, 120]~m/min}& \multirow{4}{*}{Off}\\
		G1-02 & Fluid \#3 & Low & Substrate \#3 & 37\textdegree & & \\
		G1-03 & Fluid \#3 & High & Substrate \#3 & 37\textdegree && \\
		G1-04 & Fluid \#3 & Medium & Substrate \#3 & 37\textdegree & & \\
         \hline
		\end{tabular}
\end{table} 

The large-scale printing trial yielded over 23 km of printed substrate. The printed patterns were cut from the printed substrate, then digitized using a high-resolution color-calibrated flatbed scanner and, finally, post-processed using digital image processing. Post-processing included accurate cropping and rotation of the printed patterns. 

The labeled images from the HYPA-p dataset are used for the training of the machine learning classifiers in Section~\ref{subsec:Data_set_analysis}.

\subsection{Dataset analysis}
\label{subsec:Data_set_analysis}

\subsubsection{Overview}

\begin{figure}[t]
\def\svgwidth{\textwidth}
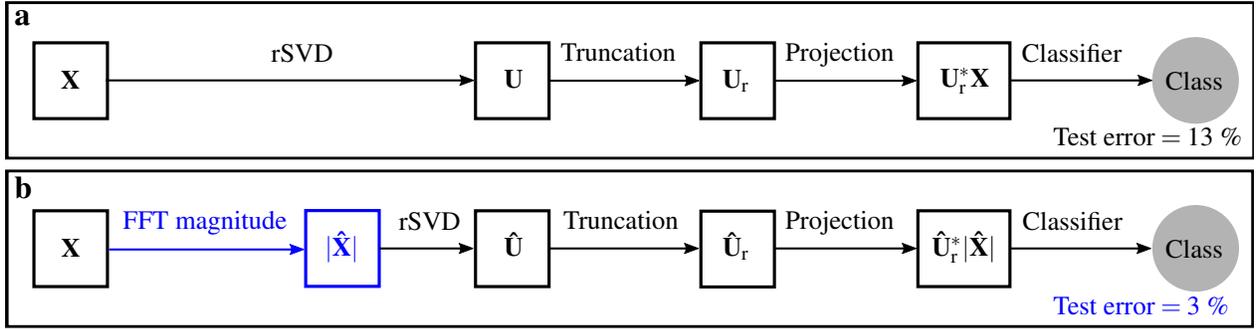
\caption{Schematic workflow. Variant (a) of the workflow performs randomized SVD (rSVD) directly on the data, which has been rearranged to a data matrix $\mathbf{X}$. The rSVD modes $\mathbf{U}$ are truncated to $\mathbf{U_{\text{r}}}$ and projected onto $\mathbf{X}$. The projected rSVD modes $\mathbf{U_{\text{r}}^*X}$ are used for training of several classifiers, leading to a minimum test error of 13\,\%. A normalization of $\mathbf{U_{\text{r}}^*X}$ to mean zero and unity standard deviation is optional and not performed by default. Variant (b) first performs a fast Fourier transform (FFT) on $\mathbf{X}$. Second, rSVD is performed on the magnitude of the FFT-transformed data $\mathbf{|\hat{X}|}$. The rest of the workflow is the same. This variant leads to a decreased minimum test error of 3\,\%. $\mathbf{X}$ can be chosen as the complete dataset $\mathbf{X_{\text{c}}}$ or as a balanced dataset $\mathbf{X_{\text{b}}}$. The train-test-split of $\mathbf{X}$ is always 80\,\%-20\,\% and leads to $\mathbf{X_{\text{train}}}$ and $\mathbf{X_{\text{test}}}$, which are fed separately through the workflow. For testing of the trained classifier, the rSVD modes from the training are reused to ensure equal transformation of $\mathbf{X_{\text{train}}}$ and $\mathbf{X_{\text{test}}}$.}
\label{fig:Schematic_workflow}
\end{figure}

The aim of the developed workflow (shown in \autoref{fig:Schematic_workflow}) is an automated pattern classification algorithm based on methods from supervised machine learning and reduced-order modeling. There are two variants of the workflow: (a) and (b). Variant (a) performs a randomized SVD (rSVD) directly on the data ($\mathbf{X}$). Variant (b) first performs a fast Fourier transform (FFT) of the data, then performs rSVD on the magnitude of the FFT-transformed data ($\mathbf{|\hat{X}|}$). We use the rSVD because it is computationally efficient for very large datasets~\cite{Halko.2011}. $\mathbf{X_{\text{c}}} \in \mathbb{R}^{\num{67600}\times\num{26880}}$ is the complete data matrix of the images from the printing experiments, which have each been converted to grayscale and then re-shaped into a column of the matrix. To create a balanced dataset, $\mathbf{X_{\text{b}}} \in \mathbb{R}^{\num{67600}\times\num{11160}}$, an equal number of dot, mixed, and finger pattern images are aggregated. The train-test-split of $\mathbf{X}$ is always chosen as 80\,\%-20\,\% and leads to $\mathbf{X_{\text{train}}}$ and $\mathbf{X_{\text{test}}}$, which are each fed separately through the workflow. 

For variant (a), the rSVD yields the rSVD modes $\mathbf{U}$, which are are truncated to $\mathbf{U_{\text{r}}}$, with $r$ being the rank of the truncation. Then, $\mathbf{X}$ is projected onto the truncated rSVD modes. The reduced-order data ($\mathbf{U_{\text{r}}^*X}$) is used to train several classifiers. A normalization of the reduced-order data ($\mathbf{U_{\text{r}}^*X}$) to mean zero and unity standard deviation is optional and not performed by default. 
For variant (b), the rSVD yields rSVD modes $\mathbf{\hat{U}}$ and the rest of the workflow is the same. Note that for testing of the trained classifier, the rSVD modes from the training are reused to ensure equal transformation of $\mathbf{X_{\text{train}}}$ and $\mathbf{X_{\text{test}}}$.

\subsubsection*{Singular Value Decomposition}

Singular value decomposition (SVD) is one of the most important matrix decompositions and the basis for many data-driven methods~\cite{brunton2022data}. For example, it has been used to find low-rank patterns in high-rank datasets in fluid mechanics~\cite{Taira2017aiaa}.

Let $\mathbf{X} \in \mathbb{C}^{n \times m}$ be a dataset from an experiment or simulation. In the case of image data, the columns $\mathbf{x}_k \in \mathbb{C}^n$ of $\mathbf{X}$ are single images that are rearranged into a column vector. Typically, the dimension of the image data $n$ is much larger than the number of images available in the dataset $m$. This leads to a data matrix $\mathbf{X}$~\cite{brunton2022data}:
\begin{equation}
    \mathbf{X}=
    \left[
  \begin{array}{cccc}
    \vrule & \vrule & & \vrule\\
    \mathbf{x}_{1} & \mathbf{x}_{2} & \ldots & \mathbf{x}_{m} \\
    \vrule & \vrule & & \vrule 
  \end{array}
\right].
    \label{eq:X}
\end{equation}

The economy SVD is given in \autoref{eq:SVD}. For the full SVD, we refer to~\cite{brunton2022data}.

\begin{equation}
    \mathbf{X} = \mathbf{U} \bm{\Sigma} \mathbf{V^*},
    \label{eq:SVD}
\end{equation}
where $\mathbf{U} \in \mathbb{C}^{n \times m}$ is a matrix containing the left singular vectors, $\bm{\Sigma}\in \mathbb{C}^{m \times m}$ is a matrix that only contains entries on the diagonal, which are called singular values $\sigma_k$. The singular values $\sigma_k$ are ordered from high to low. $\mathbf{V^*} \in \mathbb{C}^{m \times m}$ is a matrix that contains the right singular vectors, and $*$ stands for the complex conjugate transpose. For real-valued datasets, like image datasets, we have $\mathbf{V}^* = \mathbf{V}^T$.

In this paper, we perform SVD with the MATLAB function \texttt{svd}. Performing the SVD on a large dataset is computationally expensive and the randomized SVD (rSVD) can improve the computational scaling dramatically. Here, we use the rSVD approach of~\cite{Halko.2011}, which is provided by \cite{brunton2022data} in form of a MATLAB function \texttt{rsvd.m}. The rSVD uses a projection matrix to select $k$ random columns of $\mathbf{X}$ before performing the SVD, where $k$ is the target rank of the rSVD.

\subsubsection{Pattern classification}

We use four machine learning algorithms for pattern classification: A classification tree (Tree), a naive Bayes (NB) classifier, a linear discriminant (LD), and a k-nearest neighbor (kNN) classifier. The initial choice of classifiers was assisted by the MATLAB Classifier App. All classifiers were implemented using MATLAB functions: \texttt{fitctree} (Tree)~\cite{MathWorks.2023b}, \texttt{fitcnb} (NB)~\cite{MathWorks.2023d}, \texttt{fitcdiscr} (LD)~\cite{MathWorks.2023c}, and \texttt{fitcknn} (kNN)~\cite{MathWorks.2023}. 

A classification tree is one of the basic methods of machine learning. It is a hierarchical construct that aims to optimally divide the data~\cite{brunton2022data}. Classification trees have a high degree of interpretability since they can be visualized graphically. A naive Bayes classifier is based on the Bayes theorem and applies a probability distribution to the data, assuming that the predictors are conditionally independent given the class~\cite{MathWorks.2023d}. The linear discriminant aims to maximize inter-class distance and minimize intra-class distance, by projection into a sub-space~\cite{brunton2022data}. A k-nearest-neighbor classifier uses the training data as a lookup table and the k-nearest neighbors of an unlabeled data point determine its label~\cite{flach2012machine}.

For assessment of the classifier performance, we use several performance metrics. In case of our 3-class classification problem, we write down a schematic confusion matrix as given in \autoref{tab:3class_conf_mat}, from which we derive the equations for our performance metrics: 
\begin{align}
\text{accuracy} &=\frac{t_A+t_B+t_C}{t_A+f_{A,B}+f_{A,C}+f_{B,A}+t_B+f_{B,C}+f_{C,A}+f_{C,B}+t_C}
\label{eq:acc}\\
\text{error}& =1 - \text{accuracy}
\label{eq:err}\\
\text{recall A} &=\frac{t_A}{t_A+f_{A,B}+f_{A,C}}
\label{eq:RecA}\\
\text{recall B}& =\frac{t_B}{f_{B,A}+t_B+f_{B,C}}
\label{eq:RecB}\\
\text{recall C}& =\frac{t_C}{f_{C,A}+f_{C,B}+t_C}.
\label{eq:RecC}
\end{align}

\begin{table}
	\caption{3-class confusion matrix.}
	\label{tab:3class_conf_mat}
	\centering
		\begin{tabular}{l l l l l}
		& &  \multicolumn{3}{c}{\textbf{Prediction}}\\
		\cline{3-5}
		& & A & B & C \\
		\cline{2-5}	
		\multirow{3}{*}{\textbf{Ground truth}} & A & $t_A$ & $f_{A,B}$ & $f_{A,C}$\\	
		& B & $f_{B,A}$ & $t_B$ & $f_{B,C}$\\	
		& C & $f_{C,A}$ & $f_{C,B}$ & $t_C$\\			
		\cline{2-5}
		\end{tabular}
\end{table}

\subsubsection{Code availability}
\label{subsubsec:Code availability}

The complete workflow is implemented in MATLAB~2022b. The code is publicly available~\cite{published_code}.

%% file: Figures/Dots_few/dots_few.tex
\begin{tikzpicture}

\begin{axis}[%
width=1.575in,
height=1.575in,
at={(0in,1.575in)},
scale only axis,
axis on top,
xmin=0,
xmax=260,
xtick={\empty},
tick align=outside,
y dir=reverse,
ymin=0,
ymax=260,
ytick={\empty},
ticks=none
]
\addplot [forget plot] graphics [xmin=0.5, xmax=260.5, ymin=0.5, ymax=260.5] {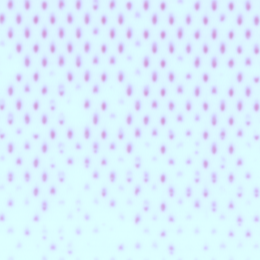};
\end{axis}

\begin{axis}[%
width=1.575in,
height=1.575in,
at={(1.575in,1.575in)},
scale only axis,
axis on top,
xmin=0,
xmax=260,
xtick={\empty},
tick align=outside,
y dir=reverse,
ymin=0,
ymax=260,
ytick={\empty},
ticks=none
]
\addplot [forget plot] graphics [xmin=0.5, xmax=260.5, ymin=0.5, ymax=260.5] {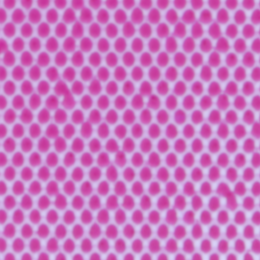};
\end{axis}

\begin{axis}[%
width=1.575in,
height=1.575in,
at={(3.15in,1.575in)},
scale only axis,
axis on top,
xmin=0,
xmax=260,
xtick={\empty},
tick align=outside,
y dir=reverse,
ymin=0,
ymax=260,
ytick={\empty},
ticks=none
]
\addplot [forget plot] graphics [xmin=0.5, xmax=260.5, ymin=0.5, ymax=260.5] {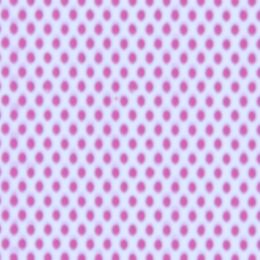};
\end{axis}

\begin{axis}[%
width=1.575in,
height=1.575in,
at={(0in,0in)},
scale only axis,
axis on top,
xmin=0,
xmax=260,
xtick={\empty},
tick align=outside,
y dir=reverse,
ymin=0,
ymax=260,
ytick={\empty},
ticks=none
]
\addplot [forget plot] graphics [xmin=0.5, xmax=260.5, ymin=0.5, ymax=260.5] {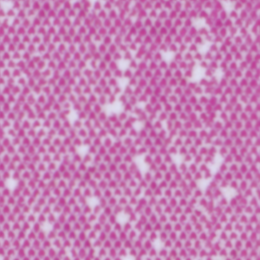};
\end{axis}

\begin{axis}[%
width=1.575in,
height=1.575in,
at={(1.575in,0in)},
scale only axis,
axis on top,
xmin=0,
xmax=260,
xtick={\empty},
tick align=outside,
y dir=reverse,
ymin=0,
ymax=260,
ytick={\empty},
ticks=none
]
\addplot [forget plot] graphics [xmin=0.5, xmax=260.5, ymin=0.5, ymax=260.5] {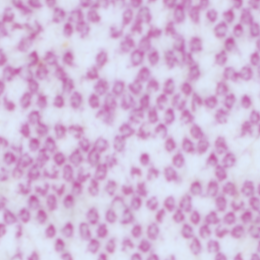};
\end{axis}

\begin{axis}[%
width=1.575in,
height=1.575in,
at={(3.15in,0in)},
scale only axis,
axis on top,
xmin=0,
xmax=260,
xtick={\empty},
tick align=outside,
y dir=reverse,
ymin=0,
ymax=260,
ytick={\empty},
ticks=none
]
\addplot [forget plot] graphics [xmin=0.5, xmax=260.5, ymin=0.5, ymax=260.5] {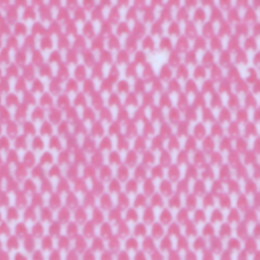};
\end{axis}
\end{tikzpicture}%

%% file: Figures/Mixed_few/mixed_few.tex
\begin{tikzpicture}

\begin{axis}[%
width=1.575in,
height=1.575in,
at={(0in,1.575in)},
scale only axis,
axis on top,
xmin=0,
xmax=260,
xtick={\empty},
tick align=outside,
y dir=reverse,
ymin=0,
ymax=260,
ytick={\empty},
ticks=none
]
\addplot [forget plot] graphics [xmin=0.5, xmax=260.5, ymin=0.5, ymax=260.5] {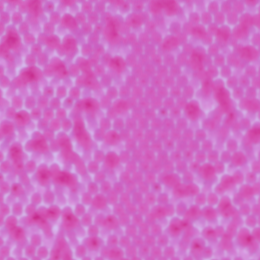};
\end{axis}

\begin{axis}[%
width=1.575in,
height=1.575in,
at={(1.575in,1.575in)},
scale only axis,
axis on top,
xmin=0,
xmax=260,
xtick={\empty},
tick align=outside,
y dir=reverse,
ymin=0,
ymax=260,
ytick={\empty},
ticks=none
]
\addplot [forget plot] graphics [xmin=0.5, xmax=260.5, ymin=0.5, ymax=260.5] {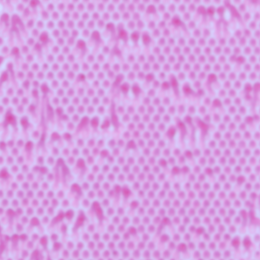};
\end{axis}

\begin{axis}[%
width=1.575in,
height=1.575in,
at={(3.15in,1.575in)},
scale only axis,
axis on top,
xmin=0,
xmax=260,
xtick={\empty},
tick align=outside,
y dir=reverse,
ymin=0,
ymax=260,
ytick={\empty},
ticks=none
]
\addplot [forget plot] graphics [xmin=0.5, xmax=260.5, ymin=0.5, ymax=260.5] {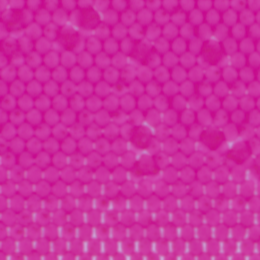};
\end{axis}

\begin{axis}[%
width=1.575in,
height=1.575in,
at={(0in,0in)},
scale only axis,
axis on top,
xmin=0,
xmax=260,
xtick={\empty},
tick align=outside,
y dir=reverse,
ymin=0,
ymax=260,
ytick={\empty},
ticks=none
]
\addplot [forget plot] graphics [xmin=0.5, xmax=260.5, ymin=0.5, ymax=260.5] {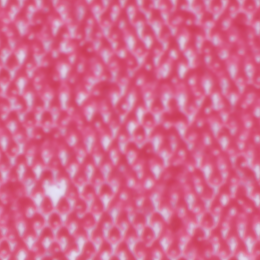};
\end{axis}

\begin{axis}[%
width=1.575in,
height=1.575in,
at={(1.575in,0in)},
scale only axis,
axis on top,
xmin=0,
xmax=260,
xtick={\empty},
tick align=outside,
y dir=reverse,
ymin=0,
ymax=260,
ytick={\empty},
ticks=none
]
\addplot [forget plot] graphics [xmin=0.5, xmax=260.5, ymin=0.5, ymax=260.5] {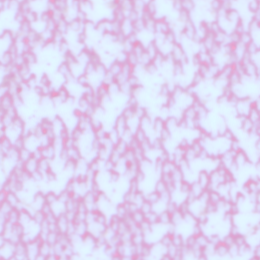};
\end{axis}

\begin{axis}[%
width=1.575in,
height=1.575in,
at={(3.15in,0in)},
scale only axis,
axis on top,
xmin=0,
xmax=260,
xtick={\empty},
tick align=outside,
y dir=reverse,
ymin=0,
ymax=260,
ytick={\empty},
ticks=none
]
\addplot [forget plot] graphics [xmin=0.5, xmax=260.5, ymin=0.5, ymax=260.5] {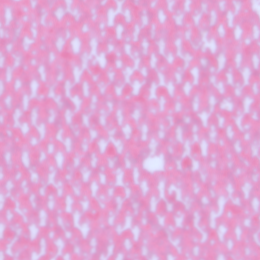};
\end{axis}
\end{tikzpicture}%

%% file: Figures/Fingers_few/fingers_few.tex
\begin{tikzpicture}

\begin{axis}[%
width=1.575in,
height=1.575in,
at={(0in,1.575in)},
scale only axis,
axis on top,
xmin=0,
xmax=260,
xtick={\empty},
tick align=outside,
y dir=reverse,
ymin=0,
ymax=260,
ytick={\empty},
ticks=none
]
\addplot [forget plot] graphics [xmin=0.5, xmax=260.5, ymin=0.5, ymax=260.5] {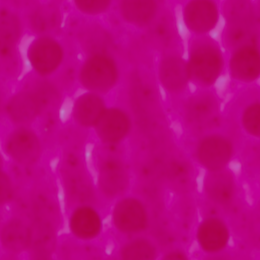};
\end{axis}

\begin{axis}[%
width=1.575in,
height=1.575in,
at={(1.575in,1.575in)},
scale only axis,
axis on top,
xmin=0,
xmax=260,
xtick={\empty},
tick align=outside,
y dir=reverse,
ymin=0,
ymax=260,
ytick={\empty},
ticks=none
]
\addplot [forget plot] graphics [xmin=0.5, xmax=260.5, ymin=0.5, ymax=260.5] {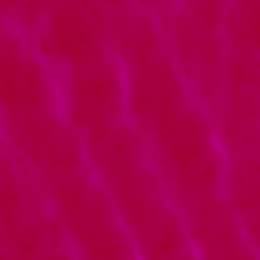};
\end{axis}

\begin{axis}[%
width=1.575in,
height=1.575in,
at={(3.15in,1.575in)},
scale only axis,
axis on top,
xmin=0,
xmax=260,
xtick={\empty},
tick align=outside,
y dir=reverse,
ymin=0,
ymax=260,
ytick={\empty},
ticks=none
]
\addplot [forget plot] graphics [xmin=0.5, xmax=260.5, ymin=0.5, ymax=260.5] {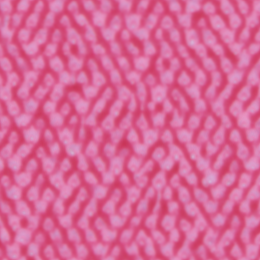};
\end{axis}

\begin{axis}[%
width=1.575in,
height=1.575in,
at={(0in,0in)},
scale only axis,
axis on top,
xmin=0,
xmax=260,
xtick={\empty},
tick align=outside,
y dir=reverse,
ymin=0,
ymax=260,
ytick={\empty},
ticks=none
]
\addplot [forget plot] graphics [xmin=0.5, xmax=260.5, ymin=0.5, ymax=260.5] {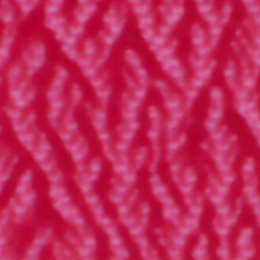};
\end{axis}

\begin{axis}[%
width=1.575in,
height=1.575in,
at={(1.575in,0in)},
scale only axis,
axis on top,
xmin=0,
xmax=260,
xtick={\empty},
tick align=outside,
y dir=reverse,
ymin=0,
ymax=260,
ytick={\empty},
ticks=none
]
\addplot [forget plot] graphics [xmin=0.5, xmax=260.5, ymin=0.5, ymax=260.5] {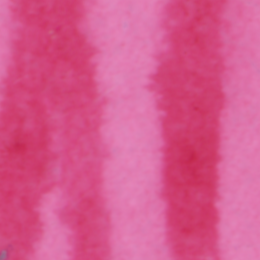};
\end{axis}

\begin{axis}[%
width=1.575in,
height=1.575in,
at={(3.15in,0in)},
scale only axis,
axis on top,
xmin=0,
xmax=260,
xtick={\empty},
tick align=outside,
y dir=reverse,
ymin=0,
ymax=260,
ytick={\empty},
ticks=none
]
\addplot [forget plot] graphics [xmin=0.5, xmax=260.5, ymin=0.5, ymax=260.5] {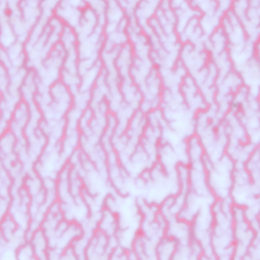};
\end{axis}
\end{tikzpicture}%

%% file: Figures/Schematic_workflow_V5.pdf_tex
%% Creator: Inkscape inkscape 0.92.5, www.inkscape.org
%% Accompanies image file 'Schematic_workflow_V2.pdf' (pdf, eps, ps)
\begingroup%
  \makeatletter%
  \providecommand\color[2][]{%
    \errmessage{(Inkscape) Color is used for the text in Inkscape, but the package 'color.sty' is not loaded}%
    \renewcommand\color[2][]{}%
  }%
  \providecommand\transparent[1]{%
    \errmessage{(Inkscape) Transparency is used (non-zero) for the text in Inkscape, but the package 'transparent.sty' is not loaded}%
    \renewcommand\transparent[1]{}%
  }%
  \providecommand\rotatebox[2]{#2}%
  \newcommand*\fsize{\dimexpr\f@size pt\relax}%
  \newcommand*\lineheight[1]{\fontsize{\fsize}{#1\fsize}\selectfont}%
  \ifx\svgwidth\undefined%
    \setlength{\unitlength}{894.10336953bp}%
    \ifx\svgscale\undefined%
      \relax%
    \else%
      \setlength{\unitlength}{\unitlength * \real{\svgscale}}%
    \fi%
  \else%
    \setlength{\unitlength}{\svgwidth}%
  \fi%
  \global\let\svgwidth\undefined%
  \global\let\svgscale\undefined%
  \makeatother%
  \begin{picture}(1,0.26249173)%
    \lineheight{1}%
    \setlength\tabcolsep{0pt}%
    \put(0,0){\includegraphics[width=\unitlength,page=1]{Schematic_workflow_V5.pdf}}%
    \put(0.0076923,0.24329023){\color[rgb]{0,0,0}\makebox(0,0)[lt]{\lineheight{1.25}\smash{\begin{tabular}[t]{l}\textbf{a}\end{tabular}}}}%
    \put(0.00748044,0.10472757){\color[rgb]{0,0,0}\makebox(0,0)[lt]{\lineheight{1.25}\smash{\begin{tabular}[t]{l}\textbf{b}\end{tabular}}}}%
    \put(0,0){\includegraphics[width=\unitlength,page=2]{Schematic_workflow_V5.pdf}}%
    \put(0.84,0.14732372){\color[rgb]{0,0,0}\makebox(0,0)[lt]{\lineheight{1.25}\smash{\begin{tabular}[t]{l}{\footnotesize $\text{Test error} = 13~\%$}\end{tabular}}}}%
    \put(0.84,0.01155679){\color[rgb]{0,0,1}\makebox(0,0)[lt]{\lineheight{1.25}\smash{\begin{tabular}[t]{l}{\footnotesize $\text{Test error} = 3~\%$}\end{tabular}}}}%
    \put(0.21341563,0.21425495){\color[rgb]{0,0,0}\makebox(0,0)[lt]{\lineheight{1.25}\smash{\begin{tabular}[t]{l}{\footnotesize rSVD}\end{tabular}}}}%
    \put(0.44527731,0.21448745){\color[rgb]{0,0,0}\makebox(0,0)[lt]{\lineheight{1.25}\smash{\begin{tabular}[t]{l}{\footnotesize Truncation}\end{tabular}}}}%
    \put(0.62535837,0.21448745){\color[rgb]{0,0,0}\makebox(0,0)[lt]{\lineheight{1.25}\smash{\begin{tabular}[t]{l}{\footnotesize Projection}\end{tabular}}}}%
    \put(0.81493049,0.21382883){\color[rgb]{0,0,0}\makebox(0,0)[lt]{\lineheight{1.25}\smash{\begin{tabular}[t]{l}{\footnotesize Classifier}\end{tabular}}}}%
    \put(0.93,0.19157806){\color[rgb]{0,0,0}\makebox(0,0)[lt]{\lineheight{1.25}\smash{\begin{tabular}[t]{l}{\footnotesize Class}\end{tabular}}}}%
    \put(0.3156124,0.07946604){\color[rgb]{0,0,0}\makebox(0,0)[lt]{\lineheight{1.25}\smash{\begin{tabular}[t]{l}{\footnotesize rSVD}\end{tabular}}}}%
    \put(0.16023113,0.07974125){\color[rgb]{0,0,1}\makebox(0,0)[t]{\lineheight{1.25}\smash{\begin{tabular}[t]{c}{\footnotesize FFT magnitude}\end{tabular}}}}%
    \put(0.44743871,0.07999431){\color[rgb]{0,0,0}\makebox(0,0)[lt]{\lineheight{1.25}\smash{\begin{tabular}[t]{l}{\footnotesize Truncation}\end{tabular}}}}%
    \put(0.62535837,0.07977575){\color[rgb]{0,0,0}\makebox(0,0)[lt]{\lineheight{1.25}\smash{\begin{tabular}[t]{l}{\footnotesize Projection}\end{tabular}}}}%
    \put(0.81546742,0.07921565){\color[rgb]{0,0,0}\makebox(0,0)[lt]{\lineheight{1.25}\smash{\begin{tabular}[t]{l}{\footnotesize Classifier}\end{tabular}}}}%
    \put(0.04476127,0.19253666){\color[rgb]{0,0,0}\makebox(0,0)[lt]{\lineheight{1.25}\smash{\begin{tabular}[t]{l}{\footnotesize $\mathbf{X}$}\end{tabular}}}}%
    \put(0.04497842,0.05806251){\color[rgb]{0,0,0}\makebox(0,0)[lt]{\lineheight{1.25}\smash{\begin{tabular}[t]{l}{\footnotesize $\mathbf{X}$}\end{tabular}}}}%
    \put(0.4,0.1932541){\color[rgb]{0,0,0}\makebox(0,0)[lt]{\lineheight{1.25}\smash{\begin{tabular}[t]{l}{\footnotesize $\mathbf{U}$}\end{tabular}}}}%
    \put(0.74937053,0.19398876){\color[rgb]{0,0,0}\makebox(0,0)[lt]{\lineheight{1.25}\smash{\begin{tabular}[t]{l}{\footnotesize $\mathbf{U_{\text{r}}^*X}$}\end{tabular}}}}%
    \put(0,0){\includegraphics[width=\unitlength,page=3]{Schematic_workflow_V5.pdf}}%
    \put(0.93,0.05690283){\color[rgb]{0,0,0}\makebox(0,0)[lt]{\lineheight{1.25}\smash{\begin{tabular}[t]{l}{\footnotesize Class}\end{tabular}}}}%
    \put(0,0){\includegraphics[width=\unitlength,page=4]{Schematic_workflow_V5.pdf}}%
    \put(0.25632407,0.05847003){\color[rgb]{0,0,1}\makebox(0,0)[lt]{\lineheight{1.25}\smash{\begin{tabular}[t]{l}{\footnotesize $\mathbf{|\hat{X}|}$}\end{tabular}}}}%
    \put(0.4,0.05824256){\color[rgb]{0,0,0}\makebox(0,0)[lt]{\lineheight{1.25}\smash{\begin{tabular}[t]{l}{\footnotesize $\mathbf{\hat{U}}$}\end{tabular}}}}%
    \put(0.575,0.05824256){\color[rgb]{0,0,0}\makebox(0,0)[lt]{\lineheight{1.25}\smash{\begin{tabular}[t]{l}{\footnotesize $\mathbf{\hat{U}_{\text{r}}}$}\end{tabular}}}}%
    \put(0.74269192,0.05847003){\color[rgb]{0,0,0}\makebox(0,0)[lt]{\lineheight{1.25}\smash{\begin{tabular}[t]{l}{\footnotesize $\mathbf{\hat{U}_{\text{r}}^*|\hat{X}|}$}\end{tabular}}}}%
    \put(0.575,0.1932541){\color[rgb]{0,0,0}\makebox(0,0)[lt]{\lineheight{1.25}\smash{\begin{tabular}[t]{l}{\footnotesize $\mathbf{U_{\text{r}}}$}\end{tabular}}}}%
  \end{picture}%
\endgroup%

%% file: 3_Results.tex
\section{Results}
\label{sec:Results}

We will first present the results of the singular value decomposition (SVD) of the complete dataset (Section~\ref{subsec:SVD}). We show that an FFT applied to the dataset before performing SVD reduces the dimensionality of our dataset. Then, we show test error and recall of four different machine learning pattern classification algorithms on our dataset following dimensionality reduction using a randomized SVD (rSVD) (Section~\ref{subsec:Classification}). We learn that a k-nearest neighbor (kNN) classifier applied to the FFT-transformed data performs best and outperforms a human observer. Additionally, we evaluate the influence of other factors, such as dataset balancing, on classification. Finally, we show how to use the trained machine learning models to classify unlabeled images from the HYPA-p dataset and create fluid splitting regime maps (Section~\ref{subsec:Regime_maps}). These regime maps provide novel insights about the dynamics of fluid splitting in gravure printing.

\subsection{Singular value decomposition}
\label{subsec:SVD}

\begin{figure}[t]
	\centering
		\begin{subfigure}[t]{0.48\textwidth}
		\vskip 0pt
		\centering
		\input{Figures/sing_values.tex}
		\caption{\textbf{All modes}}
		\label{fig:sing_values}
	\end{subfigure}	
	\hfill
\begin{subfigure}[t]{0.48\textwidth}
		\centering
		\vskip 0pt
		\input{Figures/sing_values_zoom.tex}
		\caption{\textbf{Modes 1-20}}
		\label{fig:sing_values_zoom}
	\end{subfigure}	
	\caption{Normalized singular values $\sigma_{k,n}$ and cumulative energy in \% over first $r$ modes. In the complete view (a), only every $500^{\text{th}}$ data point is plotted for better clarity. The zoomed in view (b) shows all data points for the first 20 modes.}	\label{fig:CumEngSingVal}
\end{figure}
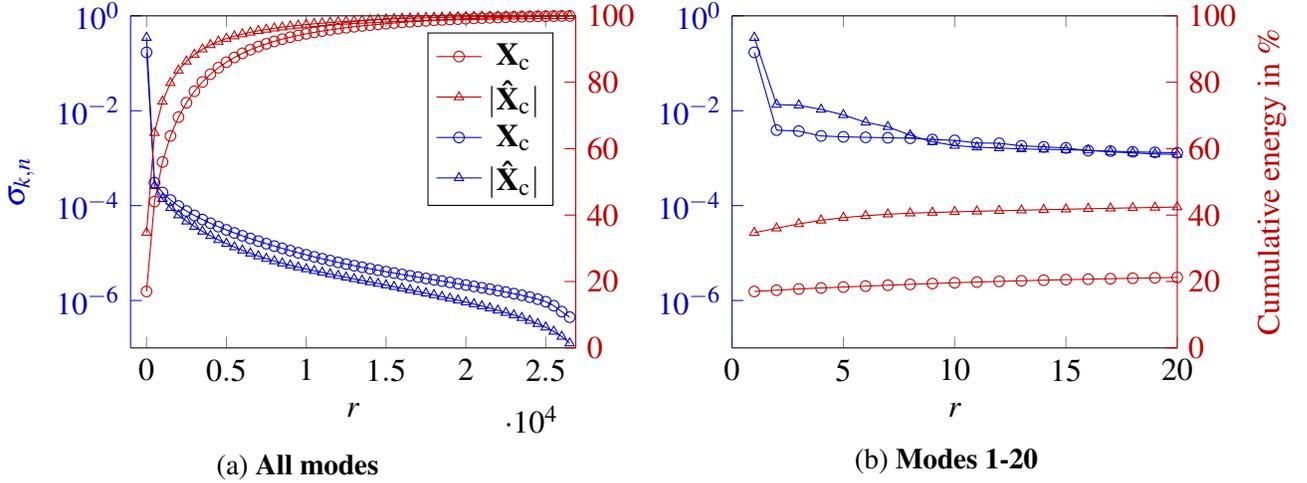

\begin{figure}[t]
\centering
		\begin{subfigure}[b]{0.48\textwidth}
		\centering
		\raisebox{-0.5cm}{\input{Figures/Modes/modes_U_few.tex}}
	\end{subfigure}	
	\hfill
\begin{subfigure}[b]{0.48\textwidth}
		\centering
		\raisebox{-0.5cm}{\input{Figures/Modes/modes_Uhat_few.tex}}
	\end{subfigure}	
\caption{Selected modes $\mathbf{U}$ (a) and selected FFT modes $\mathbf{\hat{U}}$ (b). The modes were obtained by using SVD on the complete dataset $\mathbf{X_{\text{c}}}$ and on the complete, FFT-transformed dataset $|\mathbf{\hat{X}_{\text{c}}}|$, respectively. No train-test-split was performed. The FFT modes $\mathbf{\hat{U}}$ are displayed in the spatial domain to aid in interpretation; this is achieved by taking the magnitude of the inverse FFT with phase zero for each FFT mode in $\mathbf{\hat{U}}$.}
\label{fig:Modes_U_few}
\end{figure}
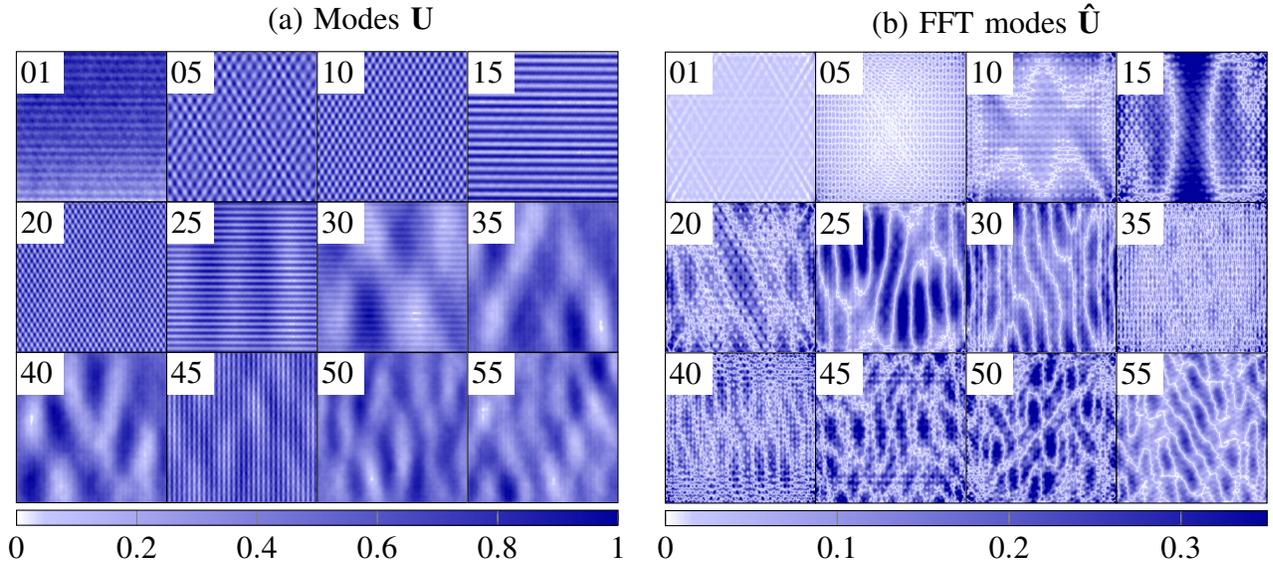

In this section, we show the decay and the cumulative energy of the singular values for the complete dataset with and without the FFT applied (\autoref{fig:CumEngSingVal}). Additionally, we explore the characteristics of the SVD modes of the complete dataset before and after the FFT (\autoref{fig:Modes_U_few}). Note that we are using a standard SVD for this analysis.

In \autoref{fig:CumEngSingVal}, the normalized singular values $\sigma_{k,n}$, along with their cumulative energy, are plotted for $r$ SVD modes. Normalized singular values are computed by dividing each singular value $\sigma_k$ by the sum of all singular values $\sum_{j=1}^m\sigma_j$:
\begin{equation}
    \sigma_{k,n} = \frac{\sigma_k}{\sum_{j=1}^m\sigma_j}
    \label{eq:norm_sing_val}
\end{equation}

\autoref{fig:CumEngSingVal}a shows the singular values of the first \num{26880} modes, while \autoref{fig:CumEngSingVal}b displays the first twenty modes. The normalized singular values $\sigma_{k,n}$ of the first eight modes for $|\mathbf{\hat{X}_{\text{c}}}|$ contain more cumulative energy than those of $\mathbf{X_{\text{c}}}$ (35\,\% versus 17\,\%). Approximately 90\,\% of the energy for $\mathbf{X_{\text{c}}}$ is contained within the first \num{7000} modes, while approximately 90\,\% of the energy for $|\mathbf{\hat{X}_{\text{c}}}|$ is contained within the first \num{3500} modes. Therefore, using the FFT can reduce the number of modes required to accurately represent the dataset by half.

\autoref{fig:Modes_U_few}a displays 12 selected SVD modes ($\mathbf{U}$) of the complete dataset $\mathbf{X_{\text{c}}}$. More SVD modes can be found in the Appendix (\autoref{fig:Modes_U}). The darker blue regions correspond to a higher concentration of printing fluid per unit area. Since no mean subtraction was performed during the SVD, mode~\#01 represents the mean of $\mathbf{X_{\text{c}}}$. In this first mode, a slight gradient of fluid distribution from top to bottom is observed, which can also be confirmed by visual examination of the dataset and is likely due to doctor blading effects in some images. Modes~\#05, \#10, \#15 and \#20 show patterns that correspond to the raster frequencies of the printing form, whereas modes~\#25, \#30, \#35, \#40, \#45, \#50 and \#55 predominantly exhibit finger patterns that are partially superposed with raster patterns.

\autoref{fig:Modes_U_few}b shows 12 selected SVD modes ($\mathbf{\hat{U}}$) of the complete, FFT-transformed dataset $|\mathbf{\hat{X}_{\text{c}}}|$, henceforth referred to as \emph{FFT modes}. More FFT modes can be found in the Appendix (\autoref{fig:Modes_Uhat}).  For interpretability, the FFT modes $\mathbf{\hat{U}}$ are displayed in the spatial domain and not in the frequency domain; this is achieved by taking the magnitude of the inverse FFT with phase zero for each FFT mode in $\mathbf{\hat{U}}$. Consequently, by using a phase of zero instead of the actual phase, we lose the phase information (compare \autoref{fig:Schematic_workflow}b). However, the phase is not important for our final goal of pattern classification. Thus, the phase loss is acceptable.
FFT mode~\#01 again corresponds to the mean of the dataset. In contrast to \autoref{fig:Modes_U_few}a, only FFT mode~\#05 primarily shows a raster pattern. All other displayed FFT modes, starting from FFT mode~\#10 predominantly exhibit finger patterns. \autoref{fig:Modes_U} and \autoref{fig:Modes_Uhat} in the Appendix confirm that there are significantly fewer modes dedicated to the raster pattern if the FFT is used (modes~\#02 to \#21 compared to FFT modes~\#02 to \#08).
Our initial aim of FFT-transforming the dataset was to reduce the number of modes exhibiting spatially repeating raster patterns. Additionally, by keeping only the FFT magnitude we removed the phase shift of the raster pattern between images. It is possible that imperfections in the dataset, such as slightly rotated raster patterns, prevent further reduction. Nevertheless, the FFT is able to reduce the dimensionality of our dataset.

\subsection{Classification}
\label{subsec:Classification}

Four different machine learning classifiers are trained and tested on the manually classified images. These classifiers are a classification tree, a naive Bayes (NB) classifier, a linear discriminant (LD), and a k-nearest neighbor (kNN) classifier. We investigate the influence of the FFT (Section~\ref{subsubsec:FFT}), dataset normalization (Section~\ref{subsubsec:Normalization}), dataset balancing (Section~\ref{subsubsec:Balancing}), and rSVD target rank, and balancing split (Section~\ref{subsubsec:targetRank_balancingSplit}). A benchmark from human observers is presented in Section~\ref{subsubsec:Benchmark} and confusion matrices from an example kNN model are shown in Section~\ref{subsubsec:Confusions_kNN}.

\subsubsection{Influence of FFT}
\label{subsubsec:FFT}

\begin{figure}[t]
\centering
\begin{subfigure}[b]{0.48\textwidth}
    \centering
    \input{Figures/balancing0_norm0_noFFT_50rSVDmodes.tex}
    \caption{\textbf{Without FFT}}
\end{subfigure}
\hfill
\hfill
\begin{subfigure}[b]{0.48\textwidth}
    \centering    \input{Figures/balancing0_norm0_FFTtransformed_50rSVDmodes_duplicate}
    \caption{\textbf{With FFT}}
\end{subfigure}
\caption{Test error over the rank of truncation $r$. Comparison of the classifier results without FFT (a) and with FFT (b). In both figures, an unbalanced, non-normalized dataset was used and the target rank of the rSVD was chosen as $k=50$. The error bands denote the standard deviation of the test error over five training cycles with different train-test-splits.}
\label{fig:noFFTvsFFT}
\end{figure}
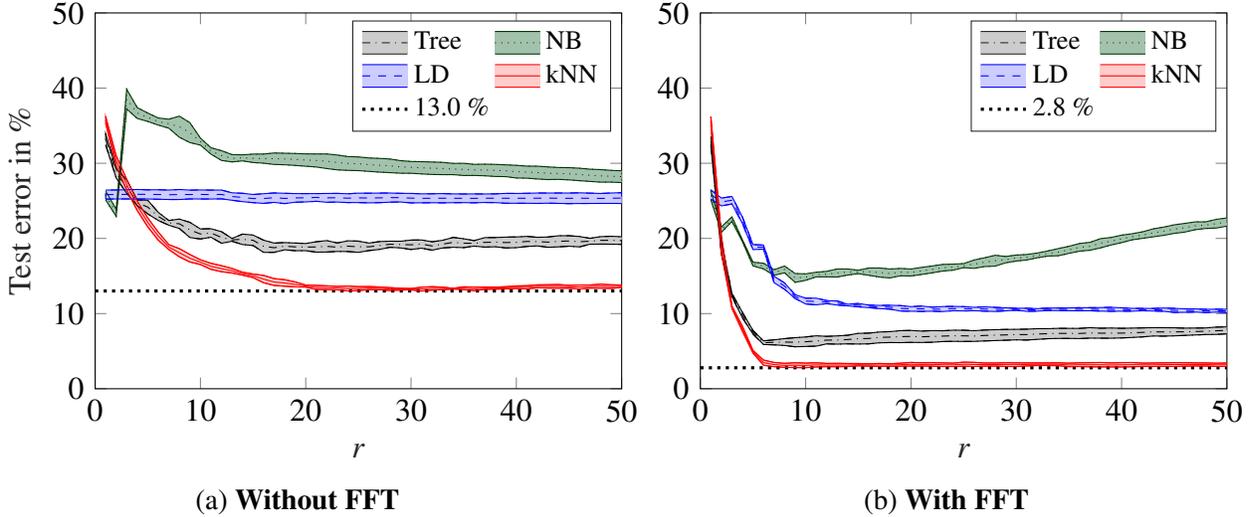

 The training of four different machine learning classifiers is performed using the two variants of the workflow outlined in \autoref{fig:Schematic_workflow}. In variant~(a), the data is used directly, while in variant~(b), a 2-dimensional FFT is applied. \autoref{fig:noFFTvsFFT} depicts the test error of all four classifiers plotted against the rank of truncation $r$ for each variant. For higher values of $r$, more modes of the dataset are utilized by the classifier for training. The shaded error bands represent the standard deviation of the test error across five training cycles. For each training cycle, a random split of \SI{80}{\percent} training and \SI{20}{\percent} test data is used. The target rank of the rSVD remains constant at $k=50$.

From \autoref{fig:noFFTvsFFT} it is evident that the kNN classifier exhibits the lowest test error for both variants. The minimum test error is $E_{\text{test}}=\SI{13}{\percent}$ for variant~(a) and $E_{\text{test}}=\SI{2.8}{\percent}$ for variant~(b). The second lowest test error is achieved by the classification tree, followed by the LD and the NB classifiers.

The test error for the kNN without the FFT converges at approximately $r=21$, while the error for the kNN classification using the FFT data converges at approximately $r=7$. 
 
Thus, variant~(a) without FFT requires more modes to reach a higher minimum test error compared to variant~(b) with FFT. This observation aligns with findings from \autoref{fig:CumEngSingVal}, which demonstrate that the cumulative energy of the first eight singular values are higher when using an FFT. This shift of energy towards the initial modes likely explains why the kNN classifier requires fewer modes to converge in the FFT-aided variant. With fewer modes containing the same amount of energy, we have a better low rank representation and so we need fewer modes for the classification. We hypothesize that the kNN classifier only requires modes associated with the raster patterns of the four raster frequencies $f_r=\SI{60}{\lines\per\cm}$, $\SI{70}{\lines\per\cm}$, $\SI{80}{\lines\per\cm}$, and $\SI{100}{\lines\per\cm}$. The SVD modes of $\mathbf{X_{\text{c}}}$ exhibit raster patterns up to mode~\#21 (see \autoref{fig:Modes_U}). For the FFT modes ($|\hat{\mathbf{X}}_{\text{c}}|$), raster patterns are only observed until around FFT mode~\#8 (see \autoref{fig:Modes_Uhat}). This closely matches the values of $r$ where the kNN classifier converges in \autoref{fig:noFFTvsFFT}. Due to the significantly lower test errors in the FFT-aided variant, this approach is used for subsequent investigations.

\subsubsection{Influence of normalization}
\label{subsubsec:Normalization}

\begin{figure}[t]
\centering
\begin{subfigure}[b]{0.48\textwidth}
    \centering    \input{Figures/balancing0_norm0_FFTtransformed_50rSVDmodes_duplicate2}
    \caption{\textbf{Non-normalized}}
\end{subfigure}
\hfill
\begin{subfigure}[b]{0.48\textwidth}
    \centering    \input{Figures/balancing0_norm1_FFTtransformed_50rSVDmodes.tex}
    \caption{\textbf{Normalized}}
\end{subfigure}
\caption{Test error over $r$. Comparison of the classifier results for a non-normalized (a) and a normalized dataset (b). In both figures, an unbalanced, FFT-transformed dataset was used and the target rank of the rSVD was chosen as $k=50$.}
\label{fig:normalized_notnormalized}
\end{figure}
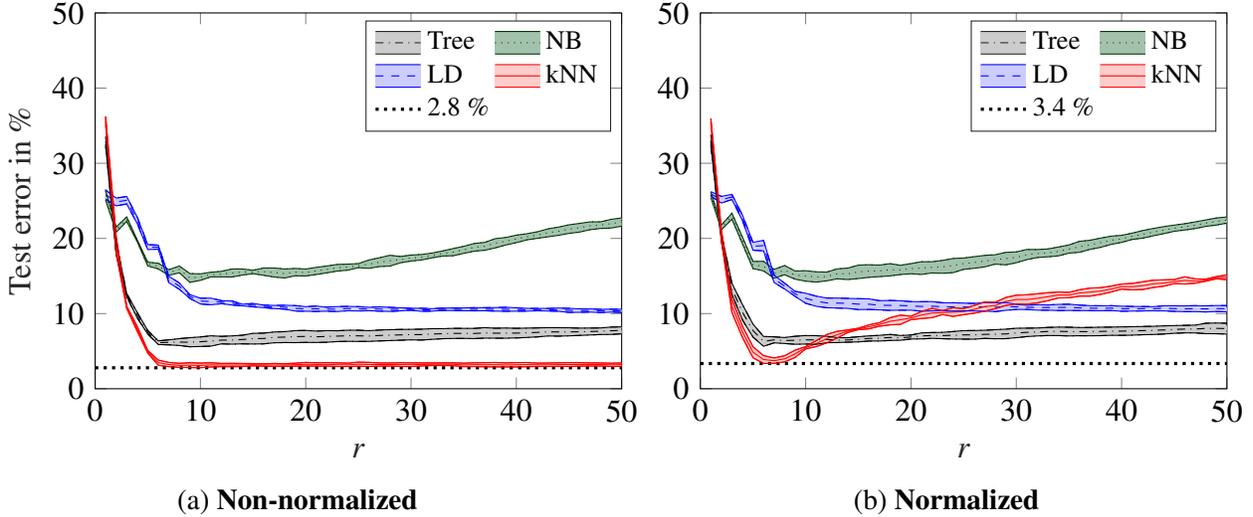

In \autoref{fig:normalized_notnormalized}, the influence of normalizing the truncated, projected modes ($\mathbf{U_{\text{r}}^*X}$) to have a mean of zero and unity standard deviation is examined for the FFT-transformed dataset. In \autoref{fig:normalized_notnormalized}a, the results without normalization are presented and \autoref{fig:normalized_notnormalized}b shows the results with normalization. In both plots, the test error is plotted against the rank of truncation $r$ for the four different classifiers with an rSVD target rank of $k=50$ for all cases. The lowest test error is achieved by the kNN classifier in both cases, with 2.8\,\% for \autoref{fig:normalized_notnormalized}a and 3.4\,\% for \autoref{fig:normalized_notnormalized}b, showing similar values.

However, the test error for the kNN classifier with data normalization is lowest at around $r=7$ and starts to rise approximately linearly starting at $r=8$. This re-increase is not observed for the kNN classifier without data normalization. However, a similar re-increase is observed both within the non-normalized and normalized results for the NB classifier as well. The classification tree shows a slight re-increase in both cases, as well. 

Normalization does not have a significant impact on the minimum test error, but it does alter the behavior of the test error for the kNN classifier across different values of $r$. When using normalized data, the ideal value for $r$ needs to be known to achieve the minimum test error for the kNN classifier. In contrast, with non-normalized data, $r$ can be intentionally set higher without increasing the test error. It is advantageous if the test error converges to a minimal value as $r$ increases. In such cases, it suffices for further investigations to use a larger value for $r$ and separate examination of the test error for small $r$ values is unnecessary. 

The test error re-increase for normalized data may be related to the \glq{}curse of dimensionality\grq ~\cite{flach2012machine}. We hypothesize that normalization alters the distances between points in the $r$-dimensional space in a way that leads to a less distinct separation and thus higher test error for the kNN classifier. Due to the re-increase for the kNN classifier beyond $r=8$, the variant without data normalization is chosen for subsequent investigations.

\subsubsection{Influence of dataset balancing}
\label{subsubsec:Balancing}

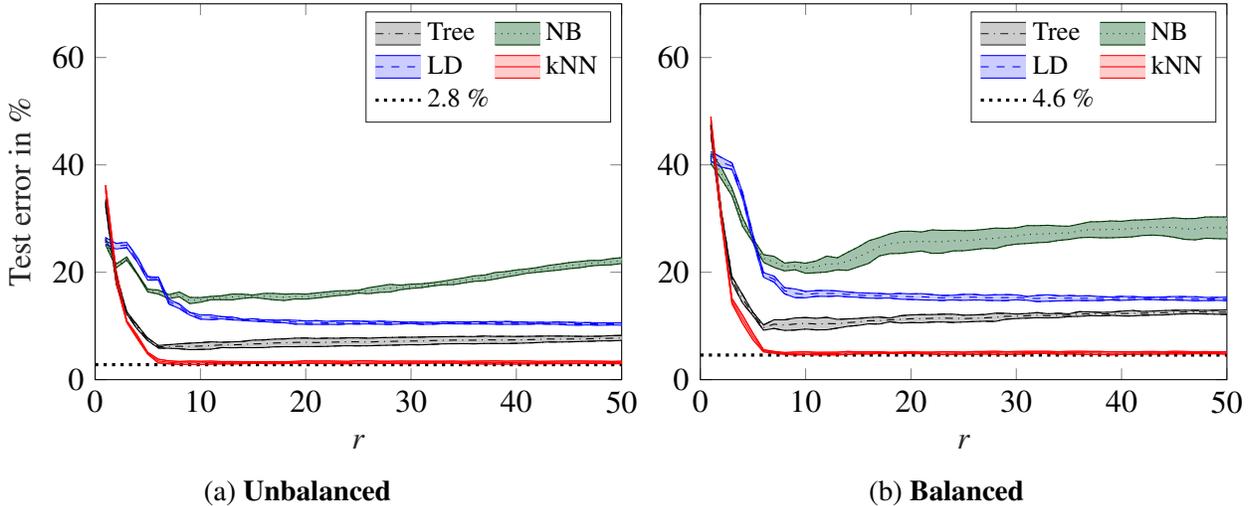
\begin{figure}[t]
\centering
\begin{subfigure}[b]{0.48\textwidth}
    \centering    \input{Figures/balancing0_norm0_FFTtransformed_50rSVDmodes}
    \caption{\textbf{Unbalanced}}
    \label{fig:error_unbalanced}
\end{subfigure}
\hfill
\begin{subfigure}[b]{0.48\textwidth}
    \centering    
    \input{Figures/balancing1_norm0_FFTtransformed_50rSVDmodes}
    \caption{\textbf{Balanced}}
    \label{fig:error_balanced}
\end{subfigure}
\caption{Test error over $r$. Comparison of the classifier results for an unbalanced (a) and a balanced dataset (b). Unbalanced dataset means 34.8~\% dots, 13.9~\% mixed, 51.3~\% fingers. Balanced dataset means 33.3~\% dots, 33.3~\% mixed, 33.3~\% fingers. In both figures, a non-normalized, FFT-transformed dataset was used and the target rank of the rSVD was chosen as $k=50$.}
\label{fig:error_unbalanced_balanced}
\end{figure}

In \autoref{fig:error_unbalanced_balanced}, the test errors for an unbalanced dataset (a) and a balanced dataset (b) are compared. In the unbalanced dataset, 34.8\,\% of the images are classified as dot patterns, 13.9\,\% as mixed patterns, and 51.3\,\% as finger patterns. This distribution applies to both the training and test datasets. In the balanced dataset, the number of images from over-represented classes has been reduced, resulting in an even percentage distribution of 33.3\,\% for each regime. The diagrams display the test error for the four different classifiers against the rank of truncation $r$.

\begin{figure}[t]
\centering
\begin{subfigure}[b]{0.48\textwidth}
    \centering    \input{Figures/balancing0_norm0_FFTtransformed_50rSVDmodes_recall_mixed_class}
    \caption{\textbf{Unbalanced}}
\end{subfigure}
\hfill
\begin{subfigure}[b]{0.48\textwidth}
    \centering    
    \input{Figures/balancing1_norm0_FFTtransformed_50rSVDmodes_recall_mixed_class}
    \caption{\textbf{Balanced}}
\end{subfigure}
\caption{Recall B over $r$ for the mixed class. Comparison of the classifier results for an unbalanced (a) and a balanced dataset (b). Unbalanced dataset means 34.8~\% dots, 13.9~\% mixed, 51.3~\% fingers. Balanced dataset means 33.3~\% dots, 33.3~\% mixed, 33.3~\% fingers. In all figures, a non-normalized, FFT-transformed dataset was used and the target rank of the rSVD was chosen as $k=50$.}
\label{fig:Recall}
\end{figure}
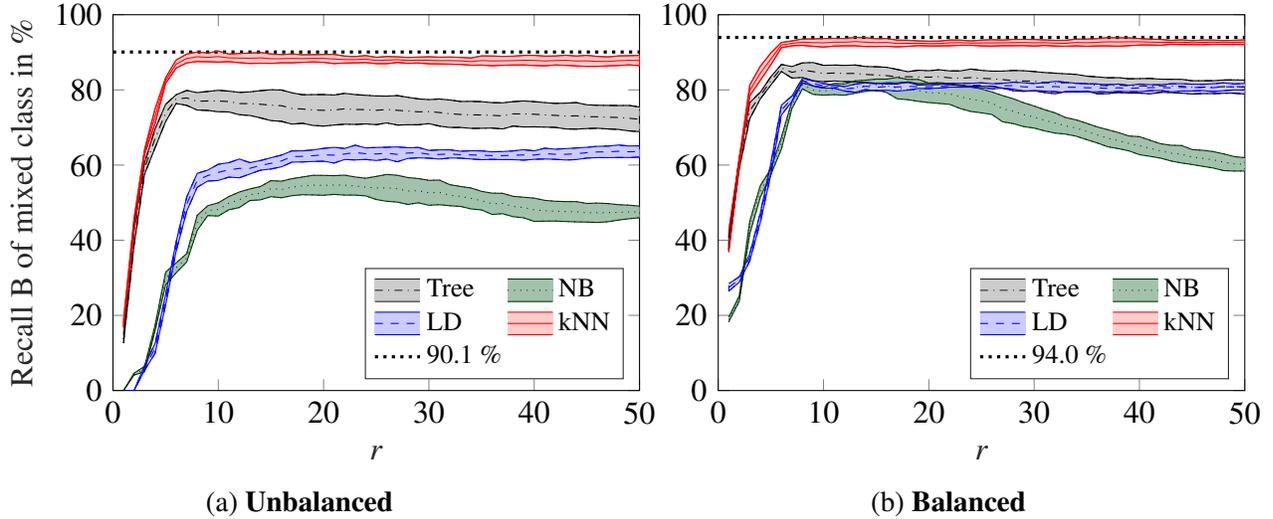

\autoref{fig:error_unbalanced_balanced} shows that the minimum test error for the balanced dataset is approximately $4.6\,\%$, which is around 2\,\% higher than the minimum test error for the unbalanced dataset. From this error increase, it would be reasonable to assume that balancing is disadvantageous. However, from previous studies we know that the mixed class is the hardest to classify~\cite{Brumm2021Ink}, therefore, we aim for a high recall of this class (recall~B). \autoref{fig:Recall} shows that dataset balancing is advantageous because it increases the maximum recall~B for all trained classifiers. The kNN model exhibits the highest recall~B, which is about 4\,\% higher for the balanced dataset.

The increased recall~B in the balanced dataset could result from reducing the representation bias of the more prevalent classes. However, with the increased recall~B for the balanced dataset, there is a slight reduction in recall~A for the dots class (approximately $-2\,\%$) and a slight reduction in recall~C for the fingers class ($-1\,\%$), see \autoref{tab:Recall} for a summary and \autoref{fig:Recall_Appendix} in the Appendix for more details.

\begin{table}[t]
    \centering
    \caption{Recall for unbalanced and balanced dataset.}
	\label{tab:Recall}
\begin{tabular}{|l|l|l|}
    \hline
         & \textbf{Unbalanced} & \textbf{Balanced} \\
     \hline
     Recall A (dots) & 98.3\,\% & 96.6\,\% \\
     Recall B (mixed) & 90.1\,\% & 94.0\,\% \\
     Recall C (fingers) & 98.9\,\% & 97.9\,\% \\
     \hline
\end{tabular}
\end{table}

\subsubsection{Influence of rSVD target rank and balancing split}
\label{subsubsec:targetRank_balancingSplit}

\begin{figure}[t]
    \centering
    \begin{subfigure}[b]{0.48\textwidth}
    \centering\input{Figures/test_error_k}
    \end{subfigure}
    \hfill    
    \begin{subfigure}[b]{0.48\textwidth}
    \centering\input{Figures/test_error_balancing_split}
    \end{subfigure}
    \caption{Test error of kNN classifier over $r$. Influence of rSVD target rank $k$ (a) and different, random balancing splits (b). Both diagrams display the average test error from five training cycles with random train-test-splits.}
    \label{fig:k_and_balancingsplit}
\end{figure}
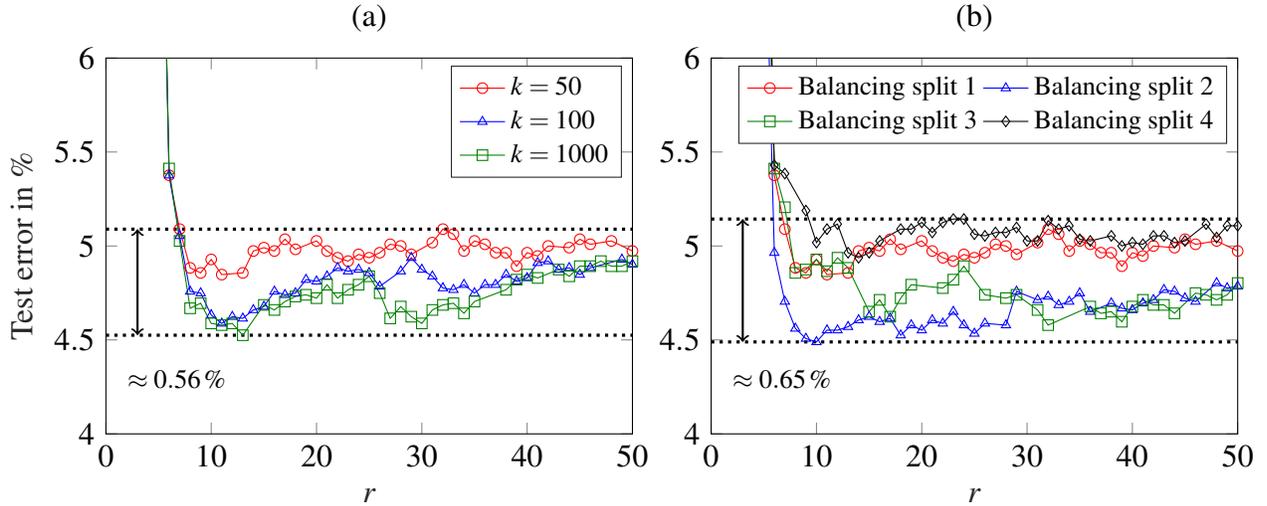

In \autoref{fig:k_and_balancingsplit} we assess how the rSVD target rank $k$ and the balancing split affect the average test error from five training cycles with random train-test-splits. 

\autoref{fig:k_and_balancingsplit}a illustrates that higher values of $k$ tend to result in lower test errors. However, test error differences between $k=50$ and $k=1000$ are expected to be small (around $ \num{0.56}\,\%$). This is determined by subtracting the highest observed test error of $k=50$ and the lowest observed test error of $k=1000$ for $r\geq10$; see horizontal lines in \autoref{fig:k_and_balancingsplit}a. Increasing the rSVD target rank $k$ results in a marginal error reduction but comes at a higher computational cost. The computational time of the rSVD at $k=100$ is approximately \num{1.5} times and at $k=1000$ is approximately \num{15} times that of $k=50$. This is why $k=50$ is chosen for all other investigations in this study.

A possible explanation why lower rSVD target ranks $k$ produce only a marginally higher test error is provided in \autoref{fig:target_rank} in the Appendix. \autoref{fig:target_rank}a shows the first 50 rSVD modes for $k=50$ and \autoref{fig:target_rank}b for $k=1000$. While modes \#01 to \#06 show raster patterns and look similar for $k=50$ and $k=1000$, modes $\geq$\#07 look quite different for the two values of $k$. The modes $\geq$\#07 for $k=50$ look noisy, whereas the ones for $k=1000$ look like finger patterns and qualitatively similar to the modes $\geq$\#07 from \autoref{fig:Modes_Uhat}, which were computed using standard SVD. Consequently, rSVD with a target rank of $k=1000$ can better approximate the results of a standard SVD, especially for modes $\geq$\#07. However, the test error for the kNN classifier converges at $r=7$ (\autoref{fig:noFFTvsFFT}b), meaning that only modes \#01 to \#07 are needed for a adequate classification. 

In \autoref{fig:k_and_balancingsplit}b, $k=50$ is held constant and the impact of various random train-test divisions of a balanced dataset is examined. Depending on how the division is performed, the resulting balanced dataset could randomly contain more or fewer \glq easy-to-classify\grq{} images, which is likely to affect the resulting test error.  \autoref{fig:k_and_balancingsplit}b emphasizes that the expected deviation of the test error for different random divisions into a balanced dataset is relatively low, at around $\num{0.65}\,\%$. This is determined by subtracting the highest observed test error (which occurs in balancing split~4) and the lowest observed test error (which occurs in balancing split~2) for $r\geq10$; see horizontal lines in \autoref{fig:k_and_balancingsplit}b.

\subsubsection{Benchmark from human observers}
\label{subsubsec:Benchmark}

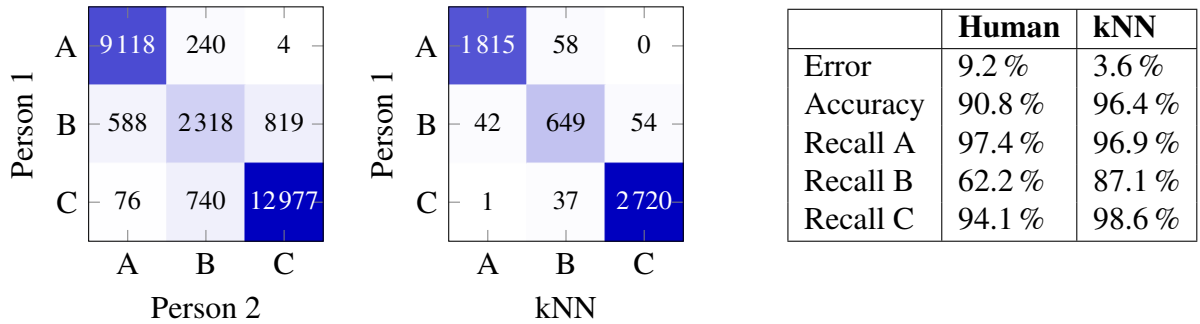
\begin{figure}[t]
\centering
\begin{subfigure}[b]{0.28\textwidth}
\centering
\input{Figures/conf_mat_person1_person2.tex}
\end{subfigure}
\begin{subfigure}[b]{0.28\textwidth}
\centering
\input{Figures/exemplary_conf_mat_knn1_hat}
\end{subfigure}
\hfill
\begin{subfigure}[b]{0.4\textwidth}
\centering
\raisebox{2.6cm}{\begin{tabular}{|l|l|l|}
    \hline
     & \textbf{Human} & \textbf{kNN}  \\
     \hline
     Error & 9.2\,\% & 3.6\,\% \\
     Accuracy & 90.8\,\% & 96.4\,\%\\
     Recall A & 97.4\,\% & 96.9\,\%\\
     Recall B & 62.2\,\% & 87.1\,\% \\
     Recall C & 94.1\,\% & 98.6\,\%\\
     \hline
\end{tabular}}
\end{subfigure}
\caption{Left: Benchmark from human observers. Confusion matrix for the manual classification of all \num{26880} labeled images from thy HYPA-p dataset by person~1 and person~2. Person~1's classifications are regarded as ground truth. Middle: Confusion matrix on test dataset with \num{5376} images for person~1 and an exemplary kNN model with $k=50$, $r=7$, trained on FFT-transformed, unbalanced, non-normalized data. Right: Performance metrics for human and kNN calculated from the two confusion matrices.}
\label{fig:confusion_matrices_metrics}
\end{figure}

As a benchmark for the test error and recall, we show results from a visual experiment with two human observers: person~1 and person~2. The classifications from person~1 are regarded as ground truth, since person~1 is an expert in the dataset and person~2 is a non-expert trained by person~1. The labels provided within the HYPA-p dataset are the labels from person~1. The confusion matrix between person~1 and person~2 on the complete labeled dataset (see \autoref{fig:confusion_matrices_metrics} left) reveals that the mixed pattern (class~B) is the most challenging to classify because most misclassifications or \emph{confusions} happen between class~B and the other two classes (A and C). The confusion matrix between person 1 and an exemplary kNN model trained on the FFT-transformed, unbalanced, non-normalized test dataset is depicted in \autoref{fig:confusion_matrices_metrics} (middle) and also shows that most confusions are related to class~B. 

From the two confusion matrices, we calculate the performance metrics as presented in the table in \autoref{fig:confusion_matrices_metrics} (right). The metrics show that the kNN performs better than the human observer in almost all cases. One exception is the recall of A, where the human performs slightly better than the kNN.

\subsubsection{Confusions of kNN model}
\label{subsubsec:Confusions_kNN}

\begin{figure}[t]
    \centering
\begin{subfigure}[t]{0.16\textwidth}     
            \centering
            \frame{\includegraphics[width=1\textwidth]{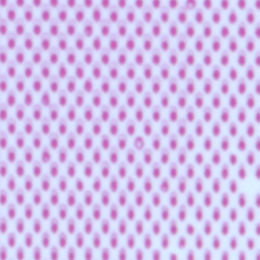}}
            \caption{{\footnotesize Truth: dots, prediction: dots}}
    \end{subfigure}
    \hfill
            \begin{subfigure}[t]{0.16\textwidth}     
            \centering\frame{\includegraphics[width=1\textwidth]{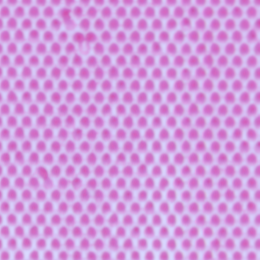}}
            \caption{{\footnotesize Truth: dots, prediction: mixed}}
    \end{subfigure}  
    \hfill
    \begin{subfigure}[t]{0.16\textwidth}     
            \centering\frame{\includegraphics[width=1\textwidth]{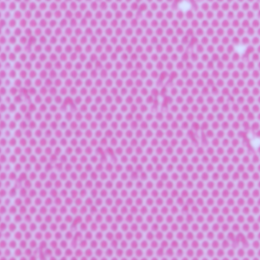}}
            \caption{{\footnotesize Truth: mixed, prediction: dots}}
    \end{subfigure}
    \hfill
        \begin{subfigure}[t]{0.16\textwidth}     
            \centering
            \frame{\includegraphics[width=1\textwidth]{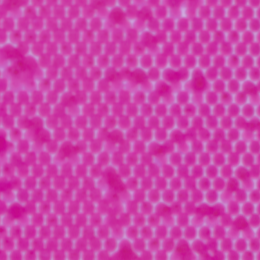}}
            \caption{{\footnotesize Truth: mixed, prediction: mixed}}
    \end{subfigure}
        \hfill
        \begin{subfigure}[t]{0.16\textwidth}     
            \centering\frame{\includegraphics[width=1\textwidth]{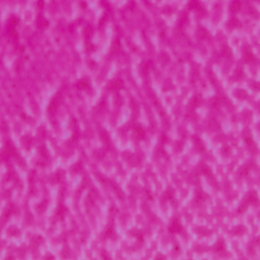}}
            \caption{{\footnotesize Truth: mixed, prediction: fingers}}
    \end{subfigure}
            \hfill
        \begin{subfigure}[t]{0.16\textwidth}     
            \centering\frame{\includegraphics[width=1\textwidth]{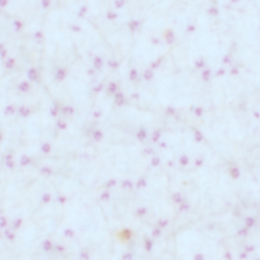}}
            \caption{{\footnotesize Truth: fingers*, prediction: dots}}
    \end{subfigure}
    \caption{Examples for correctly and incorrectly classified patterns by the kNN model. Each patch has a size of \SI{260}{\px}~x~\SI{260}{\px} (\SI{2.75}{\mm}~x~\SI{2.75}{\mm}). The ground truth labels from person~1 are called \glq truth\grq, and the classifications from the kNN are called \glq prediction\grq. The incorrectly classified patterns are borderline cases and would also be difficult for the human observer to judge. In example (f), the kNN detected a classification mistake from the human observer. *Classification mistake by human observer, prediction is indeed correct.}
    \label{fig:misclassifications}
\end{figure}

Using the same kNN model as in \autoref{fig:confusion_matrices_metrics} (middle), we show examples of correctly and incorrectly classified patterns in \autoref{fig:misclassifications}. The incorrectly classified patterns in \autoref{fig:misclassifications}b, c and e are borderline cases and would also be difficult for the human observer to judge. Interestingly, in \autoref{fig:misclassifications}f, the kNN model detected a classification mistake from person~1. The image clearly shows dot patterns, which is also the prediction from the kNN model, however, person~1 erroneously labeled it as finger patterns. This demonstrates how a trained model could be used to check the labels from the human observer. First, the model would be trained on the labeled images from the human observer. Then, the trained model could be used to predict the class of all labeled images. All images, where the prediction does not match the label from the human observer, could then be reviewed by a human observer for potential mistakes by the human observer. Using this pipeline, the data quality of the labels could be improved. 

\subsection{Regime maps}
\label{subsec:Regime_maps}

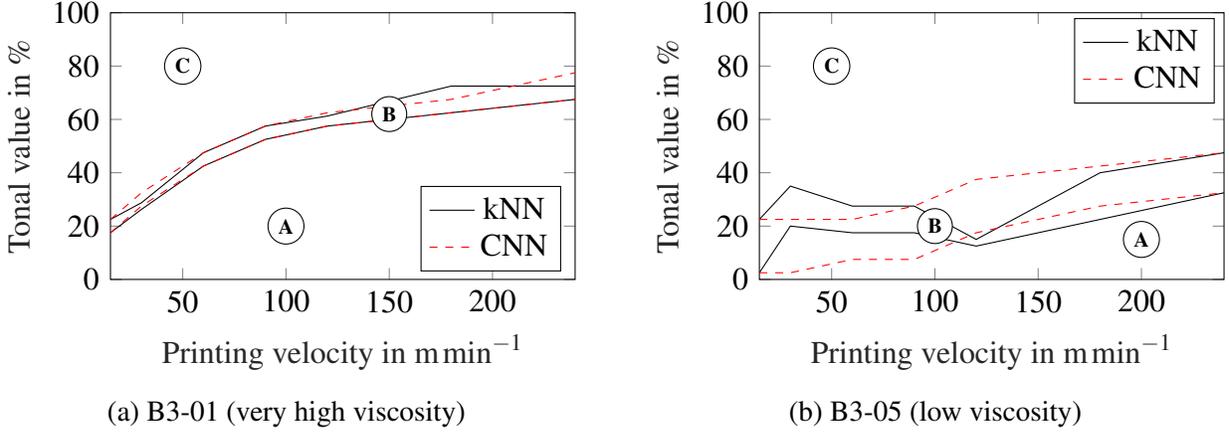
\begin{figure}[t]
	\centering
		\begin{subfigure}[b]{0.48\textwidth}
		\centering        \input{Figures/2024_B3-01_R060_ESA0_thresh_u_-4_thresh_o_4}
		\caption{B3-01 (very high viscosity)}
		\label{fig:Regime_B3-01_R060}
	\end{subfigure}	
	\hfill
\begin{subfigure}[b]{0.48\textwidth}
		\centering		\input{Figures/2024_B3-05_R060_ESA0_thresh_u_-4_thresh_o_4}
		\caption{B3-05 (low viscosity)}
		\label{fig:Regime_B3-05_R060}
	\end{subfigure}
		\caption{Regime maps for printing experiment B3-01 with very high viscosity (a) and for printing experiment B3-05 with low viscosity (b), compare \autoref{tab:DoE}. Raster frequency is \SI{60}{\lines\per\cm}. The black solid lines are the regime borders that are based on the results from our kNN classification. The red dashed lines result from a classification using CNNs by \cite{thesis_Pauline}. A~-~point splitting regime, B~-~transition regime, C~-~lamella splitting regime.}	
        \label{fig:Regime_maps}
\end{figure}

To demonstrate the use of the trained machine learning classifiers, we use the kNN model with $k=50$, $r=7$, trained on FFT-transformed, unbalanced, non-normalized data to automatically classify selected images from the HYPA-p dataset. The classification results are used to draw fluid splitting regime maps (\autoref{fig:Regime_maps}), which illustrate the correlation of fluid splitting class and printing process parameters, specifically printing speed and tonal value. 

The regime maps are created using the method from~\cite{RothmannBrumm.2023} and the code from~\cite{tudatalib/3961}. Each regime map is based on the classification results of \num{6720} images and the regime borders as lines. For details on the regime map construction, ~\cite{RothmannBrumm.2023}. The upper regime border separates the lamella splitting regime (C) and the transition regime (B) and the lower regime border separates the transition regime (B) and the point splitting regime (A).

As a plausibility check, the created regime maps are compared to regime maps from \cite{thesis_Pauline}, which used a convolutional neural network (CNN), namely MobileNetV2 \cite{Sandler.2018}, with a test error of \SI{6}{\percent} for pattern classification. \autoref{fig:Regime_maps} shows the regime maps for experiments B3-01 and B3-05 for our kNN and for the CNN. B3-01 uses a printing ink with very high viscosity whereas B3-05 uses a very low viscous ink, compare \autoref{tab:DoE}. The raster frequency is set at \SI{60}{\lines\per\cm}. In \autoref{fig:Regime_maps} we can see that the regime borders for our kNN classification are similar to the ones from the CNN classification, however, they are not the same. Especially in \autoref{fig:Regime_maps}b, deviations between the regime borders from the kNN and the CNN can be observed. 

From the regime maps, we can extract information about the fluid dynamics of fluid splitting. We learn that for the lower viscous ink (\autoref{fig:Regime_maps}b), the regime borders are shifted towards lower tonal values, i.\,e. lower transfer volumes of printing ink, compared to \autoref{fig:Regime_maps}a. We also observe that the transition regime shifts towards higher tonal values for increasing printing velocities for both inks and that the transition regime spans a larger area for the lower viscous ink. 

%% file: Figures/sing_values.tex
\definecolor{mycolor1}{rgb}{0.85000,0.32500,0.09800}%
\begin{tikzpicture}

\pgfplotsset{
xmin=-1000,xmax=26880,
    y axis style/.style={
        yticklabel style=#1,
        ylabel style=#1,
        y axis line style=#1,
        ytick style=#1
   }
}

\begin{axis}[%
width=7.5cm,
height=6cm,
ymin=0.0000001,ymax=1,
ymode = log,
axis y line*=left,
xlabel={$r$},
xlabel near ticks,
ylabel={$\sigma_{k,n}$},
ylabel near ticks,
y axis style=blue!75!black
]
\addplot [color=blue!75!black, mark=o, mark options={solid, blue!75!black}]
  table[row sep=crcr]{%
1	0.169947259917069\\
501	0.000304114261528995\\
1001	0.000188714950979205\\
1501	0.000131111664438279\\
2001	9.81928684363352e-05\\
2501	7.69233556150624e-05\\
3001	6.18812701192801e-05\\
3501	5.09029385585826e-05\\
4001	4.25127063910878e-05\\
4501	3.59830159500037e-05\\
5001	3.07776181626855e-05\\
5501	2.65663834581941e-05\\
6001	2.31137683237125e-05\\
6501	2.02414303161332e-05\\
7001	1.78188866060424e-05\\
7501	1.57691881852368e-05\\
8001	1.40200487031535e-05\\
8501	1.25251000127763e-05\\
9001	1.12436146230837e-05\\
9501	1.01357602941371e-05\\
10001	9.16795964777187e-06\\
10501	8.32382116828334e-06\\
11001	7.58310716335838e-06\\
11501	6.93742276859533e-06\\
12001	6.36500897121741e-06\\
12501	5.86141794797934e-06\\
13001	5.41639846469063e-06\\
13501	5.01863321284652e-06\\
14001	4.66004747070076e-06\\
14501	4.33852957536311e-06\\
15001	4.04616759381785e-06\\
15501	3.7804196580372e-06\\
16001	3.53636856394407e-06\\
16501	3.31256056008358e-06\\
17001	3.10483789870936e-06\\
17501	2.91156731601903e-06\\
18001	2.73112105766085e-06\\
18501	2.56269826144569e-06\\
19001	2.40415447121765e-06\\
19501	2.25469889207137e-06\\
20001	2.11367956103501e-06\\
20501	1.98000814083229e-06\\
21001	1.85278369486194e-06\\
21501	1.73177119618247e-06\\
22001	1.61460139491357e-06\\
22501	1.50181092707726e-06\\
23001	1.39190603898227e-06\\
23501	1.2839681466473e-06\\
24001	1.17595374496255e-06\\
24501	1.06486011082962e-06\\
25001	9.43110082696624e-07\\
25501	7.77240193633299e-07\\
26001	5.82506074258605e-07\\
26501	4.4479756880156e-07\\
};\label{a_plot_1_y1}

\addplot [color=blue!75!black, mark=triangle, mark options={solid, blue!75!black}]
  table[row sep=crcr]{%
1	0.347029514652115\\
501	0.000261077139799113\\
1001	0.000139341716912978\\
1501	8.980512951828e-05\\
2001	6.29843817756399e-05\\
2501	4.67854732334679e-05\\
3001	3.61050713653519e-05\\
3501	2.86120350494725e-05\\
4001	2.30931970805512e-05\\
4501	1.8984360392351e-05\\
5001	1.58280344338402e-05\\
5501	1.33661071128326e-05\\
6001	1.14333435762054e-05\\
6501	9.87335050120406e-06\\
7001	8.62288271739048e-06\\
7501	7.61333520386529e-06\\
8001	6.78427339066075e-06\\
8501	6.09882589036372e-06\\
9001	5.52309511866332e-06\\
9501	5.02998232135416e-06\\
10001	4.60322410591879e-06\\
10501	4.22787109213532e-06\\
11001	3.89479097766922e-06\\
11501	3.59567968422496e-06\\
12001	3.32524358068332e-06\\
12501	3.07826589698938e-06\\
13001	2.85120423910596e-06\\
13501	2.64306206207598e-06\\
14001	2.44961529222487e-06\\
14501	2.27084751147032e-06\\
15001	2.10428965254729e-06\\
15501	1.94847910701341e-06\\
16001	1.80296866005441e-06\\
16501	1.66665639339752e-06\\
17001	1.53957814957092e-06\\
17501	1.42011274732968e-06\\
18001	1.30799973371178e-06\\
18501	1.20276792044667e-06\\
19001	1.10389130928408e-06\\
19501	1.01075817704671e-06\\
20001	9.23104039757926e-07\\
20501	8.40337014567189e-07\\
21001	7.62316126465289e-07\\
21501	6.89130690704359e-07\\
22001	6.19439723488458e-07\\
22501	5.53777865942789e-07\\
23001	4.91640792495053e-07\\
23501	4.32603554355054e-07\\
24001	3.76795162711612e-07\\
24501	3.23255575952796e-07\\
25001	2.71802344118848e-07\\
25501	2.22098628898967e-07\\
26001	1.73816930132384e-07\\
26501	1.258089251685e-07\\
};\label{a_plot_2_y1}
\end{axis}

\begin{axis}[%
width=7.5cm,
height=6cm,
ymin=0,ymax=100,
axis y line*=right,
hide x axis,
y axis style=red!75!black,
legend style={at={(0.95,0.95)},anchor=north east},
]
\addplot [color=red!75!black, mark=o, mark options={solid, red!75!black}]
  table[row sep=crcr]{%
1	16.9947259917062\\
501	44.055458365463\\
1001	55.9866073000885\\
1501	63.829097365935\\
2001	69.491620152312\\
2501	73.8356129320491\\
3001	77.285343608768\\
3501	80.0920855180302\\
4001	82.4179547840213\\
4501	84.3733928551483\\
5001	86.0373501168833\\
5501	87.4673177113364\\
6001	88.7063448650115\\
6501	89.7877314567704\\
7001	90.7372302500553\\
7501	91.5752862691224\\
8001	92.3187912940884\\
8501	92.9814878750294\\
9001	93.5749045908087\\
9501	94.1087321801351\\
10001	94.5906459271282\\
10501	95.0274315636671\\
11001	95.4246532719262\\
11501	95.7873122524761\\
12001	96.1195593835764\\
12501	96.4250040883999\\
13001	96.7067140532563\\
13501	96.9673610091195\\
14001	97.2091329960203\\
14501	97.4339338972968\\
15001	97.6434114527183\\
15501	97.838966085139\\
16001	98.0218041719527\\
16501	98.1929695502658\\
17001	98.3533299920855\\
17501	98.5036580970118\\
18001	98.6446778273967\\
18501	98.7769735889269\\
19001	98.9010914334649\\
19501	99.0175207136344\\
20001	99.1266973912061\\
20501	99.2290130095716\\
21001	99.3248020117826\\
21501	99.4143846216648\\
22001	99.4980104287206\\
22501	99.5759014352043\\
23001	99.6482320687974\\
23501	99.7151162702539\\
24001	99.7766085791882\\
24501	99.8326417337958\\
25001	99.8829038321164\\
25501	99.9262985658861\\
26001	99.9597876427179\\
26501	99.985431840214\\
};\label{a_plot_1_y2}

\addplot [color=red!75!black, mark=triangle, mark options={solid, red!75!black}]
  table[row sep=crcr]{%
1	34.7029514652131\\
501	64.7823149739052\\
1001	74.2356750965228\\
1501	79.821359963662\\
2001	83.5820833920407\\
2501	86.2951704294137\\
3001	88.3503451329452\\
3501	89.9576321789064\\
4001	91.2430961108403\\
4501	92.2897838409262\\
5001	93.1562154315143\\
5501	93.883250352952\\
6001	94.5013259853185\\
6501	95.0324008849857\\
7001	95.4935686903591\\
7501	95.8984519603691\\
8001	96.2576831899678\\
8501	96.5791529738453\\
9001	96.8692421526175\\
9501	97.1327129834353\\
10001	97.373258153315\\
10501	97.5938385924237\\
11001	97.7967545864485\\
11501	97.9838756121062\\
12001	98.1567807575193\\
12501	98.3167580944137\\
13001	98.464905026729\\
13501	98.6021905287962\\
14001	98.729436186175\\
14501	98.847385570396\\
15001	98.9566874108459\\
15501	99.0579417210211\\
16001	99.1516730947733\\
16501	99.2383694097298\\
17001	99.3184783003817\\
17501	99.3924373637674\\
18001	99.4606100718011\\
18501	99.5233516751287\\
19001	99.5809855977641\\
19501	99.6338225052823\\
20001	99.6821418932777\\
20501	99.7262024650336\\
21001	99.7662469699426\\
21501	99.8025123158441\\
22001	99.8352052076152\\
22501	99.8645153035141\\
23001	99.8906337031112\\
23501	99.9137303148673\\
24001	99.9339464555524\\
24501	99.9514277759554\\
25001	99.9662903786666\\
25501	99.9786330346069\\
26001	99.9885267523641\\
26501	99.9960239731772\\
};\label{a_plot_2_y2}
\addlegendimage{/pgfplots/refstyle=a_plot_1_y1}\addlegendentry{$\mathbf{X_{\text{c}}}$}
\addlegendimage{/pgfplots/refstyle=a_plot_2_y1}\addlegendentry{$|\mathbf{\hat{X}_{\text{c}}}|$}
\addlegendimage{/pgfplots/refstyle=a_plot_1_y2}\addlegendentry{$\mathbf{X_{\text{c}}}$}
\addlegendimage{/pgfplots/refstyle=a_plot_2_y2}\addlegendentry{$|\mathbf{\hat{X}_{\text{c}}}|$}
\end{axis}
\end{tikzpicture}%

%% file: Figures/sing_values_zoom.tex
\definecolor{mycolor1}{rgb}{0.85000,0.32500,0.09800}%
\begin{tikzpicture}

\pgfplotsset{
xmin=0,xmax=20,
    y axis style/.style={
        yticklabel style=#1,
        ylabel style=#1,
        y axis line style=#1,
        ytick style=#1
   }
}

\begin{axis}[%
width=7.5cm,
height=6cm,
ymin=0.0000001,ymax=1,
ymode = log,
axis y line*=left,
xlabel={$r$},
xlabel near ticks,
y axis style=blue!75!black,
]
\addplot [color=blue!75!black, mark=o, mark options={solid, blue!75!black}]
  table[row sep=crcr]{%
1	0.169947259917069\\
2	0.00389948449416284\\
3	0.00370764468840267\\
4	0.00293979487846214\\
5	0.00281454244642405\\
6	0.00274495400869397\\
7	0.00268837318608179\\
8	0.00262892030720748\\
9	0.00247565204087127\\
10	0.00237124392376244\\
11	0.00209144797867873\\
12	0.00205977895637295\\
13	0.0018189272877343\\
14	0.00173380767729447\\
15	0.00164193492354064\\
16	0.001439700133966\\
17	0.00140887975434718\\
18	0.00135881399713063\\
19	0.00132664620917524\\
20	0.00130017004877097\\
}; \label{plot_1_y1}

\addplot [color=blue!75!black, mark=triangle, mark options={solid, blue!75!black}]
  table[row sep=crcr]{%
1	0.347029514652115\\
2	0.013493429286638\\
3	0.0130311961411479\\
4	0.0106404435481288\\
5	0.00816597366840304\\
6	0.00568839668106849\\
7	0.00458625186821454\\
8	0.00305683316626793\\
9	0.00221196726572151\\
10	0.00184422981796908\\
11	0.00170129701404939\\
12	0.00163604696853996\\
13	0.00156098546924501\\
14	0.00153434653983257\\
15	0.00150205447388673\\
16	0.00147008709733287\\
17	0.00135474707697749\\
18	0.00128964217795842\\
19	0.00122796725206249\\
20	0.00119477696205512\\
}; \label{plot_2_y1}
\end{axis}

\begin{axis}[%
width=7.5cm,
height=6cm,
ymin=0,ymax=100,
axis y line*=right,
hide x axis,
ylabel={Cumulative energy in \%},
ylabel near ticks,
y axis style=red!75!black,
]

\addplot [color=red!75!black, mark=o, mark options={solid, red!75!black}]
  table[row sep=crcr]{%
1	16.9947259917069\\
2	17.3846744411232\\
3	17.7554389099634\\
4	18.0494183978097\\
5	18.3308726424521\\
6	18.6053680433215\\
7	18.8742053619296\\
8	19.1370973926504\\
9	19.3846625967375\\
10	19.6217869891138\\
11	19.8309317869816\\
12	20.0369096826189\\
13	20.2188024113924\\
14	20.3921831791218\\
15	20.5563766714759\\
16	20.7003466848725\\
17	20.8412346603072\\
18	20.9771160600203\\
19	21.1097806809378\\
20	21.2397976858149\\
};\label{plot_1_y2}

\addplot [color=red!75!black, mark=triangle, mark options={solid, red!75!black}]
  table[row sep=crcr]{%
1	34.7029514652115\\
2	36.0522943938753\\
3	37.3554140079901\\
4	38.419458362803\\
5	39.2360557296433\\
6	39.8048953977502\\
7	40.2635205845716\\
8	40.5692039011984\\
9	40.7904006277706\\
10	40.9748236095675\\
11	41.1449533109724\\
12	41.3085580078264\\
13	41.4646565547509\\
14	41.6180912087342\\
15	41.7682966561228\\
16	41.9153053658561\\
17	42.0507800735539\\
18	42.1797442913497\\
19	42.302541016556\\
20	42.4220187127615\\
};\label{plot_2_y2}
\end{axis}

\end{tikzpicture}%

%% file: Figures/Modes/modes_U_few.tex
\begin{tikzpicture}

\begin{axis}[%
width=0.787in,
height=0.787in,
at={(0in,5.512in)},
scale only axis,
point meta min=0,
point meta max=1,
axis on top,
xmin=0,
xmax=260,
xtick={\empty},
y dir=reverse,
ymin=0,
ymax=260,
ytick={\empty},
axis background/.style={fill=white},
title={(a) Modes $\mathbf{U}$},
title style={at={(1.6,1.1)},anchor=west, text width=6cm},
]
\addplot [forget plot] graphics [xmin=0.5, xmax=260.5, ymin=0.5, ymax=260.5] {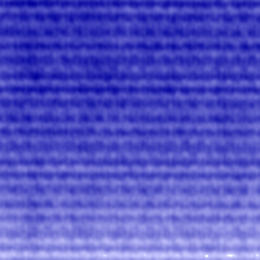};
\node[fill=white, align=center]
at (axis cs:35,35) {01};
\end{axis}

\begin{axis}[%
width=0.787in,
height=0.787in,
at={(0.787in,5.512in)},
scale only axis,
point meta min=0,
point meta max=1,
axis on top,
xmin=0,
xmax=260,
xtick={\empty},
y dir=reverse,
ymin=0,
ymax=260,
ytick={\empty},
axis background/.style={fill=white}
]
\addplot [forget plot] graphics [xmin=0.5, xmax=260.5, ymin=0.5, ymax=260.5] {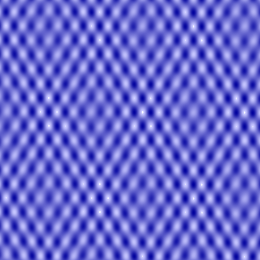};
\node[fill=white, align=center]
at (axis cs:35,35) {05};
\end{axis}

\begin{axis}[%
width=0.787in,
height=0.787in,
at={(1.575in,5.512in)},
scale only axis,
point meta min=0,
point meta max=1,
axis on top,
xmin=0,
xmax=260,
xtick={\empty},
y dir=reverse,
ymin=0,
ymax=260,
ytick={\empty},
axis background/.style={fill=white}
]
\addplot [forget plot] graphics [xmin=0.5, xmax=260.5, ymin=0.5, ymax=260.5] {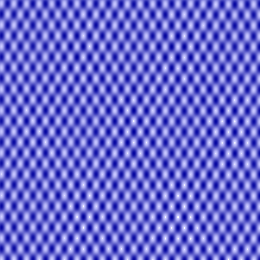};
\node[fill=white, align=center]
at (axis cs:35,35) {10};
\end{axis}

\begin{axis}[%
width=0.787in,
height=0.787in,
at={(2.362in,5.512in)},
scale only axis,
point meta min=0,
point meta max=1,
axis on top,
xmin=0,
xmax=260,
xtick={\empty},
y dir=reverse,
ymin=0,
ymax=260,
ytick={\empty},
axis background/.style={fill=white}
]
\addplot [forget plot] graphics [xmin=0.5, xmax=260.5, ymin=0.5, ymax=260.5] {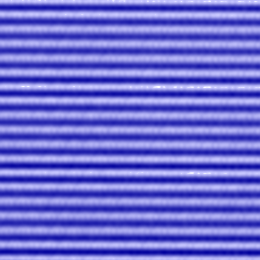};
\node[fill=white, align=center]
at (axis cs:35,35) {15};
\end{axis}

\begin{axis}[%
width=0.787in,
height=0.787in,
at={(0in,4.724in)},
scale only axis,
point meta min=0,
point meta max=1,
axis on top,
xmin=0,
xmax=260,
xtick={\empty},
y dir=reverse,
ymin=0,
ymax=260,
ytick={\empty},
axis background/.style={fill=white}
]
\addplot [forget plot] graphics [xmin=0.5, xmax=260.5, ymin=0.5, ymax=260.5] {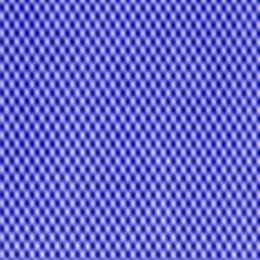};
\node[fill=white, align=center]
at (axis cs:35,35) {20};
\end{axis}

\begin{axis}[%
width=0.787in,
height=0.787in,
at={(0.787in,4.724in)},
scale only axis,
point meta min=0,
point meta max=1,
axis on top,
xmin=0,
xmax=260,
xtick={\empty},
y dir=reverse,
ymin=0,
ymax=260,
ytick={\empty},
axis background/.style={fill=white}
]
\addplot [forget plot] graphics [xmin=0.5, xmax=260.5, ymin=0.5, ymax=260.5] {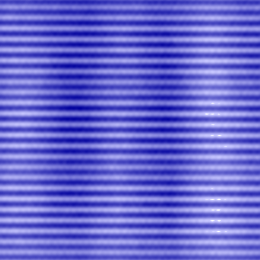};
\node[fill=white, align=center]
at (axis cs:35,35) {25};
\end{axis}

\begin{axis}[%
width=0.787in,
height=0.787in,
at={(1.575in,4.724in)},
scale only axis,
point meta min=0,
point meta max=1,
axis on top,
xmin=0,
xmax=260,
xtick={\empty},
y dir=reverse,
ymin=0,
ymax=260,
ytick={\empty},
axis background/.style={fill=white}
]
\addplot [forget plot] graphics [xmin=0.5, xmax=260.5, ymin=0.5, ymax=260.5] {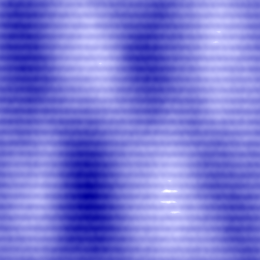};
\node[fill=white, align=center]
at (axis cs:35,35) {30};
\end{axis}

\begin{axis}[%
width=0.787in,
height=0.787in,
at={(2.362in,4.724in)},
scale only axis,
point meta min=0,
point meta max=1,
axis on top,
xmin=0,
xmax=260,
xtick={\empty},
y dir=reverse,
ymin=0,
ymax=260,
ytick={\empty},
axis background/.style={fill=white}
]
\addplot [forget plot] graphics [xmin=0.5, xmax=260.5, ymin=0.5, ymax=260.5] {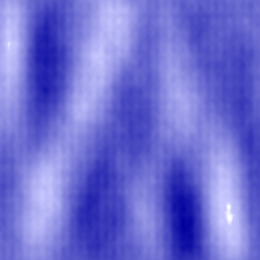};
\node[fill=white, align=center]
at (axis cs:35,35) {35};
\end{axis}

\begin{axis}[%
width=0.787in,
height=0.787in,
at={(0in,3.936in)},
scale only axis,
point meta min=0,
point meta max=1,
axis on top,
xmin=0,
xmax=260,
xtick={\empty},
y dir=reverse,
ymin=0,
ymax=260,
ytick={\empty},
axis background/.style={fill=white}
]
\addplot [forget plot] graphics [xmin=0.5, xmax=260.5, ymin=0.5, ymax=260.5] {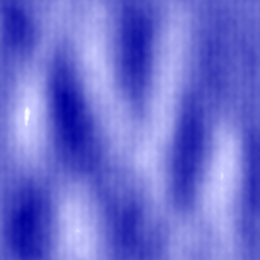};
\node[fill=white, align=center]
at (axis cs:35,35) {40};
\end{axis}

\begin{axis}[%
width=0.787in,
height=0.787in,
at={(0.787in,3.936in)},
scale only axis,
point meta min=0,
point meta max=1,
axis on top,
xmin=0,
xmax=260,
xtick={\empty},
y dir=reverse,
ymin=0,
ymax=260,
ytick={\empty},
axis background/.style={fill=white}
]
\addplot [forget plot] graphics [xmin=0.5, xmax=260.5, ymin=0.5, ymax=260.5] {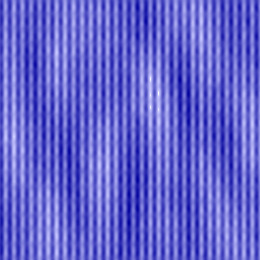};
\node[fill=white, align=center]
at (axis cs:35,35) {45};
\end{axis}

\begin{axis}[%
width=0.787in,
height=0.787in,
at={(1.575in,3.936in)},
scale only axis,
point meta min=0,
point meta max=1,
axis on top,
xmin=0,
xmax=260,
xtick={\empty},
y dir=reverse,
ymin=0,
ymax=260,
ytick={\empty},
axis background/.style={fill=white}
]
\addplot [forget plot] graphics [xmin=0.5, xmax=260.5, ymin=0.5, ymax=260.5] {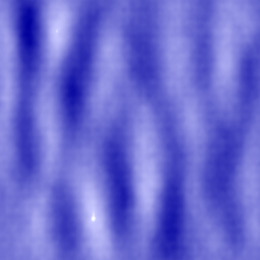};
\node[fill=white, align=center]
at (axis cs:35,35) {50};
\end{axis}

\begin{axis}[%
width=0.787in,
height=0.787in,
at={(2.362in,3.936in)},
scale only axis,
point meta min=0,
point meta max=1,
axis on top,
xmin=0,
xmax=260,
xtick={\empty},
y dir=reverse,
ymin=0,
ymax=260,
ytick={\empty},
axis background/.style={fill=white},
colormap={mymap}{[1pt] rgb(0pt)=(1,1,1); rgb(11pt)=(0.780392,0.780392,1); rgb(255pt)=(0,0,0.611765)},
colorbar horizontal,
colorbar style={
at={(1,-0.15)},anchor=south east,
width=4*
\pgfkeysvalueof{/pgfplots/parent axis width},
xticklabel pos=lower,
colorbar/width=2mm,
},
]
\addplot [forget plot] graphics [xmin=0.5, xmax=260.5, ymin=0.5, ymax=260.5] {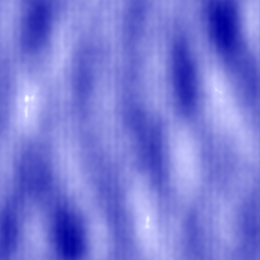};
\node[fill=white, align=center]
at (axis cs:35,35) {55};
\end{axis}

\end{tikzpicture}

%% file: Figures/Modes/modes_Uhat_few.tex
\begin{tikzpicture}

\begin{axis}[%
width=0.787in,
height=0.787in,
at={(0in,5.512in)},
scale only axis,
point meta min=0,
point meta max=1,
axis on top,
xmin=0,
xmax=260,
xtick={\empty},
y dir=reverse,
ymin=0,
ymax=260,
ytick={\empty},
axis background/.style={fill=white},
title={(b) FFT modes $\mathbf{\hat{U}}$},
title style={at={(1.3,1.1)},anchor=west, text width=6cm},
]
\addplot [forget plot] graphics [xmin=0.5, xmax=260.5, ymin=0.5, ymax=260.5] {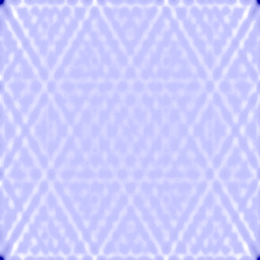};
\node[fill=white, align=center]
at (axis cs:35,35) {01};
\end{axis}

\begin{axis}[%
width=0.787in,
height=0.787in,
at={(0.787in,5.512in)},
scale only axis,
point meta min=0,
point meta max=1,
axis on top,
xmin=0,
xmax=260,
xtick={\empty},
y dir=reverse,
ymin=0,
ymax=260,
ytick={\empty},
axis background/.style={fill=white}
]
\addplot [forget plot] graphics [xmin=0.5, xmax=260.5, ymin=0.5, ymax=260.5] {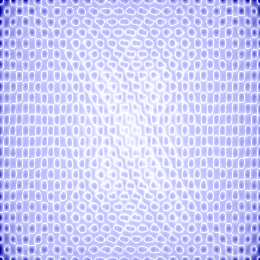};
\node[fill=white, align=center]
at (axis cs:35,35) {05};
\end{axis}

\begin{axis}[%
width=0.787in,
height=0.787in,
at={(1.575in,5.512in)},
scale only axis,
point meta min=0,
point meta max=1,
axis on top,
xmin=0,
xmax=260,
xtick={\empty},
y dir=reverse,
ymin=0,
ymax=260,
ytick={\empty},
axis background/.style={fill=white}
]
\addplot [forget plot] graphics [xmin=0.5, xmax=260.5, ymin=0.5, ymax=260.5] {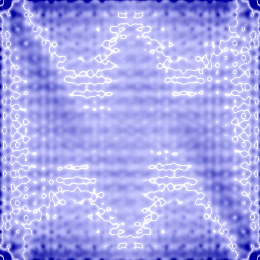};
\node[fill=white, align=center]
at (axis cs:35,35) {10};
\end{axis}

\begin{axis}[%
width=0.787in,
height=0.787in,
at={(2.362in,5.512in)},
scale only axis,
point meta min=0,
point meta max=1,
axis on top,
xmin=0,
xmax=260,
xtick={\empty},
y dir=reverse,
ymin=0,
ymax=260,
ytick={\empty},
axis background/.style={fill=white}
]
\addplot [forget plot] graphics [xmin=0.5, xmax=260.5, ymin=0.5, ymax=260.5] {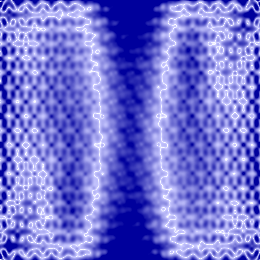};
\node[fill=white, align=center]
at (axis cs:35,35) {15};
\end{axis}

\begin{axis}[%
width=0.787in,
height=0.787in,
at={(0in,4.724in)},
scale only axis,
point meta min=0,
point meta max=1,
axis on top,
xmin=0,
xmax=260,
xtick={\empty},
y dir=reverse,
ymin=0,
ymax=260,
ytick={\empty},
axis background/.style={fill=white}
]
\addplot [forget plot] graphics [xmin=0.5, xmax=260.5, ymin=0.5, ymax=260.5] {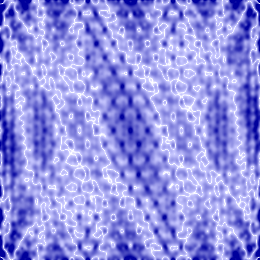};
\node[fill=white, align=center]
at (axis cs:35,35) {20};
\end{axis}

\begin{axis}[%
width=0.787in,
height=0.787in,
at={(0.787in,4.724in)},
scale only axis,
point meta min=0,
point meta max=1,
axis on top,
xmin=0,
xmax=260,
xtick={\empty},
y dir=reverse,
ymin=0,
ymax=260,
ytick={\empty},
axis background/.style={fill=white}
]
\addplot [forget plot] graphics [xmin=0.5, xmax=260.5, ymin=0.5, ymax=260.5] {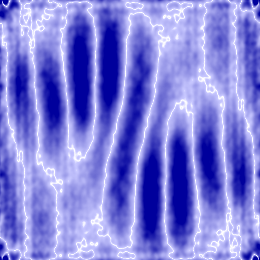};
\node[fill=white, align=center]
at (axis cs:35,35) {25};
\end{axis}

\begin{axis}[%
width=0.787in,
height=0.787in,
at={(1.575in,4.724in)},
scale only axis,
point meta min=0,
point meta max=1,
axis on top,
xmin=0,
xmax=260,
xtick={\empty},
y dir=reverse,
ymin=0,
ymax=260,
ytick={\empty},
axis background/.style={fill=white}
]
\addplot [forget plot] graphics [xmin=0.5, xmax=260.5, ymin=0.5, ymax=260.5] {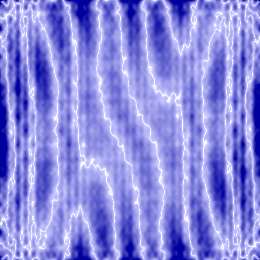};
\node[fill=white, align=center]
at (axis cs:35,35) {30};
\end{axis}

\begin{axis}[%
width=0.787in,
height=0.787in,
at={(2.362in,4.724in)},
scale only axis,
point meta min=0,
point meta max=1,
axis on top,
xmin=0,
xmax=260,
xtick={\empty},
y dir=reverse,
ymin=0,
ymax=260,
ytick={\empty},
axis background/.style={fill=white}
]
\addplot [forget plot] graphics [xmin=0.5, xmax=260.5, ymin=0.5, ymax=260.5] {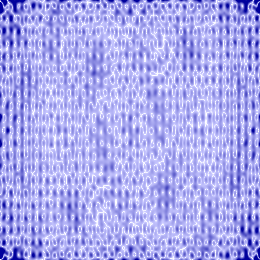};
\node[fill=white, align=center]
at (axis cs:35,35) {35};
\end{axis}

\begin{axis}[%
width=0.787in,
height=0.787in,
at={(0in,3.936in)},
scale only axis,
point meta min=0,
point meta max=1,
axis on top,
xmin=0,
xmax=260,
xtick={\empty},
y dir=reverse,
ymin=0,
ymax=260,
ytick={\empty},
axis background/.style={fill=white}
]
\addplot [forget plot] graphics [xmin=0.5, xmax=260.5, ymin=0.5, ymax=260.5] {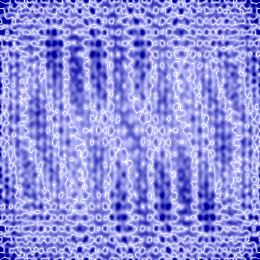};
\node[fill=white, align=center]
at (axis cs:35,35) {40};
\end{axis}

\begin{axis}[%
width=0.787in,
height=0.787in,
at={(0.787in,3.936in)},
scale only axis,
point meta min=0,
point meta max=1,
axis on top,
xmin=0,
xmax=260,
xtick={\empty},
y dir=reverse,
ymin=0,
ymax=260,
ytick={\empty},
axis background/.style={fill=white}
]
\addplot [forget plot] graphics [xmin=0.5, xmax=260.5, ymin=0.5, ymax=260.5] {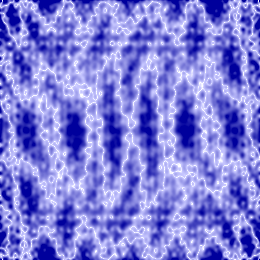};
\node[fill=white, align=center]
at (axis cs:35,35) {45};
\end{axis}

\begin{axis}[%
width=0.787in,
height=0.787in,
at={(1.575in,3.936in)},
scale only axis,
point meta min=0,
point meta max=1,
axis on top,
xmin=0,
xmax=260,
xtick={\empty},
y dir=reverse,
ymin=0,
ymax=260,
ytick={\empty},
axis background/.style={fill=white}
]
\addplot [forget plot] graphics [xmin=0.5, xmax=260.5, ymin=0.5, ymax=260.5] {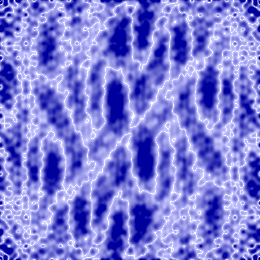};
\node[fill=white, align=center]
at (axis cs:35,35) {50};
\end{axis}

\begin{axis}[%
width=0.787in,
height=0.787in,
at={(2.362in,3.936in)},
scale only axis,
point meta min=0,
point meta max=0.35,
axis on top,
xmin=0,
xmax=260,
xtick={\empty},
y dir=reverse,
ymin=0,
ymax=260,
ytick={\empty},
axis background/.style={fill=white},
colormap={mymap}{[1pt] rgb(0pt)=(1,1,1); rgb(11pt)=(0.780392,0.780392,1); rgb(255pt)=(0,0,0.611765)},
colorbar horizontal,
colorbar style={
at={(1,-0.15)},anchor=south east,
width=4*
\pgfkeysvalueof{/pgfplots/parent axis width},
xticklabel pos=lower,
xtick={
0,
0.1,
0.2,
0.3
},
colorbar/width=2mm,
},
]
\addplot [forget plot] graphics [xmin=0.5, xmax=260.5, ymin=0.5, ymax=260.5] {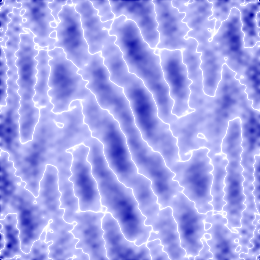};
\node[fill=white, align=center]
at (axis cs:35,35) {55};
\end{axis}

\end{tikzpicture}

%% file: Figures/balancing0_norm0_noFFT_50rSVDmodes.tex
\definecolor{mycolor1}{rgb}{0.64, 0.76, 0.68}%
\begin{tikzpicture}

\begin{axis}[%
width=7cm,
height=5cm,
scale only axis,
unbounded coords=jump,
xmin=0,
xmax=50,
xlabel style={font=\color{white!15!black}},
xlabel={$r$},
ymin=0,
ymax=50,
ylabel style={font=\color{white!15!black}},
ylabel={Test error in \%},
axis background/.style={fill=white},
title style={font=\bfseries},
legend columns=2,
legend style={legend cell align=left, align=left, draw=white!15!black, font = \footnotesize},
legend entries={Tree,
                NB,
                LD,
                kNN,
                13.0~\%},
]
\addlegendimage{legend image code/.code={
\draw [draw=none, fill=white!80!black] (0cm,-0.15cm) rectangle (0.6cm,0.15cm);
\draw[dashdotted] (0cm,0cm) -- (0.6cm,0cm);
\draw[-] (0cm,0.15cm) -- (0.6cm,0.15cm);
\draw[-] (0cm,-0.15cm) -- (0.6cm,-0.15cm);
}
} 
\addlegendimage{legend image code/.code={
\draw [draw=none, fill=mycolor1] (0cm,-0.15cm) rectangle (0.6cm,0.15cm);
\draw[dotted, black!80!green] (0cm,0cm) -- (0.6cm,0cm);
\draw[-, black!80!green] (0cm,0.15cm) -- (0.6cm,0.15cm);
\draw[-, black!80!green] (0cm,-0.15cm) -- (0.6cm,-0.15cm);
}
}
\addlegendimage{legend image code/.code={
\draw [draw=none, fill=white!80!blue] (0cm,-0.15cm) rectangle (0.6cm,0.15cm);
\draw[dashed, blue] (0cm,0cm) -- (0.6cm,0cm);
\draw[-, blue] (0cm,0.15cm) -- (0.6cm,0.15cm);
\draw[-, blue] (0cm,-0.15cm) -- (0.6cm,-0.15cm);
}
}
\addlegendimage{legend image code/.code={
\draw [draw=none, fill=white!80!red] (0cm,-0.15cm) rectangle (0.6cm,0.15cm);
\draw[-, red] (0cm,0cm) -- (0.6cm,0cm);
\draw[-, red] (0cm,0.15cm) -- (0.6cm,0.15cm);
\draw[-, red] (0cm,-0.15cm) -- (0.6cm,-0.15cm);
}
}
\addlegendimage{mark=none, black, very thick, dotted}
\addplot[area legend, draw=black, dashdotted,fill=white!80!black]
table[row sep=crcr] {%
x	y\\
1	32.4416250988252\\
2	28.0361961437281\\
3	25.9947328375039\\
4	24.3617970133233\\
5	23.2619659428818\\
6	22.3845462687488\\
7	21.7553061199775\\
8	21.3205345965905\\
9	20.1210532761127\\
10	19.9193275195994\\
11	20.156976898856\\
12	19.3595585397729\\
13	19.6215053984952\\
14	19.1859557675291\\
15	19.0599629135976\\
16	18.1379955302293\\
17	18.1451480150055\\
18	18.4889231761354\\
19	18.3633616878325\\
20	18.3686301762536\\
21	18.1724047325964\\
22	18.1887350492392\\
23	18.2641382702931\\
24	18.2542642422833\\
25	18.1419100004087\\
26	18.381194396675\\
27	18.536496674007\\
28	18.8193516883716\\
29	18.7357354774739\\
30	18.528072599494\\
31	18.6987198249374\\
32	18.433444137167\\
33	18.7292423378408\\
34	18.8200485512483\\
35	18.7720358403978\\
36	18.9795604577119\\
37	18.8400826891288\\
38	18.8896051262893\\
39	18.7296339957863\\
40	18.7670548663194\\
41	18.9318833574166\\
42	19.0045547428845\\
43	18.8204635559098\\
44	18.8760291056468\\
45	18.9610615275328\\
46	18.8724047422374\\
47	19.2320965565894\\
48	19.2398060711988\\
49	19.2322496924743\\
50	19.2066820026192\\
50	20.2189110714942\\
49	20.2900065659528\\
48	20.2378925829234\\
47	20.0149573207207\\
46	20.2556282579842\\
45	20.3604345419856\\
44	20.3561255861169\\
43	20.1885717884008\\
42	20.1531804148464\\
41	20.2708321628953\\
40	20.1972959811291\\
39	20.1752486510614\\
38	20.1193393978975\\
37	20.1094454035567\\
36	19.7097095388546\\
35	19.9544407888796\\
34	20.1370892565216\\
33	19.7519520335242\\
32	19.5566194262182\\
31	19.7972324446081\\
30	19.7819214481822\\
29	19.626275801472\\
28	19.6318769301556\\
27	19.5797811926064\\
26	19.3856086032332\\
25	19.639631562643\\
24	19.6835682645252\\
23	19.3983858659864\\
22	19.7118553253588\\
21	19.5569239034479\\
20	19.3237101429952\\
19	19.3883803971805\\
18	19.4488145635346\\
17	19.3834673797257\\
16	19.7625415743272\\
15	20.4480437712826\\
14	20.5747839994576\\
13	19.9976774273886\\
12	20.899278887606\\
11	21.292354592992\\
10	21.2623307110482\\
9	22.1390190492016\\
8	22.5767328972085\\
7	22.3868261713137\\
6	23.4612987742951\\
5	25.0767371332002\\
4	25.1296172937782\\
3	27.7600142570066\\
2	29.2670459257744\\
1	33.9381598173575\\
}--cycle;

\addplot[area legend, draw=black!80!green, dotted, fill=mycolor1]
table[row sep=crcr] {%
x	y\\
1	25.1145157920631\\
2	22.9394735091144\\
3	37.2178820988764\\
4	35.9596745715967\\
5	35.5615077824829\\
6	35.0757953328974\\
7	34.665730612813\\
8	33.4573901341895\\
9	32.7520617621147\\
10	32.4138399964724\\
11	31.2164067806549\\
12	30.3746548374132\\
13	30.1806723889983\\
14	30.3147014432029\\
15	30.100764686848\\
16	30.0057398415049\\
17	29.8421661529911\\
18	29.7984862534482\\
19	29.7092145219593\\
20	29.604248966428\\
21	29.5034952190302\\
22	29.2945093186611\\
23	29.1184539437286\\
24	29.0546681003903\\
25	29.0205784970694\\
26	28.9218997321159\\
27	28.8194869472137\\
28	28.7555327464413\\
29	28.6747083480263\\
30	28.6659665753498\\
31	28.5815059812606\\
32	28.5100917982681\\
33	28.4842921372246\\
34	28.4839416261373\\
35	28.3719798025262\\
36	28.3790175440861\\
37	28.3318018173502\\
38	28.2415019503061\\
39	28.2221960982997\\
40	28.038241227868\\
41	28.0160986209792\\
42	27.9079794969249\\
43	27.8289233140315\\
44	27.8558107688694\\
45	27.787521698424\\
46	27.6798648257835\\
47	27.5876159480658\\
48	27.4871950640288\\
49	27.4905812332711\\
50	27.4320688116131\\
50	29.0033779497056\\
49	29.0564792780967\\
48	29.1566267484925\\
47	29.2124851698209\\
46	29.343454304284\\
45	29.4144557804064\\
44	29.4354285102764\\
43	29.5144162901588\\
42	29.6139161549806\\
41	29.6248201415591\\
40	29.7290651751601\\
39	29.7906373164612\\
38	29.8531351750005\\
37	29.8819455592542\\
36	29.8570046008157\\
35	29.9979722219078\\
34	29.9827061630706\\
33	30.1237178716088\\
32	30.1723619842377\\
31	30.2124624613545\\
30	30.2396343792058\\
29	30.3499423818664\\
28	30.5667915386832\\
27	30.6516075161673\\
26	30.7054309111243\\
25	30.8448722700938\\
24	30.8926423882726\\
23	31.0966589994639\\
22	31.2405742507622\\
21	31.2399238455952\\
20	31.2507256258929\\
19	31.3020620210142\\
18	31.3393359721162\\
17	31.3700345440304\\
16	31.2286914333063\\
15	31.2229313432622\\
14	31.1203847894372\\
13	31.1575087937772\\
12	31.67801832324\\
11	32.0642107143409\\
10	33.3516504475101\\
9	35.4016764747664\\
8	36.2730387072884\\
7	35.8606549395679\\
6	36.0234911286885\\
5	36.6834726392135\\
4	37.4090086075888\\
3	39.8421031259538\\
2	23.89025078343\\
1	26.3201973703475\\
}--cycle;

\addplot[area legend, draw=blue, dashed, fill=white!80!blue]
table[row sep=crcr] {%
x	y\\
1	25.2273590504246\\
2	25.2510068878318\\
3	25.251693407179\\
4	25.2439178664139\\
5	25.2467463682518\\
6	25.2092875017966\\
7	25.1774332144084\\
8	25.1750272645472\\
9	25.2517805281826\\
10	25.2502481592879\\
11	25.240866014011\\
12	25.2653355919987\\
13	25.0809405967789\\
14	24.8905047976355\\
15	24.7611036847154\\
16	24.665084802294\\
17	24.7932929686879\\
18	24.632720031863\\
19	24.8150844914702\\
20	24.8537827045421\\
21	24.93949168546\\
22	24.8086634863506\\
23	24.8384095967529\\
24	24.7707710447178\\
25	24.7674303481208\\
26	24.7319520329259\\
27	24.7583688094461\\
28	24.8260940148326\\
29	24.8466376340385\\
30	24.7265872336557\\
31	24.7341780941817\\
32	24.7485782737733\\
33	24.7303471347784\\
34	24.7541699840052\\
35	24.7602792611864\\
36	24.7364151758326\\
37	24.7805981252532\\
38	24.7609466020203\\
39	24.6805144141152\\
40	24.6804028972913\\
41	24.6776173140277\\
42	24.6552338292186\\
43	24.6910159811017\\
44	24.661932508761\\
45	24.6256042378974\\
46	24.6447099654073\\
47	24.6519599445706\\
48	24.6596766434834\\
49	24.6605356855882\\
50	24.6515746011379\\
50	26.0837263285492\\
49	26.0449868715595\\
48	25.9863099868275\\
47	26.0237737467256\\
46	26.0682189109707\\
45	26.0426858593854\\
44	26.0435845767682\\
43	26.0368250133829\\
42	26.0428127775061\\
41	26.0055721631256\\
40	25.9879157426878\\
39	25.9952255650698\\
38	25.9296657372222\\
37	25.9174419487605\\
36	25.9244219864071\\
35	25.9154414436186\\
34	25.9289912690463\\
33	25.9677008884619\\
32	25.9420535471766\\
31	25.9341104696605\\
30	25.993772982435\\
29	26.0002475171193\\
28	26.0431338843771\\
27	26.0364168389746\\
26	26.0181762403861\\
25	26.0273249825238\\
24	25.9421352070229\\
23	25.9935488630645\\
22	25.986056999618\\
21	25.9668245624225\\
20	25.9632674536572\\
19	25.9647447008424\\
18	26.0056415886017\\
17	26.0534774302488\\
16	25.951034972359\\
15	25.8475777723532\\
14	26.0083660251032\\
13	26.0709653416987\\
12	26.4372725566935\\
11	26.4319837746708\\
10	26.4523398687912\\
9	26.4284998697378\\
8	26.5126623873394\\
7	26.4358549328588\\
6	26.4784137504359\\
5	26.4632756768294\\
4	26.5107573269664\\
3	26.4732220113915\\
2	26.4441402734059\\
1	26.4380281901749\\
}--cycle;

\addplot[area legend, draw=red, fill=white!80!red]
table[row sep=crcr] {%
x	y\\
1	35.3149316361938\\
2	29.646941420844\\
3	26.4413237345461\\
4	23.7181755070448\\
5	21.5340552305909\\
6	19.7591018303024\\
7	18.5172165367463\\
8	17.5514340395775\\
9	16.8651611072175\\
10	16.5245471871128\\
11	15.8893179971415\\
12	15.5720449231528\\
13	15.3995203874374\\
14	15.1106213463129\\
15	14.7893215201234\\
16	14.098891030755\\
17	13.7439815881468\\
18	13.6365578855833\\
19	13.5515404786987\\
20	13.5234390558074\\
21	13.2025809528304\\
22	13.1850918153088\\
23	13.1510448075151\\
24	13.0155388906032\\
25	13.0508510396169\\
26	13.191518536449\\
27	13.1685482154454\\
28	13.0639251400442\\
29	13.1283852123776\\
30	13.1542845551113\\
31	12.9900049264216\\
32	13.1182034210753\\
33	13.1759130865974\\
34	13.2608120038142\\
35	13.1970814225669\\
36	13.0431623557207\\
37	13.2261082232078\\
38	13.2431575062201\\
39	13.2815629747676\\
40	13.3405757718181\\
41	13.26425090804\\
42	13.4336568324064\\
43	13.3846035060142\\
44	13.3822791875291\\
45	13.371932493718\\
46	13.2604153925207\\
47	13.384189831469\\
48	13.3849954556206\\
49	13.3447952436554\\
50	13.4749300563875\\
50	13.8005303368076\\
49	13.8338761609353\\
48	13.9127020490425\\
47	13.809432898014\\
46	13.8067800231966\\
45	13.7622586278891\\
44	13.8709256747126\\
43	13.8388106293857\\
42	13.827006586975\\
41	13.7137076438365\\
40	13.6821362389152\\
39	13.5774911406283\\
38	13.5266529536687\\
37	13.5436307371375\\
36	13.5256957600351\\
35	13.5652115470534\\
34	13.3676530455456\\
33	13.5195649640804\\
32	13.584858571858\\
31	13.5122271585378\\
30	13.3405408998559\\
29	13.3664963534954\\
28	13.4532535125702\\
27	13.5420548502789\\
26	13.6011943188906\\
25	13.5855183298742\\
24	13.6655743349347\\
23	13.6117769128329\\
22	13.6969719144156\\
21	13.7539489270271\\
20	13.7977231583079\\
19	14.208457709027\\
18	14.5104372294717\\
17	14.6561905237356\\
16	15.0746424661879\\
15	15.262302928867\\
14	15.5064083602265\\
13	16.005756267916\\
12	16.3766893634856\\
11	16.6919188652176\\
10	17.1578546542977\\
9	17.8662936432488\\
8	18.6382779529801\\
7	19.1754231770746\\
6	20.8424194917223\\
5	22.7128275596996\\
4	24.962876199452\\
3	27.8569440416326\\
2	31.0053244166358\\
1	36.2148703278764\\
}--cycle;

\addplot[mark=none, black, very thick, dotted, domain=0:50] {13.01};

\addplot [color=black, forget plot, dashdotted]
  table[row sep=crcr]{%
1	33.1898924580914\\
2	28.6516210347513\\
3	26.8773735472552\\
4	24.7457071535508\\
5	24.169351538041\\
6	22.9229225215219\\
7	22.0710661456456\\
8	21.9486337468995\\
9	21.1300361626571\\
10	20.5908291153238\\
11	20.724665745924\\
12	20.1294187136894\\
13	19.8095914129419\\
14	19.8803698834934\\
15	19.7540033424401\\
16	18.9502685522782\\
17	18.7643076973656\\
18	18.968868869835\\
19	18.8758710425065\\
20	18.8461701596244\\
21	18.8646643180222\\
22	18.950295187299\\
23	18.8312620681398\\
24	18.9689162534043\\
25	18.8907707815258\\
26	18.8834014999541\\
28	19.2256143092636\\
29	19.181005639473\\
30	19.1549970238381\\
31	19.2479761347727\\
32	18.9950317816926\\
34	19.4785689038849\\
35	19.3632383146387\\
36	19.3446349982832\\
37	19.4747640463428\\
38	19.5044722620934\\
39	19.4524413234238\\
40	19.4821754237243\\
41	19.601357760156\\
42	19.5788675788654\\
43	19.5045176721553\\
44	19.6160773458818\\
45	19.6607480347592\\
46	19.5640165001108\\
47	19.623526938655\\
48	19.7388493270611\\
49	19.7611281292136\\
50	19.7127965370567\\
};
\addplot [color=black!80!green, forget plot, dotted]
  table[row sep=crcr]{%
1	25.7173565812053\\
2	23.4148621462722\\
3	38.5299926124151\\
4	36.6843415895928\\
6	35.549643230793\\
7	35.2631927761904\\
8	34.865214420739\\
9	34.0768691184406\\
10	32.8827452219913\\
11	31.6403087474979\\
12	31.0263365803266\\
13	30.6690905913878\\
14	30.71754311632\\
15	30.6618480150551\\
16	30.6172156374056\\
17	30.6061003485108\\
18	30.5689111127822\\
19	30.5056382714867\\
20	30.4274872961605\\
21	30.3717095323127\\
22	30.2675417847116\\
23	30.1075564715963\\
24	29.9736552443314\\
25	29.9327253835816\\
26	29.8136653216201\\
28	29.6611621425622\\
29	29.5123253649463\\
32	29.3412268912529\\
33	29.3040050044167\\
34	29.2333238946039\\
35	29.184976012217\\
36	29.1180110724509\\
37	29.1068736883022\\
38	29.0473185626533\\
39	29.0064167073804\\
40	28.883653201514\\
42	28.7609478259528\\
43	28.6716698020951\\
44	28.6456196395729\\
45	28.6009887394152\\
46	28.5116595650338\\
47	28.4000505589433\\
48	28.3219109062606\\
50	28.2177233806594\\
};
\addplot [color=blue, forget plot, dashed]
  table[row sep=crcr]{%
1	25.8326936202997\\
4	25.8773375966901\\
5	25.8550110225407\\
6	25.8438506261163\\
7	25.8066440736336\\
8	25.8438448259433\\
9	25.8401401989602\\
10	25.8512940140395\\
11	25.8364248943409\\
12	25.8513040743461\\
13	25.5759529692388\\
14	25.4494354113694\\
15	25.3043407285343\\
16	25.3080598873265\\
17	25.4233851994684\\
18	25.3191808102323\\
19	25.3899145961563\\
20	25.4085250790997\\
21	25.4531581239413\\
22	25.3973602429843\\
23	25.4159792299087\\
24	25.3564531258704\\
25	25.3973776653223\\
26	25.375064136656\\
27	25.3973928242103\\
28	25.4346139496049\\
29	25.4234425755789\\
30	25.3601801080454\\
31	25.3341442819211\\
33	25.3490240116202\\
36	25.3304185811198\\
37	25.3490200370069\\
41	25.3415947385766\\
43	25.3639204972423\\
45	25.3341450486414\\
46	25.356464438189\\
48	25.3229933151555\\
49	25.3527612785738\\
50	25.3676504648436\\
};
\addplot [color=red, forget plot]
  table[row sep=crcr]{%
1	35.7649009820351\\
2	30.3261329187399\\
3	27.1491338880893\\
4	24.3405258532484\\
5	22.1234413951452\\
6	20.3007606610124\\
7	18.8463198569104\\
8	18.0948559962788\\
9	17.3657273752332\\
10	16.8412009207053\\
11	16.2906184311795\\
12	15.9743671433192\\
13	15.7026383276767\\
14	15.3085148532697\\
15	15.0258122244952\\
16	14.5867667484715\\
17	14.2000860559412\\
18	14.0734975575275\\
19	13.8799990938628\\
20	13.6605811070577\\
21	13.4782649399287\\
22	13.4410318648622\\
23	13.381410860174\\
24	13.340556612769\\
25	13.3181846847456\\
26	13.3963564276698\\
27	13.3553015328621\\
28	13.2585893263072\\
29	13.2474407829365\\
31	13.2511160424797\\
32	13.3515309964666\\
33	13.3477390253389\\
34	13.3142325246799\\
35	13.3811464848101\\
36	13.2844290578779\\
37	13.3848694801726\\
38	13.3849052299444\\
39	13.4295270576979\\
40	13.5113560053667\\
41	13.4889792759382\\
42	13.6303317096907\\
43	13.6117070676999\\
44	13.6266024311208\\
45	13.5670955608036\\
46	13.5335977078587\\
47	13.5968113647415\\
48	13.6488487523316\\
49	13.5893357022953\\
50	13.6377301965975\\
};
\addplot [color=black, forget plot]
  table[row sep=crcr]{%
1	32.4416250988252\\
2	28.0361961437281\\
3	25.9947328375039\\
4	24.3617970133233\\
5	23.2619659428818\\
6	22.3845462687488\\
7	21.7553061199775\\
8	21.3205345965905\\
9	20.1210532761127\\
10	19.9193275195994\\
11	20.156976898856\\
12	19.3595585397729\\
13	19.6215053984952\\
14	19.1859557675291\\
15	19.0599629135976\\
16	18.1379955302293\\
17	18.1451480150055\\
18	18.4889231761354\\
19	18.3633616878325\\
20	18.3686301762536\\
21	18.1724047325964\\
22	18.1887350492392\\
23	18.2641382702931\\
24	18.2542642422833\\
25	18.1419100004087\\
26	18.381194396675\\
27	18.536496674007\\
28	18.8193516883716\\
29	18.7357354774739\\
30	18.528072599494\\
31	18.6987198249374\\
32	18.433444137167\\
33	18.7292423378408\\
34	18.8200485512483\\
35	18.7720358403978\\
36	18.9795604577119\\
37	18.8400826891288\\
38	18.8896051262893\\
39	18.7296339957863\\
40	18.7670548663194\\
41	18.9318833574166\\
42	19.0045547428845\\
43	18.8204635559098\\
44	18.8760291056468\\
45	18.9610615275328\\
46	18.8724047422374\\
47	19.2320965565894\\
48	19.2398060711988\\
49	19.2322496924743\\
50	19.2066820026192\\
nan	nan\\
50	20.2189110714942\\
49	20.2900065659528\\
48	20.2378925829234\\
47	20.0149573207207\\
46	20.2556282579842\\
45	20.3604345419856\\
44	20.3561255861169\\
43	20.1885717884008\\
42	20.1531804148464\\
41	20.2708321628953\\
40	20.1972959811291\\
39	20.1752486510614\\
38	20.1193393978975\\
37	20.1094454035567\\
36	19.7097095388546\\
35	19.9544407888796\\
34	20.1370892565216\\
33	19.7519520335242\\
32	19.5566194262182\\
31	19.7972324446081\\
30	19.7819214481822\\
29	19.626275801472\\
28	19.6318769301556\\
27	19.5797811926064\\
26	19.3856086032332\\
25	19.639631562643\\
24	19.6835682645252\\
23	19.3983858659864\\
22	19.7118553253588\\
21	19.5569239034479\\
20	19.3237101429952\\
18	19.4488145635346\\
17	19.3834673797257\\
16	19.7625415743272\\
15	20.4480437712826\\
14	20.5747839994576\\
13	19.9976774273886\\
12	20.899278887606\\
11	21.292354592992\\
10	21.2623307110482\\
9	22.1390190492016\\
8	22.5767328972085\\
7	22.3868261713137\\
6	23.4612987742951\\
5	25.0767371332002\\
4	25.1296172937782\\
3	27.7600142570066\\
2	29.2670459257744\\
1	33.9381598173575\\
};
\addplot [color=black!80!green, forget plot]
  table[row sep=crcr]{%
1	25.1145157920631\\
2	22.9394735091144\\
3	37.2178820988764\\
4	35.9596745715967\\
5	35.5615077824829\\
6	35.0757953328974\\
7	34.665730612813\\
8	33.4573901341895\\
9	32.7520617621147\\
10	32.4138399964724\\
11	31.2164067806549\\
12	30.3746548374132\\
13	30.1806723889983\\
14	30.3147014432029\\
15	30.100764686848\\
16	30.0057398415049\\
17	29.8421661529911\\
18	29.7984862534482\\
19	29.7092145219593\\
21	29.5034952190302\\
22	29.2945093186611\\
23	29.1184539437286\\
24	29.0546681003903\\
25	29.0205784970694\\
27	28.8194869472137\\
28	28.7555327464413\\
29	28.6747083480263\\
30	28.6659665753498\\
31	28.5815059812606\\
32	28.5100917982681\\
33	28.4842921372246\\
34	28.4839416261373\\
35	28.3719798025262\\
36	28.3790175440861\\
37	28.3318018173502\\
38	28.2415019503061\\
39	28.2221960982997\\
40	28.038241227868\\
41	28.0160986209792\\
42	27.9079794969249\\
43	27.8289233140315\\
44	27.8558107688694\\
45	27.787521698424\\
46	27.6798648257835\\
48	27.4871950640288\\
49	27.4905812332711\\
50	27.4320688116131\\
nan	nan\\
50	29.0033779497056\\
49	29.0564792780967\\
48	29.1566267484925\\
47	29.2124851698209\\
46	29.3434543042841\\
45	29.4144557804064\\
44	29.4354285102764\\
43	29.5144162901588\\
42	29.6139161549806\\
41	29.6248201415591\\
40	29.7290651751601\\
38	29.8531351750005\\
37	29.8819455592542\\
36	29.8570046008157\\
35	29.9979722219078\\
34	29.9827061630706\\
33	30.1237178716088\\
31	30.2124624613545\\
30	30.2396343792058\\
29	30.3499423818664\\
28	30.5667915386832\\
27	30.6516075161673\\
26	30.7054309111243\\
25	30.8448722700938\\
24	30.8926423882726\\
23	31.0966589994639\\
22	31.2405742507622\\
21	31.2399238455952\\
20	31.2507256258929\\
19	31.3020620210142\\
17	31.3700345440304\\
16	31.2286914333063\\
15	31.2229313432622\\
14	31.1203847894372\\
13	31.1575087937772\\
12	31.67801832324\\
11	32.0642107143409\\
10	33.3516504475101\\
9	35.4016764747664\\
8	36.2730387072884\\
7	35.8606549395679\\
6	36.0234911286885\\
5	36.6834726392135\\
4	37.4090086075888\\
3	39.8421031259538\\
2	23.89025078343\\
1	26.3201973703475\\
};
\addplot [color=blue, forget plot]
  table[row sep=crcr]{%
1	25.2273590504246\\
2	25.2510068878318\\
5	25.2467463682518\\
7	25.1774332144084\\
8	25.1750272645472\\
9	25.2517805281826\\
11	25.240866014011\\
12	25.2653355919987\\
14	24.8905047976355\\
15	24.7611036847155\\
16	24.665084802294\\
17	24.7932929686879\\
18	24.632720031863\\
19	24.8150844914702\\
20	24.8537827045421\\
21	24.93949168546\\
22	24.8086634863506\\
23	24.8384095967529\\
24	24.7707710447178\\
25	24.7674303481208\\
26	24.7319520329259\\
27	24.7583688094461\\
28	24.8260940148326\\
29	24.8466376340385\\
30	24.7265872336557\\
32	24.7485782737733\\
33	24.7303471347784\\
34	24.7541699840052\\
35	24.7602792611864\\
36	24.7364151758326\\
37	24.7805981252532\\
38	24.7609466020203\\
39	24.6805144141152\\
41	24.6776173140277\\
42	24.6552338292186\\
43	24.6910159811017\\
45	24.6256042378974\\
46	24.6447099654073\\
49	24.6605356855882\\
50	24.6515746011379\\
nan	nan\\
50	26.0837263285492\\
49	26.0449868715595\\
48	25.9863099868275\\
46	26.0682189109707\\
45	26.0426858593854\\
42	26.0428127775061\\
41	26.0055721631256\\
40	25.9879157426878\\
39	25.9952255650698\\
38	25.9296657372222\\
37	25.9174419487605\\
36	25.9244219864071\\
35	25.9154414436186\\
34	25.9289912690463\\
33	25.9677008884619\\
32	25.9420535471766\\
31	25.9341104696605\\
30	25.993772982435\\
29	26.0002475171193\\
28	26.0431338843771\\
27	26.0364168389746\\
26	26.0181762403861\\
25	26.0273249825238\\
24	25.9421352070229\\
23	25.9935488630645\\
22	25.986056999618\\
21	25.9668245624225\\
19	25.9647447008424\\
17	26.0534774302488\\
15	25.8475777723532\\
14	26.0083660251032\\
13	26.0709653416987\\
12	26.4372725566935\\
11	26.4319837746708\\
10	26.4523398687912\\
9	26.4284998697378\\
8	26.5126623873394\\
7	26.4358549328588\\
6	26.4784137504359\\
5	26.4632756768294\\
4	26.5107573269664\\
2	26.4441402734059\\
1	26.4380281901749\\
};
\addplot [color=red, forget plot]
  table[row sep=crcr]{%
1	35.3149316361938\\
2	29.646941420844\\
3	26.4413237345461\\
4	23.7181755070448\\
5	21.5340552305909\\
6	19.7591018303024\\
7	18.5172165367463\\
8	17.5514340395775\\
9	16.8651611072175\\
10	16.5245471871128\\
11	15.8893179971415\\
12	15.5720449231528\\
13	15.3995203874374\\
14	15.1106213463129\\
15	14.7893215201234\\
16	14.098891030755\\
17	13.7439815881468\\
18	13.6365578855834\\
19	13.5515404786987\\
20	13.5234390558074\\
21	13.2025809528304\\
22	13.1850918153088\\
23	13.1510448075151\\
24	13.0155388906032\\
25	13.0508510396169\\
26	13.191518536449\\
27	13.1685482154454\\
28	13.0639251400442\\
29	13.1283852123776\\
30	13.1542845551113\\
31	12.9900049264216\\
32	13.1182034210753\\
33	13.1759130865974\\
34	13.2608120038142\\
35	13.1970814225669\\
36	13.0431623557207\\
37	13.2261082232078\\
38	13.2431575062202\\
39	13.2815629747676\\
40	13.3405757718181\\
41	13.26425090804\\
42	13.4336568324064\\
43	13.3846035060142\\
45	13.371932493718\\
46	13.2604153925207\\
47	13.384189831469\\
48	13.3849954556206\\
49	13.3447952436554\\
50	13.4749300563875\\
nan	nan\\
50	13.8005303368076\\
49	13.8338761609353\\
48	13.9127020490425\\
47	13.809432898014\\
46	13.8067800231966\\
45	13.7622586278891\\
44	13.8709256747126\\
43	13.8388106293857\\
42	13.827006586975\\
41	13.7137076438365\\
40	13.6821362389152\\
39	13.5774911406283\\
38	13.5266529536687\\
37	13.5436307371375\\
36	13.5256957600351\\
35	13.5652115470534\\
34	13.3676530455456\\
33	13.5195649640804\\
32	13.584858571858\\
31	13.5122271585378\\
30	13.3405408998559\\
29	13.3664963534954\\
27	13.5420548502789\\
26	13.6011943188906\\
25	13.5855183298742\\
24	13.6655743349347\\
23	13.6117769128329\\
22	13.6969719144156\\
21	13.7539489270271\\
20	13.7977231583079\\
19	14.208457709027\\
18	14.5104372294717\\
17	14.6561905237356\\
16	15.0746424661879\\
15	15.262302928867\\
14	15.5064083602265\\
13	16.005756267916\\
12	16.3766893634856\\
11	16.6919188652176\\
10	17.1578546542977\\
9	17.8662936432489\\
8	18.6382779529801\\
7	19.1754231770746\\
6	20.8424194917223\\
5	22.7128275596996\\
4	24.962876199452\\
3	27.8569440416326\\
2	31.0053244166358\\
1	36.2148703278764\\
};
\end{axis}
\end{tikzpicture}%

%% file: Figures/balancing0_norm0_FFTtransformed_50rSVDmodes_duplicate.tex
\definecolor{mycolor1}{rgb}{0.64, 0.76, 0.68}%
\begin{tikzpicture}

\begin{axis}[%
width=7cm,
height=5cm,
scale only axis,
unbounded coords=jump,
xmin=0,
xmax=50,
xlabel style={font=\color{white!15!black}},
xlabel={$r$},
ymin=0,
ymax=50,
ylabel style={font=\color{white!15!black}},
axis background/.style={fill=white},
title style={font=\bfseries},
legend columns=2,
legend style={legend cell align=left, align=left, draw=white!15!black, font = \footnotesize},
legend entries={Tree,
                NB,
                LD,
                kNN,
                2.8~\%},
]
\addlegendimage{legend image code/.code={
\draw [draw=none, fill=white!80!black] (0cm,-0.15cm) rectangle (0.6cm,0.15cm);
\draw[dashdotted] (0cm,0cm) -- (0.6cm,0cm);
\draw[-] (0cm,0.15cm) -- (0.6cm,0.15cm);
\draw[-] (0cm,-0.15cm) -- (0.6cm,-0.15cm);
}
} 
\addlegendimage{legend image code/.code={
\draw [draw=none, fill=mycolor1] (0cm,-0.15cm) rectangle (0.6cm,0.15cm);
\draw[dotted, black!80!green] (0cm,0cm) -- (0.6cm,0cm);
\draw[-, black!80!green] (0cm,0.15cm) -- (0.6cm,0.15cm);
\draw[-, black!80!green] (0cm,-0.15cm) -- (0.6cm,-0.15cm);
}
}
\addlegendimage{legend image code/.code={
\draw [draw=none, fill=white!80!blue] (0cm,-0.15cm) rectangle (0.6cm,0.15cm);
\draw[dashed, blue] (0cm,0cm) -- (0.6cm,0cm);
\draw[-, blue] (0cm,0.15cm) -- (0.6cm,0.15cm);
\draw[-, blue] (0cm,-0.15cm) -- (0.6cm,-0.15cm);
}
}
\addlegendimage{legend image code/.code={
\draw [draw=none, fill=white!80!red] (0cm,-0.15cm) rectangle (0.6cm,0.15cm);
\draw[-, red] (0cm,0cm) -- (0.6cm,0cm);
\draw[-, red] (0cm,0.15cm) -- (0.6cm,0.15cm);
\draw[-, red] (0cm,-0.15cm) -- (0.6cm,-0.15cm);
}
}
\addlegendimage{mark=none, black, very thick, dotted}

\addplot[area legend, draw=black, dashdotted, fill=white!80!black]
table[row sep=crcr] {%
x	y\\
1	32.2653176205512\\
2	17.8532964684552\\
3	12.2408056287417\\
4	9.41623563841375\\
5	7.26559816492054\\
6	5.88400848899825\\
7	5.84068622316195\\
8	5.75314963809708\\
9	5.58213099656179\\
10	5.62019633651235\\
11	5.66695267030981\\
12	5.97117659984233\\
13	5.93335740375575\\
14	5.97402050780451\\
15	5.94879923825127\\
16	5.97246428835041\\
17	6.1113049708149\\
18	6.14195941484675\\
19	6.1335297960797\\
20	6.16468595583544\\
21	6.15446425858792\\
22	6.18359351884024\\
23	6.27730452459262\\
24	6.30666695633937\\
25	6.29307976860633\\
26	6.2996765817743\\
27	6.36194338705992\\
28	6.40665346460722\\
29	6.45621230953474\\
30	6.5041086866459\\
31	6.47924265088407\\
32	6.57378440534831\\
33	6.66098209009066\\
34	6.69108459947119\\
35	6.6947339748358\\
36	6.67558850405993\\
37	6.73234356204454\\
38	6.72494628047895\\
39	6.72070451858107\\
40	6.76094728630891\\
41	6.8225580704983\\
42	6.89741239978608\\
43	7.01686284725347\\
44	6.99937425148352\\
45	7.03390081180594\\
46	7.09242277732984\\
47	7.14587703875906\\
48	7.21230799642535\\
49	7.22680387976835\\
50	7.30694302104934\\
50	8.23193922910968\\
49	8.20790976259494\\
48	8.0885217240728\\
47	8.0805561931555\\
46	8.0670903504769\\
45	8.11070781519347\\
44	8.1229680021143\\
43	8.12778115130227\\
42	8.09849997334392\\
41	8.06169300267325\\
40	8.04145111014155\\
39	8.08173735144742\\
38	8.05518591730211\\
37	8.12955344407876\\
36	8.03010277955879\\
35	8.03333287251847\\
34	8.03705089095514\\
33	7.99275581726382\\
32	7.99065259628762\\
31	7.92901022186128\\
30	7.90423687557456\\
29	7.89264589965725\\
28	7.85291183649545\\
27	7.81581826989545\\
26	7.72183001157924\\
25	7.78794675587584\\
24	7.78927848398085\\
23	7.71451967280156\\
22	7.73390509153567\\
21	7.67368711212826\\
20	7.76023291425019\\
19	7.73186714617926\\
18	7.59693925531041\\
17	7.55312037078379\\
16	7.40210148544412\\
15	7.37363964519574\\
14	7.12511969801469\\
13	7.03179939342797\\
12	6.97177352874856\\
11	7.04539455143798\\
10	6.89138607353659\\
9	6.79542332225676\\
8	6.66902841999913\\
7	6.50731043688273\\
6	6.37437337049488\\
5	7.74541062083036\\
4	10.1550360803122\\
3	12.6788595414542\\
2	19.6236078403228\\
1	33.5866802643059\\
}--cycle;

\addplot[area legend,  draw=black!80!green, dotted, fill=mycolor1]
table[row sep=crcr] {%
x	y\\
1	25.0826501102209\\
2	20.8182733915379\\
3	22.258218178962\\
4	19.5567708218547\\
5	16.3385148544372\\
6	15.9384258125764\\
7	15.0903596792752\\
8	15.424762293534\\
9	14.1518684639514\\
10	14.4209392414241\\
11	14.8827182389319\\
12	14.8856768497377\\
13	14.8636250900147\\
14	15.1260737879082\\
15	15.3115759747783\\
16	15.0916766850302\\
17	14.8608589042193\\
18	14.7363291821513\\
19	15.0274562439833\\
20	14.9998772069099\\
21	15.1970983422739\\
22	15.4149319212358\\
23	15.6557321066243\\
24	15.7748222449526\\
25	16.1091813521924\\
26	16.0153334855208\\
27	16.2536904415015\\
28	16.3123396855553\\
29	16.7601227384171\\
30	16.9877664358718\\
31	17.1310305504999\\
32	17.4267897310412\\
33	17.6531350450409\\
34	17.6928950598513\\
35	18.0127879844733\\
36	18.3012178059147\\
37	18.4241605809402\\
38	18.7419142929375\\
39	19.1266651812885\\
40	19.4373755857742\\
41	19.8110304829552\\
42	20.0102572838273\\
43	20.2169503765104\\
44	20.4100874411426\\
45	20.6680780327509\\
46	20.8422024998065\\
47	21.1597907134693\\
48	21.3237878933239\\
49	21.3805249051278\\
50	21.6325376123172\\
50	22.7283949564154\\
49	22.5414206720856\\
48	22.2113168325563\\
47	22.2041998334895\\
46	22.0009315468056\\
45	21.8103295456805\\
44	21.5102109188894\\
43	21.2718160253832\\
42	20.8832127445509\\
41	20.6211884026663\\
40	20.4368163532011\\
39	20.1300089539158\\
38	19.9716175257859\\
37	19.4784346913923\\
36	19.325997481529\\
35	18.9522911210457\\
34	18.6548607175061\\
33	18.33753805386\\
32	18.0579564708158\\
31	17.8627173809822\\
30	17.7902040640123\\
29	17.6013289460237\\
28	17.7588367213156\\
27	17.4306108277364\\
26	17.1332263461256\\
25	16.6524822732683\\
24	16.5776140964988\\
23	16.5405355480113\\
22	16.3349287351674\\
21	16.1508017729408\\
20	15.9313313684522\\
19	16.0079794272282\\
18	15.9344915907845\\
17	16.0703291566632\\
16	16.0626173828204\\
15	15.7681320523886\\
14	15.9314483986846\\
13	15.9407265405264\\
12	15.620468547244\\
11	15.5044036212584\\
10	15.2884771822851\\
9	15.3268439223092\\
8	16.3304738048086\\
7	15.844206813925\\
6	16.699498825581\\
5	16.8420744753101\\
4	19.8046131941855\\
3	22.867749580424\\
2	21.5390254162298\\
1	26.2999740020083\\
}--cycle;

\addplot[area legend, draw=blue, dashed, fill=white!80!blue]
table[row sep=crcr] {%
x	y\\
1	25.2222348392988\\
2	24.3305177517844\\
3	24.5831777921692\\
4	21.970420362519\\
5	18.5118964151506\\
6	18.5744954543389\\
7	14.1116280081771\\
8	13.2254324060553\\
9	12.0105512268702\\
10	11.2756571539852\\
11	11.169065472239\\
12	11.3219596704668\\
13	11.1295076953704\\
14	11.0230829091474\\
15	10.8931648215753\\
16	10.7884591300485\\
17	10.6630815596603\\
18	10.5887332747165\\
19	10.336300226329\\
20	10.311265319769\\
21	10.2862460581086\\
22	10.3137814777313\\
23	10.2697527098138\\
24	10.3764888221194\\
25	10.3466482225096\\
26	10.3815345235983\\
27	10.3223232361421\\
28	10.2820567038392\\
29	10.360930633376\\
30	10.4450000194838\\
31	10.296891696696\\
32	10.3196547893401\\
33	10.41763237503\\
34	10.3930966203781\\
35	10.3753930781208\\
36	10.3665781235427\\
37	10.3744937129501\\
38	10.3228968361005\\
39	10.3014425827525\\
40	10.3153922387687\\
41	10.3163095458351\\
42	10.3744408990628\\
43	10.3389935860452\\
44	10.3019109356742\\
45	10.1533834356693\\
46	10.163167227288\\
47	10.1986939686107\\
48	10.1273736052463\\
49	10.1974110331533\\
50	10.1392412575041\\
50	10.5769363798504\\
49	10.5857097929694\\
48	10.5664441893966\\
47	10.6216305540203\\
46	10.6348257456061\\
45	10.6966372398558\\
44	10.741461075603\\
43	10.6819944566872\\
42	10.6912455650382\\
41	10.6972851744657\\
40	10.7725324005107\\
39	10.8386373456948\\
38	10.8023017083876\\
37	10.7879327121953\\
36	10.7883663466863\\
35	10.7498489969523\\
34	10.7470301005894\\
33	10.7373856013405\\
32	10.6567802971679\\
31	10.7092498284541\\
30	10.7100806338478\\
29	10.7644079265902\\
28	10.8283274469651\\
27	10.8995979855572\\
26	10.8329865186925\\
25	10.9274234144452\\
24	10.9570913710994\\
23	10.9374329984843\\
22	10.8560898143965\\
21	11.0026327373698\\
20	10.9255332760062\\
19	11.0641420053804\\
18	11.1093700190604\\
17	11.0870120581226\\
16	11.1326148385265\\
15	11.3477843628924\\
14	11.2399270826816\\
13	11.6766742344222\\
12	11.729593365327\\
11	12.0536177730451\\
10	12.0510877120869\\
9	12.5139769364673\\
8	14.0742887699671\\
7	14.9652837978786\\
6	19.1093250942162\\
5	19.184116155042\\
4	22.8367798139562\\
3	25.5804668230785\\
2	25.3703588079262\\
1	26.4431508389564\\
}--cycle;

\addplot[area legend, draw=red, fill=white!80!red]
table[row sep=crcr] {%
x	y\\
1	35.0625689186529\\
2	17.9990422352818\\
3	10.6772695528429\\
4	7.63270203474922\\
5	4.74313097517599\\
6	3.12392785995105\\
7	2.94030873862707\\
8	2.83601419619886\\
9	2.81203661528382\\
10	2.79740735232772\\
11	2.89820953274571\\
12	2.9390528070987\\
13	2.92647317821287\\
14	2.95398319540093\\
15	3.05128392945722\\
16	3.00045164281072\\
17	2.96200760994402\\
18	2.98919545425854\\
19	3.00247647208016\\
20	3.01689386675747\\
21	3.01677513743703\\
22	3.01185622961899\\
23	2.95431821124029\\
24	2.95777613848422\\
25	2.99653039678491\\
26	2.96001792122389\\
27	3.0052969860974\\
28	3.01954438724652\\
29	2.96334846320636\\
30	2.9520303535229\\
31	3.03024838406274\\
32	3.08855886935185\\
33	3.08854646454756\\
34	3.06649180212396\\
35	2.93626018013401\\
36	2.9568217002129\\
37	2.9149170289027\\
38	2.90761849039002\\
39	2.8980293366581\\
40	2.89597168740448\\
41	2.92770685428403\\
42	2.96593378754616\\
43	2.95811619663729\\
44	2.97925569726442\\
45	2.9629550590707\\
46	3.00522174279367\\
47	3.02975732704222\\
48	3.02077127676315\\
49	2.96359269517029\\
50	2.97496551031483\\
50	3.42951022275927\\
49	3.39627039913576\\
48	3.43578859288302\\
47	3.44163188532314\\
46	3.47364240256289\\
45	3.47132998602919\\
44	3.44757917321705\\
43	3.41665774114697\\
42	3.40879301392458\\
41	3.49163227579326\\
40	3.47874349774182\\
39	3.43952143305995\\
38	3.4968626084432\\
37	3.46721262884506\\
36	3.45508274782822\\
35	3.4012175407279\\
34	3.45692858727065\\
33	3.43491856643613\\
32	3.4349287197128\\
31	3.44113772527902\\
30	3.44499575752028\\
29	3.4634225931322\\
28	3.42947589994005\\
27	3.42888142074444\\
26	3.5113365475697\\
25	3.54177610807207\\
24	3.40947282862812\\
23	3.44268712882549\\
22	3.48194792277232\\
21	3.46216771334188\\
20	3.50674098369648\\
19	3.40207733439867\\
18	3.33348094678829\\
17	3.36816440466878\\
16	3.32228775969922\\
15	3.3607711208299\\
14	3.36134975235759\\
13	3.38138518971046\\
12	3.36880522470634\\
11	3.45422652634394\\
10	3.41378806532614\\
9	3.45864423501227\\
8	3.38263551161586\\
7	3.50150482674176\\
6	3.81599517342961\\
5	5.09055833228946\\
4	8.29345105637951\\
3	10.9913953078811\\
2	20.304136698731\\
1	36.2221737765557\\
}--cycle;

\addplot[mark=none, black, very thick, dotted, domain=0:50] {2.79};

\addplot [color=black, forget plot, dashdotted]
  table[row sep=crcr]{%
1	32.9259989424285\\
2	18.738452154389\\
3	12.459832585098\\
4	9.78563585936298\\
5	7.50550439287544\\
6	6.12919092974657\\
8	6.2110890290481\\
9	6.18877715940928\\
10	6.25579120502447\\
11	6.3561736108739\\
12	6.47147506429545\\
13	6.48257839859186\\
14	6.5495701029096\\
15	6.6612194417235\\
16	6.68728288689726\\
17	6.83221267079934\\
18	6.86944933507858\\
19	6.93269847112948\\
20	6.96245943504282\\
21	6.91407568535809\\
24	7.04797272016012\\
25	7.04051326224108\\
26	7.01075329667677\\
27	7.08888082847768\\
30	7.20417278111024\\
31	7.20412643637268\\
32	7.28221850081797\\
34	7.36406774521316\\
35	7.36403342367714\\
36	7.35284564180936\\
37	7.43094850306165\\
38	7.39006609889053\\
39	7.40122093501424\\
40	7.40119919822523\\
41	7.44212553658578\\
42	7.497956186565\\
43	7.57232199927787\\
44	7.56117112679891\\
46	7.57975656390337\\
48	7.65041486024908\\
49	7.71735682118165\\
50	7.76944112507951\\
};
\addplot [color=black!80!green, forget plot, dotted]
  table[row sep=crcr]{%
1	25.6913120561146\\
2	21.1786494038839\\
3	22.562983879693\\
4	19.6806920080201\\
5	16.5902946648737\\
6	16.3189623190787\\
7	15.4672832466001\\
8	15.8776180491713\\
9	14.7393561931303\\
10	14.8547082118546\\
11	15.1935609300951\\
12	15.2530726984908\\
13	15.4021758152705\\
14	15.5287610932964\\
15	15.5398540135835\\
16	15.5771470339253\\
17	15.4655940304412\\
18	15.3354103864679\\
19	15.5177178356058\\
20	15.465604287681\\
23	16.0981338273178\\
24	16.1762181707257\\
25	16.3808318127303\\
26	16.5742799158232\\
27	16.8421506346189\\
28	17.0355882034355\\
29	17.1807258422204\\
30	17.3889852499421\\
31	17.496873965741\\
33	17.9953365494505\\
34	18.1738778886787\\
35	18.4825395527595\\
36	18.8136076437219\\
37	18.9512976361662\\
38	19.3567659093617\\
39	19.6283370676021\\
40	19.9370959694876\\
41	20.2161094428107\\
42	20.4467350141891\\
43	20.7443832009468\\
44	20.960149180016\\
45	21.2392037892157\\
46	21.4215670233061\\
47	21.6819952734794\\
48	21.7675523629401\\
49	21.9609727886067\\
50	22.1804662843663\\
};
\addplot [color=blue, forget plot, dashed]
  table[row sep=crcr]{%
1	25.8326928391276\\
2	24.8504382798553\\
3	25.0818223076238\\
4	22.4036000882376\\
5	18.8480062850963\\
6	18.8419102742776\\
7	14.5384559030279\\
8	13.6498605880112\\
9	12.2622640816687\\
10	11.6633724330361\\
11	11.611341622642\\
12	11.5257765178969\\
13	11.4030909648963\\
14	11.1315049959145\\
15	11.1204745922338\\
16	10.9605369842875\\
17	10.8750468088914\\
18	10.8490516468884\\
19	10.7002211158547\\
20	10.6183992978876\\
21	10.6444393977392\\
22	10.5849356460639\\
23	10.603592854149\\
24	10.6667900966094\\
26	10.6072605211454\\
27	10.6109606108496\\
28	10.5551920754022\\
30	10.5775403266658\\
31	10.5030707625751\\
32	10.488217543254\\
33	10.5775089881853\\
35	10.5626210375366\\
36	10.5774722351145\\
37	10.5812132125727\\
38	10.562599272244\\
39	10.5700399642236\\
40	10.5439623196397\\
41	10.5067973601504\\
42	10.5328432320505\\
43	10.5104940213662\\
44	10.5216860056386\\
45	10.4250103377625\\
46	10.3989964864471\\
47	10.4101622613155\\
48	10.3469088973215\\
49	10.3915604130614\\
50	10.3580888186773\\
};
\addplot [color=red, forget plot]
  table[row sep=crcr]{%
1	35.6423713476043\\
2	19.1515894670064\\
3	10.834332430362\\
4	7.96307654556437\\
5	4.91684465373272\\
6	3.46996151669033\\
7	3.22090678268442\\
8	3.10932485390736\\
9	3.13534042514804\\
10	3.10559770882693\\
11	3.17621802954483\\
12	3.15392901590252\\
14	3.15766647387926\\
15	3.20602752514356\\
16	3.16136970125497\\
18	3.16133820052342\\
19	3.20227690323942\\
20	3.26181742522697\\
21	3.23947142538945\\
22	3.24690207619565\\
23	3.19850267003289\\
24	3.18362448355617\\
25	3.2691532524285\\
26	3.2356772343968\\
27	3.21708920342092\\
28	3.22451014359329\\
30	3.19851305552159\\
31	3.23569305467088\\
32	3.26174379453232\\
34	3.2617101946973\\
35	3.16873886043096\\
36	3.20595222402056\\
37	3.19106482887388\\
38	3.20224054941662\\
39	3.16877538485902\\
41	3.20966956503865\\
42	3.18736340073537\\
43	3.18738696889213\\
44	3.21341743524074\\
45	3.21714252254995\\
46	3.23943207267828\\
48	3.22827993482309\\
49	3.17993154715302\\
50	3.20223786653705\\
};
\addplot [color=black, forget plot]
  table[row sep=crcr]{%
1	32.2653176205512\\
2	17.8532964684552\\
3	12.2408056287417\\
4	9.41623563841375\\
5	7.26559816492054\\
6	5.88400848899825\\
7	5.84068622316195\\
8	5.75314963809708\\
9	5.58213099656179\\
11	5.66695267030981\\
12	5.97117659984233\\
13	5.93335740375575\\
14	5.9740205078045\\
15	5.94879923825127\\
16	5.97246428835041\\
17	6.1113049708149\\
18	6.14195941484675\\
19	6.1335297960797\\
20	6.16468595583544\\
21	6.15446425858792\\
22	6.18359351884024\\
23	6.27730452459262\\
24	6.30666695633938\\
25	6.29307976860633\\
26	6.2996765817743\\
27	6.36194338705992\\
30	6.50410868664591\\
31	6.47924265088407\\
33	6.66098209009066\\
34	6.69108459947119\\
35	6.69473397483581\\
36	6.67558850405993\\
37	6.73234356204453\\
39	6.72070451858107\\
40	6.76094728630891\\
41	6.8225580704983\\
42	6.89741239978608\\
43	7.01686284725347\\
44	6.99937425148352\\
45	7.03390081180594\\
48	7.21230799642535\\
49	7.22680387976835\\
50	7.30694302104934\\
nan	nan\\
50	8.23193922910968\\
49	8.20790976259494\\
48	8.08852172407281\\
46	8.0670903504769\\
45	8.11070781519347\\
43	8.12778115130227\\
40	8.04145111014154\\
39	8.08173735144742\\
38	8.05518591730211\\
37	8.12955344407877\\
36	8.03010277955879\\
34	8.03705089095514\\
33	7.99275581726382\\
32	7.99065259628762\\
31	7.92901022186128\\
30	7.90423687557457\\
29	7.89264589965725\\
27	7.81581826989544\\
26	7.72183001157924\\
25	7.78794675587584\\
24	7.78927848398086\\
23	7.71451967280156\\
22	7.73390509153567\\
21	7.67368711212826\\
20	7.7602329142502\\
19	7.73186714617926\\
18	7.59693925531041\\
17	7.55312037078379\\
16	7.40210148544411\\
15	7.37363964519574\\
14	7.12511969801469\\
13	7.03179939342797\\
12	6.97177352874856\\
11	7.04539455143798\\
10	6.89138607353659\\
9	6.79542332225677\\
8	6.66902841999913\\
7	6.50731043688273\\
6	6.37437337049489\\
5	7.74541062083036\\
4	10.1550360803122\\
3	12.6788595414542\\
2	19.6236078403228\\
1	33.5866802643059\\
};
\addplot [color=black!80!green, forget plot]
  table[row sep=crcr]{%
1	25.0826501102209\\
2	20.8182733915379\\
3	22.258218178962\\
4	19.5567708218547\\
5	16.3385148544372\\
6	15.9384258125764\\
7	15.0903596792752\\
8	15.4247622935341\\
9	14.1518684639514\\
10	14.4209392414241\\
11	14.8827182389319\\
12	14.8856768497377\\
13	14.8636250900147\\
14	15.1260737879082\\
15	15.3115759747783\\
16	15.0916766850302\\
17	14.8608589042193\\
18	14.7363291821513\\
19	15.0274562439833\\
20	14.9998772069099\\
21	15.1970983422739\\
22	15.4149319212358\\
23	15.6557321066243\\
24	15.7748222449526\\
25	16.1091813521924\\
26	16.0153334855208\\
27	16.2536904415015\\
28	16.3123396855553\\
29	16.7601227384171\\
30	16.9877664358718\\
31	17.1310305504999\\
32	17.4267897310412\\
33	17.6531350450409\\
34	17.6928950598513\\
35	18.0127879844733\\
36	18.3012178059147\\
37	18.4241605809402\\
38	18.7419142929375\\
39	19.1266651812885\\
40	19.4373755857742\\
41	19.8110304829552\\
44	20.4100874411426\\
45	20.6680780327509\\
46	20.8422024998065\\
47	21.1597907134693\\
48	21.3237878933239\\
49	21.3805249051278\\
50	21.6325376123172\\
nan	nan\\
50	22.7283949564154\\
49	22.5414206720856\\
48	22.2113168325563\\
47	22.2041998334895\\
46	22.0009315468056\\
45	21.8103295456805\\
44	21.5102109188894\\
43	21.2718160253832\\
42	20.8832127445509\\
41	20.6211884026663\\
40	20.4368163532011\\
39	20.1300089539158\\
38	19.9716175257859\\
37	19.4784346913923\\
36	19.325997481529\\
35	18.9522911210457\\
34	18.6548607175061\\
33	18.33753805386\\
32	18.0579564708158\\
31	17.8627173809822\\
30	17.7902040640123\\
29	17.6013289460237\\
28	17.7588367213156\\
27	17.4306108277364\\
26	17.1332263461256\\
25	16.6524822732683\\
24	16.5776140964988\\
23	16.5405355480113\\
22	16.3349287351674\\
21	16.1508017729408\\
20	15.9313313684521\\
19	16.0079794272282\\
18	15.9344915907845\\
17	16.0703291566632\\
16	16.0626173828204\\
15	15.7681320523886\\
14	15.9314483986847\\
13	15.9407265405264\\
12	15.620468547244\\
11	15.5044036212584\\
10	15.2884771822851\\
9	15.3268439223092\\
8	16.3304738048086\\
7	15.844206813925\\
6	16.699498825581\\
5	16.8420744753101\\
4	19.8046131941855\\
3	22.867749580424\\
2	21.5390254162298\\
1	26.2999740020083\\
};
\addplot [color=blue, forget plot]
  table[row sep=crcr]{%
1	25.2222348392988\\
2	24.3305177517844\\
3	24.5831777921692\\
4	21.970420362519\\
5	18.5118964151506\\
6	18.5744954543389\\
7	14.1116280081771\\
8	13.2254324060553\\
9	12.0105512268702\\
10	11.2756571539852\\
11	11.169065472239\\
12	11.3219596704668\\
13	11.1295076953704\\
14	11.0230829091474\\
15	10.8931648215753\\
16	10.7884591300485\\
17	10.6630815596603\\
18	10.5887332747165\\
19	10.336300226329\\
21	10.2862460581086\\
22	10.3137814777313\\
23	10.2697527098138\\
24	10.3764888221194\\
25	10.3466482225096\\
26	10.3815345235983\\
27	10.3223232361421\\
28	10.2820567038393\\
30	10.4450000194838\\
31	10.296891696696\\
32	10.3196547893401\\
33	10.41763237503\\
35	10.3753930781209\\
36	10.3665781235427\\
37	10.3744937129501\\
38	10.3228968361005\\
39	10.3014425827525\\
40	10.3153922387687\\
41	10.3163095458351\\
42	10.3744408990628\\
44	10.3019109356742\\
45	10.1533834356693\\
46	10.163167227288\\
47	10.1986939686107\\
48	10.1273736052463\\
49	10.1974110331533\\
50	10.1392412575041\\
nan	nan\\
50	10.5769363798504\\
49	10.5857097929694\\
48	10.5664441893966\\
47	10.6216305540203\\
46	10.6348257456061\\
45	10.6966372398558\\
44	10.741461075603\\
43	10.6819944566872\\
41	10.6972851744657\\
39	10.8386373456948\\
38	10.8023017083876\\
37	10.7879327121953\\
36	10.7883663466863\\
35	10.7498489969523\\
33	10.7373856013405\\
32	10.6567802971679\\
31	10.7092498284541\\
30	10.7100806338478\\
28	10.8283274469651\\
27	10.8995979855572\\
26	10.8329865186925\\
25	10.9274234144452\\
24	10.9570913710994\\
23	10.9374329984843\\
22	10.8560898143965\\
21	11.0026327373698\\
20	10.9255332760062\\
19	11.0641420053804\\
18	11.1093700190604\\
17	11.0870120581226\\
16	11.1326148385265\\
15	11.3477843628924\\
14	11.2399270826816\\
13	11.6766742344222\\
12	11.729593365327\\
11	12.0536177730451\\
10	12.0510877120869\\
9	12.5139769364673\\
8	14.0742887699671\\
7	14.9652837978786\\
6	19.1093250942162\\
5	19.184116155042\\
4	22.8367798139562\\
3	25.5804668230785\\
2	25.3703588079262\\
1	26.4431508389564\\
};
\addplot [color=red, forget plot]
  table[row sep=crcr]{%
1	35.0625689186529\\
2	17.9990422352818\\
3	10.6772695528429\\
4	7.63270203474922\\
5	4.74313097517599\\
6	3.12392785995105\\
7	2.94030873862707\\
8	2.83601419619886\\
10	2.79740735232772\\
11	2.89820953274571\\
12	2.9390528070987\\
13	2.92647317821287\\
14	2.95398319540093\\
15	3.05128392945722\\
16	3.00045164281072\\
17	2.96200760994402\\
18	2.98919545425854\\
20	3.01689386675746\\
22	3.01185622961899\\
23	2.95431821124029\\
24	2.95777613848422\\
25	2.99653039678491\\
26	2.96001792122389\\
27	3.0052969860974\\
28	3.01954438724653\\
29	2.96334846320637\\
30	2.9520303535229\\
31	3.03024838406274\\
32	3.08855886935185\\
33	3.08854646454756\\
34	3.06649180212396\\
35	2.93626018013401\\
36	2.95682170021291\\
37	2.9149170289027\\
40	2.89597168740448\\
42	2.96593378754616\\
43	2.95811619663729\\
44	2.97925569726441\\
45	2.9629550590707\\
46	3.00522174279367\\
47	3.02975732704222\\
48	3.02077127676316\\
49	2.96359269517029\\
50	2.97496551031482\\
nan	nan\\
50	3.42951022275927\\
49	3.39627039913576\\
48	3.43578859288302\\
47	3.44163188532314\\
46	3.47364240256289\\
45	3.47132998602919\\
42	3.40879301392457\\
41	3.49163227579326\\
40	3.47874349774182\\
39	3.43952143305994\\
38	3.4968626084432\\
37	3.46721262884506\\
36	3.45508274782822\\
35	3.4012175407279\\
34	3.45692858727065\\
33	3.43491856643613\\
30	3.44499575752028\\
29	3.4634225931322\\
28	3.42947589994005\\
27	3.42888142074444\\
26	3.51133654756971\\
25	3.54177610807208\\
24	3.40947282862812\\
22	3.48194792277232\\
21	3.46216771334188\\
20	3.50674098369648\\
19	3.40207733439867\\
18	3.33348094678828\\
17	3.36816440466878\\
16	3.32228775969922\\
15	3.3607711208299\\
14	3.36134975235758\\
13	3.38138518971046\\
12	3.36880522470634\\
11	3.45422652634394\\
10	3.41378806532614\\
9	3.45864423501227\\
8	3.38263551161586\\
7	3.50150482674177\\
6	3.81599517342961\\
5	5.09055833228946\\
4	8.29345105637951\\
3	10.9913953078811\\
2	20.304136698731\\
1	36.2221737765557\\
};
\end{axis}
\end{tikzpicture}%

%% file: Figures/balancing0_norm0_FFttransformed_50rSVDmodes_duplicate2.tex
\definecolor{mycolor1}{rgb}{0.64, 0.76, 0.68}%
\begin{tikzpicture}

\begin{axis}[%
width=7cm,
height=5cm,
scale only axis,
unbounded coords=jump,
xmin=0,
xmax=50,
xlabel style={font=\color{white!15!black}},
xlabel={$r$},
ymin=0,
ymax=50,
ylabel style={font=\color{white!15!black}},
ylabel={Test error in \%},
axis background/.style={fill=white},
title style={font=\bfseries},
legend columns=2,
legend style={legend cell align=left, align=left, draw=white!15!black, font = \footnotesize},
legend entries={Tree,
                NB,
                LD,
                kNN,
                2.8~\%},
]
\addlegendimage{legend image code/.code={
\draw [draw=none, fill=white!80!black] (0cm,-0.15cm) rectangle (0.6cm,0.15cm);
\draw[dashdotted] (0cm,0cm) -- (0.6cm,0cm);
\draw[-] (0cm,0.15cm) -- (0.6cm,0.15cm);
\draw[-] (0cm,-0.15cm) -- (0.6cm,-0.15cm);
}
} 
\addlegendimage{legend image code/.code={
\draw [draw=none, fill=mycolor1] (0cm,-0.15cm) rectangle (0.6cm,0.15cm);
\draw[dotted, black!80!green] (0cm,0cm) -- (0.6cm,0cm);
\draw[-, black!80!green] (0cm,0.15cm) -- (0.6cm,0.15cm);
\draw[-, black!80!green] (0cm,-0.15cm) -- (0.6cm,-0.15cm);
}
}
\addlegendimage{legend image code/.code={
\draw [draw=none, fill=white!80!blue] (0cm,-0.15cm) rectangle (0.6cm,0.15cm);
\draw[dashed, blue] (0cm,0cm) -- (0.6cm,0cm);
\draw[-, blue] (0cm,0.15cm) -- (0.6cm,0.15cm);
\draw[-, blue] (0cm,-0.15cm) -- (0.6cm,-0.15cm);
}
}
\addlegendimage{legend image code/.code={
\draw [draw=none, fill=white!80!red] (0cm,-0.15cm) rectangle (0.6cm,0.15cm);
\draw[-, red] (0cm,0cm) -- (0.6cm,0cm);
\draw[-, red] (0cm,0.15cm) -- (0.6cm,0.15cm);
\draw[-, red] (0cm,-0.15cm) -- (0.6cm,-0.15cm);
}
}
\addlegendimage{mark=none, black, very thick, dotted}

\addplot[area legend, draw=black, fill=white!80!black, dashdotted]
table[row sep=crcr] {%
x	y\\
1	32.2653176205512\\
2	17.8532964684552\\
3	12.2408056287417\\
4	9.41623563841375\\
5	7.26559816492054\\
6	5.88400848899825\\
7	5.84068622316195\\
8	5.75314963809708\\
9	5.58213099656179\\
10	5.62019633651235\\
11	5.66695267030981\\
12	5.97117659984233\\
13	5.93335740375575\\
14	5.97402050780451\\
15	5.94879923825127\\
16	5.97246428835041\\
17	6.1113049708149\\
18	6.14195941484675\\
19	6.1335297960797\\
20	6.16468595583544\\
21	6.15446425858792\\
22	6.18359351884024\\
23	6.27730452459262\\
24	6.30666695633937\\
25	6.29307976860633\\
26	6.2996765817743\\
27	6.36194338705992\\
28	6.40665346460722\\
29	6.45621230953474\\
30	6.5041086866459\\
31	6.47924265088407\\
32	6.57378440534831\\
33	6.66098209009066\\
34	6.69108459947119\\
35	6.6947339748358\\
36	6.67558850405993\\
37	6.73234356204454\\
38	6.72494628047895\\
39	6.72070451858107\\
40	6.76094728630891\\
41	6.8225580704983\\
42	6.89741239978608\\
43	7.01686284725347\\
44	6.99937425148352\\
45	7.03390081180594\\
46	7.09242277732984\\
47	7.14587703875906\\
48	7.21230799642535\\
49	7.22680387976835\\
50	7.30694302104934\\
50	8.23193922910968\\
49	8.20790976259494\\
48	8.0885217240728\\
47	8.0805561931555\\
46	8.0670903504769\\
45	8.11070781519347\\
44	8.1229680021143\\
43	8.12778115130227\\
42	8.09849997334392\\
41	8.06169300267325\\
40	8.04145111014155\\
39	8.08173735144742\\
38	8.05518591730211\\
37	8.12955344407876\\
36	8.03010277955879\\
35	8.03333287251847\\
34	8.03705089095514\\
33	7.99275581726382\\
32	7.99065259628762\\
31	7.92901022186128\\
30	7.90423687557456\\
29	7.89264589965725\\
28	7.85291183649545\\
27	7.81581826989545\\
26	7.72183001157924\\
25	7.78794675587584\\
24	7.78927848398085\\
23	7.71451967280156\\
22	7.73390509153567\\
21	7.67368711212826\\
20	7.76023291425019\\
19	7.73186714617926\\
18	7.59693925531041\\
17	7.55312037078379\\
16	7.40210148544412\\
15	7.37363964519574\\
14	7.12511969801469\\
13	7.03179939342797\\
12	6.97177352874856\\
11	7.04539455143798\\
10	6.89138607353659\\
9	6.79542332225676\\
8	6.66902841999913\\
7	6.50731043688273\\
6	6.37437337049488\\
5	7.74541062083036\\
4	10.1550360803122\\
3	12.6788595414542\\
2	19.6236078403228\\
1	33.5866802643059\\
}--cycle;
\addlegendentry{Tree}

\addplot[area legend, draw=black!80!green, fill=mycolor1, dotted]
table[row sep=crcr] {%
x	y\\
1	25.0826501102209\\
2	20.8182733915379\\
3	22.258218178962\\
4	19.5567708218547\\
5	16.3385148544372\\
6	15.9384258125764\\
7	15.0903596792752\\
8	15.424762293534\\
9	14.1518684639514\\
10	14.4209392414241\\
11	14.8827182389319\\
12	14.8856768497377\\
13	14.8636250900147\\
14	15.1260737879082\\
15	15.3115759747783\\
16	15.0916766850302\\
17	14.8608589042193\\
18	14.7363291821513\\
19	15.0274562439833\\
20	14.9998772069099\\
21	15.1970983422739\\
22	15.4149319212358\\
23	15.6557321066243\\
24	15.7748222449526\\
25	16.1091813521924\\
26	16.0153334855208\\
27	16.2536904415015\\
28	16.3123396855553\\
29	16.7601227384171\\
30	16.9877664358718\\
31	17.1310305504999\\
32	17.4267897310412\\
33	17.6531350450409\\
34	17.6928950598513\\
35	18.0127879844733\\
36	18.3012178059147\\
37	18.4241605809402\\
38	18.7419142929375\\
39	19.1266651812885\\
40	19.4373755857742\\
41	19.8110304829552\\
42	20.0102572838273\\
43	20.2169503765104\\
44	20.4100874411426\\
45	20.6680780327509\\
46	20.8422024998065\\
47	21.1597907134693\\
48	21.3237878933239\\
49	21.3805249051278\\
50	21.6325376123172\\
50	22.7283949564154\\
49	22.5414206720856\\
48	22.2113168325563\\
47	22.2041998334895\\
46	22.0009315468056\\
45	21.8103295456805\\
44	21.5102109188894\\
43	21.2718160253832\\
42	20.8832127445509\\
41	20.6211884026663\\
40	20.4368163532011\\
39	20.1300089539158\\
38	19.9716175257859\\
37	19.4784346913923\\
36	19.325997481529\\
35	18.9522911210457\\
34	18.6548607175061\\
33	18.33753805386\\
32	18.0579564708158\\
31	17.8627173809822\\
30	17.7902040640123\\
29	17.6013289460237\\
28	17.7588367213156\\
27	17.4306108277364\\
26	17.1332263461256\\
25	16.6524822732683\\
24	16.5776140964988\\
23	16.5405355480113\\
22	16.3349287351674\\
21	16.1508017729408\\
20	15.9313313684522\\
19	16.0079794272282\\
18	15.9344915907845\\
17	16.0703291566632\\
16	16.0626173828204\\
15	15.7681320523886\\
14	15.9314483986846\\
13	15.9407265405264\\
12	15.620468547244\\
11	15.5044036212584\\
10	15.2884771822851\\
9	15.3268439223092\\
8	16.3304738048086\\
7	15.844206813925\\
6	16.699498825581\\
5	16.8420744753101\\
4	19.8046131941855\\
3	22.867749580424\\
2	21.5390254162298\\
1	26.2999740020083\\
}--cycle;
\addlegendentry{NB}

\addplot[area legend, draw=blue, fill=white!80!blue, dashed]
table[row sep=crcr] {%
x	y\\
1	25.2222348392988\\
2	24.3305177517844\\
3	24.5831777921692\\
4	21.970420362519\\
5	18.5118964151506\\
6	18.5744954543389\\
7	14.1116280081771\\
8	13.2254324060553\\
9	12.0105512268702\\
10	11.2756571539852\\
11	11.169065472239\\
12	11.3219596704668\\
13	11.1295076953704\\
14	11.0230829091474\\
15	10.8931648215753\\
16	10.7884591300485\\
17	10.6630815596603\\
18	10.5887332747165\\
19	10.336300226329\\
20	10.311265319769\\
21	10.2862460581086\\
22	10.3137814777313\\
23	10.2697527098138\\
24	10.3764888221194\\
25	10.3466482225096\\
26	10.3815345235983\\
27	10.3223232361421\\
28	10.2820567038392\\
29	10.360930633376\\
30	10.4450000194838\\
31	10.296891696696\\
32	10.3196547893401\\
33	10.41763237503\\
34	10.3930966203781\\
35	10.3753930781208\\
36	10.3665781235427\\
37	10.3744937129501\\
38	10.3228968361005\\
39	10.3014425827525\\
40	10.3153922387687\\
41	10.3163095458351\\
42	10.3744408990628\\
43	10.3389935860452\\
44	10.3019109356742\\
45	10.1533834356693\\
46	10.163167227288\\
47	10.1986939686107\\
48	10.1273736052463\\
49	10.1974110331533\\
50	10.1392412575041\\
50	10.5769363798504\\
49	10.5857097929694\\
48	10.5664441893966\\
47	10.6216305540203\\
46	10.6348257456061\\
45	10.6966372398558\\
44	10.741461075603\\
43	10.6819944566872\\
42	10.6912455650382\\
41	10.6972851744657\\
40	10.7725324005107\\
39	10.8386373456948\\
38	10.8023017083876\\
37	10.7879327121953\\
36	10.7883663466863\\
35	10.7498489969523\\
34	10.7470301005894\\
33	10.7373856013405\\
32	10.6567802971679\\
31	10.7092498284541\\
30	10.7100806338478\\
29	10.7644079265902\\
28	10.8283274469651\\
27	10.8995979855572\\
26	10.8329865186925\\
25	10.9274234144452\\
24	10.9570913710994\\
23	10.9374329984843\\
22	10.8560898143965\\
21	11.0026327373698\\
20	10.9255332760062\\
19	11.0641420053804\\
18	11.1093700190604\\
17	11.0870120581226\\
16	11.1326148385265\\
15	11.3477843628924\\
14	11.2399270826816\\
13	11.6766742344222\\
12	11.729593365327\\
11	12.0536177730451\\
10	12.0510877120869\\
9	12.5139769364673\\
8	14.0742887699671\\
7	14.9652837978786\\
6	19.1093250942162\\
5	19.184116155042\\
4	22.8367798139562\\
3	25.5804668230785\\
2	25.3703588079262\\
1	26.4431508389564\\
}--cycle;
\addlegendentry{LD}

\addplot[area legend, draw=red, fill=white!80!red]
table[row sep=crcr] {%
x	y\\
1	35.0625689186529\\
2	17.9990422352818\\
3	10.6772695528429\\
4	7.63270203474922\\
5	4.74313097517599\\
6	3.12392785995105\\
7	2.94030873862707\\
8	2.83601419619886\\
9	2.81203661528382\\
10	2.79740735232772\\
11	2.89820953274571\\
12	2.9390528070987\\
13	2.92647317821287\\
14	2.95398319540093\\
15	3.05128392945722\\
16	3.00045164281072\\
17	2.96200760994402\\
18	2.98919545425854\\
19	3.00247647208016\\
20	3.01689386675747\\
21	3.01677513743703\\
22	3.01185622961899\\
23	2.95431821124029\\
24	2.95777613848422\\
25	2.99653039678491\\
26	2.96001792122389\\
27	3.0052969860974\\
28	3.01954438724652\\
29	2.96334846320636\\
30	2.9520303535229\\
31	3.03024838406274\\
32	3.08855886935185\\
33	3.08854646454756\\
34	3.06649180212396\\
35	2.93626018013401\\
36	2.9568217002129\\
37	2.9149170289027\\
38	2.90761849039002\\
39	2.8980293366581\\
40	2.89597168740448\\
41	2.92770685428403\\
42	2.96593378754616\\
43	2.95811619663729\\
44	2.97925569726442\\
45	2.9629550590707\\
46	3.00522174279367\\
47	3.02975732704222\\
48	3.02077127676315\\
49	2.96359269517029\\
50	2.97496551031483\\
50	3.42951022275927\\
49	3.39627039913576\\
48	3.43578859288302\\
47	3.44163188532314\\
46	3.47364240256289\\
45	3.47132998602919\\
44	3.44757917321705\\
43	3.41665774114697\\
42	3.40879301392458\\
41	3.49163227579326\\
40	3.47874349774182\\
39	3.43952143305995\\
38	3.4968626084432\\
37	3.46721262884506\\
36	3.45508274782822\\
35	3.4012175407279\\
34	3.45692858727065\\
33	3.43491856643613\\
32	3.4349287197128\\
31	3.44113772527902\\
30	3.44499575752028\\
29	3.4634225931322\\
28	3.42947589994005\\
27	3.42888142074444\\
26	3.5113365475697\\
25	3.54177610807207\\
24	3.40947282862812\\
23	3.44268712882549\\
22	3.48194792277232\\
21	3.46216771334188\\
20	3.50674098369648\\
19	3.40207733439867\\
18	3.33348094678829\\
17	3.36816440466878\\
16	3.32228775969922\\
15	3.3607711208299\\
14	3.36134975235759\\
13	3.38138518971046\\
12	3.36880522470634\\
11	3.45422652634394\\
10	3.41378806532614\\
9	3.45864423501227\\
8	3.38263551161586\\
7	3.50150482674176\\
6	3.81599517342961\\
5	5.09055833228946\\
4	8.29345105637951\\
3	10.9913953078811\\
2	20.304136698731\\
1	36.2221737765557\\
}--cycle;
\addlegendentry{kNN}

\addplot[mark=none, black, very thick, dotted, domain=0:50] {2.79};
\addlegendentry{2.8~\%}

\addplot [color=black, forget plot, dashdotted]
  table[row sep=crcr]{%
1	32.9259989424285\\
2	18.738452154389\\
3	12.459832585098\\
4	9.78563585936298\\
5	7.50550439287544\\
6	6.12919092974657\\
8	6.2110890290481\\
9	6.18877715940928\\
10	6.25579120502447\\
11	6.3561736108739\\
12	6.47147506429545\\
13	6.48257839859186\\
14	6.5495701029096\\
15	6.6612194417235\\
16	6.68728288689726\\
17	6.83221267079934\\
18	6.86944933507858\\
19	6.93269847112948\\
20	6.96245943504282\\
21	6.91407568535809\\
24	7.04797272016012\\
25	7.04051326224108\\
26	7.01075329667677\\
27	7.08888082847768\\
30	7.20417278111024\\
31	7.20412643637268\\
32	7.28221850081797\\
34	7.36406774521316\\
35	7.36403342367714\\
36	7.35284564180936\\
37	7.43094850306165\\
38	7.39006609889053\\
39	7.40122093501424\\
40	7.40119919822523\\
41	7.44212553658578\\
42	7.497956186565\\
43	7.57232199927787\\
44	7.56117112679891\\
46	7.57975656390337\\
48	7.65041486024908\\
49	7.71735682118165\\
50	7.76944112507951\\
};
\addplot [color=black!80!green, forget plot, dotted]
  table[row sep=crcr]{%
1	25.6913120561146\\
2	21.1786494038839\\
3	22.562983879693\\
4	19.6806920080201\\
5	16.5902946648737\\
6	16.3189623190787\\
7	15.4672832466001\\
8	15.8776180491713\\
9	14.7393561931303\\
10	14.8547082118546\\
11	15.1935609300951\\
12	15.2530726984908\\
13	15.4021758152705\\
14	15.5287610932964\\
15	15.5398540135835\\
16	15.5771470339253\\
17	15.4655940304412\\
18	15.3354103864679\\
19	15.5177178356058\\
20	15.465604287681\\
23	16.0981338273178\\
24	16.1762181707257\\
25	16.3808318127303\\
26	16.5742799158232\\
27	16.8421506346189\\
28	17.0355882034355\\
29	17.1807258422204\\
30	17.3889852499421\\
31	17.496873965741\\
33	17.9953365494505\\
34	18.1738778886787\\
35	18.4825395527595\\
36	18.8136076437219\\
37	18.9512976361662\\
38	19.3567659093617\\
39	19.6283370676021\\
40	19.9370959694876\\
41	20.2161094428107\\
42	20.4467350141891\\
43	20.7443832009468\\
44	20.960149180016\\
45	21.2392037892157\\
46	21.4215670233061\\
47	21.6819952734794\\
48	21.7675523629401\\
49	21.9609727886067\\
50	22.1804662843663\\
};
\addplot [color=blue, forget plot, dashed]
  table[row sep=crcr]{%
1	25.8326928391276\\
2	24.8504382798553\\
3	25.0818223076238\\
4	22.4036000882376\\
5	18.8480062850963\\
6	18.8419102742776\\
7	14.5384559030279\\
8	13.6498605880112\\
9	12.2622640816687\\
10	11.6633724330361\\
11	11.611341622642\\
12	11.5257765178969\\
13	11.4030909648963\\
14	11.1315049959145\\
15	11.1204745922338\\
16	10.9605369842875\\
17	10.8750468088914\\
18	10.8490516468884\\
19	10.7002211158547\\
20	10.6183992978876\\
21	10.6444393977392\\
22	10.5849356460639\\
23	10.603592854149\\
24	10.6667900966094\\
26	10.6072605211454\\
27	10.6109606108496\\
28	10.5551920754022\\
30	10.5775403266658\\
31	10.5030707625751\\
32	10.488217543254\\
33	10.5775089881853\\
35	10.5626210375366\\
36	10.5774722351145\\
37	10.5812132125727\\
38	10.562599272244\\
39	10.5700399642236\\
40	10.5439623196397\\
41	10.5067973601504\\
42	10.5328432320505\\
43	10.5104940213662\\
44	10.5216860056386\\
45	10.4250103377625\\
46	10.3989964864471\\
47	10.4101622613155\\
48	10.3469088973215\\
49	10.3915604130614\\
50	10.3580888186773\\
};
\addplot [color=red, forget plot]
  table[row sep=crcr]{%
1	35.6423713476043\\
2	19.1515894670064\\
3	10.834332430362\\
4	7.96307654556437\\
5	4.91684465373272\\
6	3.46996151669033\\
7	3.22090678268442\\
8	3.10932485390736\\
9	3.13534042514804\\
10	3.10559770882693\\
11	3.17621802954483\\
12	3.15392901590252\\
14	3.15766647387926\\
15	3.20602752514356\\
16	3.16136970125497\\
18	3.16133820052342\\
19	3.20227690323942\\
20	3.26181742522697\\
21	3.23947142538945\\
22	3.24690207619565\\
23	3.19850267003289\\
24	3.18362448355617\\
25	3.2691532524285\\
26	3.2356772343968\\
27	3.21708920342092\\
28	3.22451014359329\\
30	3.19851305552159\\
31	3.23569305467088\\
32	3.26174379453232\\
34	3.2617101946973\\
35	3.16873886043096\\
36	3.20595222402056\\
37	3.19106482887388\\
38	3.20224054941662\\
39	3.16877538485902\\
41	3.20966956503865\\
42	3.18736340073537\\
43	3.18738696889213\\
44	3.21341743524074\\
45	3.21714252254995\\
46	3.23943207267828\\
48	3.22827993482309\\
49	3.17993154715302\\
50	3.20223786653705\\
};
\addplot [color=black, forget plot]
  table[row sep=crcr]{%
1	32.2653176205512\\
2	17.8532964684552\\
3	12.2408056287417\\
4	9.41623563841375\\
5	7.26559816492054\\
6	5.88400848899825\\
7	5.84068622316195\\
8	5.75314963809708\\
9	5.58213099656179\\
11	5.66695267030981\\
12	5.97117659984233\\
13	5.93335740375575\\
14	5.9740205078045\\
15	5.94879923825127\\
16	5.97246428835041\\
17	6.1113049708149\\
18	6.14195941484675\\
19	6.1335297960797\\
20	6.16468595583544\\
21	6.15446425858792\\
22	6.18359351884024\\
23	6.27730452459262\\
24	6.30666695633938\\
25	6.29307976860633\\
26	6.2996765817743\\
27	6.36194338705992\\
30	6.50410868664591\\
31	6.47924265088407\\
33	6.66098209009066\\
34	6.69108459947119\\
35	6.69473397483581\\
36	6.67558850405993\\
37	6.73234356204453\\
39	6.72070451858107\\
40	6.76094728630891\\
41	6.8225580704983\\
42	6.89741239978608\\
43	7.01686284725347\\
44	6.99937425148352\\
45	7.03390081180594\\
48	7.21230799642535\\
49	7.22680387976835\\
50	7.30694302104934\\
nan	nan\\
50	8.23193922910968\\
49	8.20790976259494\\
48	8.08852172407281\\
46	8.0670903504769\\
45	8.11070781519347\\
43	8.12778115130227\\
40	8.04145111014154\\
39	8.08173735144742\\
38	8.05518591730211\\
37	8.12955344407877\\
36	8.03010277955879\\
34	8.03705089095514\\
33	7.99275581726382\\
32	7.99065259628762\\
31	7.92901022186128\\
30	7.90423687557457\\
29	7.89264589965725\\
27	7.81581826989544\\
26	7.72183001157924\\
25	7.78794675587584\\
24	7.78927848398086\\
23	7.71451967280156\\
22	7.73390509153567\\
21	7.67368711212826\\
20	7.7602329142502\\
19	7.73186714617926\\
18	7.59693925531041\\
17	7.55312037078379\\
16	7.40210148544411\\
15	7.37363964519574\\
14	7.12511969801469\\
13	7.03179939342797\\
12	6.97177352874856\\
11	7.04539455143798\\
10	6.89138607353659\\
9	6.79542332225677\\
8	6.66902841999913\\
7	6.50731043688273\\
6	6.37437337049489\\
5	7.74541062083036\\
4	10.1550360803122\\
3	12.6788595414542\\
2	19.6236078403228\\
1	33.5866802643059\\
};
\addplot [color=black!80!green, forget plot]
  table[row sep=crcr]{%
1	25.0826501102209\\
2	20.8182733915379\\
3	22.258218178962\\
4	19.5567708218547\\
5	16.3385148544372\\
6	15.9384258125764\\
7	15.0903596792752\\
8	15.4247622935341\\
9	14.1518684639514\\
10	14.4209392414241\\
11	14.8827182389319\\
12	14.8856768497377\\
13	14.8636250900147\\
14	15.1260737879082\\
15	15.3115759747783\\
16	15.0916766850302\\
17	14.8608589042193\\
18	14.7363291821513\\
19	15.0274562439833\\
20	14.9998772069099\\
21	15.1970983422739\\
22	15.4149319212358\\
23	15.6557321066243\\
24	15.7748222449526\\
25	16.1091813521924\\
26	16.0153334855208\\
27	16.2536904415015\\
28	16.3123396855553\\
29	16.7601227384171\\
30	16.9877664358718\\
31	17.1310305504999\\
32	17.4267897310412\\
33	17.6531350450409\\
34	17.6928950598513\\
35	18.0127879844733\\
36	18.3012178059147\\
37	18.4241605809402\\
38	18.7419142929375\\
39	19.1266651812885\\
40	19.4373755857742\\
41	19.8110304829552\\
44	20.4100874411426\\
45	20.6680780327509\\
46	20.8422024998065\\
47	21.1597907134693\\
48	21.3237878933239\\
49	21.3805249051278\\
50	21.6325376123172\\
nan	nan\\
50	22.7283949564154\\
49	22.5414206720856\\
48	22.2113168325563\\
47	22.2041998334895\\
46	22.0009315468056\\
45	21.8103295456805\\
44	21.5102109188894\\
43	21.2718160253832\\
42	20.8832127445509\\
41	20.6211884026663\\
40	20.4368163532011\\
39	20.1300089539158\\
38	19.9716175257859\\
37	19.4784346913923\\
36	19.325997481529\\
35	18.9522911210457\\
34	18.6548607175061\\
33	18.33753805386\\
32	18.0579564708158\\
31	17.8627173809822\\
30	17.7902040640123\\
29	17.6013289460237\\
28	17.7588367213156\\
27	17.4306108277364\\
26	17.1332263461256\\
25	16.6524822732683\\
24	16.5776140964988\\
23	16.5405355480113\\
22	16.3349287351674\\
21	16.1508017729408\\
20	15.9313313684521\\
19	16.0079794272282\\
18	15.9344915907845\\
17	16.0703291566632\\
16	16.0626173828204\\
15	15.7681320523886\\
14	15.9314483986847\\
13	15.9407265405264\\
12	15.620468547244\\
11	15.5044036212584\\
10	15.2884771822851\\
9	15.3268439223092\\
8	16.3304738048086\\
7	15.844206813925\\
6	16.699498825581\\
5	16.8420744753101\\
4	19.8046131941855\\
3	22.867749580424\\
2	21.5390254162298\\
1	26.2999740020083\\
};
\addplot [color=blue, forget plot]
  table[row sep=crcr]{%
1	25.2222348392988\\
2	24.3305177517844\\
3	24.5831777921692\\
4	21.970420362519\\
5	18.5118964151506\\
6	18.5744954543389\\
7	14.1116280081771\\
8	13.2254324060553\\
9	12.0105512268702\\
10	11.2756571539852\\
11	11.169065472239\\
12	11.3219596704668\\
13	11.1295076953704\\
14	11.0230829091474\\
15	10.8931648215753\\
16	10.7884591300485\\
17	10.6630815596603\\
18	10.5887332747165\\
19	10.336300226329\\
21	10.2862460581086\\
22	10.3137814777313\\
23	10.2697527098138\\
24	10.3764888221194\\
25	10.3466482225096\\
26	10.3815345235983\\
27	10.3223232361421\\
28	10.2820567038393\\
30	10.4450000194838\\
31	10.296891696696\\
32	10.3196547893401\\
33	10.41763237503\\
35	10.3753930781209\\
36	10.3665781235427\\
37	10.3744937129501\\
38	10.3228968361005\\
39	10.3014425827525\\
40	10.3153922387687\\
41	10.3163095458351\\
42	10.3744408990628\\
44	10.3019109356742\\
45	10.1533834356693\\
46	10.163167227288\\
47	10.1986939686107\\
48	10.1273736052463\\
49	10.1974110331533\\
50	10.1392412575041\\
nan	nan\\
50	10.5769363798504\\
49	10.5857097929694\\
48	10.5664441893966\\
47	10.6216305540203\\
46	10.6348257456061\\
45	10.6966372398558\\
44	10.741461075603\\
43	10.6819944566872\\
41	10.6972851744657\\
39	10.8386373456948\\
38	10.8023017083876\\
37	10.7879327121953\\
36	10.7883663466863\\
35	10.7498489969523\\
33	10.7373856013405\\
32	10.6567802971679\\
31	10.7092498284541\\
30	10.7100806338478\\
28	10.8283274469651\\
27	10.8995979855572\\
26	10.8329865186925\\
25	10.9274234144452\\
24	10.9570913710994\\
23	10.9374329984843\\
22	10.8560898143965\\
21	11.0026327373698\\
20	10.9255332760062\\
19	11.0641420053804\\
18	11.1093700190604\\
17	11.0870120581226\\
16	11.1326148385265\\
15	11.3477843628924\\
14	11.2399270826816\\
13	11.6766742344222\\
12	11.729593365327\\
11	12.0536177730451\\
10	12.0510877120869\\
9	12.5139769364673\\
8	14.0742887699671\\
7	14.9652837978786\\
6	19.1093250942162\\
5	19.184116155042\\
4	22.8367798139562\\
3	25.5804668230785\\
2	25.3703588079262\\
1	26.4431508389564\\
};
\addplot [color=red, forget plot]
  table[row sep=crcr]{%
1	35.0625689186529\\
2	17.9990422352818\\
3	10.6772695528429\\
4	7.63270203474922\\
5	4.74313097517599\\
6	3.12392785995105\\
7	2.94030873862707\\
8	2.83601419619886\\
10	2.79740735232772\\
11	2.89820953274571\\
12	2.9390528070987\\
13	2.92647317821287\\
14	2.95398319540093\\
15	3.05128392945722\\
16	3.00045164281072\\
17	2.96200760994402\\
18	2.98919545425854\\
20	3.01689386675746\\
22	3.01185622961899\\
23	2.95431821124029\\
24	2.95777613848422\\
25	2.99653039678491\\
26	2.96001792122389\\
27	3.0052969860974\\
28	3.01954438724653\\
29	2.96334846320637\\
30	2.9520303535229\\
31	3.03024838406274\\
32	3.08855886935185\\
33	3.08854646454756\\
34	3.06649180212396\\
35	2.93626018013401\\
36	2.95682170021291\\
37	2.9149170289027\\
40	2.89597168740448\\
42	2.96593378754616\\
43	2.95811619663729\\
44	2.97925569726441\\
45	2.9629550590707\\
46	3.00522174279367\\
47	3.02975732704222\\
48	3.02077127676316\\
49	2.96359269517029\\
50	2.97496551031482\\
nan	nan\\
50	3.42951022275927\\
49	3.39627039913576\\
48	3.43578859288302\\
47	3.44163188532314\\
46	3.47364240256289\\
45	3.47132998602919\\
42	3.40879301392457\\
41	3.49163227579326\\
40	3.47874349774182\\
39	3.43952143305994\\
38	3.4968626084432\\
37	3.46721262884506\\
36	3.45508274782822\\
35	3.4012175407279\\
34	3.45692858727065\\
33	3.43491856643613\\
30	3.44499575752028\\
29	3.4634225931322\\
28	3.42947589994005\\
27	3.42888142074444\\
26	3.51133654756971\\
25	3.54177610807208\\
24	3.40947282862812\\
22	3.48194792277232\\
21	3.46216771334188\\
20	3.50674098369648\\
19	3.40207733439867\\
18	3.33348094678828\\
17	3.36816440466878\\
16	3.32228775969922\\
15	3.3607711208299\\
14	3.36134975235758\\
13	3.38138518971046\\
12	3.36880522470634\\
11	3.45422652634394\\
10	3.41378806532614\\
9	3.45864423501227\\
8	3.38263551161586\\
7	3.50150482674177\\
6	3.81599517342961\\
5	5.09055833228946\\
4	8.29345105637951\\
3	10.9913953078811\\
2	20.304136698731\\
1	36.2221737765557\\
};
\end{axis}
\end{tikzpicture}%

%% file: Figures/balancing0_norm1_FFTtransformed_50rSVDmodes.tex
\definecolor{mycolor1}{rgb}{0.64, 0.76, 0.68}%
\begin{tikzpicture}

\begin{axis}[%
width=7cm,
height=5cm,
scale only axis,
unbounded coords=jump,
xmin=0,
xmax=50,
xlabel style={font=\color{white!15!black}},
xlabel={$r$},
ymin=0,
ymax=50,
ylabel style={font=\color{white!15!black}},
axis background/.style={fill=white},
title style={font=\bfseries},
legend columns = 2,
legend style={legend cell align=left, align=left, draw=white!15!black, font = \footnotesize},
legend entries={Tree,
                NB,
                LD,
                kNN,
                3.4~\%},
]
\addlegendimage{legend image code/.code={
\draw [draw=none, fill=white!80!black] (0cm,-0.15cm) rectangle (0.6cm,0.15cm);
\draw[dashdotted] (0cm,0cm) -- (0.6cm,0cm);
\draw[-] (0cm,0.15cm) -- (0.6cm,0.15cm);
\draw[-] (0cm,-0.15cm) -- (0.6cm,-0.15cm);
}
} 
\addlegendimage{legend image code/.code={
\draw [draw=none, fill=mycolor1] (0cm,-0.15cm) rectangle (0.6cm,0.15cm);
\draw[dotted, black!80!green] (0cm,0cm) -- (0.6cm,0cm);
\draw[-, black!80!green] (0cm,0.15cm) -- (0.6cm,0.15cm);
\draw[-, black!80!green] (0cm,-0.15cm) -- (0.6cm,-0.15cm);
}
}
\addlegendimage{legend image code/.code={
\draw [draw=none, fill=white!80!blue] (0cm,-0.15cm) rectangle (0.6cm,0.15cm);
\draw[dashed, blue] (0cm,0cm) -- (0.6cm,0cm);
\draw[-, blue] (0cm,0.15cm) -- (0.6cm,0.15cm);
\draw[-, blue] (0cm,-0.15cm) -- (0.6cm,-0.15cm);
}
}
\addlegendimage{legend image code/.code={
\draw [draw=none, fill=white!80!red] (0cm,-0.15cm) rectangle (0.6cm,0.15cm);
\draw[-, red] (0cm,0cm) -- (0.6cm,0cm);
\draw[-, red] (0cm,0.15cm) -- (0.6cm,0.15cm);
\draw[-, red] (0cm,-0.15cm) -- (0.6cm,-0.15cm);
}
}
\addlegendimage{mark=none, black, very thick, dotted}

\addplot[area legend, draw=black, fill=white!80!black, dashdotted]
table[row sep=crcr] {%
x	y\\
1	32.2621262976457\\
2	19.537710871243\\
3	12.5481599070386\\
4	9.13120319828073\\
5	6.86806478146147\\
6	5.64891271442786\\
7	6.1045515323992\\
8	6.01766920156828\\
9	5.91674608812642\\
10	5.94513991841254\\
11	5.91126999342652\\
12	6.12555519328891\\
13	6.1447458146752\\
14	6.19746879862158\\
15	6.23087937192761\\
16	6.3489890464151\\
17	6.34611342351325\\
18	6.42230038726623\\
19	6.45117173228119\\
20	6.61380283241868\\
21	6.62824727262272\\
22	6.70280340932478\\
23	6.65375953833175\\
24	6.6147498245906\\
25	6.60094401860985\\
26	6.59173570521862\\
27	6.69415757306319\\
28	6.64650296108793\\
29	6.66346301398579\\
30	6.79179470973912\\
31	6.93817050869826\\
32	7.02660860594637\\
33	7.0757113865389\\
34	7.07056449986065\\
35	7.02367337220971\\
36	7.05246484572747\\
37	7.04384509069328\\
38	7.05878334951837\\
39	7.10779709143629\\
40	7.10241166601004\\
41	7.12504912297023\\
42	7.18905264719553\\
43	7.21250858371217\\
44	7.28111290681829\\
45	7.37415189480373\\
46	7.43626052502011\\
47	7.42627353224843\\
48	7.32934965844573\\
49	7.32139276524321\\
50	7.27462233192782\\
50	8.72015753555289\\
49	8.74777788730562\\
48	8.71750361220111\\
47	8.46437704506665\\
46	8.43953125401152\\
45	8.40498732036533\\
44	8.37157733339352\\
43	8.3062783007427\\
42	8.28505437612588\\
41	8.25974768685546\\
40	8.23031210017171\\
39	8.20262437945359\\
38	8.19212109978018\\
37	8.15495800595682\\
36	8.1463259378202\\
35	8.19004689134952\\
34	8.20263504461687\\
33	8.16024964510737\\
32	8.08282347637939\\
31	8.06713660752618\\
30	8.04983037001601\\
29	7.92512864886318\\
28	7.86027080355356\\
27	7.7828676380799\\
26	7.74395582056216\\
25	7.73469217974189\\
24	7.60928144624485\\
23	7.50331646391739\\
22	7.47663335587785\\
21	7.43211405048065\\
20	7.14901793971347\\
19	7.08847385241546\\
18	7.22896932503223\\
17	7.10423180021946\\
16	7.03433139248834\\
15	6.92181514451671\\
14	6.79164238989085\\
13	6.79980550284209\\
12	6.98266582052238\\
11	7.06303742628784\\
10	7.08120227933135\\
9	6.98321374449767\\
8	6.81552174276594\\
7	6.90710983925099\\
6	6.96127227912644\\
5	8.59862546225885\\
4	11.2084510148837\\
3	14.0260042354997\\
2	20.64366891462\\
1	33.7577182843052\\
}--cycle;
\addlegendentry{Tree}

\addplot[area legend, draw=black!50!blue, fill=mycolor1, dotted]
table[row sep=crcr] {%
x	y\\
1	25.3989820094337\\
2	21.1517209720294\\
3	22.5545544421895\\
4	19.258407539917\\
5	15.9880514845308\\
6	15.6728599847001\\
7	14.7196553570295\\
8	15.1632010826265\\
9	14.5799587750621\\
10	14.3099903411923\\
11	14.1872786047513\\
12	14.3545259557285\\
13	14.6240008677943\\
14	14.5299166492261\\
15	14.6590214587237\\
16	14.8202550403049\\
17	14.9193420600217\\
18	15.0103580023115\\
19	15.1946942334316\\
20	15.2962298135129\\
21	15.2071313616749\\
22	15.4022053397998\\
23	15.3612189650255\\
24	15.4406268878853\\
25	15.8836240180881\\
26	16.0248246002956\\
27	16.209151012026\\
28	16.5801968369824\\
29	16.4798898532601\\
30	16.7665124254786\\
31	17.1723240433379\\
32	17.4561459752823\\
33	17.5508264342784\\
34	17.6791203528859\\
35	18.0367212706587\\
36	18.1571003042156\\
37	18.5390760925495\\
38	18.8127071775664\\
39	19.2030270871202\\
40	19.466963232459\\
41	19.8640926449256\\
42	20.0284984609788\\
43	20.3458107184417\\
44	20.6308428942562\\
45	20.9470916379374\\
46	21.1219489359903\\
47	21.314402829764\\
48	21.6037924345937\\
49	21.7767267350278\\
50	22.0330818174521\\
50	22.8603117492627\\
49	22.722251662148\\
48	22.4636311774163\\
47	22.2024811087241\\
46	21.963305159689\\
45	21.773615062404\\
44	21.5689428487848\\
43	21.3256188995594\\
42	21.047614242682\\
41	20.7061027499245\\
40	20.4112373470349\\
39	20.303135407137\\
38	20.0461485150163\\
37	19.7542404629935\\
36	19.6154319715439\\
35	19.1332226703768\\
34	18.8583637780512\\
33	18.5997382464531\\
32	18.5233246734649\\
31	18.5318687713773\\
30	18.2085396757576\\
29	18.145360226956\\
28	17.8291198716483\\
27	17.7463351324547\\
26	17.5735170008685\\
25	17.3946988531828\\
24	17.3317514934707\\
23	17.0762541651037\\
22	16.9013221863178\\
21	16.9027962134618\\
20	16.6499333298019\\
19	16.42407261001\\
18	16.6455177454768\\
17	16.4538985267717\\
16	16.5006622545875\\
15	16.4459670770414\\
14	16.054306362714\\
13	15.9525038298868\\
12	15.5224106950296\\
11	15.5556768244509\\
10	15.7081790991428\\
9	15.6613953142805\\
8	16.4021465632191\\
7	15.8776003809914\\
6	16.9553411622523\\
5	16.9375142858909\\
4	20.1683341580436\\
3	23.3520406309439\\
2	21.8014742926857\\
1	26.1777459081948\\
}--cycle;
\addlegendentry{NB}

\addplot[area legend, draw=blue, fill=white!80!blue, dashed]
table[row sep=crcr] {%
x	y\\
1	25.6171061771513\\
2	24.7101012338156\\
3	25.1769905995514\\
4	22.3594092930103\\
5	18.428643013557\\
6	18.3432363584772\\
7	14.1321634793093\\
8	13.2666806209379\\
9	12.290775712998\\
10	11.3693373631875\\
11	10.9701353127454\\
12	10.7738447068218\\
13	10.7361368805021\\
14	10.6915937262892\\
15	10.5302795724697\\
16	10.5477174109989\\
17	10.4871563620671\\
18	10.4266132174227\\
19	10.4506442462741\\
20	10.4467788268339\\
21	10.3100360073901\\
22	10.3608114840678\\
23	10.376890509325\\
24	10.340642353941\\
25	10.373120622117\\
26	10.3528261640374\\
27	10.3274346731571\\
28	10.2562668009132\\
29	10.4338686445384\\
30	10.5444969173374\\
31	10.5454671818705\\
32	10.4478818564622\\
33	10.4495722487719\\
34	10.4017696438126\\
35	10.4143669259482\\
36	10.3962899098392\\
37	10.3871268103711\\
38	10.3775708599052\\
39	10.401148478745\\
40	10.3128826817548\\
41	10.3251268387011\\
42	10.3547684754273\\
43	10.3696279434503\\
44	10.3532609560655\\
45	10.293717023072\\
46	10.2533194389161\\
47	10.2587237531806\\
48	10.2171884283497\\
49	10.2672515135945\\
50	10.232388137478\\
50	11.0669589922294\\
49	11.0692153395835\\
48	11.0300439419415\\
47	11.0554622991906\\
46	11.0236588601278\\
45	11.0353857626413\\
44	11.0353423113271\\
43	10.9222817949663\\
42	10.9445438697159\\
41	10.9518816354868\\
40	10.9863900018071\\
39	11.0543726404056\\
38	11.0407628074584\\
37	11.0535055470931\\
36	11.1335635726459\\
35	11.2494175735953\\
34	11.2247906119306\\
33	11.1472509827705\\
32	11.193559439289\\
31	11.1033598179539\\
30	11.2530863582014\\
29	11.4604032439573\\
28	11.4296942430343\\
27	11.4700699432678\\
26	11.3554218916502\\
25	11.3945885480878\\
24	11.4420106396007\\
23	11.4429268979285\\
22	11.6004263573318\\
21	11.5842147654415\\
20	11.5963396654954\\
19	11.7040735642072\\
18	11.750494344391\\
17	11.7196832900089\\
16	11.9567555970961\\
15	11.9815960132121\\
14	12.0285367111993\\
13	12.065829291879\\
12	12.0652077231388\\
11	12.3227696706458\\
10	12.7047259853148\\
9	12.959254098719\\
8	14.0447338872017\\
7	15.1133187726181\\
6	19.7403210919528\\
5	19.4821718984117\\
4	23.0868488228573\\
3	25.8568882023696\\
2	25.5944973695822\\
1	26.2350728444954\\
}--cycle;
\addlegendentry{LD}

\addplot[area legend, draw=red, fill=white!80!red]
table[row sep=crcr] {%
x	y\\
1	35.318868957119\\
2	19.7000044130406\\
3	10.1376221311176\\
4	6.81320824759194\\
5	4.12333973878777\\
6	3.35638458020164\\
7	3.379018628305\\
8	3.57074992929982\\
9	4.1642739137047\\
10	5.29921958987186\\
11	5.60360268364831\\
12	6.10469436423505\\
13	6.75414278944396\\
14	7.48231410781185\\
15	7.86017700460439\\
16	7.96310832466095\\
17	8.31025065753874\\
18	8.4945362147363\\
19	8.56642382279114\\
20	9.19364342436208\\
21	9.33715078970357\\
22	9.45783117313074\\
23	9.8676658934417\\
24	9.62266333526735\\
25	9.85556412585923\\
26	10.049462151396\\
27	10.3209487408047\\
28	10.6559380190006\\
29	11.0809681223902\\
30	11.294449009216\\
31	11.2366732853917\\
32	11.4252734677328\\
33	11.7136014576258\\
34	12.0774824828003\\
35	12.097095637936\\
36	12.3185037537388\\
37	12.3212215192972\\
38	12.54972047739\\
39	12.8710079132757\\
40	13.2583569033782\\
41	13.1503506902067\\
42	13.1946675249705\\
43	13.4682568822572\\
44	13.6375538347217\\
45	13.9888029578767\\
46	13.9300140272813\\
47	14.0391877917628\\
48	14.309475877602\\
49	14.6242775463788\\
50	14.5089790856295\\
50	15.1431186086746\\
49	14.9235020119943\\
48	14.5912428038901\\
47	14.6160548551548\\
46	14.8964929007589\\
45	14.5251067293388\\
44	14.4897044743917\\
43	14.4432408250707\\
42	14.2257110373096\\
41	14.017106150891\\
40	13.9313429491577\\
39	13.6567118477609\\
38	13.442225816132\\
37	13.3956683138612\\
36	13.1675211902918\\
35	12.8234244824742\\
34	12.8728694921952\\
33	12.7756205753207\\
32	12.6325334425352\\
31	12.5012481968893\\
30	12.4509427467026\\
29	12.1361542363746\\
28	11.5047677986011\\
27	11.5349156620809\\
26	11.0772420491817\\
25	10.6686151120466\\
24	10.7748705994229\\
23	10.3962818434616\\
22	10.6946035578501\\
21	10.2796161595566\\
20	9.64948534913308\\
19	9.79325070829372\\
18	9.44845443821188\\
17	8.7850315990225\\
16	8.49225130655384\\
15	8.05229501269508\\
14	7.82751932505708\\
13	7.64073497832515\\
12	7.21901152278914\\
11	6.39582580808061\\
10	5.60682723737088\\
9	5.1350580175528\\
8	4.38932209121775\\
7	4.09773858262168\\
6	4.41036501353532\\
5	5.58507064170059\\
4	8.31096885470028\\
3	11.9055439331715\\
2	21.1217879002854\\
1	35.9619191419042\\
}--cycle;
\addlegendentry{kNN}

\addplot[mark=none, black, very thick, dotted, domain=0:50] {3.356};
\addlegendentry{3.4~\%}

\addplot [color=black, forget plot, dashdotted]
  table[row sep=crcr]{%
1	33.0099222909754\\
2	20.0906898929315\\
3	13.2870820712691\\
4	10.1698271065822\\
5	7.73334512186016\\
6	6.30509249677715\\
7	6.5058306858251\\
8	6.41659547216711\\
9	6.44997991631205\\
10	6.51317109887195\\
11	6.48715370985717\\
12	6.55411050690565\\
13	6.47227565875865\\
14	6.49455559425621\\
15	6.57634725822216\\
16	6.69166021945172\\
17	6.72517261186636\\
18	6.82563485614923\\
19	6.76982279234832\\
20	6.88141038606607\\
21	7.03018066155168\\
22	7.08971838260132\\
23	7.07853800112457\\
24	7.11201563541773\\
25	7.16781809917588\\
26	7.16784576289039\\
27	7.23851260557155\\
28	7.25338688232074\\
29	7.29429583142448\\
30	7.42081253987757\\
31	7.50265355811221\\
32	7.55471604116288\\
33	7.61798051582313\\
34	7.63659977223876\\
35	7.60686013177962\\
37	7.59940154832505\\
39	7.65521073544495\\
40	7.66636188309087\\
41	7.69239840491284\\
42	7.7370535116607\\
43	7.75939344222743\\
45	7.88956960758453\\
46	7.93789588951582\\
47	7.94532528865754\\
48	8.02342663532342\\
49	8.03458532627442\\
50	7.99738993374035\\
};
\addplot [color=black!80!green, forget plot, dotted]
  table[row sep=crcr]{%
1	25.7883639588143\\
2	21.4765976323576\\
3	22.9532975365667\\
5	16.4627828852108\\
6	16.3141005734762\\
7	15.2986278690104\\
8	15.7826738229228\\
9	15.1206770446713\\
10	15.0090847201675\\
11	14.8714777146011\\
12	14.938468325379\\
13	15.2882523488405\\
14	15.29211150597\\
15	15.5524942678826\\
16	15.6604586474462\\
17	15.6866202933967\\
18	15.8279378738942\\
19	15.8093834217208\\
20	15.9730815716574\\
21	16.0549637875684\\
22	16.1517637630588\\
23	16.2187365650646\\
24	16.386189190678\\
25	16.6391614356354\\
26	16.7991708005821\\
27	16.9777430722403\\
28	17.2046583543154\\
29	17.3126250401081\\
30	17.4875260506181\\
31	17.8520964073576\\
32	17.9897353243736\\
33	18.0752823403658\\
34	18.2687420654686\\
35	18.5849719705177\\
36	18.8862661378798\\
37	19.1466582777715\\
38	19.4294278462913\\
39	19.7530812471286\\
40	19.9391002897469\\
41	20.2850976974251\\
42	20.5380563518304\\
43	20.8357148090005\\
45	21.3603533501707\\
46	21.5426270478396\\
47	21.758441969244\\
48	22.033711806005\\
49	22.2494891985879\\
50	22.4466967833574\\
};
\addplot [color=blue, forget plot, dashed]
  table[row sep=crcr]{%
1	25.9260895108234\\
2	25.1522993016989\\
3	25.5169394009605\\
4	22.7231290579338\\
5	18.9554074559844\\
6	19.041778725215\\
7	14.6227411259637\\
8	13.6557072540698\\
9	12.6250149058585\\
10	12.0370316742511\\
11	11.6464524916956\\
12	11.4195262149803\\
13	11.4009830861906\\
14	11.3600652187442\\
15	11.2559377928409\\
16	11.2522365040475\\
17	11.103419826038\\
19	11.0773589052407\\
20	11.0215592461647\\
21	10.9471253864158\\
22	10.9806189206998\\
23	10.9099087036268\\
24	10.8913264967708\\
25	10.8838545851024\\
26	10.8541240278438\\
27	10.8987523082125\\
28	10.8429805219738\\
29	10.9471359442479\\
30	10.8987916377694\\
31	10.8244134999122\\
32	10.8207206478756\\
33	10.7984116157712\\
35	10.8318922497718\\
36	10.7649267412426\\
37	10.7203161787321\\
38	10.7091668336818\\
39	10.7277605595753\\
40	10.649636341781\\
41	10.6385042370939\\
42	10.6496561725716\\
43	10.6459548692083\\
44	10.6943016336963\\
46	10.6384891495219\\
47	10.6570930261856\\
48	10.6236161851456\\
49	10.668233426589\\
50	10.6496735648537\\
};
\addplot [color=red, forget plot]
  table[row sep=crcr]{%
1	35.6403940495116\\
2	20.410896156663\\
3	11.0215830321445\\
4	7.56208855114611\\
5	4.85420519024418\\
6	3.88337479686848\\
7	3.73837860546334\\
8	3.98003601025879\\
9	4.64966596562875\\
10	5.45302341362137\\
11	5.99971424586446\\
12	6.6618529435121\\
13	7.19743888388455\\
14	7.65491671643446\\
15	7.95623600864974\\
16	8.2276798156074\\
17	8.54764112828062\\
18	8.97149532647408\\
19	9.17983726554243\\
20	9.42156438674758\\
21	9.80838347463011\\
22	10.0762173654904\\
23	10.1319738684516\\
25	10.2620896189529\\
26	10.5633521002889\\
27	10.9279322014428\\
28	11.0803529088008\\
29	11.6085611793824\\
30	11.8726958779593\\
31	11.8689607411405\\
32	12.028903455134\\
33	12.2446110164732\\
34	12.4751759874977\\
35	12.4602600602051\\
36	12.7430124720153\\
37	12.8584449165792\\
38	12.995973146761\\
39	13.2638598805183\\
40	13.5948499262679\\
41	13.5837284205489\\
42	13.7101892811401\\
43	13.9557488536639\\
44	14.0636291545567\\
45	14.2569548436077\\
46	14.4132534640201\\
47	14.3276213234588\\
48	14.4503593407461\\
49	14.7738897791866\\
50	14.826048847152\\
};
\addplot [color=black, forget plot]
  table[row sep=crcr]{%
1	32.2621262976457\\
2	19.537710871243\\
3	12.5481599070386\\
4	9.13120319828074\\
5	6.86806478146147\\
6	5.64891271442786\\
7	6.1045515323992\\
8	6.01766920156828\\
9	5.91674608812642\\
10	5.94513991841254\\
11	5.91126999342652\\
12	6.12555519328891\\
13	6.1447458146752\\
14	6.19746879862159\\
15	6.23087937192761\\
16	6.3489890464151\\
17	6.34611342351325\\
18	6.42230038726623\\
19	6.45117173228119\\
20	6.61380283241868\\
21	6.62824727262272\\
22	6.70280340932478\\
24	6.6147498245906\\
26	6.59173570521862\\
27	6.69415757306319\\
28	6.64650296108793\\
29	6.66346301398578\\
30	6.79179470973912\\
31	6.93817050869826\\
32	7.02660860594637\\
33	7.07571138653891\\
34	7.07056449986065\\
35	7.02367337220971\\
36	7.05246484572746\\
37	7.04384509069328\\
38	7.05878334951837\\
39	7.10779709143629\\
40	7.10241166601004\\
41	7.12504912297023\\
42	7.18905264719553\\
43	7.21250858371217\\
44	7.28111290681829\\
45	7.37415189480372\\
46	7.43626052502012\\
47	7.42627353224843\\
48	7.32934965844574\\
49	7.32139276524322\\
50	7.27462233192782\\
nan	nan\\
50	8.72015753555289\\
49	8.74777788730562\\
48	8.71750361220111\\
47	8.46437704506665\\
45	8.40498732036533\\
44	8.37157733339352\\
43	8.3062783007427\\
41	8.25974768685546\\
39	8.2026243794536\\
38	8.19212109978018\\
37	8.15495800595682\\
36	8.14632593782019\\
35	8.19004689134952\\
34	8.20263504461688\\
33	8.16024964510737\\
32	8.08282347637939\\
30	8.04983037001601\\
29	7.92512864886318\\
28	7.86027080355356\\
27	7.78286763807991\\
26	7.74395582056216\\
25	7.73469217974189\\
24	7.60928144624486\\
23	7.50331646391739\\
22	7.47663335587785\\
21	7.43211405048065\\
20	7.14901793971347\\
19	7.08847385241545\\
18	7.22896932503222\\
17	7.10423180021946\\
16	7.03433139248834\\
15	6.92181514451671\\
14	6.79164238989085\\
13	6.79980550284209\\
12	6.98266582052239\\
11	7.06303742628783\\
10	7.08120227933136\\
9	6.98321374449767\\
8	6.81552174276595\\
7	6.90710983925099\\
6	6.96127227912644\\
5	8.59862546225885\\
4	11.2084510148837\\
3	14.0260042354997\\
2	20.6436689146201\\
1	33.7577182843052\\
};
\addplot [color=black!80!green, forget plot]
  table[row sep=crcr]{%
1	25.3989820094337\\
2	21.1517209720294\\
3	22.5545544421895\\
5	15.9880514845308\\
6	15.6728599847001\\
7	14.7196553570295\\
8	15.1632010826265\\
9	14.5799587750621\\
10	14.3099903411923\\
11	14.1872786047513\\
12	14.3545259557285\\
13	14.6240008677943\\
14	14.5299166492261\\
15	14.6590214587237\\
16	14.8202550403049\\
18	15.0103580023115\\
19	15.1946942334316\\
20	15.296229813513\\
21	15.2071313616749\\
22	15.4022053397998\\
23	15.3612189650255\\
24	15.4406268878853\\
25	15.8836240180881\\
26	16.0248246002956\\
27	16.209151012026\\
28	16.5801968369824\\
29	16.4798898532601\\
30	16.7665124254786\\
31	17.172324043338\\
32	17.4561459752823\\
33	17.5508264342784\\
34	17.6791203528859\\
35	18.0367212706587\\
36	18.1571003042156\\
37	18.5390760925495\\
38	18.8127071775664\\
39	19.2030270871202\\
40	19.466963232459\\
41	19.8640926449256\\
42	20.0284984609788\\
43	20.3458107184417\\
44	20.6308428942562\\
45	20.9470916379374\\
46	21.1219489359903\\
47	21.314402829764\\
48	21.6037924345937\\
49	21.7767267350278\\
50	22.0330818174521\\
nan	nan\\
50	22.8603117492627\\
49	22.722251662148\\
47	22.2024811087241\\
46	21.963305159689\\
45	21.773615062404\\
44	21.5689428487848\\
43	21.3256188995594\\
42	21.047614242682\\
41	20.7061027499245\\
40	20.4112373470349\\
39	20.303135407137\\
38	20.0461485150163\\
37	19.7542404629935\\
36	19.6154319715439\\
35	19.1332226703768\\
34	18.8583637780512\\
33	18.5997382464531\\
32	18.5233246734649\\
31	18.5318687713773\\
30	18.2085396757576\\
29	18.145360226956\\
28	17.8291198716483\\
27	17.7463351324547\\
25	17.3946988531828\\
24	17.3317514934707\\
23	17.0762541651037\\
22	16.9013221863178\\
21	16.9027962134618\\
20	16.6499333298019\\
19	16.42407261001\\
18	16.6455177454768\\
17	16.4538985267717\\
16	16.5006622545875\\
15	16.4459670770414\\
14	16.054306362714\\
13	15.9525038298868\\
12	15.5224106950296\\
11	15.5556768244509\\
10	15.7081790991428\\
9	15.6613953142805\\
8	16.4021465632191\\
7	15.8776003809914\\
6	16.9553411622523\\
5	16.9375142858909\\
4	20.1683341580436\\
3	23.3520406309439\\
2	21.8014742926857\\
1	26.1777459081949\\
};
\addplot [color=blue, forget plot]
  table[row sep=crcr]{%
1	25.6171061771513\\
2	24.7101012338156\\
3	25.1769905995514\\
4	22.3594092930103\\
5	18.428643013557\\
6	18.3432363584772\\
7	14.1321634793093\\
8	13.2666806209379\\
9	12.290775712998\\
10	11.3693373631875\\
11	10.9701353127454\\
12	10.7738447068218\\
14	10.6915937262892\\
15	10.5302795724697\\
16	10.5477174109989\\
18	10.4266132174227\\
19	10.4506442462741\\
20	10.4467788268339\\
21	10.3100360073901\\
22	10.3608114840678\\
23	10.376890509325\\
24	10.340642353941\\
25	10.373120622117\\
27	10.3274346731571\\
28	10.2562668009132\\
29	10.4338686445384\\
30	10.5444969173374\\
31	10.5454671818705\\
32	10.4478818564622\\
33	10.449572248772\\
34	10.4017696438126\\
35	10.4143669259482\\
37	10.3871268103711\\
38	10.3775708599052\\
39	10.401148478745\\
40	10.3128826817548\\
41	10.3251268387011\\
42	10.3547684754273\\
43	10.3696279434503\\
44	10.3532609560655\\
45	10.293717023072\\
46	10.2533194389161\\
47	10.2587237531806\\
48	10.2171884283497\\
49	10.2672515135945\\
50	10.232388137478\\
nan	nan\\
50	11.0669589922294\\
49	11.0692153395835\\
48	11.0300439419415\\
47	11.0554622991906\\
46	11.0236588601278\\
45	11.0353857626413\\
44	11.0353423113271\\
43	10.9222817949663\\
42	10.9445438697159\\
41	10.9518816354868\\
40	10.9863900018072\\
39	11.0543726404056\\
38	11.0407628074584\\
37	11.0535055470931\\
36	11.1335635726459\\
35	11.2494175735953\\
34	11.2247906119306\\
33	11.1472509827705\\
32	11.193559439289\\
31	11.1033598179539\\
30	11.2530863582014\\
29	11.4604032439573\\
28	11.4296942430343\\
27	11.4700699432678\\
26	11.3554218916502\\
24	11.4420106396007\\
23	11.4429268979285\\
22	11.6004263573318\\
21	11.5842147654415\\
20	11.5963396654954\\
19	11.7040735642072\\
18	11.750494344391\\
17	11.7196832900089\\
16	11.9567555970961\\
15	11.9815960132121\\
13	12.065829291879\\
12	12.0652077231388\\
11	12.3227696706458\\
10	12.7047259853148\\
9	12.959254098719\\
8	14.0447338872017\\
7	15.1133187726181\\
6	19.7403210919528\\
5	19.4821718984117\\
4	23.0868488228573\\
3	25.8568882023696\\
2	25.5944973695822\\
1	26.2350728444954\\
};
\addplot [color=red, forget plot]
  table[row sep=crcr]{%
1	35.318868957119\\
2	19.7000044130406\\
3	10.1376221311176\\
4	6.81320824759194\\
5	4.12333973878777\\
6	3.35638458020164\\
7	3.379018628305\\
8	3.57074992929982\\
9	4.1642739137047\\
10	5.29921958987186\\
11	5.60360268364831\\
12	6.10469436423505\\
13	6.75414278944396\\
14	7.48231410781185\\
15	7.8601770046044\\
16	7.96310832466094\\
17	8.31025065753874\\
18	8.4945362147363\\
19	8.56642382279114\\
20	9.19364342436208\\
21	9.33715078970357\\
22	9.45783117313074\\
23	9.8676658934417\\
24	9.62266333526735\\
25	9.85556412585922\\
26	10.049462151396\\
27	10.3209487408047\\
28	10.6559380190006\\
29	11.0809681223902\\
30	11.294449009216\\
31	11.2366732853917\\
32	11.4252734677328\\
33	11.7136014576258\\
34	12.0774824828003\\
35	12.097095637936\\
36	12.3185037537388\\
37	12.3212215192972\\
38	12.54972047739\\
39	12.8710079132757\\
40	13.2583569033782\\
41	13.1503506902067\\
42	13.1946675249705\\
43	13.4682568822572\\
44	13.6375538347217\\
45	13.9888029578767\\
46	13.9300140272813\\
47	14.0391877917628\\
48	14.309475877602\\
49	14.6242775463788\\
50	14.5089790856295\\
nan	nan\\
50	15.1431186086746\\
49	14.9235020119943\\
48	14.5912428038901\\
47	14.6160548551548\\
46	14.8964929007589\\
45	14.5251067293388\\
44	14.4897044743917\\
43	14.4432408250707\\
41	14.017106150891\\
40	13.9313429491577\\
39	13.6567118477609\\
38	13.442225816132\\
37	13.3956683138612\\
36	13.1675211902918\\
35	12.8234244824742\\
34	12.8728694921952\\
33	12.7756205753207\\
32	12.6325334425352\\
31	12.5012481968893\\
30	12.4509427467026\\
29	12.1361542363746\\
28	11.5047677986011\\
27	11.5349156620809\\
26	11.0772420491817\\
25	10.6686151120466\\
24	10.7748705994229\\
23	10.3962818434616\\
22	10.6946035578501\\
21	10.2796161595566\\
20	9.64948534913308\\
19	9.79325070829372\\
18	9.44845443821188\\
17	8.7850315990225\\
16	8.49225130655385\\
15	8.05229501269508\\
14	7.82751932505708\\
13	7.64073497832515\\
12	7.21901152278915\\
11	6.39582580808061\\
10	5.60682723737088\\
9	5.1350580175528\\
8	4.38932209121775\\
7	4.09773858262167\\
6	4.41036501353532\\
5	5.58507064170059\\
4	8.31096885470028\\
3	11.9055439331715\\
2	21.1217879002854\\
1	35.9619191419042\\
};
\end{axis}
\end{tikzpicture}%

%% file: Figures/balancing0_norm0_FFTtransformed_50rSVDmodes.tex
\definecolor{mycolor1}{rgb}{0.64, 0.76, 0.68}%
\begin{tikzpicture}

\begin{axis}[%
width=7cm,
height=5cm,
scale only axis,
unbounded coords=jump,
xmin=0,
xmax=50,
xlabel style={font=\color{white!15!black}},
xlabel={$r$},
ymin=0,
ymax=70,
ylabel style={font=\color{white!15!black}},
ylabel={Test error in \%},
axis background/.style={fill=white},
title style={font=\bfseries},
legend columns=2,
legend style={legend cell align=left, align=left, draw=white!15!black, font = \footnotesize},
legend entries={Tree,
                NB,
                LD,
                kNN,
                2.8~\%},
]
\addlegendimage{legend image code/.code={
\draw [draw=none, fill=white!80!black] (0cm,-0.15cm) rectangle (0.6cm,0.15cm);
\draw[dashdotted] (0cm,0cm) -- (0.6cm,0cm);
\draw[-] (0cm,0.15cm) -- (0.6cm,0.15cm);
\draw[-] (0cm,-0.15cm) -- (0.6cm,-0.15cm);
}
} 
\addlegendimage{legend image code/.code={
\draw [draw=none, fill=mycolor1] (0cm,-0.15cm) rectangle (0.6cm,0.15cm);
\draw[dotted, black!80!green] (0cm,0cm) -- (0.6cm,0cm);
\draw[-, black!80!green] (0cm,0.15cm) -- (0.6cm,0.15cm);
\draw[-, black!80!green] (0cm,-0.15cm) -- (0.6cm,-0.15cm);
}
}
\addlegendimage{legend image code/.code={
\draw [draw=none, fill=white!80!blue] (0cm,-0.15cm) rectangle (0.6cm,0.15cm);
\draw[dashed, blue] (0cm,0cm) -- (0.6cm,0cm);
\draw[-, blue] (0cm,0.15cm) -- (0.6cm,0.15cm);
\draw[-, blue] (0cm,-0.15cm) -- (0.6cm,-0.15cm);
}
}
\addlegendimage{legend image code/.code={
\draw [draw=none, fill=white!80!red] (0cm,-0.15cm) rectangle (0.6cm,0.15cm);
\draw[-, red] (0cm,0cm) -- (0.6cm,0cm);
\draw[-, red] (0cm,0.15cm) -- (0.6cm,0.15cm);
\draw[-, red] (0cm,-0.15cm) -- (0.6cm,-0.15cm);
}
}
%\addlegendimage{mark=none, orange, thick, dashed}
\addlegendimage{mark=none, black, very thick, dotted}

\addplot[area legend, draw=black, fill=white!80!black, dashdotted]
table[row sep=crcr] {%
x	y\\
1	32.2653176205512\\
2	17.8532964684552\\
3	12.2408056287417\\
4	9.41623563841375\\
5	7.26559816492054\\
6	5.88400848899825\\
7	5.84068622316195\\
8	5.75314963809708\\
9	5.58213099656179\\
10	5.62019633651235\\
11	5.66695267030981\\
12	5.97117659984233\\
13	5.93335740375575\\
14	5.97402050780451\\
15	5.94879923825127\\
16	5.97246428835041\\
17	6.1113049708149\\
18	6.14195941484675\\
19	6.1335297960797\\
20	6.16468595583544\\
21	6.15446425858792\\
22	6.18359351884024\\
23	6.27730452459262\\
24	6.30666695633937\\
25	6.29307976860633\\
26	6.2996765817743\\
27	6.36194338705992\\
28	6.40665346460722\\
29	6.45621230953474\\
30	6.5041086866459\\
31	6.47924265088407\\
32	6.57378440534831\\
33	6.66098209009066\\
34	6.69108459947119\\
35	6.6947339748358\\
36	6.67558850405993\\
37	6.73234356204454\\
38	6.72494628047895\\
39	6.72070451858107\\
40	6.76094728630891\\
41	6.8225580704983\\
42	6.89741239978608\\
43	7.01686284725347\\
44	6.99937425148352\\
45	7.03390081180594\\
46	7.09242277732984\\
47	7.14587703875906\\
48	7.21230799642535\\
49	7.22680387976835\\
50	7.30694302104934\\
50	8.23193922910968\\
49	8.20790976259494\\
48	8.0885217240728\\
47	8.0805561931555\\
46	8.0670903504769\\
45	8.11070781519347\\
44	8.1229680021143\\
43	8.12778115130227\\
42	8.09849997334392\\
41	8.06169300267325\\
40	8.04145111014155\\
39	8.08173735144742\\
38	8.05518591730211\\
37	8.12955344407876\\
36	8.03010277955879\\
35	8.03333287251847\\
34	8.03705089095514\\
33	7.99275581726382\\
32	7.99065259628762\\
31	7.92901022186128\\
30	7.90423687557456\\
29	7.89264589965725\\
28	7.85291183649545\\
27	7.81581826989545\\
26	7.72183001157924\\
25	7.78794675587584\\
24	7.78927848398085\\
23	7.71451967280156\\
22	7.73390509153567\\
21	7.67368711212826\\
20	7.76023291425019\\
19	7.73186714617926\\
18	7.59693925531041\\
17	7.55312037078379\\
16	7.40210148544412\\
15	7.37363964519574\\
14	7.12511969801469\\
13	7.03179939342797\\
12	6.97177352874856\\
11	7.04539455143798\\
10	6.89138607353659\\
9	6.79542332225676\\
8	6.66902841999913\\
7	6.50731043688273\\
6	6.37437337049488\\
5	7.74541062083036\\
4	10.1550360803122\\
3	12.6788595414542\\
2	19.6236078403228\\
1	33.5866802643059\\
}--cycle;
\addlegendentry{Tree}

\addplot[area legend, draw=black!80!green, fill=mycolor1, dotted]
table[row sep=crcr] {%
x	y\\
1	25.0826501102209\\
2	20.8182733915379\\
3	22.258218178962\\
4	19.5567708218547\\
5	16.3385148544372\\
6	15.9384258125764\\
7	15.0903596792752\\
8	15.424762293534\\
9	14.1518684639514\\
10	14.4209392414241\\
11	14.8827182389319\\
12	14.8856768497377\\
13	14.8636250900147\\
14	15.1260737879082\\
15	15.3115759747783\\
16	15.0916766850302\\
17	14.8608589042193\\
18	14.7363291821513\\
19	15.0274562439833\\
20	14.9998772069099\\
21	15.1970983422739\\
22	15.4149319212358\\
23	15.6557321066243\\
24	15.7748222449526\\
25	16.1091813521924\\
26	16.0153334855208\\
27	16.2536904415015\\
28	16.3123396855553\\
29	16.7601227384171\\
30	16.9877664358718\\
31	17.1310305504999\\
32	17.4267897310412\\
33	17.6531350450409\\
34	17.6928950598513\\
35	18.0127879844733\\
36	18.3012178059147\\
37	18.4241605809402\\
38	18.7419142929375\\
39	19.1266651812885\\
40	19.4373755857742\\
41	19.8110304829552\\
42	20.0102572838273\\
43	20.2169503765104\\
44	20.4100874411426\\
45	20.6680780327509\\
46	20.8422024998065\\
47	21.1597907134693\\
48	21.3237878933239\\
49	21.3805249051278\\
50	21.6325376123172\\
50	22.7283949564154\\
49	22.5414206720856\\
48	22.2113168325563\\
47	22.2041998334895\\
46	22.0009315468056\\
45	21.8103295456805\\
44	21.5102109188894\\
43	21.2718160253832\\
42	20.8832127445509\\
41	20.6211884026663\\
40	20.4368163532011\\
39	20.1300089539158\\
38	19.9716175257859\\
37	19.4784346913923\\
36	19.325997481529\\
35	18.9522911210457\\
34	18.6548607175061\\
33	18.33753805386\\
32	18.0579564708158\\
31	17.8627173809822\\
30	17.7902040640123\\
29	17.6013289460237\\
28	17.7588367213156\\
27	17.4306108277364\\
26	17.1332263461256\\
25	16.6524822732683\\
24	16.5776140964988\\
23	16.5405355480113\\
22	16.3349287351674\\
21	16.1508017729408\\
20	15.9313313684522\\
19	16.0079794272282\\
18	15.9344915907845\\
17	16.0703291566632\\
16	16.0626173828204\\
15	15.7681320523886\\
14	15.9314483986846\\
13	15.9407265405264\\
12	15.620468547244\\
11	15.5044036212584\\
10	15.2884771822851\\
9	15.3268439223092\\
8	16.3304738048086\\
7	15.844206813925\\
6	16.699498825581\\
5	16.8420744753101\\
4	19.8046131941855\\
3	22.867749580424\\
2	21.5390254162298\\
1	26.2999740020083\\
}--cycle;
\addlegendentry{NB}

\addplot[area legend, draw=blue, fill=white!80!blue, dashed]
table[row sep=crcr] {%
x	y\\
1	25.2222348392988\\
2	24.3305177517844\\
3	24.5831777921692\\
4	21.970420362519\\
5	18.5118964151506\\
6	18.5744954543389\\
7	14.1116280081771\\
8	13.2254324060553\\
9	12.0105512268702\\
10	11.2756571539852\\
11	11.169065472239\\
12	11.3219596704668\\
13	11.1295076953704\\
14	11.0230829091474\\
15	10.8931648215753\\
16	10.7884591300485\\
17	10.6630815596603\\
18	10.5887332747165\\
19	10.336300226329\\
20	10.311265319769\\
21	10.2862460581086\\
22	10.3137814777313\\
23	10.2697527098138\\
24	10.3764888221194\\
25	10.3466482225096\\
26	10.3815345235983\\
27	10.3223232361421\\
28	10.2820567038392\\
29	10.360930633376\\
30	10.4450000194838\\
31	10.296891696696\\
32	10.3196547893401\\
33	10.41763237503\\
34	10.3930966203781\\
35	10.3753930781208\\
36	10.3665781235427\\
37	10.3744937129501\\
38	10.3228968361005\\
39	10.3014425827525\\
40	10.3153922387687\\
41	10.3163095458351\\
42	10.3744408990628\\
43	10.3389935860452\\
44	10.3019109356742\\
45	10.1533834356693\\
46	10.163167227288\\
47	10.1986939686107\\
48	10.1273736052463\\
49	10.1974110331533\\
50	10.1392412575041\\
50	10.5769363798504\\
49	10.5857097929694\\
48	10.5664441893966\\
47	10.6216305540203\\
46	10.6348257456061\\
45	10.6966372398558\\
44	10.741461075603\\
43	10.6819944566872\\
42	10.6912455650382\\
41	10.6972851744657\\
40	10.7725324005107\\
39	10.8386373456948\\
38	10.8023017083876\\
37	10.7879327121953\\
36	10.7883663466863\\
35	10.7498489969523\\
34	10.7470301005894\\
33	10.7373856013405\\
32	10.6567802971679\\
31	10.7092498284541\\
30	10.7100806338478\\
29	10.7644079265902\\
28	10.8283274469651\\
27	10.8995979855572\\
26	10.8329865186925\\
25	10.9274234144452\\
24	10.9570913710994\\
23	10.9374329984843\\
22	10.8560898143965\\
21	11.0026327373698\\
20	10.9255332760062\\
19	11.0641420053804\\
18	11.1093700190604\\
17	11.0870120581226\\
16	11.1326148385265\\
15	11.3477843628924\\
14	11.2399270826816\\
13	11.6766742344222\\
12	11.729593365327\\
11	12.0536177730451\\
10	12.0510877120869\\
9	12.5139769364673\\
8	14.0742887699671\\
7	14.9652837978786\\
6	19.1093250942162\\
5	19.184116155042\\
4	22.8367798139562\\
3	25.5804668230785\\
2	25.3703588079262\\
1	26.4431508389564\\
}--cycle;
\addlegendentry{LD}

\addplot[area legend, draw=red, fill=white!80!red]
table[row sep=crcr] {%
x	y\\
1	35.0625689186529\\
2	17.9990422352818\\
3	10.6772695528429\\
4	7.63270203474922\\
5	4.74313097517599\\
6	3.12392785995105\\
7	2.94030873862707\\
8	2.83601419619886\\
9	2.81203661528382\\
10	2.79740735232772\\
11	2.89820953274571\\
12	2.9390528070987\\
13	2.92647317821287\\
14	2.95398319540093\\
15	3.05128392945722\\
16	3.00045164281072\\
17	2.96200760994402\\
18	2.98919545425854\\
19	3.00247647208016\\
20	3.01689386675747\\
21	3.01677513743703\\
22	3.01185622961899\\
23	2.95431821124029\\
24	2.95777613848422\\
25	2.99653039678491\\
26	2.96001792122389\\
27	3.0052969860974\\
28	3.01954438724652\\
29	2.96334846320636\\
30	2.9520303535229\\
31	3.03024838406274\\
32	3.08855886935185\\
33	3.08854646454756\\
34	3.06649180212396\\
35	2.93626018013401\\
36	2.9568217002129\\
37	2.9149170289027\\
38	2.90761849039002\\
39	2.8980293366581\\
40	2.89597168740448\\
41	2.92770685428403\\
42	2.96593378754616\\
43	2.95811619663729\\
44	2.97925569726442\\
45	2.9629550590707\\
46	3.00522174279367\\
47	3.02975732704222\\
48	3.02077127676315\\
49	2.96359269517029\\
50	2.97496551031483\\
50	3.42951022275927\\
49	3.39627039913576\\
48	3.43578859288302\\
47	3.44163188532314\\
46	3.47364240256289\\
45	3.47132998602919\\
44	3.44757917321705\\
43	3.41665774114697\\
42	3.40879301392458\\
41	3.49163227579326\\
40	3.47874349774182\\
39	3.43952143305995\\
38	3.4968626084432\\
37	3.46721262884506\\
36	3.45508274782822\\
35	3.4012175407279\\
34	3.45692858727065\\
33	3.43491856643613\\
32	3.4349287197128\\
31	3.44113772527902\\
30	3.44499575752028\\
29	3.4634225931322\\
28	3.42947589994005\\
27	3.42888142074444\\
26	3.5113365475697\\
25	3.54177610807207\\
24	3.40947282862812\\
23	3.44268712882549\\
22	3.48194792277232\\
21	3.46216771334188\\
20	3.50674098369648\\
19	3.40207733439867\\
18	3.33348094678829\\
17	3.36816440466878\\
16	3.32228775969922\\
15	3.3607711208299\\
14	3.36134975235759\\
13	3.38138518971046\\
12	3.36880522470634\\
11	3.45422652634394\\
10	3.41378806532614\\
9	3.45864423501227\\
8	3.38263551161586\\
7	3.50150482674176\\
6	3.81599517342961\\
5	5.09055833228946\\
4	8.29345105637951\\
3	10.9913953078811\\
2	20.304136698731\\
1	36.2221737765557\\
}--cycle;
\addlegendentry{kNN}

\addplot[mark=none, black, very thick, dotted, domain=0:50] {2.79};
\addlegendentry{2.8~\%}

\addplot [color=black, forget plot, dashdotted]
  table[row sep=crcr]{%
1	32.9259989424285\\
2	18.738452154389\\
3	12.459832585098\\
4	9.78563585936298\\
5	7.50550439287544\\
6	6.12919092974657\\
8	6.2110890290481\\
9	6.18877715940928\\
10	6.25579120502447\\
11	6.3561736108739\\
12	6.47147506429545\\
13	6.48257839859186\\
14	6.5495701029096\\
15	6.6612194417235\\
16	6.68728288689726\\
17	6.83221267079934\\
18	6.86944933507858\\
19	6.93269847112948\\
20	6.96245943504282\\
21	6.91407568535809\\
24	7.04797272016012\\
25	7.04051326224108\\
26	7.01075329667677\\
27	7.08888082847768\\
30	7.20417278111024\\
31	7.20412643637268\\
32	7.28221850081797\\
34	7.36406774521316\\
35	7.36403342367714\\
36	7.35284564180936\\
37	7.43094850306165\\
38	7.39006609889053\\
39	7.40122093501424\\
40	7.40119919822523\\
41	7.44212553658578\\
42	7.497956186565\\
43	7.57232199927787\\
44	7.56117112679891\\
46	7.57975656390337\\
48	7.65041486024908\\
49	7.71735682118165\\
50	7.76944112507951\\
};
\addplot [color=black!80!green, forget plot, dotted]
  table[row sep=crcr]{%
1	25.6913120561146\\
2	21.1786494038839\\
3	22.562983879693\\
4	19.6806920080201\\
5	16.5902946648737\\
6	16.3189623190787\\
7	15.4672832466001\\
8	15.8776180491713\\
9	14.7393561931303\\
10	14.8547082118546\\
11	15.1935609300951\\
12	15.2530726984908\\
13	15.4021758152705\\
14	15.5287610932964\\
15	15.5398540135835\\
16	15.5771470339253\\
17	15.4655940304412\\
18	15.3354103864679\\
19	15.5177178356058\\
20	15.465604287681\\
23	16.0981338273178\\
24	16.1762181707257\\
25	16.3808318127303\\
26	16.5742799158232\\
27	16.8421506346189\\
28	17.0355882034355\\
29	17.1807258422204\\
30	17.3889852499421\\
31	17.496873965741\\
33	17.9953365494505\\
34	18.1738778886787\\
35	18.4825395527595\\
36	18.8136076437219\\
37	18.9512976361662\\
38	19.3567659093617\\
39	19.6283370676021\\
40	19.9370959694876\\
41	20.2161094428107\\
42	20.4467350141891\\
43	20.7443832009468\\
44	20.960149180016\\
45	21.2392037892157\\
46	21.4215670233061\\
47	21.6819952734794\\
48	21.7675523629401\\
49	21.9609727886067\\
50	22.1804662843663\\
};
\addplot [color=blue, forget plot, dashed]
  table[row sep=crcr]{%
1	25.8326928391276\\
2	24.8504382798553\\
3	25.0818223076238\\
4	22.4036000882376\\
5	18.8480062850963\\
6	18.8419102742776\\
7	14.5384559030279\\
8	13.6498605880112\\
9	12.2622640816687\\
10	11.6633724330361\\
11	11.611341622642\\
12	11.5257765178969\\
13	11.4030909648963\\
14	11.1315049959145\\
15	11.1204745922338\\
16	10.9605369842875\\
17	10.8750468088914\\
18	10.8490516468884\\
19	10.7002211158547\\
20	10.6183992978876\\
21	10.6444393977392\\
22	10.5849356460639\\
23	10.603592854149\\
24	10.6667900966094\\
26	10.6072605211454\\
27	10.6109606108496\\
28	10.5551920754022\\
30	10.5775403266658\\
31	10.5030707625751\\
32	10.488217543254\\
33	10.5775089881853\\
35	10.5626210375366\\
36	10.5774722351145\\
37	10.5812132125727\\
38	10.562599272244\\
39	10.5700399642236\\
40	10.5439623196397\\
41	10.5067973601504\\
42	10.5328432320505\\
43	10.5104940213662\\
44	10.5216860056386\\
45	10.4250103377625\\
46	10.3989964864471\\
47	10.4101622613155\\
48	10.3469088973215\\
49	10.3915604130614\\
50	10.3580888186773\\
};
\addplot [color=red, forget plot]
  table[row sep=crcr]{%
1	35.6423713476043\\
2	19.1515894670064\\
3	10.834332430362\\
4	7.96307654556437\\
5	4.91684465373272\\
6	3.46996151669033\\
7	3.22090678268442\\
8	3.10932485390736\\
9	3.13534042514804\\
10	3.10559770882693\\
11	3.17621802954483\\
12	3.15392901590252\\
14	3.15766647387926\\
15	3.20602752514356\\
16	3.16136970125497\\
18	3.16133820052342\\
19	3.20227690323942\\
20	3.26181742522697\\
21	3.23947142538945\\
22	3.24690207619565\\
23	3.19850267003289\\
24	3.18362448355617\\
25	3.2691532524285\\
26	3.2356772343968\\
27	3.21708920342092\\
28	3.22451014359329\\
30	3.19851305552159\\
31	3.23569305467088\\
32	3.26174379453232\\
34	3.2617101946973\\
35	3.16873886043096\\
36	3.20595222402056\\
37	3.19106482887388\\
38	3.20224054941662\\
39	3.16877538485902\\
41	3.20966956503865\\
42	3.18736340073537\\
43	3.18738696889213\\
44	3.21341743524074\\
45	3.21714252254995\\
46	3.23943207267828\\
48	3.22827993482309\\
49	3.17993154715302\\
50	3.20223786653705\\
};
\addplot [color=black, forget plot]
  table[row sep=crcr]{%
1	32.2653176205512\\
2	17.8532964684552\\
3	12.2408056287417\\
4	9.41623563841375\\
5	7.26559816492054\\
6	5.88400848899825\\
7	5.84068622316195\\
8	5.75314963809708\\
9	5.58213099656179\\
11	5.66695267030981\\
12	5.97117659984233\\
13	5.93335740375575\\
14	5.9740205078045\\
15	5.94879923825127\\
16	5.97246428835041\\
17	6.1113049708149\\
18	6.14195941484675\\
19	6.1335297960797\\
20	6.16468595583544\\
21	6.15446425858792\\
22	6.18359351884024\\
23	6.27730452459262\\
24	6.30666695633938\\
25	6.29307976860633\\
26	6.2996765817743\\
27	6.36194338705992\\
30	6.50410868664591\\
31	6.47924265088407\\
33	6.66098209009066\\
34	6.69108459947119\\
35	6.69473397483581\\
36	6.67558850405993\\
37	6.73234356204453\\
39	6.72070451858107\\
40	6.76094728630891\\
41	6.8225580704983\\
42	6.89741239978608\\
43	7.01686284725347\\
44	6.99937425148352\\
45	7.03390081180594\\
48	7.21230799642535\\
49	7.22680387976835\\
50	7.30694302104934\\
nan	nan\\
50	8.23193922910968\\
49	8.20790976259494\\
48	8.08852172407281\\
46	8.0670903504769\\
45	8.11070781519347\\
43	8.12778115130227\\
40	8.04145111014154\\
39	8.08173735144742\\
38	8.05518591730211\\
37	8.12955344407877\\
36	8.03010277955879\\
34	8.03705089095514\\
33	7.99275581726382\\
32	7.99065259628762\\
31	7.92901022186128\\
30	7.90423687557457\\
29	7.89264589965725\\
27	7.81581826989544\\
26	7.72183001157924\\
25	7.78794675587584\\
24	7.78927848398086\\
23	7.71451967280156\\
22	7.73390509153567\\
21	7.67368711212826\\
20	7.7602329142502\\
19	7.73186714617926\\
18	7.59693925531041\\
17	7.55312037078379\\
16	7.40210148544411\\
15	7.37363964519574\\
14	7.12511969801469\\
13	7.03179939342797\\
12	6.97177352874856\\
11	7.04539455143798\\
10	6.89138607353659\\
9	6.79542332225677\\
8	6.66902841999913\\
7	6.50731043688273\\
6	6.37437337049489\\
5	7.74541062083036\\
4	10.1550360803122\\
3	12.6788595414542\\
2	19.6236078403228\\
1	33.5866802643059\\
};
\addplot [color=black!80!green, forget plot]
  table[row sep=crcr]{%
1	25.0826501102209\\
2	20.8182733915379\\
3	22.258218178962\\
4	19.5567708218547\\
5	16.3385148544372\\
6	15.9384258125764\\
7	15.0903596792752\\
8	15.4247622935341\\
9	14.1518684639514\\
10	14.4209392414241\\
11	14.8827182389319\\
12	14.8856768497377\\
13	14.8636250900147\\
14	15.1260737879082\\
15	15.3115759747783\\
16	15.0916766850302\\
17	14.8608589042193\\
18	14.7363291821513\\
19	15.0274562439833\\
20	14.9998772069099\\
21	15.1970983422739\\
22	15.4149319212358\\
23	15.6557321066243\\
24	15.7748222449526\\
25	16.1091813521924\\
26	16.0153334855208\\
27	16.2536904415015\\
28	16.3123396855553\\
29	16.7601227384171\\
30	16.9877664358718\\
31	17.1310305504999\\
32	17.4267897310412\\
33	17.6531350450409\\
34	17.6928950598513\\
35	18.0127879844733\\
36	18.3012178059147\\
37	18.4241605809402\\
38	18.7419142929375\\
39	19.1266651812885\\
40	19.4373755857742\\
41	19.8110304829552\\
44	20.4100874411426\\
45	20.6680780327509\\
46	20.8422024998065\\
47	21.1597907134693\\
48	21.3237878933239\\
49	21.3805249051278\\
50	21.6325376123172\\
nan	nan\\
50	22.7283949564154\\
49	22.5414206720856\\
48	22.2113168325563\\
47	22.2041998334895\\
46	22.0009315468056\\
45	21.8103295456805\\
44	21.5102109188894\\
43	21.2718160253832\\
42	20.8832127445509\\
41	20.6211884026663\\
40	20.4368163532011\\
39	20.1300089539158\\
38	19.9716175257859\\
37	19.4784346913923\\
36	19.325997481529\\
35	18.9522911210457\\
34	18.6548607175061\\
33	18.33753805386\\
32	18.0579564708158\\
31	17.8627173809822\\
30	17.7902040640123\\
29	17.6013289460237\\
28	17.7588367213156\\
27	17.4306108277364\\
26	17.1332263461256\\
25	16.6524822732683\\
24	16.5776140964988\\
23	16.5405355480113\\
22	16.3349287351674\\
21	16.1508017729408\\
20	15.9313313684521\\
19	16.0079794272282\\
18	15.9344915907845\\
17	16.0703291566632\\
16	16.0626173828204\\
15	15.7681320523886\\
14	15.9314483986847\\
13	15.9407265405264\\
12	15.620468547244\\
11	15.5044036212584\\
10	15.2884771822851\\
9	15.3268439223092\\
8	16.3304738048086\\
7	15.844206813925\\
6	16.699498825581\\
5	16.8420744753101\\
4	19.8046131941855\\
3	22.867749580424\\
2	21.5390254162298\\
1	26.2999740020083\\
};
\addplot [color=blue, forget plot]
  table[row sep=crcr]{%
1	25.2222348392988\\
2	24.3305177517844\\
3	24.5831777921692\\
4	21.970420362519\\
5	18.5118964151506\\
6	18.5744954543389\\
7	14.1116280081771\\
8	13.2254324060553\\
9	12.0105512268702\\
10	11.2756571539852\\
11	11.169065472239\\
12	11.3219596704668\\
13	11.1295076953704\\
14	11.0230829091474\\
15	10.8931648215753\\
16	10.7884591300485\\
17	10.6630815596603\\
18	10.5887332747165\\
19	10.336300226329\\
21	10.2862460581086\\
22	10.3137814777313\\
23	10.2697527098138\\
24	10.3764888221194\\
25	10.3466482225096\\
26	10.3815345235983\\
27	10.3223232361421\\
28	10.2820567038393\\
30	10.4450000194838\\
31	10.296891696696\\
32	10.3196547893401\\
33	10.41763237503\\
35	10.3753930781209\\
36	10.3665781235427\\
37	10.3744937129501\\
38	10.3228968361005\\
39	10.3014425827525\\
40	10.3153922387687\\
41	10.3163095458351\\
42	10.3744408990628\\
44	10.3019109356742\\
45	10.1533834356693\\
46	10.163167227288\\
47	10.1986939686107\\
48	10.1273736052463\\
49	10.1974110331533\\
50	10.1392412575041\\
nan	nan\\
50	10.5769363798504\\
49	10.5857097929694\\
48	10.5664441893966\\
47	10.6216305540203\\
46	10.6348257456061\\
45	10.6966372398558\\
44	10.741461075603\\
43	10.6819944566872\\
41	10.6972851744657\\
39	10.8386373456948\\
38	10.8023017083876\\
37	10.7879327121953\\
36	10.7883663466863\\
35	10.7498489969523\\
33	10.7373856013405\\
32	10.6567802971679\\
31	10.7092498284541\\
30	10.7100806338478\\
28	10.8283274469651\\
27	10.8995979855572\\
26	10.8329865186925\\
25	10.9274234144452\\
24	10.9570913710994\\
23	10.9374329984843\\
22	10.8560898143965\\
21	11.0026327373698\\
20	10.9255332760062\\
19	11.0641420053804\\
18	11.1093700190604\\
17	11.0870120581226\\
16	11.1326148385265\\
15	11.3477843628924\\
14	11.2399270826816\\
13	11.6766742344222\\
12	11.729593365327\\
11	12.0536177730451\\
10	12.0510877120869\\
9	12.5139769364673\\
8	14.0742887699671\\
7	14.9652837978786\\
6	19.1093250942162\\
5	19.184116155042\\
4	22.8367798139562\\
3	25.5804668230785\\
2	25.3703588079262\\
1	26.4431508389564\\
};
\addplot [color=red, forget plot]
  table[row sep=crcr]{%
1	35.0625689186529\\
2	17.9990422352818\\
3	10.6772695528429\\
4	7.63270203474922\\
5	4.74313097517599\\
6	3.12392785995105\\
7	2.94030873862707\\
8	2.83601419619886\\
10	2.79740735232772\\
11	2.89820953274571\\
12	2.9390528070987\\
13	2.92647317821287\\
14	2.95398319540093\\
15	3.05128392945722\\
16	3.00045164281072\\
17	2.96200760994402\\
18	2.98919545425854\\
20	3.01689386675746\\
22	3.01185622961899\\
23	2.95431821124029\\
24	2.95777613848422\\
25	2.99653039678491\\
26	2.96001792122389\\
27	3.0052969860974\\
28	3.01954438724653\\
29	2.96334846320637\\
30	2.9520303535229\\
31	3.03024838406274\\
32	3.08855886935185\\
33	3.08854646454756\\
34	3.06649180212396\\
35	2.93626018013401\\
36	2.95682170021291\\
37	2.9149170289027\\
40	2.89597168740448\\
42	2.96593378754616\\
43	2.95811619663729\\
44	2.97925569726441\\
45	2.9629550590707\\
46	3.00522174279367\\
47	3.02975732704222\\
48	3.02077127676316\\
49	2.96359269517029\\
50	2.97496551031482\\
nan	nan\\
50	3.42951022275927\\
49	3.39627039913576\\
48	3.43578859288302\\
47	3.44163188532314\\
46	3.47364240256289\\
45	3.47132998602919\\
42	3.40879301392457\\
41	3.49163227579326\\
40	3.47874349774182\\
39	3.43952143305994\\
38	3.4968626084432\\
37	3.46721262884506\\
36	3.45508274782822\\
35	3.4012175407279\\
34	3.45692858727065\\
33	3.43491856643613\\
30	3.44499575752028\\
29	3.4634225931322\\
28	3.42947589994005\\
27	3.42888142074444\\
26	3.51133654756971\\
25	3.54177610807208\\
24	3.40947282862812\\
22	3.48194792277232\\
21	3.46216771334188\\
20	3.50674098369648\\
19	3.40207733439867\\
18	3.33348094678828\\
17	3.36816440466878\\
16	3.32228775969922\\
15	3.3607711208299\\
14	3.36134975235758\\
13	3.38138518971046\\
12	3.36880522470634\\
11	3.45422652634394\\
10	3.41378806532614\\
9	3.45864423501227\\
8	3.38263551161586\\
7	3.50150482674177\\
6	3.81599517342961\\
5	5.09055833228946\\
4	8.29345105637951\\
3	10.9913953078811\\
2	20.304136698731\\
1	36.2221737765557\\
};
\end{axis}
\end{tikzpicture}%

%% file: Figures/balancing1_norm0_FFTtransformed_50rSVDmodes.tex
\definecolor{mycolor1}{rgb}{0.64, 0.76, 0.68}%
\begin{tikzpicture}

\begin{axis}[%
width=7cm,
height=5cm,
scale only axis,
unbounded coords=jump,
xmin=0,
xmax=50,
xlabel style={font=\color{white!15!black}},
xlabel={$r$},
ymin=0,
ymax=70,
ylabel style={font=\color{white!15!black}},
axis background/.style={fill=white},
title style={font=\bfseries},
legend columns = 2,
legend style={legend cell align=left, align=left, draw=white!15!black, font = \footnotesize},
legend entries={Tree,
                NB,
                LD,
                kNN,
                4.6~\%},
]
\addlegendimage{legend image code/.code={
\draw [draw=none, fill=white!80!black] (0cm,-0.15cm) rectangle (0.6cm,0.15cm);
\draw[dashdotted] (0cm,0cm) -- (0.6cm,0cm);
\draw[-] (0cm,0.15cm) -- (0.6cm,0.15cm);
\draw[-] (0cm,-0.15cm) -- (0.6cm,-0.15cm);
}
} 
\addlegendimage{legend image code/.code={
\draw [draw=none, fill=mycolor1] (0cm,-0.15cm) rectangle (0.6cm,0.15cm);
\draw[dotted, black!80!green] (0cm,0cm) -- (0.6cm,0cm);
\draw[-, black!80!green] (0cm,0.15cm) -- (0.6cm,0.15cm);
\draw[-, black!80!green] (0cm,-0.15cm) -- (0.6cm,-0.15cm);
}
}
\addlegendimage{legend image code/.code={
\draw [draw=none, fill=white!80!blue] (0cm,-0.15cm) rectangle (0.6cm,0.15cm);
\draw[dashed, blue] (0cm,0cm) -- (0.6cm,0cm);
\draw[-, blue] (0cm,0.15cm) -- (0.6cm,0.15cm);
\draw[-, blue] (0cm,-0.15cm) -- (0.6cm,-0.15cm);
}
}
\addlegendimage{legend image code/.code={
\draw [draw=none, fill=white!80!red] (0cm,-0.15cm) rectangle (0.6cm,0.15cm);
\draw[-, red] (0cm,0cm) -- (0.6cm,0cm);
\draw[-, red] (0cm,0.15cm) -- (0.6cm,0.15cm);
\draw[-, red] (0cm,-0.15cm) -- (0.6cm,-0.15cm);
}
}
\addlegendimage{mark=none, black, very thick, dotted}

\addplot[area legend, draw=black, fill=white!80!black, dashdotted]
table[row sep=crcr] {%
x	y\\
1	45.6121235646344\\
2	29.3663656409692\\
3	18.0969537261858\\
4	14.3037016286169\\
5	11.8865479708515\\
6	9.20781737716161\\
7	9.5257545347663\\
8	9.15926128236174\\
9	9.24078959893916\\
10	9.41784277840579\\
11	9.26767952798853\\
12	9.29076098272682\\
13	9.59179217028558\\
14	9.93214563897704\\
15	10.0972787258815\\
16	10.3934850915136\\
17	10.3446621360845\\
18	10.7854786825686\\
19	10.5365152565174\\
20	10.6220783957659\\
21	10.5923208206695\\
22	10.7256054420106\\
23	10.7381349589018\\
24	10.6519628893075\\
25	10.7323107641147\\
26	10.8117543658556\\
27	10.873830701117\\
28	10.8424762656381\\
29	11.249071941532\\
30	11.3741530734922\\
31	11.4708707249062\\
32	11.5277568423158\\
33	11.5057933405931\\
34	11.5374374771172\\
35	11.7497955803524\\
36	11.779375849708\\
37	11.9303931760633\\
38	11.8609658036374\\
39	11.973773800107\\
40	11.8712537717228\\
41	11.8469607122408\\
42	11.9856253790174\\
43	12.0247164335203\\
44	12.0579560230596\\
45	12.1699551852448\\
46	12.2469296921871\\
47	12.2702496243629\\
48	12.2266734174163\\
49	12.1717549507656\\
50	12.162271773121\\
50	12.99901854946\\
49	12.9357719309553\\
48	12.9166957582113\\
47	12.926882992125\\
46	12.9322817773474\\
45	12.9196505495226\\
44	12.8703593891272\\
43	12.7781509499926\\
42	12.709715122775\\
41	12.8842220834585\\
40	12.7524021422561\\
39	12.5781975260583\\
38	12.5297151999472\\
37	12.6215781501019\\
36	12.6650685947368\\
35	12.2645413372107\\
34	12.3693725587255\\
33	12.2755686665757\\
32	12.4865800752473\\
31	12.3104912822626\\
30	12.2279974641425\\
29	12.3530785961028\\
28	12.3833301859752\\
27	12.1369219870554\\
26	12.0556291466896\\
25	12.0096247197567\\
24	12.1974994762843\\
23	12.2009331414569\\
22	12.1238569235812\\
21	12.113772369295\\
20	12.0660936472452\\
19	11.900760729146\\
18	11.4725858335608\\
17	11.6625063226972\\
16	11.5778410733613\\
15	11.4080976182048\\
14	11.3223346477616\\
13	11.2325805895714\\
12	11.6052963649438\\
11	11.4312452031946\\
10	11.5678203040318\\
9	11.4939774261508\\
8	11.1453982158465\\
7	11.1552490494634\\
6	10.2007847733763\\
5	12.5220541796865\\
4	15.8755098409176\\
3	19.1969530838506\\
2	31.3504802371679\\
1	47.3986291235391\\
}--cycle;
\addlegendentry{Tree}

\addplot[area legend, draw=black!80!green, fill=mycolor1, dotted]
table[row sep=crcr] {%
x	y\\
1	40.215814100108\\
2	37.3595413632489\\
3	34.2511020902393\\
4	28.5857648318247\\
5	25.5559578641439\\
6	21.8419685465166\\
7	20.9176862716395\\
8	20.253852437667\\
9	20.4690433598842\\
10	19.8084205752076\\
11	19.9137322160906\\
12	20.0508473419412\\
13	20.4468644043889\\
14	20.3538178744169\\
15	21.0145262778478\\
16	21.6643801021343\\
17	23.0480264576396\\
18	23.4674374750865\\
19	23.5046558809273\\
20	23.7379098280223\\
21	23.9599287041056\\
22	23.5181752401485\\
23	23.5675615766845\\
24	23.9786509190683\\
25	24.0029341437137\\
26	24.2256423607837\\
27	24.4521948481193\\
28	24.4183306271592\\
29	24.9551971326167\\
30	25.2417378369844\\
31	25.3554988652129\\
32	25.509778461739\\
33	25.5989208043244\\
34	25.9276500870174\\
35	26.1143294178904\\
36	26.4327140411446\\
37	26.5131825229104\\
38	26.4169444754308\\
39	26.4166236218619\\
40	26.4949811568515\\
41	26.5871099527687\\
42	26.848670677602\\
43	26.924912369981\\
44	26.7983493658567\\
45	26.5182365193067\\
46	26.1271056561803\\
47	26.0344177334534\\
48	26.3410170717716\\
49	26.2573767025905\\
50	26.2169978444291\\
50	30.3062996466108\\
49	30.3017630823562\\
48	30.3077284479421\\
47	30.022929936798\\
46	29.9840054549312\\
45	30.1484301473604\\
44	30.0116864764376\\
43	29.86720232536\\
42	29.7642325482049\\
41	29.7211337748303\\
40	29.4906819255863\\
39	29.5690394605759\\
38	29.4253494313797\\
37	29.4007959717137\\
36	28.9794723387841\\
35	28.5810110839021\\
34	28.6243212391479\\
33	28.7559179053535\\
32	28.6837699253582\\
31	28.6946803462571\\
30	28.2887281128368\\
29	28.1810035842296\\
28	28.2339991219451\\
27	28.0030022844975\\
26	27.8173683919048\\
25	27.7533382577204\\
24	27.831384923226\\
23	27.7944455917746\\
22	27.8975953691709\\
21	27.509605346074\\
20	27.6599396343436\\
19	27.6243763771375\\
18	27.159802668283\\
17	26.4681025746189\\
16	25.8804227652496\\
15	24.7202407472421\\
14	23.7322036309597\\
13	22.6893363124573\\
12	23.0136687870913\\
11	22.1113573896445\\
10	21.6969557688787\\
9	21.9323903318724\\
8	21.7174737272079\\
7	22.9712026172497\\
6	23.2834794821575\\
5	26.2540779781504\\
4	30.2314394692511\\
3	35.6772133219478\\
2	39.5221790668595\\
1	42.221461885556\\
}--cycle;
\addlegendentry{NB}

\addplot[area legend, draw=blue, fill=white!80!blue, dashed]
table[row sep=crcr] {%
x	y\\
1	40.7092304002609\\
2	39.5690500389897\\
3	39.1612744164465\\
4	33.8719019466531\\
5	25.8427013459557\\
6	19.0802462634564\\
7	18.1622881155458\\
8	15.9785255230203\\
9	15.2697285410979\\
10	15.2443843467688\\
11	15.5247774348704\\
12	15.5008174234103\\
13	15.5759345402702\\
14	15.4151388508905\\
15	15.3293070016156\\
16	15.1684789936105\\
17	15.1707998457621\\
18	15.2059293135845\\
19	14.9741107194134\\
20	14.9432193126713\\
21	14.8526808875146\\
22	14.8134914150696\\
23	14.7488024824106\\
24	14.8565374951916\\
25	14.7724919083401\\
26	14.8487528269361\\
27	14.7900576748882\\
28	14.7354964858016\\
29	14.7226921284337\\
30	14.8785757555993\\
31	14.7896884576763\\
32	14.5468141831197\\
33	14.6093030902813\\
34	14.6681630054751\\
35	14.7257434251771\\
36	14.7398168055745\\
37	14.7226704751348\\
38	14.7415462040658\\
39	14.8427990986455\\
40	14.7603650275137\\
41	14.8403941831638\\
42	14.9497182983314\\
43	14.9023315484233\\
44	14.8182099589434\\
45	14.8341987052252\\
46	14.7714762847998\\
47	14.6907284647048\\
48	14.7217474214869\\
49	14.7733847449668\\
50	14.695222619915\\
50	15.3764619678989\\
49	15.3162209898005\\
48	15.3140948724202\\
47	15.4526407109227\\
46	15.4614986255947\\
45	15.4883819399364\\
44	15.3968438044977\\
43	15.3485645089247\\
42	15.3728623468302\\
41	15.4821864619978\\
40	15.5084521767876\\
39	15.4080969587024\\
38	15.4555864124219\\
37	15.6178313169801\\
36	15.6006849865405\\
35	15.6685218077981\\
34	15.6185753457796\\
33	15.8208044366007\\
32	15.7040818742282\\
31	15.4791287466251\\
30	15.8024278286303\\
29	15.8328634271222\\
28	15.8200590697542\\
27	15.7117344398074\\
26	15.8860141981538\\
25	15.8905905289361\\
24	15.7169392073176\\
23	15.6813050444714\\
22	15.8495910222067\\
21	16.0254553132025\\
20	15.9707591819526\\
19	16.0653158038844\\
18	15.9410240914337\\
17	16.0478381470697\\
16	16.1218435870349\\
15	16.1760693424707\\
14	16.32321240359\\
13	16.5208396532784\\
12	16.6138779170917\\
11	16.5899179056317\\
10	16.4222823198982\\
9	16.8987302402644\\
8	17.121832899919\\
7	19.2749878701175\\
6	19.9340906541067\\
5	27.0246821665895\\
4	34.9453023544229\\
3	40.3727757627659\\
2	41.3807707495417\\
1	42.4986549043995\\
}--cycle;
\addlegendentry{LD}

\addplot[area legend, draw=red, fill=white!80!red]
table[row sep=crcr] {%
x	y\\
1	46.5607072317721\\
2	29.3155165981336\\
3	14.1774482060498\\
4	10.5100063437428\\
5	7.40459094006522\\
6	5.17597957190867\\
7	4.92621341820067\\
8	4.71587556510874\\
9	4.5793593468322\\
10	4.66900006938198\\
11	4.60765062301117\\
12	4.61001148844946\\
13	4.60597489729934\\
14	4.68799457747859\\
15	4.80468965642401\\
16	4.76059961961239\\
17	4.86348222166966\\
18	4.87991259631733\\
19	4.8789493182249\\
20	4.84379196468013\\
21	4.81791659430391\\
22	4.7541862299131\\
23	4.64244557096129\\
24	4.71685598426043\\
25	4.7487841938355\\
26	4.69273958367637\\
27	4.77529740483149\\
28	4.71033638216412\\
29	4.70056859787849\\
30	4.7364108917853\\
31	4.76646567741219\\
32	4.90872196023135\\
33	4.84785701060432\\
34	4.73604378933118\\
35	4.78478035687986\\
36	4.73386950836906\\
37	4.67658052803085\\
38	4.67833086186856\\
39	4.61375750869187\\
40	4.72205634254294\\
41	4.72312007888914\\
42	4.77023510026961\\
43	4.75542394150621\\
44	4.77477565616228\\
45	4.79332672754979\\
46	4.77529740483149\\
47	4.83319419240069\\
48	4.8181753638199\\
49	4.81911622546432\\
50	4.74248140365177\\
50	5.20375515548811\\
49	5.18088377453577\\
48	5.23558807704041\\
47	5.20264810150622\\
46	5.24262374212201\\
45	5.27835786026393\\
44	5.20730319688441\\
43	5.24457605849389\\
42	5.22976489973048\\
41	5.16935303939053\\
40	5.20625906964353\\
39	5.17118872786736\\
38	5.24998455031791\\
37	5.25173488415563\\
36	5.28405163858444\\
35	5.26898308398045\\
34	5.2101927698087\\
33	5.27759101806962\\
32	5.2704895093028\\
31	5.26937661649472\\
30	5.24566796126139\\
29	5.20982566735458\\
28	5.28966361783598\\
27	5.24262374212201\\
26	5.2355758285101\\
25	5.12576777749075\\
24	5.19353828097264\\
23	5.19626410645816\\
22	5.12036574141315\\
21	5.12831996483597\\
20	5.20997147618018\\
19	5.12105068177519\\
18	5.08424510977596\\
17	5.20820236614406\\
16	5.18563693952749\\
15	5.17738919662268\\
14	5.25824198166129\\
13	5.10728675144627\\
12	5.10325016029616\\
11	5.08768987878103\\
10	5.18763075499087\\
9	5.13390230191342\\
8	5.05114952449709\\
7	5.25299805133348\\
6	5.57670860013444\\
5	8.63483558323243\\
4	12.0168753766875\\
3	15.1236270627677\\
2	30.8995371653078\\
1	48.9410848829244\\
}--cycle;
\addlegendentry{kNN}

\addplot[mark=none, black, very thick, dotted, domain=0:50] {4.579};
\addlegendentry{4.6~\%}

\addplot [color=black, forget plot, dashdotted]
  table[row sep=crcr]{%
1	46.5053763440867\\
2	30.3584229390686\\
3	18.6469534050182\\
4	15.0896057347673\\
5	12.204301075269\\
6	9.70430107526897\\
7	10.3405017921149\\
8	10.1523297491041\\
9	10.367383512545\\
10	10.4928315412188\\
11	10.3494623655916\\
12	10.4480286738353\\
13	10.4121863799285\\
14	10.6272401433693\\
15	10.7526881720432\\
16	10.9856630824374\\
17	11.0035842293909\\
18	11.1290322580647\\
19	11.2186379928317\\
20	11.3440860215056\\
21	11.3530465949823\\
22	11.4247311827959\\
23	11.4695340501794\\
25	11.3709677419357\\
27	11.5053763440862\\
28	11.6129032258066\\
29	11.8010752688174\\
30	11.8010752688174\\
31	11.8906810035844\\
32	12.0071684587815\\
33	11.8906810035844\\
35	12.0071684587815\\
36	12.2222222222224\\
37	12.2759856630826\\
38	12.1953405017923\\
39	12.2759856630826\\
40	12.3118279569894\\
41	12.3655913978497\\
42	12.3476702508962\\
44	12.4641577060934\\
45	12.5448028673837\\
46	12.5896057347672\\
47	12.5985663082439\\
49	12.5537634408604\\
50	12.5806451612905\\
};
\addplot [color=black!80!green, forget plot, dotted]
  table[row sep=crcr]{%
1	41.218637992832\\
2	38.4408602150542\\
3	34.9641577060935\\
4	29.4086021505379\\
5	25.9050179211471\\
6	22.5627240143371\\
7	21.9444444444446\\
8	20.9856630824374\\
9	21.2007168458783\\
10	20.7526881720432\\
11	21.0125448028675\\
12	21.5322580645163\\
13	21.5681003584231\\
14	22.0430107526883\\
15	22.867383512545\\
16	23.7724014336919\\
17	24.7580645161292\\
18	25.3136200716848\\
19	25.5645161290324\\
20	25.698924731183\\
21	25.7347670250898\\
23	25.6810035842296\\
24	25.9050179211471\\
25	25.878136200717\\
26	26.0215053763443\\
27	26.2275985663084\\
28	26.3261648745522\\
29	26.5681003584231\\
30	26.7652329749106\\
31	27.025089605735\\
33	27.1774193548389\\
34	27.2759856630827\\
35	27.3476702508963\\
36	27.7060931899644\\
37	27.9569892473121\\
38	27.9211469534052\\
39	27.9928315412189\\
40	27.9928315412189\\
42	28.3064516129035\\
43	28.3960573476705\\
44	28.4050179211472\\
45	28.3333333333336\\
46	28.0555555555558\\
47	28.0286738351257\\
48	28.3243727598569\\
49	28.2795698924733\\
50	28.2616487455199\\
};
\addplot [color=blue, forget plot, dashed]
  table[row sep=crcr]{%
1	41.6039426523302\\
2	40.4749103942657\\
3	39.7670250896062\\
4	34.408602150538\\
5	26.4336917562726\\
6	19.5071684587815\\
7	18.7186379928317\\
8	16.5501792114697\\
9	16.0842293906811\\
10	15.8333333333335\\
11	16.057347670251\\
13	16.0483870967743\\
14	15.8691756272403\\
16	15.6451612903227\\
18	15.5734767025091\\
20	15.456989247312\\
21	15.4390681003586\\
23	15.215053763441\\
24	15.2867383512546\\
26	15.3673835125449\\
27	15.2508960573478\\
28	15.2777777777779\\
29	15.2777777777779\\
30	15.3405017921148\\
31	15.1344086021507\\
32	15.125448028674\\
33	15.215053763441\\
34	15.1433691756274\\
35	15.1971326164876\\
36	15.1702508960575\\
37	15.1702508960575\\
38	15.0985663082439\\
39	15.125448028674\\
40	15.1344086021507\\
41	15.1612903225808\\
42	15.1612903225808\\
43	15.125448028674\\
44	15.1075268817206\\
45	15.1612903225808\\
48	15.0179211469535\\
49	15.0448028673837\\
50	15.0358422939069\\
};
\addplot [color=red, forget plot]
  table[row sep=crcr]{%
1	47.7508960573482\\
2	30.1075268817207\\
3	14.6505376344087\\
4	11.2634408602152\\
5	8.01971326164882\\
6	5.37634408602156\\
7	5.08960573476708\\
8	4.88351254480291\\
9	4.85663082437281\\
10	4.92831541218643\\
11	4.8476702508961\\
13	4.85663082437281\\
14	4.97311827956994\\
15	4.99103942652334\\
16	4.97311827956994\\
17	5.03584229390686\\
18	4.98207885304664\\
20	5.02688172043015\\
21	4.97311827956994\\
22	4.93727598566313\\
23	4.91935483870972\\
24	4.95519713261653\\
25	4.93727598566313\\
26	4.96415770609324\\
27	5.00896057347675\\
28	5.00000000000005\\
29	4.95519713261653\\
31	5.01792114695346\\
32	5.08960573476708\\
33	5.06272401433696\\
34	4.97311827956994\\
35	5.02688172043015\\
36	5.00896057347675\\
37	4.96415770609324\\
38	4.96415770609324\\
39	4.89247311827962\\
40	4.96415770609324\\
41	4.94623655913983\\
42	5.00000000000005\\
44	4.99103942652334\\
45	5.03584229390686\\
46	5.00896057347676\\
48	5.02688172043015\\
50	4.97311827956994\\
};
\addplot [color=black, forget plot]
  table[row sep=crcr]{%
1	45.6121235646344\\
2	29.3663656409692\\
3	18.0969537261858\\
4	14.3037016286169\\
5	11.8865479708515\\
6	9.20781737716161\\
7	9.5257545347663\\
8	9.15926128236175\\
9	9.24078959893916\\
10	9.41784277840579\\
11	9.26767952798853\\
12	9.29076098272682\\
13	9.59179217028558\\
14	9.93214563897705\\
15	10.0972787258815\\
16	10.3934850915136\\
17	10.3446621360845\\
18	10.7854786825686\\
19	10.5365152565174\\
20	10.6220783957659\\
21	10.5923208206695\\
22	10.7256054420106\\
23	10.7381349589018\\
24	10.6519628893075\\
26	10.8117543658556\\
27	10.873830701117\\
28	10.8424762656381\\
29	11.249071941532\\
30	11.3741530734922\\
31	11.4708707249062\\
32	11.5277568423158\\
33	11.5057933405931\\
34	11.5374374771172\\
35	11.7497955803524\\
36	11.779375849708\\
37	11.9303931760633\\
38	11.8609658036374\\
39	11.973773800107\\
40	11.8712537717228\\
41	11.8469607122408\\
42	11.9856253790174\\
44	12.0579560230596\\
45	12.1699551852448\\
46	12.2469296921871\\
47	12.2702496243629\\
48	12.2266734174163\\
49	12.1717549507656\\
50	12.162271773121\\
nan	nan\\
50	12.99901854946\\
49	12.9357719309553\\
48	12.9166957582113\\
46	12.9322817773474\\
45	12.9196505495226\\
44	12.8703593891272\\
43	12.7781509499926\\
42	12.7097151227751\\
41	12.8842220834585\\
40	12.7524021422561\\
39	12.5781975260583\\
38	12.5297151999472\\
37	12.6215781501019\\
36	12.6650685947368\\
35	12.2645413372107\\
34	12.3693725587255\\
33	12.2755686665757\\
32	12.4865800752473\\
31	12.3104912822626\\
30	12.2279974641425\\
29	12.3530785961028\\
28	12.3833301859752\\
27	12.1369219870554\\
26	12.0556291466896\\
25	12.0096247197567\\
24	12.1974994762842\\
23	12.2009331414569\\
22	12.1238569235812\\
21	12.113772369295\\
20	12.0660936472453\\
19	11.900760729146\\
18	11.4725858335608\\
17	11.6625063226972\\
16	11.5778410733613\\
15	11.4080976182048\\
13	11.2325805895714\\
12	11.6052963649438\\
11	11.4312452031946\\
10	11.5678203040318\\
9	11.4939774261508\\
8	11.1453982158465\\
7	11.1552490494634\\
6	10.2007847733763\\
5	12.5220541796865\\
3	19.1969530838506\\
2	31.3504802371679\\
1	47.3986291235391\\
};
\addplot [color=black!80!green, forget plot]
  table[row sep=crcr]{%
1	40.215814100108\\
2	37.3595413632489\\
3	34.2511020902393\\
4	28.5857648318247\\
5	25.5559578641439\\
6	21.8419685465166\\
7	20.9176862716395\\
8	20.253852437667\\
9	20.4690433598842\\
10	19.8084205752076\\
11	19.9137322160906\\
12	20.0508473419412\\
13	20.4468644043889\\
14	20.3538178744169\\
16	21.6643801021343\\
17	23.0480264576396\\
18	23.4674374750865\\
19	23.5046558809274\\
20	23.7379098280223\\
21	23.9599287041056\\
22	23.5181752401485\\
23	23.5675615766845\\
24	23.9786509190683\\
25	24.0029341437137\\
27	24.4521948481193\\
28	24.4183306271592\\
29	24.9551971326167\\
30	25.2417378369844\\
31	25.3554988652129\\
32	25.509778461739\\
33	25.5989208043244\\
34	25.9276500870174\\
35	26.1143294178904\\
36	26.4327140411446\\
37	26.5131825229104\\
38	26.4169444754308\\
39	26.4166236218619\\
40	26.4949811568515\\
41	26.5871099527687\\
42	26.848670677602\\
43	26.924912369981\\
44	26.7983493658567\\
45	26.5182365193067\\
46	26.1271056561803\\
47	26.0344177334534\\
48	26.3410170717716\\
49	26.2573767025905\\
50	26.2169978444291\\
nan	nan\\
50	30.3062996466108\\
48	30.3077284479421\\
47	30.022929936798\\
46	29.9840054549312\\
45	30.1484301473604\\
43	29.86720232536\\
42	29.7642325482049\\
41	29.7211337748303\\
40	29.4906819255863\\
39	29.5690394605759\\
38	29.4253494313797\\
37	29.4007959717137\\
36	28.9794723387841\\
35	28.5810110839021\\
34	28.6243212391479\\
33	28.7559179053535\\
32	28.6837699253582\\
31	28.6946803462571\\
30	28.2887281128368\\
29	28.1810035842296\\
28	28.2339991219451\\
27	28.0030022844975\\
26	27.8173683919048\\
25	27.7533382577204\\
24	27.831384923226\\
23	27.7944455917746\\
22	27.8975953691709\\
21	27.509605346074\\
20	27.6599396343436\\
19	27.6243763771375\\
18	27.159802668283\\
17	26.4681025746189\\
16	25.8804227652496\\
15	24.7202407472421\\
14	23.7322036309597\\
13	22.6893363124573\\
12	23.0136687870913\\
11	22.1113573896445\\
10	21.6969557688787\\
9	21.9323903318724\\
8	21.7174737272079\\
7	22.9712026172497\\
6	23.2834794821575\\
5	26.2540779781504\\
4	30.2314394692511\\
3	35.6772133219478\\
2	39.5221790668595\\
1	42.221461885556\\
};
\addplot [color=blue, forget plot]
  table[row sep=crcr]{%
1	40.7092304002609\\
2	39.5690500389897\\
3	39.1612744164465\\
4	33.8719019466531\\
5	25.8427013459557\\
6	19.0802462634564\\
7	18.1622881155458\\
8	15.9785255230203\\
9	15.2697285410979\\
10	15.2443843467688\\
11	15.5247774348704\\
12	15.5008174234103\\
13	15.5759345402702\\
14	15.4151388508905\\
15	15.3293070016156\\
16	15.1684789936105\\
17	15.1707998457621\\
18	15.2059293135845\\
19	14.9741107194134\\
20	14.9432193126713\\
21	14.8526808875146\\
22	14.8134914150696\\
23	14.7488024824106\\
24	14.8565374951916\\
25	14.7724919083401\\
26	14.8487528269361\\
28	14.7354964858016\\
29	14.7226921284337\\
30	14.8785757555993\\
31	14.7896884576763\\
32	14.5468141831197\\
35	14.7257434251771\\
36	14.7398168055745\\
37	14.7226704751348\\
38	14.7415462040658\\
39	14.8427990986455\\
40	14.7603650275137\\
41	14.8403941831638\\
42	14.9497182983314\\
43	14.9023315484233\\
44	14.8182099589434\\
45	14.8341987052252\\
46	14.7714762847998\\
47	14.6907284647048\\
48	14.7217474214869\\
49	14.7733847449668\\
50	14.695222619915\\
nan	nan\\
50	15.3764619678989\\
49	15.3162209898005\\
48	15.3140948724202\\
47	15.4526407109227\\
46	15.4614986255947\\
45	15.4883819399364\\
44	15.3968438044977\\
43	15.3485645089247\\
42	15.3728623468302\\
41	15.4821864619978\\
40	15.5084521767876\\
39	15.4080969587024\\
38	15.4555864124219\\
37	15.6178313169801\\
36	15.6006849865405\\
35	15.6685218077981\\
34	15.6185753457796\\
33	15.8208044366007\\
32	15.7040818742282\\
31	15.4791287466251\\
30	15.8024278286303\\
29	15.8328634271222\\
28	15.8200590697542\\
27	15.7117344398074\\
26	15.8860141981538\\
25	15.8905905289361\\
24	15.7169392073176\\
23	15.6813050444714\\
21	16.0254553132025\\
20	15.9707591819526\\
19	16.0653158038844\\
18	15.9410240914337\\
17	16.0478381470697\\
16	16.1218435870349\\
15	16.1760693424707\\
14	16.32321240359\\
13	16.5208396532784\\
12	16.6138779170917\\
11	16.5899179056317\\
10	16.4222823198982\\
9	16.8987302402644\\
8	17.121832899919\\
7	19.2749878701175\\
6	19.9340906541067\\
5	27.0246821665895\\
4	34.9453023544229\\
3	40.3727757627659\\
2	41.3807707495417\\
1	42.4986549043995\\
};
\addplot [color=red, forget plot]
  table[row sep=crcr]{%
1	46.5607072317721\\
2	29.3155165981336\\
3	14.1774482060498\\
4	10.5100063437428\\
5	7.40459094006522\\
6	5.17597957190867\\
7	4.92621341820067\\
8	4.71587556510874\\
9	4.5793593468322\\
10	4.66900006938199\\
11	4.60765062301117\\
13	4.60597489729934\\
14	4.68799457747859\\
15	4.80468965642401\\
16	4.76059961961239\\
17	4.86348222166966\\
18	4.87991259631733\\
19	4.8789493182249\\
21	4.81791659430391\\
22	4.7541862299131\\
23	4.64244557096129\\
24	4.71685598426043\\
25	4.7487841938355\\
26	4.69273958367637\\
27	4.77529740483149\\
28	4.71033638216412\\
29	4.70056859787849\\
31	4.76646567741219\\
32	4.90872196023135\\
33	4.84785701060432\\
34	4.73604378933118\\
35	4.78478035687986\\
37	4.67658052803085\\
38	4.67833086186857\\
39	4.61375750869188\\
40	4.72205634254294\\
41	4.72312007888914\\
42	4.77023510026962\\
43	4.75542394150621\\
45	4.79332672754979\\
46	4.77529740483149\\
47	4.83319419240069\\
48	4.8181753638199\\
49	4.81911622546433\\
50	4.74248140365177\\
nan	nan\\
50	5.20375515548811\\
49	5.18088377453577\\
48	5.23558807704041\\
47	5.20264810150622\\
45	5.27835786026393\\
44	5.20730319688441\\
43	5.24457605849389\\
42	5.22976489973048\\
41	5.16935303939053\\
40	5.20625906964353\\
39	5.17118872786736\\
38	5.24998455031791\\
37	5.25173488415563\\
36	5.28405163858444\\
35	5.26898308398045\\
34	5.21019276980871\\
33	5.27759101806962\\
31	5.26937661649472\\
30	5.24566796126139\\
29	5.20982566735458\\
28	5.28966361783598\\
27	5.24262374212201\\
26	5.23557582851011\\
25	5.12576777749076\\
24	5.19353828097264\\
23	5.19626410645816\\
22	5.12036574141315\\
21	5.12831996483597\\
20	5.20997147618018\\
19	5.12105068177519\\
18	5.08424510977596\\
17	5.20820236614406\\
16	5.1856369395275\\
15	5.17738919662268\\
14	5.25824198166129\\
13	5.10728675144627\\
12	5.10325016029616\\
11	5.08768987878103\\
10	5.18763075499087\\
9	5.13390230191342\\
8	5.05114952449709\\
7	5.25299805133348\\
6	5.57670860013445\\
5	8.63483558323244\\
4	12.0168753766875\\
3	15.1236270627677\\
2	30.8995371653078\\
1	48.9410848829244\\
};
\end{axis}
\end{tikzpicture}%

%% file: Figures/balancing0_norm0_FFTtransformed_50rSVDmodes_recall_mixed_class.tex
\definecolor{mycolor1}{rgb}{0.64, 0.76, 0.68}%
\begin{tikzpicture}

\begin{axis}[%
width=7cm,
height=5cm,
scale only axis,
unbounded coords=jump,
xmin=0,
xmax=50,
xlabel style={font=\color{white!15!black}},
xlabel={$r$},
ymin=0,
ymax=100,
ylabel style={font=\color{white!15!black}},
ylabel={Recall B of mixed class in \%},
axis background/.style={fill=white},
title style={font=\bfseries},
legend pos = south east,
legend columns=2,
legend style={legend cell align=left, align=left, draw=white!15!black, font = \footnotesize},
legend entries={Tree,
                NB,
                LD,
                kNN,
                90.1~\%},
]
\addlegendimage{legend image code/.code={
\draw [draw=none, fill=white!80!black] (0cm,-0.15cm) rectangle (0.6cm,0.15cm);
\draw[dashdotted] (0cm,0cm) -- (0.6cm,0cm);
\draw[-] (0cm,0.15cm) -- (0.6cm,0.15cm);
\draw[-] (0cm,-0.15cm) -- (0.6cm,-0.15cm);
}
} 
\addlegendimage{legend image code/.code={
\draw [draw=none, fill=mycolor1] (0cm,-0.15cm) rectangle (0.6cm,0.15cm);
\draw[dotted, black!80!green] (0cm,0cm) -- (0.6cm,0cm);
\draw[-, black!80!green] (0cm,0.15cm) -- (0.6cm,0.15cm);
\draw[-, black!80!green] (0cm,-0.15cm) -- (0.6cm,-0.15cm);
}
}
\addlegendimage{legend image code/.code={
\draw [draw=none, fill=white!80!blue] (0cm,-0.15cm) rectangle (0.6cm,0.15cm);
\draw[dashed, blue] (0cm,0cm) -- (0.6cm,0cm);
\draw[-, blue] (0cm,0.15cm) -- (0.6cm,0.15cm);
\draw[-, blue] (0cm,-0.15cm) -- (0.6cm,-0.15cm);
}
}
\addlegendimage{legend image code/.code={
\draw [draw=none, fill=white!80!red] (0cm,-0.15cm) rectangle (0.6cm,0.15cm);
\draw[-, red] (0cm,0cm) -- (0.6cm,0cm);
\draw[-, red] (0cm,0.15cm) -- (0.6cm,0.15cm);
\draw[-, red] (0cm,-0.15cm) -- (0.6cm,-0.15cm);
}
}
\addlegendimage{mark=none, black, very thick, dotted}

\addplot[area legend, draw=black, fill=white!80!black, dashdotted]
table[row sep=crcr] {%
x	y\\
1	12.6003833420696\\
2	38.5917632499777\\
3	57.5604192386609\\
4	64.6290929811649\\
5	71.5684997987279\\
6	76.369406257331\\
7	75.876367592005\\
8	74.6229910361392\\
9	74.5360637000887\\
10	74.1777251678387\\
11	74.1240557272962\\
12	73.38835653385\\
13	73.1941249207104\\
14	72.5508601371673\\
15	71.8234902438931\\
16	71.6251432868488\\
17	70.883464077159\\
18	71.1139712687373\\
19	70.538220586096\\
20	70.3681621264586\\
21	70.6392523708178\\
22	70.8608766139804\\
23	70.8096519348035\\
24	70.9407275560885\\
25	70.7429513420445\\
26	71.2134191294946\\
27	70.6442569062136\\
28	70.722080736984\\
29	70.7174986789607\\
30	71.0217490454343\\
31	70.3643773597168\\
32	69.9978823914426\\
33	70.0077301140118\\
34	69.7572794447684\\
35	69.7786198789717\\
36	69.8390258507247\\
37	69.6082361927037\\
38	70.3776114525286\\
39	70.1331556033981\\
40	70.1650517344455\\
41	70.0932456592733\\
42	70.0266398375036\\
43	69.5961917859735\\
44	69.5570934356421\\
45	69.7093394628615\\
46	69.7821653948664\\
47	69.6061021130837\\
48	69.4757351270809\\
49	69.0819588007924\\
50	68.8996521125042\\
50	75.4523640092728\\
49	75.8605893756115\\
48	75.8423640866226\\
47	75.8729989539045\\
46	75.9114848561367\\
45	76.0381460077468\\
44	76.6196343474336\\
43	76.4732973912041\\
42	76.4181844603175\\
41	76.6202509310428\\
40	76.8705205346587\\
39	77.0632745745532\\
38	76.8724380283718\\
37	76.568491590372\\
36	76.6057264748776\\
35	76.8806096584269\\
34	76.9553534789231\\
33	76.9195959381321\\
32	77.3052825868758\\
31	77.5825791150839\\
30	77.4079970725641\\
29	77.6047209442647\\
28	78.029597115365\\
27	78.482971983435\\
26	78.3432679892776\\
25	78.3842775476041\\
24	78.5621963152971\\
23	78.746675322875\\
22	79.0708577365603\\
21	78.6484745665843\\
20	78.7580591521279\\
19	78.7490745179938\\
18	78.7101646148116\\
17	79.2096319877599\\
16	79.9694372050813\\
15	80.0396905683134\\
14	79.7957580682054\\
13	79.4747129078878\\
12	79.3335248199475\\
11	79.6711473243258\\
10	79.8851425652163\\
9	79.5809991136652\\
8	79.2800263991308\\
7	79.8507137545952\\
6	78.7149631760678\\
5	76.1637635476204\\
4	69.632993853317\\
3	60.6571167924311\\
2	44.0459465940225\\
1	14.7666929664753\\
}--cycle;
\addlegendentry{Tree}

\addplot[area legend, draw=black!80!green, fill=mycolor1, dotted]
table[row sep=crcr] {%
x	y\\
1	0\\
2	4.1273070732233\\
3	4.98347768624006\\
4	15.9150257732465\\
5	28.1522151448091\\
6	31.323748027686\\
7	34.2071000852216\\
8	42.6026865422175\\
9	46.550512953491\\
10	46.4181019342266\\
11	48.0914484852921\\
12	48.6906680447498\\
13	50.5673307557979\\
14	50.7292480409364\\
15	51.5825287610245\\
16	51.2946876686995\\
17	51.8577800215672\\
18	52.1037637885409\\
19	52.040915602125\\
20	51.9508911555315\\
21	52.1870710146123\\
22	51.9967370460995\\
23	52.0022339949163\\
24	51.7982033746268\\
25	50.8263991849709\\
26	50.3531060317639\\
27	50.1543872864647\\
28	49.5180856777196\\
29	49.9463842882511\\
30	49.2883106037275\\
31	49.4445627303771\\
32	49.1652439295248\\
33	48.9958099120789\\
34	48.3247247982583\\
35	48.0516508643628\\
36	46.8499624360612\\
37	47.057304685305\\
38	46.2563664430739\\
39	45.952634437509\\
40	45.0108318800249\\
41	45.0716842462653\\
42	45.0926303522375\\
43	44.9715298270243\\
44	44.8658220153792\\
45	44.8260895785834\\
46	44.7155574490125\\
47	44.9088316225992\\
48	45.3615097948\\
49	45.709502294308\\
50	45.9931819946975\\
50	49.0963694384493\\
49	49.3257820859212\\
48	49.4585056792271\\
47	49.7498941092683\\
46	49.889261091031\\
45	49.7791607947725\\
44	50.276485054092\\
43	50.2777279594702\\
42	50.3714645071468\\
41	50.6067438804777\\
40	51.2042930817399\\
39	51.443194772416\\
38	51.6760156574353\\
37	52.4848800313943\\
36	52.5315083162284\\
35	52.9936016681596\\
34	53.9002243713623\\
33	54.3026768720224\\
32	54.7771782954963\\
31	56.0003515327173\\
30	55.941766586477\\
29	56.4102020694149\\
28	56.6777867155366\\
27	57.2760965467756\\
26	57.5068360306\\
25	56.9797796300065\\
24	56.4911969168607\\
23	56.8237911593741\\
22	57.31200577917\\
21	57.1757949191373\\
20	57.2511168693709\\
19	57.2683310286755\\
18	56.8300757493975\\
17	56.3786487169397\\
16	55.7071296298233\\
15	55.8496104332465\\
14	54.286855743003\\
13	53.6442675672494\\
12	52.8387416031261\\
11	51.6131055568476\\
10	49.9599681307283\\
9	49.1831898372318\\
8	46.6366030919837\\
7	36.4118250096918\\
6	33.9828356310218\\
5	31.5195915378114\\
4	17.4087592457362\\
3	6.44643036925052\\
2	4.61990850156483\\
1	0\\
}--cycle;
\addlegendentry{NB}

\addplot[area legend, draw=blue, fill=white!80!blue, dashed]
table[row sep=crcr] {%
x	y\\
1	0\\
2	0\\
3	5.39316151866159\\
4	9.88821595922064\\
5	22.7724225863211\\
6	37.0976342255316\\
7	47.6776858179345\\
8	54.081448972372\\
9	55.7462666335807\\
10	55.9029959451718\\
11	56.7822356180214\\
12	56.6100578818957\\
13	57.6247378055709\\
14	58.6255192052794\\
15	58.9526985599892\\
16	59.5314938297078\\
17	60.7738499571072\\
18	60.8286377270022\\
19	60.9538906686748\\
20	61.0155372197392\\
21	60.5355549651551\\
22	61.4757107645331\\
23	61.3579296368351\\
24	61.4859500392972\\
25	61.2122373338556\\
26	61.5778552732547\\
27	60.8595773591944\\
28	61.7630455337151\\
29	61.9195005325602\\
30	61.8302065307413\\
31	61.5892633028\\
32	61.8785306692256\\
33	61.8362795401693\\
34	61.6070469892959\\
35	61.4166258216035\\
36	61.5072425130641\\
37	61.4330261793231\\
38	61.427169496301\\
39	61.3161595780853\\
40	61.5199705209853\\
41	61.6202670100204\\
42	61.5137560512569\\
43	61.3437722390045\\
44	61.4521639453053\\
45	61.8649004374024\\
46	61.9831751378246\\
47	61.9604005610703\\
48	62.0104612489426\\
49	61.9980932157436\\
50	62.1103008879635\\
50	65.1226911770994\\
49	65.0739689682534\\
48	65.3300573108933\\
47	65.0579703477589\\
46	64.9815764680555\\
45	64.7243721033969\\
44	64.5470083741794\\
43	64.3872315935161\\
42	64.2171038368263\\
41	64.4864318042373\\
40	63.7817909990679\\
39	64.0926246312099\\
38	63.7668496123231\\
37	63.5997751593601\\
36	63.5259906589315\\
35	63.7772493425832\\
34	63.8015932755618\\
33	63.6796713028053\\
32	64.0668784028726\\
31	64.0344299516038\\
30	63.9002215240295\\
29	64.6697720082391\\
28	64.6118937397254\\
27	64.8177351981584\\
26	64.7434646972367\\
25	64.679408490856\\
24	64.566985527574\\
23	65.2849622069132\\
22	64.4162949212722\\
21	64.7122993630742\\
20	64.4470102369425\\
19	64.0257232003717\\
18	63.7752091880886\\
17	63.7767375161282\\
16	62.6573252861135\\
15	61.9479617851176\\
14	61.2026831086275\\
13	60.6464175284701\\
12	61.6084857602584\\
11	60.6310108688347\\
10	60.2222357154072\\
9	58.6620677493474\\
8	57.8056266164507\\
7	51.4906751947146\\
6	39.5873253800607\\
5	25.5771464800193\\
4	12.5419979781995\\
3	6.62706267479974\\
2	0\\
1	0\\
}--cycle;
\addlegendentry{LD}

\addplot[area legend, draw=red, fill=white!80!red]
table[row sep=crcr] {%
x	y\\
1	16.8488892220627\\
2	42.2459898366468\\
3	62.8532619010418\\
4	71.0861288403606\\
5	81.7287381044262\\
6	85.7995338268804\\
7	87.1857557879542\\
8	87.5392859203172\\
9	87.4863177648082\\
10	87.3957390023301\\
11	87.4991782795015\\
12	87.2411332900459\\
13	87.019124753559\\
14	87.0936941063486\\
15	87.0506393594983\\
16	87.2086216193164\\
17	87.1978885193957\\
18	87.3167622570919\\
19	87.4447027387499\\
20	87.3314660066855\\
21	86.9934218081045\\
22	87.129765474327\\
23	87.1827910578898\\
24	87.2617818804738\\
25	86.9753506181729\\
26	86.9702250750959\\
27	87.2742306523047\\
28	87.0504446331128\\
29	87.1374426797777\\
30	86.9371134658172\\
31	86.8913384494166\\
32	86.8467386779329\\
33	86.7733555832964\\
34	86.555387294122\\
35	86.5969501749637\\
36	86.4167319809284\\
37	86.5527630384156\\
38	86.6156816656081\\
39	86.6288397995449\\
40	86.4965763087405\\
41	86.3909133474556\\
42	86.6230469481226\\
43	86.5258356129352\\
44	86.3716060287977\\
45	86.2361469830918\\
46	86.2543406716911\\
47	86.3808838003043\\
48	86.4886703433809\\
49	86.6525880662821\\
50	86.4539351553525\\
50	89.1823892234373\\
49	89.0909749184058\\
48	89.040343457292\\
47	88.772578963069\\
46	88.8989061750261\\
45	89.1851963783707\\
44	88.9961180297157\\
43	89.2176553995341\\
42	89.0133494028859\\
41	88.9770985999327\\
40	89.1933673729984\\
39	89.1682705158733\\
38	88.9668794657952\\
37	88.9225594870895\\
36	89.004899269409\\
35	89.0394461760448\\
34	88.4909088355719\\
33	88.6485635559159\\
32	88.7361103423453\\
31	88.6915825430803\\
30	88.8065214910894\\
29	88.7137907441207\\
28	88.6400748263759\\
27	88.7919118167023\\
26	88.8812962376774\\
25	88.6614775661479\\
24	89.1262923228838\\
23	88.883135494461\\
22	88.9897084087542\\
21	89.4480557815459\\
20	89.4320152895342\\
19	89.3726137770749\\
18	89.3395524054484\\
17	89.5654488323865\\
16	89.3937138791812\\
15	89.5516241667805\\
14	89.9380276490539\\
13	89.7980478178285\\
12	89.6834938038958\\
11	89.5330472814319\\
10	90.3336814413786\\
9	89.8138603664331\\
8	90.0826800008372\\
7	89.5237463442227\\
6	87.9593232542862\\
5	83.2292840990033\\
4	74.9819928646613\\
3	64.1656128133185\\
2	45.8665747134369\\
1	18.8904993739393\\
}--cycle;
\addlegendentry{kNN}

\addplot[mark=none, black, very thick, dotted, domain=0:50] {90.08};
\addlegendentry{90.1~\%}

\addplot [color=black, forget plot, dashdotted]
  table[row sep=crcr]{%
1	13.6835381542725\\
2	41.3188549220001\\
3	59.108768015546\\
4	67.1310434172409\\
5	73.8661316731742\\
6	77.5421847166994\\
7	77.8635406733001\\
8	76.951508717635\\
9	77.058531406877\\
10	77.0314338665275\\
11	76.897601525811\\
12	76.3609406768987\\
13	76.3344189142991\\
14	76.1733091026864\\
15	75.9315904061032\\
16	75.7972902459651\\
17	75.0465480324595\\
18	74.9120679417745\\
19	74.6436475520449\\
20	74.5631106392932\\
21	74.6438634687011\\
22	74.9658671752703\\
23	74.7781636288393\\
24	74.7514619356928\\
25	74.5636144448243\\
26	74.7783435593861\\
27	74.5636144448243\\
28	74.3758389261745\\
29	74.1611098116127\\
30	74.2148730589992\\
31	73.9734782374004\\
32	73.6515824891592\\
33	73.4636630260719\\
34	73.3563164618457\\
35	73.3296147686993\\
36	73.2223761628012\\
37	73.0883638915379\\
38	73.6250247404502\\
39	73.5982150889756\\
40	73.5177861345521\\
41	73.3567482951581\\
42	73.2224121489105\\
43	73.0347445885888\\
44	73.0883638915379\\
45	72.8737427353042\\
46	72.8468251255016\\
47	72.7395505334941\\
48	72.6590496068518\\
49	72.471274088202\\
50	72.1760080608885\\
};
\addplot [color=black!80!green, forget plot, dotted]
  table[row sep=crcr]{%
1	0\\
2	4.37360778739406\\
3	5.71495402774529\\
4	16.6618925094913\\
5	29.8359033413103\\
6	32.6532918293539\\
7	35.3094625474567\\
8	44.6196448171006\\
9	47.8668513953614\\
10	48.1890350324775\\
11	49.8522770210699\\
12	50.764704823938\\
13	52.1057991615237\\
14	52.5080518919697\\
15	53.7160695971355\\
16	53.5009086492614\\
17	54.1182143692535\\
18	54.4669197689692\\
19	54.6546233154003\\
20	54.6010040124512\\
21	54.6814329668748\\
22	54.6543714126347\\
23	54.4130125771452\\
24	54.1447001457437\\
25	53.9030894074887\\
26	53.929971031182\\
27	53.7152419166202\\
28	53.0979361966281\\
29	53.178293178833\\
30	52.6150385951023\\
31	52.7224571315472\\
32	51.9712111125106\\
33	51.6492433920507\\
34	51.1124745848103\\
35	50.5226262662612\\
36	49.6907353761448\\
37	49.7710923583497\\
38	48.9661910502546\\
39	48.6979146049625\\
40	48.1075624808824\\
41	47.8392140633715\\
43	47.6246288932472\\
44	47.5711535347356\\
45	47.302625186678\\
46	47.3024092700218\\
47	47.3293628659338\\
48	47.4100077370135\\
49	47.5176421901146\\
50	47.5447757165734\\
};
\addplot [color=blue, forget plot, dashed]
  table[row sep=crcr]{%
1	0\\
2	0\\
3	6.01011209673067\\
4	11.2151069687101\\
5	24.1747845331702\\
6	38.3424798027961\\
7	49.5841805063246\\
8	55.9435377944114\\
9	57.2041671914641\\
10	58.0626158302895\\
11	58.706623243428\\
12	59.1092718210771\\
13	59.1355776670205\\
14	59.9141011569534\\
15	60.4503301725534\\
16	61.0944095579107\\
17	62.2752937366177\\
18	62.3019234575454\\
19	62.4898069345233\\
20	62.7312737283409\\
21	62.6239271641147\\
22	62.9460028429026\\
23	63.3214459218742\\
24	63.0264677834356\\
25	62.9458229123558\\
26	63.1606599852457\\
27	62.8386562786764\\
28	63.1874696367202\\
29	63.2946362703996\\
30	62.8652140273854\\
31	62.8118466272019\\
32	62.9727045360491\\
33	62.7579754214873\\
34	62.7043201324289\\
35	62.5969375820933\\
36	62.5166165859978\\
37	62.5164006693416\\
38	62.597009554312\\
39	62.7043921046476\\
40	62.6508807600266\\
41	63.0533494071289\\
42	62.8654299440416\\
43	62.8655019162603\\
44	62.9995861597423\\
45	63.2946362703996\\
46	63.4823758029401\\
47	63.5091854544146\\
48	63.6702592799179\\
49	63.5360310919985\\
50	63.6164960325314\\
};
\addplot [color=red, forget plot]
  table[row sep=crcr]{%
1	17.869694298001\\
2	44.0562822750418\\
3	63.5094373571801\\
5	82.4790111017147\\
6	86.8794285405833\\
7	88.3547510660885\\
8	88.8109829605772\\
9	88.6500890656207\\
10	88.8647102218544\\
11	88.5161127804667\\
13	88.4085862856937\\
14	88.5158608777012\\
15	88.3011317631394\\
16	88.3011677492488\\
17	88.3816686758911\\
18	88.3281573312701\\
19	88.4086582579124\\
20	88.3817406481098\\
22	88.0597369415406\\
23	88.0329632761754\\
24	88.1940371016788\\
25	87.8184140921604\\
27	88.0330712345035\\
28	87.8452597297443\\
29	87.9256167119492\\
30	87.8718174784533\\
31	87.7914604962484\\
32	87.7914245101391\\
33	87.7109595696061\\
34	87.5231480648469\\
35	87.8181981755043\\
36	87.7108156251687\\
37	87.7376612627526\\
38	87.7912805657016\\
39	87.8985551577091\\
40	87.8449718408694\\
41	87.6840059736942\\
42	87.8181981755043\\
43	87.8717455062346\\
44	87.6838620292567\\
45	87.7106716807312\\
46	87.5766234233586\\
47	87.5767313816867\\
48	87.7645069003365\\
49	87.871781492344\\
50	87.8181621893949\\
};
\addplot [color=black, forget plot]
  table[row sep=crcr]{%
1	12.6003833420696\\
2	38.5917632499777\\
3	57.5604192386609\\
4	64.6290929811649\\
5	71.5684997987279\\
6	76.369406257331\\
7	75.876367592005\\
8	74.6229910361392\\
9	74.5360637000887\\
10	74.1777251678387\\
11	74.1240557272962\\
12	73.38835653385\\
13	73.1941249207104\\
14	72.5508601371673\\
15	71.8234902438931\\
16	71.6251432868488\\
17	70.883464077159\\
18	71.1139712687373\\
19	70.538220586096\\
20	70.3681621264586\\
21	70.6392523708178\\
22	70.8608766139804\\
23	70.8096519348035\\
24	70.9407275560885\\
25	70.7429513420445\\
26	71.2134191294946\\
27	70.6442569062136\\
28	70.722080736984\\
29	70.7174986789607\\
30	71.0217490454343\\
31	70.3643773597168\\
32	69.9978823914426\\
33	70.0077301140118\\
34	69.7572794447684\\
35	69.7786198789717\\
36	69.8390258507247\\
37	69.6082361927037\\
38	70.3776114525286\\
39	70.1331556033981\\
40	70.1650517344455\\
42	70.0266398375036\\
43	69.5961917859735\\
44	69.5570934356421\\
45	69.7093394628615\\
46	69.7821653948664\\
47	69.6061021130837\\
48	69.4757351270809\\
49	69.0819588007924\\
50	68.8996521125042\\
nan	nan\\
50	75.4523640092728\\
49	75.8605893756115\\
48	75.8423640866226\\
46	75.9114848561367\\
45	76.0381460077468\\
44	76.6196343474336\\
43	76.4732973912041\\
42	76.4181844603175\\
41	76.6202509310428\\
40	76.8705205346587\\
39	77.0632745745532\\
38	76.8724380283718\\
37	76.568491590372\\
36	76.6057264748776\\
35	76.8806096584269\\
34	76.9553534789231\\
33	76.9195959381321\\
32	77.3052825868758\\
31	77.5825791150839\\
30	77.4079970725641\\
29	77.6047209442647\\
28	78.029597115365\\
27	78.482971983435\\
26	78.3432679892776\\
25	78.3842775476041\\
23	78.746675322875\\
22	79.0708577365603\\
21	78.6484745665843\\
20	78.7580591521279\\
19	78.7490745179938\\
18	78.7101646148116\\
17	79.2096319877599\\
16	79.9694372050813\\
15	80.0396905683134\\
14	79.7957580682054\\
13	79.4747129078878\\
12	79.3335248199475\\
11	79.6711473243258\\
10	79.8851425652163\\
8	79.2800263991308\\
7	79.8507137545952\\
6	78.7149631760678\\
5	76.1637635476204\\
4	69.632993853317\\
3	60.6571167924311\\
2	44.0459465940225\\
1	14.7666929664753\\
};
\addplot [color=black!80!green, forget plot]
  table[row sep=crcr]{%
1	0\\
2	4.1273070732233\\
3	4.98347768624006\\
4	15.9150257732465\\
5	28.1522151448091\\
6	31.323748027686\\
7	34.2071000852216\\
8	42.6026865422175\\
9	46.550512953491\\
10	46.4181019342266\\
11	48.0914484852921\\
12	48.6906680447498\\
13	50.5673307557979\\
14	50.7292480409364\\
15	51.5825287610245\\
16	51.2946876686995\\
17	51.8577800215672\\
18	52.1037637885409\\
19	52.040915602125\\
20	51.9508911555315\\
21	52.1870710146123\\
22	51.9967370460995\\
23	52.0022339949163\\
24	51.7982033746268\\
25	50.8263991849709\\
26	50.3531060317639\\
27	50.1543872864647\\
28	49.5180856777196\\
29	49.9463842882511\\
30	49.2883106037275\\
31	49.4445627303771\\
32	49.1652439295248\\
33	48.9958099120789\\
34	48.3247247982583\\
35	48.0516508643628\\
36	46.8499624360612\\
37	47.057304685305\\
38	46.2563664430739\\
39	45.952634437509\\
40	45.0108318800249\\
41	45.0716842462653\\
42	45.0926303522375\\
43	44.9715298270243\\
44	44.8658220153792\\
45	44.8260895785834\\
46	44.7155574490125\\
47	44.9088316225992\\
48	45.3615097948\\
49	45.709502294308\\
50	45.9931819946975\\
nan	nan\\
50	49.0963694384493\\
49	49.3257820859212\\
48	49.4585056792271\\
47	49.7498941092683\\
46	49.889261091031\\
45	49.7791607947725\\
44	50.276485054092\\
43	50.2777279594702\\
42	50.3714645071468\\
41	50.6067438804777\\
40	51.2042930817399\\
38	51.6760156574353\\
37	52.4848800313943\\
36	52.5315083162284\\
35	52.9936016681596\\
34	53.9002243713623\\
33	54.3026768720224\\
32	54.7771782954963\\
31	56.0003515327173\\
30	55.941766586477\\
29	56.4102020694149\\
28	56.6777867155366\\
27	57.2760965467756\\
26	57.5068360306\\
25	56.9797796300065\\
24	56.4911969168607\\
23	56.8237911593741\\
22	57.31200577917\\
21	57.1757949191373\\
20	57.2511168693709\\
19	57.2683310286755\\
18	56.8300757493975\\
17	56.3786487169397\\
16	55.7071296298233\\
15	55.8496104332465\\
14	54.286855743003\\
13	53.6442675672494\\
12	52.8387416031261\\
11	51.6131055568476\\
10	49.9599681307283\\
9	49.1831898372318\\
8	46.6366030919837\\
7	36.4118250096918\\
6	33.9828356310218\\
5	31.5195915378114\\
4	17.4087592457362\\
3	6.44643036925052\\
2	4.61990850156484\\
1	0\\
};
\addplot [color=blue, forget plot]
  table[row sep=crcr]{%
1	0\\
2	0\\
3	5.39316151866159\\
4	9.88821595922064\\
5	22.7724225863211\\
6	37.0976342255316\\
7	47.6776858179345\\
8	54.081448972372\\
9	55.7462666335807\\
10	55.9029959451718\\
11	56.7822356180214\\
12	56.6100578818957\\
14	58.6255192052794\\
15	58.9526985599892\\
16	59.5314938297078\\
17	60.7738499571072\\
18	60.8286377270022\\
19	60.9538906686748\\
20	61.0155372197392\\
21	60.5355549651551\\
22	61.4757107645331\\
23	61.3579296368351\\
24	61.4859500392972\\
25	61.2122373338556\\
26	61.5778552732547\\
27	60.8595773591944\\
28	61.7630455337151\\
29	61.9195005325602\\
30	61.8302065307413\\
31	61.5892633028\\
32	61.8785306692256\\
33	61.8362795401693\\
34	61.6070469892959\\
35	61.4166258216035\\
36	61.5072425130641\\
37	61.4330261793231\\
38	61.427169496301\\
39	61.3161595780853\\
40	61.5199705209853\\
41	61.6202670100204\\
42	61.5137560512569\\
43	61.3437722390045\\
44	61.4521639453053\\
45	61.8649004374024\\
46	61.9831751378246\\
47	61.9604005610703\\
48	62.0104612489426\\
49	61.9980932157436\\
50	62.1103008879635\\
nan	nan\\
50	65.1226911770994\\
49	65.0739689682534\\
48	65.3300573108933\\
47	65.0579703477589\\
46	64.9815764680555\\
45	64.7243721033969\\
44	64.5470083741794\\
42	64.2171038368263\\
41	64.4864318042373\\
40	63.781790999068\\
39	64.0926246312099\\
38	63.7668496123231\\
37	63.5997751593601\\
36	63.5259906589315\\
35	63.7772493425832\\
34	63.8015932755618\\
33	63.6796713028053\\
32	64.0668784028726\\
31	64.0344299516038\\
30	63.9002215240295\\
29	64.6697720082391\\
28	64.6118937397254\\
27	64.8177351981584\\
25	64.679408490856\\
24	64.566985527574\\
23	65.2849622069132\\
22	64.4162949212722\\
21	64.7122993630742\\
20	64.4470102369425\\
19	64.0257232003717\\
18	63.7752091880886\\
17	63.7767375161282\\
16	62.6573252861135\\
15	61.9479617851176\\
14	61.2026831086275\\
13	60.6464175284701\\
12	61.6084857602584\\
11	60.6310108688347\\
10	60.2222357154072\\
9	58.6620677493474\\
8	57.8056266164507\\
7	51.4906751947146\\
6	39.5873253800607\\
5	25.5771464800193\\
4	12.5419979781995\\
3	6.62706267479973\\
2	0\\
1	0\\
};
\addplot [color=red, forget plot]
  table[row sep=crcr]{%
1	16.8488892220627\\
2	42.2459898366468\\
3	62.8532619010418\\
4	71.0861288403606\\
5	81.7287381044262\\
6	85.7995338268804\\
7	87.1857557879542\\
8	87.5392859203172\\
9	87.4863177648082\\
10	87.3957390023301\\
11	87.4991782795015\\
12	87.2411332900459\\
13	87.019124753559\\
14	87.0936941063486\\
15	87.0506393594983\\
16	87.2086216193164\\
17	87.1978885193957\\
19	87.4447027387499\\
20	87.3314660066855\\
21	86.9934218081045\\
22	87.129765474327\\
23	87.1827910578898\\
24	87.2617818804738\\
25	86.9753506181729\\
26	86.9702250750959\\
27	87.2742306523047\\
28	87.0504446331128\\
29	87.1374426797777\\
30	86.9371134658172\\
32	86.8467386779329\\
33	86.7733555832964\\
34	86.555387294122\\
35	86.5969501749637\\
36	86.4167319809284\\
37	86.5527630384156\\
38	86.6156816656081\\
39	86.6288397995449\\
40	86.4965763087405\\
41	86.3909133474556\\
42	86.6230469481226\\
43	86.5258356129352\\
44	86.3716060287977\\
45	86.2361469830918\\
46	86.2543406716911\\
47	86.3808838003043\\
48	86.4886703433809\\
49	86.6525880662821\\
50	86.4539351553525\\
nan	nan\\
50	89.1823892234373\\
49	89.0909749184058\\
48	89.040343457292\\
47	88.772578963069\\
46	88.8989061750261\\
45	89.1851963783707\\
44	88.9961180297157\\
43	89.2176553995341\\
42	89.0133494028859\\
41	88.9770985999327\\
40	89.1933673729984\\
39	89.1682705158733\\
38	88.9668794657952\\
37	88.9225594870895\\
36	89.004899269409\\
35	89.0394461760448\\
34	88.4909088355719\\
33	88.6485635559159\\
32	88.7361103423453\\
31	88.6915825430803\\
30	88.8065214910894\\
29	88.7137907441207\\
28	88.6400748263759\\
27	88.7919118167023\\
26	88.8812962376774\\
25	88.6614775661479\\
24	89.1262923228838\\
23	88.883135494461\\
22	88.9897084087542\\
21	89.4480557815459\\
20	89.4320152895342\\
19	89.3726137770749\\
18	89.3395524054484\\
17	89.5654488323865\\
16	89.3937138791812\\
15	89.5516241667805\\
14	89.9380276490539\\
13	89.7980478178285\\
12	89.6834938038958\\
11	89.5330472814319\\
10	90.3336814413786\\
9	89.8138603664331\\
8	90.0826800008372\\
7	89.5237463442227\\
6	87.9593232542862\\
5	83.2292840990033\\
4	74.9819928646613\\
3	64.1656128133185\\
2	45.8665747134369\\
1	18.8904993739393\\
};
\end{axis}
\end{tikzpicture}%

%% file: Figures/balancing1_norm0_FFTtransformed_50rSVDmodes_recall_mixed_class.tex
\definecolor{mycolor1}{rgb}{0.64, 0.76, 0.68}%
\begin{tikzpicture}

\begin{axis}[%
width=7cm,
height=5cm,
scale only axis,
unbounded coords=jump,
xmin=0,
xmax=50,
xlabel style={font=\color{white!15!black}},
xlabel={$r$},
ymin=0,
ymax=100,
ylabel style={font=\color{white!15!black}},
axis background/.style={fill=white},
title style={font=\bfseries},
legend pos = south east,
legend columns=2,
legend style={legend cell align=left, align=left, draw=white!15!black, font=\footnotesize},
legend entries={Tree,
                NB,
                LD,
                kNN,
                94.0~\%},
]
\addlegendimage{legend image code/.code={
\draw [draw=none, fill=white!80!black] (0cm,-0.15cm) rectangle (0.6cm,0.15cm);
\draw[dashdotted] (0cm,0cm) -- (0.6cm,0cm);
\draw[-] (0cm,0.15cm) -- (0.6cm,0.15cm);
\draw[-] (0cm,-0.15cm) -- (0.6cm,-0.15cm);
}
} 
\addlegendimage{legend image code/.code={
\draw [draw=none, fill=mycolor1] (0cm,-0.15cm) rectangle (0.6cm,0.15cm);
\draw[dotted, black!80!green] (0cm,0cm) -- (0.6cm,0cm);
\draw[-, black!80!green] (0cm,0.15cm) -- (0.6cm,0.15cm);
\draw[-, black!80!green] (0cm,-0.15cm) -- (0.6cm,-0.15cm);
}
}
\addlegendimage{legend image code/.code={
\draw [draw=none, fill=white!80!blue] (0cm,-0.15cm) rectangle (0.6cm,0.15cm);
\draw[dashed, blue] (0cm,0cm) -- (0.6cm,0cm);
\draw[-, blue] (0cm,0.15cm) -- (0.6cm,0.15cm);
\draw[-, blue] (0cm,-0.15cm) -- (0.6cm,-0.15cm);
}
}
\addlegendimage{legend image code/.code={
\draw [draw=none, fill=white!80!red] (0cm,-0.15cm) rectangle (0.6cm,0.15cm);
\draw[-, red] (0cm,0cm) -- (0.6cm,0cm);
\draw[-, red] (0cm,0.15cm) -- (0.6cm,0.15cm);
\draw[-, red] (0cm,-0.15cm) -- (0.6cm,-0.15cm);
}
}
\addlegendimage{mark=none, black, very thick, dotted}

\addplot[area legend, draw=black, fill=white!80!black, dashdotted]
table[row sep=crcr] {%
x	y\\
1	37.5145325910684\\
2	59.0420754043394\\
3	72.3526155236361\\
4	77.6081380404386\\
5	81.2881937684708\\
6	84.9040652678794\\
7	83.144925503899\\
8	83.2955615506214\\
9	82.4525919795687\\
10	82.1759264377248\\
11	82.3712986722121\\
12	82.186315131748\\
13	82.1418907029601\\
14	82.0017894523874\\
15	81.7541247429938\\
16	81.5642466788205\\
17	81.557553666634\\
18	82.1391610499485\\
19	81.64913758769\\
20	81.6774109014884\\
21	81.6869387366542\\
22	81.348669368557\\
23	81.1265481917629\\
24	81.0192663430561\\
25	80.9372054043892\\
26	80.9354910203741\\
27	80.4796346701698\\
28	79.7772288198204\\
29	79.6317719737152\\
30	79.794514245347\\
31	79.3073278621046\\
32	78.9498345425477\\
33	79.5966197060047\\
34	79.4237717566347\\
35	79.6223122756872\\
36	78.965818493543\\
37	78.9842018980185\\
38	79.3185891590438\\
39	79.1404476389008\\
40	79.1498919807022\\
41	78.9350557731852\\
42	79.3481311540043\\
43	79.2677901479178\\
44	79.1209070927616\\
45	78.9031522713589\\
46	79.0082138655129\\
47	79.0082138655129\\
48	79.0882630647659\\
49	79.1054855290721\\
50	78.8965837262415\\
50	82.6087926178445\\
49	82.668708019315\\
48	82.632167042761\\
47	82.4433990377129\\
46	82.4433990377129\\
45	82.4409337501465\\
44	82.8145767782062\\
43	82.8289840456306\\
42	82.8561699212645\\
41	83.0541915386428\\
40	83.0006456537065\\
39	82.9563265546476\\
38	83.2620560022465\\
37	83.2200991772503\\
36	83.4535363451667\\
35	83.6572576167859\\
34	84.1246153401395\\
33	84.2743480359308\\
32	84.4372622316458\\
31	84.617403320691\\
30	84.8829051094917\\
29	85.0994108219837\\
28	85.1690077393194\\
27	85.0042362975721\\
26	84.9784874742495\\
25	85.1918268536753\\
24	85.5398734418901\\
23	85.0562475071618\\
22	85.0491800938086\\
21	85.356072016034\\
20	85.0967826468987\\
19	85.2325828424175\\
18	84.8500862618795\\
17	85.431693645194\\
16	86.0163984824698\\
15	86.1491010634579\\
14	86.2240169992255\\
13	86.621550157255\\
12	86.5233622876068\\
11	86.6071959514439\\
10	86.53375098163\\
9	87.2785908161302\\
8	87.1883094171205\\
7	86.2099132057784\\
6	86.7626013987873\\
5	83.4967524680884\\
4	79.3811092713893\\
3	76.6258791000198\\
2	62.0869568537252\\
1	40.6575104196843\\
}--cycle;
\addlegendentry{Tree}

\addplot[area legend, draw=black!80!green, fill=mycolor1, dotted]
table[row sep=crcr] {%
x	y\\
1	18.2631333896173\\
2	23.7691884281732\\
3	41.7449666348831\\
4	51.4415646373445\\
5	57.4455416519516\\
6	64.2333627806502\\
7	75.0041863748957\\
8	80.2789432586953\\
9	78.7923561273476\\
10	78.5981946703729\\
11	78.7332026666234\\
12	78.5737095249275\\
13	79.0944977942698\\
14	79.6248783156953\\
15	79.9727449703434\\
16	79.2885947279885\\
17	77.6335845176236\\
18	77.3800689412119\\
19	76.9510624255222\\
20	76.6217865541082\\
21	76.390876824185\\
22	76.1351068829174\\
23	75.0190223187284\\
24	74.2574080137181\\
25	73.8940359173227\\
26	73.4485505282217\\
27	71.5092556684891\\
28	71.0387303356152\\
29	70.2426759566657\\
30	69.8277555061058\\
31	69.0908259014467\\
32	68.4936157803391\\
33	68.2256425447773\\
34	67.6639469279375\\
35	66.7462335074165\\
36	65.8749665884313\\
37	64.851395111935\\
38	64.3645211246365\\
39	63.7718136100064\\
40	62.7618050085543\\
41	62.3168742000823\\
42	61.8469985352003\\
43	61.0746657883969\\
44	60.5514249128559\\
45	60.1336822017934\\
46	59.6602200493677\\
47	59.2110747903208\\
48	58.5258008701907\\
49	58.3901794262015\\
50	58.4141821557101\\
50	61.9621619303114\\
49	62.254981864121\\
48	62.9795754738953\\
47	63.5846241344104\\
46	63.9419304882667\\
45	64.1136296261636\\
44	65.2012632591871\\
43	65.2694202331085\\
42	65.6798831852298\\
41	66.6078569827134\\
40	67.560775636607\\
39	67.7335627340796\\
38	68.6462315635355\\
37	69.395916716022\\
36	70.1465387879127\\
35	71.1032288581749\\
34	72.0134724269012\\
33	72.6345725089861\\
32	73.8719756175104\\
31	74.7263783996286\\
30	75.7098789024963\\
29	76.5852810325816\\
28	77.6709470837397\\
27	78.4907443315109\\
26	79.4546752782299\\
25	80.2457490289139\\
24	80.796355427142\\
23	81.325063702777\\
22	81.6605920418138\\
21	81.9424565091483\\
20	81.8190736609456\\
19	82.7801203701768\\
18	82.7812213813688\\
17	83.3341574178603\\
16	82.8619429064202\\
15	82.876717395248\\
14	82.5794227595735\\
13	82.4108785498162\\
12	81.5875807976531\\
11	81.7506683011185\\
10	80.7028805984443\\
9	81.100116990932\\
8	82.5167556660359\\
7	76.9850609369323\\
6	67.4333038860164\\
5	59.0060712512742\\
4	54.5799407389996\\
3	45.1367537952245\\
2	25.2093061954827\\
1	19.5325655351139\\
}--cycle;
\addlegendentry{NB}

\addplot[area legend, draw=blue, fill=white!80!blue,dashed]
table[row sep=crcr] {%
x	y\\
1	26.4567117032631\\
2	28.8167066620373\\
3	34.5287183868581\\
4	44.7620745403094\\
5	57.447770056182\\
6	73.3019333581786\\
7	76.9475312877768\\
8	81.875037756417\\
9	81.0220225238297\\
10	80.4711772834881\\
11	80.4578301230465\\
12	79.6875874110097\\
13	79.3792032152296\\
14	80.083296565787\\
15	79.6916891113543\\
16	79.9113606767268\\
17	79.6671407690022\\
18	79.9198855368369\\
19	80.3815124646782\\
20	80.3431348370518\\
21	80.486867101613\\
22	80.7463821900857\\
23	80.7282884300756\\
24	80.7282884300755\\
25	80.5947668609509\\
26	80.2594735561486\\
27	80.4533592265512\\
28	80.2144609765482\\
29	79.7087356457476\\
30	79.7584379454061\\
31	79.6964611385988\\
32	79.508507375811\\
33	79.3697058800993\\
34	79.6465817671471\\
35	79.499375658271\\
36	79.4991159262322\\
37	79.4321294694522\\
38	79.5786494182316\\
39	79.5423606420705\\
40	79.5300180207531\\
41	79.7928252270154\\
42	79.8675139761415\\
43	79.4481782726258\\
44	79.6963069132859\\
45	79.7297583934926\\
46	79.8111778463716\\
47	79.55636433827\\
48	79.7896092119184\\
49	79.9900981099146\\
50	79.9695554294054\\
50	81.6971112372613\\
49	81.5690416750316\\
48	81.8770574547482\\
47	81.6264313606548\\
46	81.4791447342736\\
45	81.5068007462924\\
44	81.6477791082194\\
43	81.6270905445785\\
42	81.3690451636435\\
41	81.3899704719093\\
40	81.5452507964511\\
39	81.2103275299725\\
38	81.5503828398329\\
37	81.374322143451\\
36	81.5761528909722\\
35	81.6296565997935\\
34	81.7512676952185\\
33	82.0281435822663\\
32	82.2656861725761\\
31	81.9164420872077\\
30	81.6931749578197\\
29	81.9041675800589\\
28	82.0436035395809\\
27	81.7509418487176\\
26	81.622246873959\\
25	81.878351418619\\
24	81.7448298494943\\
23	81.7448298494944\\
22	81.7267360894841\\
21	81.4486167693548\\
20	81.377295270475\\
19	81.2851542019885\\
18	81.9618348932706\\
17	81.891999015944\\
16	81.8628328716603\\
15	81.6523969101511\\
14	81.8521873051807\\
13	81.5347752793941\\
12	80.8500469975924\\
11	81.3163634253406\\
10	81.6793603509205\\
9	82.418837691224\\
8	83.0711988027228\\
7	78.3750493573845\\
6	75.9453784697785\\
5	60.5092191911299\\
4	47.2809362123788\\
3	36.4390235486258\\
2	30.269314843339\\
1	28.4357614150165\\
}--cycle;
\addlegendentry{LD}

\addplot[area legend, draw=red, fill=white!80!red]
table[row sep=crcr] {%
x	y\\
1	36.8726685669795\\
2	59.5560563618559\\
3	78.4492226133821\\
4	82.5745192323848\\
5	86.6720292956487\\
6	91.2140854471632\\
7	91.7436818071964\\
8	91.8349155780589\\
9	91.6049090638236\\
10	91.3919445251299\\
11	91.6575113165225\\
12	91.7068971909439\\
13	91.7313637449954\\
14	91.525712456428\\
15	91.5334649902805\\
16	91.5915502827274\\
17	91.4471438735604\\
18	91.5715897051289\\
19	91.7577570455313\\
20	91.6714446640015\\
21	91.6941086662214\\
22	91.731399007769\\
23	91.8579473599169\\
24	91.8650067658484\\
25	91.6951722511565\\
26	91.6478028234329\\
27	91.7432425666527\\
28	91.7444983325271\\
29	91.7789766125042\\
30	91.6527090031447\\
31	91.8139054330902\\
32	91.6554692344955\\
33	91.5882647687284\\
34	91.7109096731098\\
35	91.5910750226091\\
36	91.6892642617331\\
37	91.688968341385\\
38	91.6133833586192\\
39	91.7606654135388\\
40	91.9215262074458\\
41	91.940598909092\\
42	91.8369679046133\\
43	92.1530196128713\\
44	92.0534559702625\\
45	92.0480749259097\\
46	92.0082895323198\\
47	92.1211015293084\\
48	92.095714172993\\
49	92.076977382975\\
50	92.0678504968164\\
50	93.3084935892051\\
49	93.191839821326\\
48	93.1193395904479\\
47	93.2552425567132\\
46	93.2605276719813\\
45	93.1132153966709\\
44	93.322888115759\\
43	93.16956103229\\
42	93.4318492996877\\
41	93.4895086177897\\
40	93.6698716420166\\
39	93.8307324359236\\
38	93.8704876091228\\
37	93.9024295080774\\
36	93.6333163834282\\
35	93.5164518591113\\
34	93.5041440903311\\
33	93.1966814678307\\
32	93.1832404429238\\
31	93.2936214486303\\
30	93.4010544377155\\
29	93.4360771509367\\
28	93.4167919900535\\
27	93.2567574333473\\
26	93.3521971765671\\
25	93.2510643079832\\
24	93.0274663524312\\
23	93.2495795218035\\
22	93.268600992231\\
21	92.9833106886173\\
20	92.952211249977\\
19	93.0271891910278\\
18	93.2133565314302\\
17	93.2302754812783\\
16	93.6772669215737\\
15	93.6278253323001\\
14	93.8506316295935\\
13	93.9675609861874\\
12	93.6694468950776\\
11	93.718832769499\\
10	93.6618189157303\\
9	93.5563812587571\\
8	93.5414285079626\\
7	93.0412644293627\\
6	92.6031188539121\\
5	89.8333470484373\\
4	86.2964485095507\\
3	81.335723623177\\
2	61.8417931005097\\
1	42.5896970244184\\
}--cycle;
\addlegendentry{kNN}

\addplot[mark=none, black, very thick, dotted, domain=0:50] {93.96};
\addlegendentry{94.0~\%}

\addplot [color=black, forget plot,dashdotted]
  table[row sep=crcr]{%
1	39.0860215053764\\
2	60.5645161290323\\
3	74.489247311828\\
4	78.494623655914\\
5	82.3924731182796\\
6	85.8333333333333\\
7	84.6774193548387\\
8	85.241935483871\\
9	84.8655913978495\\
10	84.3548387096774\\
11	84.489247311828\\
12	84.3548387096774\\
13	84.3817204301075\\
14	84.1129032258064\\
16	83.7903225806452\\
17	83.494623655914\\
18	83.494623655914\\
20	83.3870967741936\\
21	83.5215053763441\\
22	83.1989247311828\\
23	83.0913978494624\\
24	83.2795698924731\\
25	83.0645161290323\\
26	82.9569892473118\\
27	82.741935483871\\
28	82.4731182795699\\
29	82.3655913978495\\
30	82.3387096774194\\
31	81.9623655913978\\
32	81.6935483870968\\
33	81.9354838709677\\
34	81.7741935483871\\
35	81.6397849462366\\
36	81.2096774193548\\
37	81.1021505376344\\
38	81.2903225806452\\
39	81.0483870967742\\
40	81.0752688172043\\
41	80.994623655914\\
42	81.1021505376344\\
43	81.0483870967742\\
44	80.9677419354839\\
45	80.6720430107527\\
46	80.7258064516129\\
47	80.7258064516129\\
48	80.8602150537634\\
49	80.8870967741936\\
50	80.752688172043\\
};
\addplot [color=black!80!green, forget plot,dotted]
  table[row sep=crcr]{%
1	18.8978494623656\\
2	24.489247311828\\
3	43.4408602150538\\
4	53.010752688172\\
5	58.2258064516129\\
6	65.8333333333333\\
7	75.994623655914\\
8	81.3978494623656\\
9	79.9462365591398\\
10	79.6505376344086\\
11	80.241935483871\\
12	80.0806451612903\\
13	80.752688172043\\
14	81.1021505376344\\
15	81.4247311827957\\
16	81.0752688172043\\
17	80.483870967742\\
18	80.0806451612903\\
19	79.8655913978495\\
20	79.2204301075269\\
21	79.1666666666667\\
22	78.8978494623656\\
23	78.1720430107527\\
24	77.5268817204301\\
25	77.0698924731183\\
26	76.4516129032258\\
27	75\\
28	74.3548387096774\\
29	73.4139784946236\\
30	72.7688172043011\\
31	71.9086021505376\\
32	71.1827956989247\\
33	70.4301075268817\\
34	69.8387096774194\\
36	68.010752688172\\
37	67.1236559139785\\
38	66.505376344086\\
39	65.752688172043\\
40	65.1612903225806\\
42	63.7634408602151\\
43	63.1720430107527\\
44	62.8763440860215\\
45	62.1236559139785\\
46	61.8010752688172\\
47	61.3978494623656\\
48	60.752688172043\\
49	60.3225806451613\\
50	60.1881720430108\\
};
\addplot [color=blue, forget plot,dashed]
  table[row sep=crcr]{%
1	27.4462365591398\\
2	29.5430107526882\\
3	35.4838709677419\\
4	46.0215053763441\\
5	58.9784946236559\\
6	74.6236559139785\\
7	77.6612903225806\\
8	82.4731182795699\\
9	81.7204301075269\\
10	81.0752688172043\\
11	80.8870967741936\\
12	80.2688172043011\\
13	80.4569892473118\\
14	80.9677419354839\\
15	80.6720430107527\\
16	80.8870967741936\\
17	80.7795698924731\\
18	80.9408602150538\\
19	80.8333333333333\\
20	80.8602150537634\\
21	80.9677419354839\\
22	81.2365591397849\\
25	81.2365591397849\\
26	80.9408602150538\\
27	81.1021505376344\\
28	81.1290322580645\\
29	80.8064516129032\\
30	80.7258064516129\\
32	80.8870967741936\\
33	80.6989247311828\\
34	80.6989247311828\\
35	80.5645161290323\\
36	80.5376344086022\\
37	80.4032258064516\\
38	80.5645161290323\\
39	80.3763440860215\\
40	80.5376344086021\\
41	80.5913978494624\\
42	80.6182795698925\\
43	80.5376344086022\\
44	80.6720430107527\\
45	80.6182795698925\\
46	80.6451612903226\\
47	80.5913978494624\\
48	80.8333333333333\\
49	80.7795698924731\\
50	80.8333333333333\\
};
\addplot [color=red, forget plot]
  table[row sep=crcr]{%
1	39.7311827956989\\
2	60.6989247311828\\
3	79.8924731182796\\
4	84.4354838709677\\
5	88.252688172043\\
6	91.9086021505376\\
7	92.3924731182796\\
8	92.6881720430108\\
9	92.5806451612903\\
10	92.5268817204301\\
11	92.6881720430108\\
12	92.6881720430107\\
13	92.8494623655914\\
14	92.6881720430107\\
15	92.5806451612903\\
16	92.6344086021505\\
17	92.3387096774194\\
18	92.3924731182796\\
19	92.3924731182796\\
20	92.3118279569892\\
21	92.3387096774194\\
22	92.5\\
23	92.5537634408602\\
24	92.4462365591398\\
26	92.5\\
27	92.5\\
28	92.5806451612903\\
29	92.6075268817204\\
30	92.5268817204301\\
31	92.5537634408602\\
32	92.4193548387097\\
33	92.3924731182796\\
34	92.6075268817204\\
35	92.5537634408602\\
36	92.6612903225806\\
37	92.7956989247312\\
38	92.741935483871\\
39	92.7956989247312\\
40	92.7956989247312\\
42	92.6344086021505\\
44	92.6881720430108\\
45	92.5806451612903\\
47	92.6881720430108\\
48	92.6075268817204\\
49	92.6344086021505\\
50	92.6881720430108\\
};
\addplot [color=black, forget plot]
  table[row sep=crcr]{%
1	37.5145325910684\\
2	59.0420754043394\\
3	72.3526155236361\\
4	77.6081380404386\\
5	81.2881937684708\\
6	84.9040652678794\\
7	83.144925503899\\
8	83.2955615506214\\
9	82.4525919795687\\
10	82.1759264377248\\
11	82.3712986722121\\
12	82.186315131748\\
13	82.1418907029601\\
14	82.0017894523874\\
15	81.7541247429938\\
16	81.5642466788205\\
17	81.557553666634\\
18	82.1391610499485\\
19	81.64913758769\\
20	81.6774109014884\\
21	81.6869387366542\\
22	81.348669368557\\
23	81.1265481917629\\
24	81.0192663430561\\
25	80.9372054043892\\
26	80.9354910203741\\
27	80.4796346701698\\
28	79.7772288198204\\
29	79.6317719737152\\
30	79.794514245347\\
31	79.3073278621046\\
32	78.9498345425477\\
33	79.5966197060047\\
34	79.4237717566347\\
35	79.6223122756872\\
36	78.965818493543\\
37	78.9842018980185\\
38	79.3185891590438\\
39	79.1404476389008\\
40	79.1498919807022\\
41	78.9350557731852\\
42	79.3481311540043\\
43	79.2677901479178\\
44	79.1209070927616\\
45	78.9031522713589\\
46	79.0082138655129\\
47	79.0082138655129\\
48	79.0882630647659\\
49	79.1054855290721\\
50	78.8965837262415\\
nan	nan\\
50	82.6087926178445\\
49	82.668708019315\\
48	82.632167042761\\
47	82.4433990377129\\
45	82.4409337501465\\
44	82.8145767782062\\
43	82.8289840456306\\
42	82.8561699212645\\
41	83.0541915386428\\
39	82.9563265546476\\
38	83.2620560022465\\
37	83.2200991772503\\
36	83.4535363451667\\
35	83.6572576167859\\
34	84.1246153401395\\
33	84.2743480359308\\
32	84.4372622316458\\
31	84.617403320691\\
30	84.8829051094917\\
29	85.0994108219837\\
28	85.1690077393194\\
27	85.0042362975721\\
26	84.9784874742495\\
25	85.1918268536753\\
24	85.5398734418901\\
23	85.0562475071618\\
22	85.0491800938086\\
21	85.356072016034\\
20	85.0967826468987\\
19	85.2325828424175\\
18	84.8500862618795\\
16	86.0163984824698\\
15	86.1491010634579\\
14	86.2240169992255\\
13	86.621550157255\\
12	86.5233622876068\\
11	86.6071959514439\\
10	86.53375098163\\
9	87.2785908161302\\
8	87.1883094171205\\
7	86.2099132057784\\
6	86.7626013987873\\
5	83.4967524680884\\
4	79.3811092713893\\
3	76.6258791000198\\
2	62.0869568537252\\
1	40.6575104196843\\
};
\addplot [color=black!80!green, forget plot]
  table[row sep=crcr]{%
1	18.2631333896173\\
2	23.7691884281732\\
3	41.7449666348831\\
4	51.4415646373445\\
5	57.4455416519516\\
6	64.2333627806502\\
7	75.0041863748957\\
8	80.2789432586953\\
9	78.7923561273476\\
10	78.5981946703729\\
11	78.7332026666234\\
12	78.5737095249275\\
14	79.6248783156953\\
15	79.9727449703434\\
16	79.2885947279885\\
17	77.6335845176236\\
18	77.3800689412119\\
19	76.9510624255222\\
20	76.6217865541082\\
21	76.390876824185\\
22	76.1351068829174\\
23	75.0190223187284\\
24	74.2574080137181\\
25	73.8940359173227\\
26	73.4485505282217\\
27	71.5092556684891\\
28	71.0387303356152\\
29	70.2426759566657\\
30	69.8277555061058\\
31	69.0908259014467\\
32	68.4936157803391\\
33	68.2256425447773\\
34	67.6639469279375\\
35	66.7462335074165\\
36	65.8749665884313\\
37	64.851395111935\\
38	64.3645211246365\\
39	63.7718136100064\\
40	62.7618050085543\\
41	62.3168742000823\\
42	61.8469985352003\\
43	61.0746657883969\\
44	60.5514249128559\\
45	60.1336822017934\\
46	59.6602200493677\\
47	59.2110747903208\\
48	58.5258008701907\\
49	58.3901794262016\\
50	58.4141821557101\\
nan	nan\\
50	61.9621619303115\\
49	62.254981864121\\
48	62.9795754738953\\
47	63.5846241344104\\
46	63.9419304882667\\
45	64.1136296261636\\
44	65.2012632591871\\
43	65.2694202331085\\
42	65.6798831852298\\
41	66.6078569827134\\
40	67.560775636607\\
39	67.7335627340796\\
38	68.6462315635355\\
36	70.1465387879127\\
35	71.1032288581749\\
34	72.0134724269012\\
33	72.6345725089861\\
32	73.8719756175104\\
31	74.7263783996286\\
30	75.7098789024963\\
29	76.5852810325816\\
28	77.6709470837397\\
27	78.4907443315109\\
26	79.4546752782299\\
25	80.2457490289139\\
24	80.796355427142\\
23	81.325063702777\\
22	81.6605920418138\\
21	81.9424565091483\\
20	81.8190736609456\\
19	82.7801203701768\\
18	82.7812213813688\\
17	83.3341574178603\\
16	82.8619429064202\\
15	82.876717395248\\
14	82.5794227595735\\
13	82.4108785498162\\
12	81.5875807976531\\
11	81.7506683011185\\
10	80.7028805984443\\
9	81.100116990932\\
8	82.5167556660359\\
7	76.9850609369323\\
6	67.4333038860164\\
5	59.0060712512742\\
4	54.5799407389996\\
3	45.1367537952245\\
2	25.2093061954827\\
1	19.5325655351138\\
};
\addplot [color=blue, forget plot]
  table[row sep=crcr]{%
1	26.4567117032631\\
2	28.8167066620373\\
3	34.5287183868581\\
4	44.7620745403094\\
5	57.447770056182\\
6	73.3019333581786\\
7	76.9475312877768\\
8	81.875037756417\\
9	81.0220225238297\\
10	80.4711772834881\\
11	80.4578301230465\\
12	79.6875874110097\\
13	79.3792032152296\\
14	80.083296565787\\
15	79.6916891113543\\
16	79.9113606767268\\
17	79.6671407690022\\
18	79.9198855368369\\
19	80.3815124646782\\
20	80.3431348370518\\
21	80.486867101613\\
22	80.7463821900857\\
23	80.7282884300756\\
24	80.7282884300755\\
25	80.5947668609509\\
26	80.2594735561486\\
27	80.4533592265512\\
28	80.2144609765482\\
29	79.7087356457476\\
30	79.7584379454061\\
31	79.6964611385988\\
32	79.508507375811\\
33	79.3697058800993\\
34	79.6465817671471\\
35	79.499375658271\\
36	79.4991159262322\\
37	79.4321294694522\\
38	79.5786494182316\\
39	79.5423606420705\\
40	79.5300180207531\\
41	79.7928252270154\\
42	79.8675139761415\\
43	79.4481782726258\\
44	79.6963069132859\\
45	79.7297583934926\\
46	79.8111778463716\\
47	79.55636433827\\
48	79.7896092119184\\
49	79.9900981099146\\
50	79.9695554294054\\
nan	nan\\
50	81.6971112372613\\
49	81.5690416750316\\
48	81.8770574547482\\
47	81.6264313606548\\
46	81.4791447342736\\
45	81.5068007462924\\
44	81.6477791082194\\
43	81.6270905445785\\
42	81.3690451636435\\
41	81.3899704719093\\
40	81.5452507964511\\
39	81.2103275299725\\
38	81.5503828398329\\
37	81.374322143451\\
36	81.5761528909722\\
35	81.6296565997935\\
34	81.7512676952185\\
33	82.0281435822663\\
32	82.2656861725761\\
31	81.9164420872077\\
30	81.6931749578197\\
29	81.9041675800589\\
28	82.0436035395809\\
27	81.7509418487176\\
26	81.622246873959\\
25	81.878351418619\\
24	81.7448298494943\\
23	81.7448298494944\\
22	81.7267360894841\\
21	81.4486167693548\\
20	81.377295270475\\
19	81.2851542019885\\
18	81.9618348932706\\
17	81.891999015944\\
16	81.8628328716603\\
15	81.6523969101511\\
14	81.8521873051807\\
13	81.5347752793941\\
12	80.8500469975924\\
11	81.3163634253406\\
10	81.6793603509205\\
9	82.418837691224\\
8	83.0711988027228\\
7	78.3750493573845\\
6	75.9453784697785\\
5	60.5092191911299\\
4	47.2809362123788\\
3	36.4390235486258\\
2	30.269314843339\\
1	28.4357614150165\\
};
\addplot [color=red, forget plot]
  table[row sep=crcr]{%
1	36.8726685669795\\
2	59.5560563618559\\
3	78.4492226133821\\
5	86.6720292956487\\
6	91.2140854471632\\
7	91.7436818071964\\
8	91.8349155780589\\
9	91.6049090638236\\
10	91.3919445251299\\
11	91.6575113165225\\
12	91.7068971909439\\
13	91.7313637449954\\
14	91.525712456428\\
15	91.5334649902805\\
16	91.5915502827274\\
17	91.4471438735604\\
18	91.5715897051289\\
19	91.7577570455313\\
20	91.6714446640015\\
21	91.6941086662214\\
22	91.731399007769\\
23	91.8579473599169\\
24	91.8650067658484\\
25	91.6951722511565\\
26	91.6478028234329\\
27	91.7432425666527\\
28	91.7444983325271\\
29	91.7789766125042\\
30	91.6527090031447\\
31	91.8139054330902\\
32	91.6554692344955\\
33	91.5882647687284\\
34	91.7109096731098\\
35	91.5910750226091\\
36	91.6892642617331\\
37	91.688968341385\\
38	91.6133833586192\\
39	91.7606654135388\\
40	91.9215262074458\\
41	91.940598909092\\
42	91.8369679046133\\
43	92.1530196128713\\
44	92.0534559702625\\
45	92.0480749259097\\
46	92.0082895323198\\
47	92.1211015293084\\
49	92.076977382975\\
50	92.0678504968164\\
nan	nan\\
50	93.3084935892051\\
49	93.191839821326\\
48	93.1193395904479\\
47	93.2552425567132\\
46	93.2605276719813\\
45	93.1132153966709\\
44	93.322888115759\\
43	93.16956103229\\
42	93.4318492996877\\
41	93.4895086177897\\
40	93.6698716420166\\
39	93.8307324359236\\
37	93.9024295080774\\
36	93.6333163834282\\
35	93.5164518591113\\
34	93.5041440903311\\
33	93.1966814678307\\
32	93.1832404429238\\
30	93.4010544377155\\
29	93.4360771509367\\
28	93.4167919900535\\
27	93.2567574333473\\
26	93.3521971765671\\
25	93.2510643079832\\
24	93.0274663524312\\
23	93.2495795218035\\
22	93.268600992231\\
21	92.9833106886173\\
20	92.952211249977\\
19	93.0271891910278\\
18	93.2133565314302\\
17	93.2302754812783\\
16	93.6772669215737\\
15	93.6278253323001\\
14	93.8506316295935\\
13	93.9675609861874\\
12	93.6694468950776\\
11	93.718832769499\\
10	93.6618189157303\\
9	93.5563812587571\\
8	93.5414285079626\\
7	93.0412644293627\\
6	92.6031188539121\\
5	89.8333470484373\\
4	86.2964485095507\\
3	81.335723623177\\
2	61.8417931005097\\
1	42.5896970244184\\
};
\end{axis}
\end{tikzpicture}%

%% file: Figures/test_error_k.tex
\definecolor{mycolor1}{rgb}{0.80000,0.80000,0.90000}%
\begin{tikzpicture}

\begin{axis}[%
width=7cm,
height=5cm,
scale only axis,
unbounded coords=jump,
xmin=0,
xmax=50,
xlabel style={font=\color{white!15!black}},
xlabel={$r$},
ymin=4,
ymax=6,
ylabel style={font=\color{white!15!black}},
ylabel={Test error in \%},
axis background/.style={fill=white},
title={(a)},
legend columns = 1,
legend style={legend cell align=left, align=left, draw=white!15!black, font = \footnotesize}
]

\addplot[mark=none, black, very thick, dotted, domain=0:50, forget plot] {4.52508960573481};

\addplot[mark=none, black, very thick, dotted, domain=0:50, forget plot] {5.08960573476708};

\draw[<->, black, thick] (3,4.52508960573481) to (3, 5.08960573476708);
\node[anchor = north west] at (1,4.4) {{\footnotesize $\approx 0.56\,\%$}};

\addplot [mark=o,color=red]
  table[row sep=crcr]{%
1	47.7508960573482\\
2	30.1075268817207\\
3	14.6505376344087\\
4	11.2634408602152\\
5	8.01971326164882\\
6	5.37634408602156\\
7	5.08960573476708\\
8	4.88351254480291\\
9	4.85663082437281\\
10	4.92831541218643\\
11	4.8476702508961\\
13	4.85663082437281\\
14	4.97311827956994\\
15	4.99103942652334\\
16	4.97311827956994\\
17	5.03584229390686\\
18	4.98207885304664\\
20	5.02688172043015\\
21	4.97311827956994\\
22	4.93727598566313\\
23	4.91935483870972\\
24	4.95519713261653\\
25	4.93727598566313\\
26	4.96415770609324\\
27	5.00896057347675\\
28	5.00000000000005\\
29	4.95519713261653\\
31	5.01792114695346\\
32	5.08960573476708\\
33	5.06272401433696\\
34	4.97311827956994\\
35	5.02688172043015\\
36	5.00896057347675\\
37	4.96415770609324\\
38	4.96415770609324\\
39	4.89247311827962\\
40	4.96415770609324\\
41	4.94623655913983\\
42	5.00000000000005\\
44	4.99103942652334\\
45	5.03584229390686\\
46	5.00896057347676\\
48	5.02688172043015\\
50	4.97311827956994\\
};
\addlegendentry{$k=50$}

\addplot [mark=triangle, color=blue]
  table[row sep=crcr]{%
1	48.3691756272407\\
2	29.9910394265236\\
3	14.8118279569894\\
4	11.1827956989248\\
5	8.01075268817213\\
6	5.37634408602156\\
7	5.05376344086027\\
8	4.75806451612907\\
9	4.74910394265238\\
10	4.63261648745524\\
11	4.58781362007173\\
12	4.62365591397854\\
13	4.61469534050184\\
14	4.65949820788535\\
15	4.67741935483875\\
16	4.75806451612907\\
17	4.74014336917567\\
18	4.74910394265238\\
19	4.820788530466\\
20	4.81182795698929\\
21	4.8387096774194\\
22	4.88351254480291\\
23	4.86559139784951\\
24	4.87455197132621\\
25	4.85663082437281\\
26	4.78494623655919\\
28	4.86559139784951\\
29	4.93727598566313\\
30	4.87455197132621\\
31	4.8387096774194\\
32	4.77598566308248\\
33	4.76702508960578\\
34	4.79390681003589\\
35	4.74910394265238\\
36	4.79390681003589\\
37	4.79390681003589\\
38	4.8476702508961\\
39	4.81182795698929\\
40	4.8297491039427\\
41	4.91039426523302\\
42	4.91935483870972\\
43	4.87455197132621\\
44	4.88351254480292\\
45	4.8476702508961\\
46	4.88351254480291\\
49	4.92831541218643\\
50	4.90143369175632\\
};
\addlegendentry{$k=100$}

\addplot [mark=square, color=black!50!green]
  table[row sep=crcr]{%
1	48.028673835126\\
2	29.7401433691759\\
3	14.6774193548388\\
4	11.0931899641578\\
5	7.96594982078862\\
6	5.41218637992836\\
7	5.02688172043015\\
8	4.66845878136206\\
9	4.69534050179216\\
10	4.58781362007173\\
11	4.57885304659503\\
12	4.58781362007173\\
13	4.52508960573481\\
15	4.68637992831546\\
16	4.65949820788535\\
17	4.70430107526886\\
18	4.73118279569897\\
19	4.74014336917567\\
20	4.72222222222227\\
21	4.79390681003589\\
22	4.72222222222227\\
23	4.76702508960578\\
24	4.79390681003589\\
25	4.8387096774194\\
26	4.74910394265238\\
27	4.61469534050184\\
28	4.67741935483875\\
29	4.62365591397854\\
30	4.58781362007173\\
31	4.65949820788535\\
32	4.68637992831546\\
33	4.69534050179216\\
34	4.64157706093194\\
35	4.70430107526886\\
38	4.76702508960578\\
39	4.820788530466\\
40	4.8476702508961\\
41	4.8297491039427\\
43	4.87455197132621\\
44	4.8387096774194\\
45	4.89247311827962\\
46	4.89247311827962\\
47	4.91935483870972\\
48	4.89247311827962\\
49	4.89247311827962\\
50	4.91935483870972\\
};
\addlegendentry{$k=1000$}

\end{axis}
\end{tikzpicture}%

%% file: Figures/test_error_balancing_split.tex
\definecolor{mycolor1}{rgb}{0.80000,0.80000,0.90000}%
\begin{tikzpicture}

\begin{axis}[%
width=7cm,
height=5cm,
scale only axis,
unbounded coords=jump,
xmin=0,
xmax=50,
xlabel style={font=\color{white!15!black}},
xlabel={$r$},
ymin=4,
ymax=6,
ylabel style={font=\color{white!15!black}},
axis background/.style={fill=white},
title={(b)},
legend columns = 2,
legend style={legend cell align=left, align=left, draw=white!15!black, font = \footnotesize}
]

\addplot[mark=none, black, very thick, dotted, domain=0:50, forget plot] {4.489247311828};

\addplot[mark=none, black, very thick, dotted, domain=0:50, forget plot] {5.14336917562729};

\draw[<->, black, thick] (3,4.489247311828) to (3, 5.14336917562729);
\node[anchor = north west] at (1,4.4) {{\footnotesize $\approx 0.65\,\%$}};

\addplot [mark=o, color=red]
  table[row sep=crcr]{%
1	47.7508960573482\\
2	30.1075268817207\\
3	14.6505376344087\\
4	11.2634408602152\\
5	8.01971326164882\\
6	5.37634408602156\\
7	5.08960573476708\\
8	4.88351254480291\\
9	4.85663082437281\\
10	4.92831541218643\\
11	4.8476702508961\\
13	4.85663082437281\\
14	4.97311827956994\\
15	4.99103942652334\\
16	4.97311827956994\\
17	5.03584229390686\\
18	4.98207885304664\\
20	5.02688172043015\\
21	4.97311827956994\\
22	4.93727598566313\\
23	4.91935483870972\\
24	4.95519713261653\\
25	4.93727598566313\\
26	4.96415770609324\\
27	5.00896057347675\\
28	5.00000000000005\\
29	4.95519713261653\\
31	5.01792114695346\\
32	5.08960573476708\\
33	5.06272401433696\\
34	4.97311827956994\\
35	5.02688172043015\\
36	5.00896057347675\\
37	4.96415770609324\\
38	4.96415770609324\\
39	4.89247311827962\\
40	4.96415770609324\\
41	4.94623655913983\\
42	5.00000000000005\\
44	4.99103942652334\\
45	5.03584229390686\\
46	5.00896057347676\\
48	5.02688172043015\\
50	4.97311827956994\\
};
\addlegendentry{Balancing split 1}

\addplot [mark=triangle, color=blue]
  table[row sep=crcr]{%
1	48.1272401433698\\
2	27.7867383512547\\
3	14.8387096774195\\
4	10.3405017921148\\
5	7.07885304659505\\
6	4.96415770609324\\
7	4.70430107526886\\
8	4.56093189964162\\
9	4.50716845878141\\
10	4.489247311828\\
11	4.55197132616492\\
12	4.55197132616492\\
13	4.56989247311832\\
14	4.60573476702513\\
15	4.62365591397854\\
16	4.59677419354843\\
17	4.61469534050184\\
18	4.52508960573481\\
19	4.57885304659503\\
20	4.55197132616492\\
21	4.60573476702513\\
22	4.58781362007173\\
23	4.65053763440865\\
24	4.57885304659503\\
25	4.53405017921151\\
26	4.58781362007173\\
28	4.57885304659503\\
29	4.75806451612907\\
31	4.71326164874556\\
32	4.73118279569897\\
33	4.68637992831546\\
34	4.70430107526887\\
35	4.74910394265238\\
36	4.65053763440865\\
38	4.69534050179216\\
39	4.66845878136206\\
40	4.65949820788535\\
41	4.69534050179216\\
42	4.71326164874556\\
43	4.76702508960578\\
44	4.75806451612907\\
45	4.72222222222227\\
46	4.70430107526887\\
48	4.80286738351259\\
49	4.77598566308248\\
50	4.78494623655919\\
};
\addlegendentry{Balancing split 2}

\addplot [mark=square, color=black!50!green]
  table[row sep=crcr]{%
1	48.4139784946243\\
2	28.1451612903228\\
3	14.97311827957\\
4	11.3709677419356\\
5	7.40143369175635\\
6	5.41218637992836\\
7	5.20609318996421\\
8	4.85663082437281\\
9	4.87455197132621\\
10	4.92831541218643\\
11	4.86559139784951\\
12	4.93727598566313\\
13	4.88351254480291\\
15	4.65053763440865\\
16	4.71326164874556\\
17	4.62365591397854\\
18	4.72222222222227\\
19	4.79390681003589\\
22	4.77598566308248\\
23	4.820788530466\\
24	4.89247311827962\\
26	4.74014336917567\\
28	4.72222222222227\\
29	4.74014336917567\\
31	4.65949820788535\\
32	4.57885304659503\\
36	4.67741935483875\\
37	4.64157706093194\\
38	4.65053763440865\\
39	4.59677419354843\\
40	4.67741935483875\\
41	4.71326164874556\\
42	4.68637992831546\\
43	4.68637992831546\\
44	4.64157706093194\\
46	4.74910394265238\\
47	4.74014336917567\\
48	4.71326164874556\\
49	4.74014336917567\\
50	4.80286738351259\\
};
\addlegendentry{Balancing split 3}

\addplot [mark=diamond, color=black]
  table[row sep=crcr]{%
1	48.602150537635\\
2	28.5842293906812\\
3	15.6182795698926\\
4	11.9086021505378\\
5	7.79569892473126\\
6	5.43010752688178\\
7	5.38530465949826\\
9	5.1881720430108\\
10	5.01792114695346\\
11	5.08960573476708\\
12	5.11648745519718\\
13	4.96415770609324\\
14	4.93727598566313\\
15	4.96415770609324\\
16	5.02688172043015\\
18	5.08960573476708\\
19	5.08960573476708\\
20	5.12544802867389\\
21	5.07168458781367\\
22	5.12544802867389\\
23	5.14336917562729\\
24	5.14336917562729\\
25	5.06272401433696\\
26	5.05376344086027\\
27	5.07168458781367\\
28	5.07168458781367\\
29	5.09856630824378\\
30	5.02688172043015\\
31	5.02688172043015\\
32	5.13440860215059\\
33	5.08960573476708\\
34	5.10752688172048\\
35	5.03584229390686\\
36	5.02688172043015\\
38	5.05376344086027\\
39	5.00000000000005\\
40	5.01792114695346\\
41	5.00896057347675\\
42	5.05376344086027\\
43	5.05376344086027\\
44	5.01792114695346\\
45	5.02688172043015\\
47	5.11648745519718\\
48	5.04480286738356\\
49	5.10752688172048\\
50	5.10752688172048\\
};
\addlegendentry{Balancing split 4}
\end{axis}
\end{tikzpicture}%

%% file: Figures/conf_mat_person1_person2.tex
\begin{tikzpicture}
\definecolor{oceanboatblue}{rgb}{0.0, 0.47, 0.75}
\definecolor{cornflowerblue}{rgb}{0.39, 0.58, 0.93}
\pgfplotsset{compat=1.15,
width=4.7cm,
height=4.7cm,
xticklabels={A, B, C},
xtick={0,...,2},
xlabel={Person 2},
ylabel={Person 1},
yticklabels={A, B, C},
ytick={0,...,2},
    colormap={mycolormap}{
        color(0)=(white) color(13000)=(blue!75!black)
    }
}
\begin{axis}[enlargelimits=false,
colormap name=mycolormap,
        point meta min=0,
        point meta max=13000,
        nodes near coords={\pgfmathprintnumber\pgfplotspointmeta\,\%},
        nodes near coords black white/.style={
            small value/.style={
                font=\footnotesize,
                yshift=-7pt,
                text=black,
            },
            large value/.style={
                font=\footnotesize,
                yshift=-7pt,
                text=white,
            },
            every node near coord/.style={
                check for zero/.code={
                    \pgfmathfloatifflags{\pgfplotspointmeta}{-1}{
                        \pgfkeys{/tikz/coordinate}
                    }{
                        \begingroup
                        \pgfkeys{/pgf/fpu}
                        \pgfmathparse{\pgfplotspointmeta<#1}
                        \global\let\result=\pgfmathresult
                        \endgroup
                        \pgfmathfloatcreate{1}{1.0}{0}
                        \let\ONE=\pgfmathresult
                        \ifx\result\ONE
                            \pgfkeysalso{/pgfplots/small value}
                        \else
                            \pgfkeysalso{/pgfplots/large value}
                        \fi
                    }
                },
                check for zero,
            },
        },
        nodes near coords black white=7500,
    ]
\addplot [matrix plot,
nodes near coords={\pgfmathprintnumber\pgfplotspointmeta},mark=none,
mesh/cols=3,
point meta=explicit,
] coordinates {
(0,0) [9118] (1,0) [240] (2,0) [4]
(0,1) [588] (1,1) [2318] (2,1) [819]
(0,2) [76] (1,2) [740] (2,2) [12977]
};
\end{axis}
\end{tikzpicture}

%% file: Figures/exemplary_conf_mat_knn1_hat.tex
\begin{tikzpicture}
\definecolor{oceanboatblue}{rgb}{0.0, 0.47, 0.75}
\definecolor{cornflowerblue}{rgb}{0.39, 0.58, 0.93}
\pgfplotsset{compat=1.15,
width=4.7cm,
height=4.7cm,
xticklabels={A, B, C},
xtick={0,...,2},
xlabel={kNN},
ylabel={Person 1},
yticklabels={A, B, C},
ytick={0,...,2},
    colormap={mycolormap}{
        color(0)=(white) color(2720)=(blue!75!black)
    }
}
\begin{axis}[enlargelimits=false,
colormap name=mycolormap,
        point meta min=0,
        point meta max=2720,
        nodes near coords={\pgfmathprintnumber\pgfplotspointmeta\,\%},
        nodes near coords black white/.style={
            small value/.style={
                font=\footnotesize,
                yshift=-7pt,
                text=black,
            },
            large value/.style={
                font=\footnotesize,
                yshift=-7pt,
                text=white,
            },
            every node near coord/.style={
                check for zero/.code={
                    \pgfmathfloatifflags{\pgfplotspointmeta}{-1}{
                        \pgfkeys{/tikz/coordinate}
                    }{
                        \begingroup
                        \pgfkeys{/pgf/fpu}
                        \pgfmathparse{\pgfplotspointmeta<#1}
                        \global\let\result=\pgfmathresult
                        \endgroup
                        \pgfmathfloatcreate{1}{1.0}{0}
                        \let\ONE=\pgfmathresult
                        \ifx\result\ONE
                            \pgfkeysalso{/pgfplots/small value}
                        \else
                            \pgfkeysalso{/pgfplots/large value}
                        \fi
                    }
                },
                check for zero,
            },
        },
        nodes near coords black white=1300,
    ]
\addplot [matrix plot,
nodes near coords={\pgfmathprintnumber\pgfplotspointmeta},mark=none,
mesh/cols=3,
point meta=explicit,
] coordinates {
(0,0) [1815] (1,0) [58] (2,0) [0]
(0,1) [42] (1,1) [649] (2,1) [54]
(0,2) [1] (1,2) [37] (2,2) [2720]
};
\end{axis}
\end{tikzpicture}

%% file: Figures/2024_B3-01_R060_ESA0_thresh_u_-4_thresh_o_4.tex
\begin{tikzpicture}

\begin{axis}[%
width=2.432in,
height=1.397in,
at={(0.419in,0.362in)},
scale only axis,
xmin=15,
xmax=240,
xlabel style={font=\color{white!15!black}},
xlabel={Printing velocity in \SI{}{\meter\per\minute}},
ymin=0,
ymax=100,
ylabel style={font=\color{white!15!black}},
ylabel={Tonal value in \%},
axis background/.style={fill=white},
title style={font=\bfseries},
legend pos=south east
]
\addplot [color=black]
  table[row sep=crcr]{%
15	22.5\\
30	28.75\\
60	47.5\\
90	57.5\\
120	61.25\\
180	72.5\\
240	72.5\\
};
\addlegendentry{kNN}
\addplot [color=black, forget plot]
  table[row sep=crcr]{%
15	17.5\\
30	26.25\\
60	42.5\\
90	52.5\\
120	57.5\\
240	67.5\\
};
\addplot [color=red, dashed]
  table[row sep=crcr]{%
15	22.5\\
30	32.5\\
60	47.5\\
90	57.5\\
120	62.5\\
180	67.5\\
240	77.5\\
};
\addlegendentry{CNN}
\addplot [color=red, dashed, forget plot]
  table[row sep=crcr]{%
15	17.5\\
30	27.5\\
60	42.5\\
90	52.5\\
120	57.5\\
240	67.5\\
};
\node[] at (50,80) {\scriptsize{\circled{C}}};
\node[] at (150,62) {\scriptsize{\circledFill{B}}};
\node[] at (100,20) {\scriptsize{\circled{A}}};
\end{axis}
\end{tikzpicture}%

%% file: Figures/2024_B3-05_R060_ESA0_thresh_u_-4_thresh_o_4.tex
\begin{tikzpicture}

\begin{axis}[%
width=2.432in,
height=1.397in,
at={(0.419in,0.362in)},
scale only axis,
xmin=15,
xmax=240,
xlabel style={font=\color{white!15!black}},
xlabel={Printing velocity in \SI{}{\meter\per\minute}},
ymin=0,
ymax=100,
ylabel style={font=\color{white!15!black}},
ylabel={Tonal value in \%},
axis background/.style={fill=white},
title style={font=\bfseries}
]
\addplot [color=black]
  table[row sep=crcr]{%
15	22.5\\
30	35\\
60	27.5\\
90	27.5\\
120	15\\
180	40\\
240	47.5\\
};
\addlegendentry{kNN}
\addplot [color=black, forget plot]
  table[row sep=crcr]{%
15	2.5\\
30	20\\
60	17.5\\
90	17.5\\
120	12.5\\
240	32.5\\
};
\addplot [color=red, dashed, forget plot]
  table[row sep=crcr]{%
15	22.5\\
60	22.5\\
90	27.5\\
120	37.5\\
240	47.5\\
};
\addplot [color=red, dashed]
  table[row sep=crcr]{%
15	2.5\\
30	2.5\\
60	7.5\\
90	7.5\\
120	17.5\\
180	27.5\\
240	32.5\\
};
\addlegendentry{CNN}
\node[] at (50,80) {\scriptsize{\circled{C}}};
\node[] at (100,20) {\scriptsize{\circledFill{B}}};
\node[] at (200,15) {\scriptsize{\circled{A}}};
\end{axis}
\end{tikzpicture}%

%% file: 4_Conclusion.tex
\section{Conclusions and discussion}
\label{sec:Conclusions}

In this work, we developed an automated pattern classification algorithm for gravure printed patterns based on methods from supervised machine learning and reduced-order modeling. We have demonstrated that the SVD works very well for dimensionality reduction of the labeled images from the HYPA-p dataset~\cite{HYPA-p}. However, the size of the dataset presents a challenge and requires the use of a randomized SVD (rSVD) approach. Because of spatially repeating patterns in the dataset, we applied the FFT to the data before performing the rSVD and thus we were able to reduce the number of rSVD modes to only $r=7$. This is very low rank, compared to the dimensionality of the original data. We trained selected machine learning classifiers on the low-rank data and found that a kNN model works best. It achieves a minimum test error of \SI{3}{\percent}. Compared to the benchmark from human observers with a classification error of \SI{10}{\percent}, the performance of the kNN is regarded as very good. Compared to previous approaches that use CNNs for pattern classification, the kNN model provides a higher degree of interpretability and the test error is even comparable. For example, a CNN trained on the HYPA-p dataset by \cite{thesis_Pauline} has a test error of \SI{6}{\percent}. Furthermore, we found that data balancing increased the minimum test error of the kNN model to \SI{5}{\percent}, but increases the recall of the mixed pattern class, which is hardest to classify, from \SI{90}{\percent} to \SI{94}{\percent}. Data normalization, however, does not have a appreciable impact on the performance of the kNN model. 

Finally, we have demonstrated how the trained kNN model can be used for automated classification of unlabeled images from the HYPA-p dataset and for creation of fluid splitting regime maps. These regime maps provide new insights into the fluid dynamics of fluid splitting in gravure printing. Moreover, the regime maps can be used to predict how the printed patterns from a certain set of printing process parameters will look. With the trained kNN model, all unlabeled images of the HYPA-p dataset can be classified without the need for further time-consuming manual classification by a human observer. The trained kNN model can even be used to refine the labels from the human observer by checking all the images where the prediction of the kNN differs from the ground truth label. 

However, the regime maps created from the HYPA-p dataset cannot yet provide a full understanding of the physics of fluid splitting in gravure printing. To achieve this, we additionally need improved theoretical models or numerical simulations of the flow dynamics in the printing nip, for example as given in \cite{rothmann2024gravure,rieckmann2023pressure}. With these models it might be possible to create a digital twin of a gravure printing machine and a 3D-simulation of the gravure printing process. Another limitation of our approach is that our kNN model would have to be retrained for new sets of printing parameters (e.g. different printing inks, different raster frequencies, etc.). Patterns that were created with completely different printing process parameters than those from the HYPA-p dataset are likely to be misclassified by our trained kNN model since our kNN model would have to perform extrapolation. In future research, it would be interesting to test our model on other datasets than the HYPA-p dataset. Additionally, unsupervised learning methods could be tested on the HYPA-p dataset. The key advantage would be that the time-consuming labeling step would not be needed. Additionally, more than three fluid splitting classes could be considered. Tuning of hyperparameters, e.\,g. number of neighbors and distance metric, could also improve the performance of the trained kNN model.

Printing machines are currently controlled by trained machine operators. Future directions of our research could include a real-time implementation of our developed pattern classification algorithm. By employing pattern classification online during the printing process, the results of the classification could be used to tune the printing process parameters to reduce production waste to a minimum.

%% file: Acknowledgements.tex
\FloatBarrier
\section*{Acknowledgments}

PRB kindly acknowledges the financial support by the Deutsche Forschungsgemeinschaft (DFG, German Research Foundation) -- Project-ID 265191195 -- Collaborative Research Center 1194 (CRC 1194) ‘Interaction between Transport and Wetting Processes,’ project C01.

SLB acknowledges support from the National Science Foundation AI Institute in Dynamic Systems (grant number 2112085) and from The Boeing Company.

%% file: Appendix.tex
\FloatBarrier
\section{Appendix}

\begin{figure}[ht]
\centering
\begin{subfigure}[b]{1\textwidth}
\centering
\input{Figures/Dots/dots}
\caption{Dot patterns}
\end{subfigure}
\par \bigskip
\begin{subfigure}[b]{1\textwidth}
\centering
\input{Figures/Mixed/mixed}
\caption{Mixed patterns}
\end{subfigure}
\par \bigskip
\begin{subfigure}[b]{1\textwidth}
\centering
\input{Figures/Fingers/fingers}
\caption{Finger patterns}
\end{subfigure}
\caption{Examples for dot pattern images (a), mixed pattern images (b) and finger pattern images (c) from the labeled dataset. Each printed example has a size of \SI{260}{\px}~x~\SI{260}{\px} (\SI{2.75}{\mm}~x~\SI{2.75}{\mm}).}
\label{fig:patterns_many}
\end{figure}
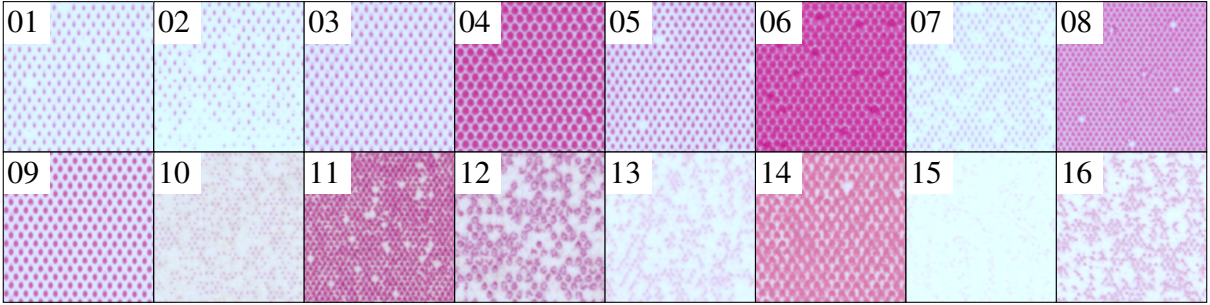
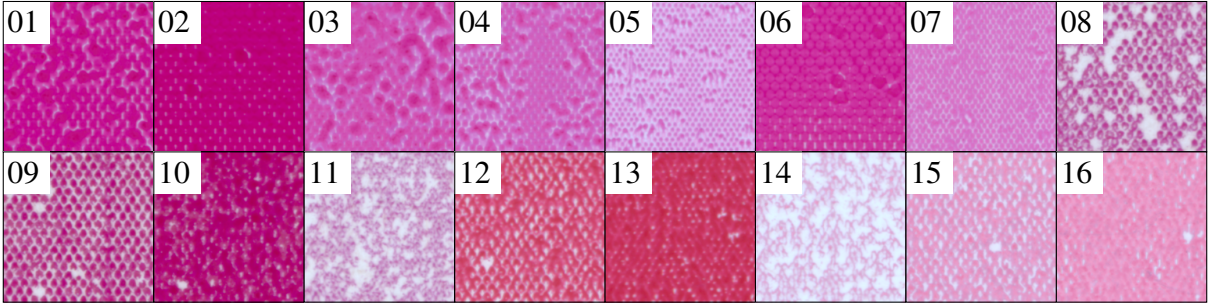
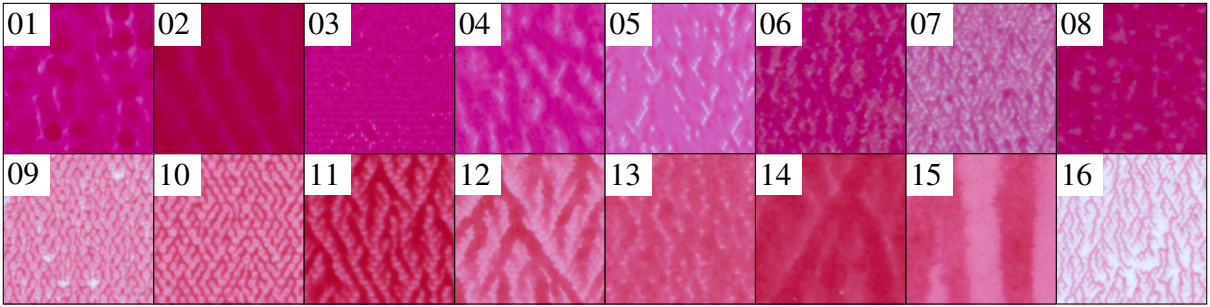

\begin{figure}[t]
\input{Figures/Modes/modes_U.tex}
\caption{First 64 modes ($\mathbf{U}$). The modes were obtained by using SVD on the complete dataset $\mathbf{X_{\text{c}}}$, which consists of \num{26880} images.}
\label{fig:Modes_U}
\end{figure}
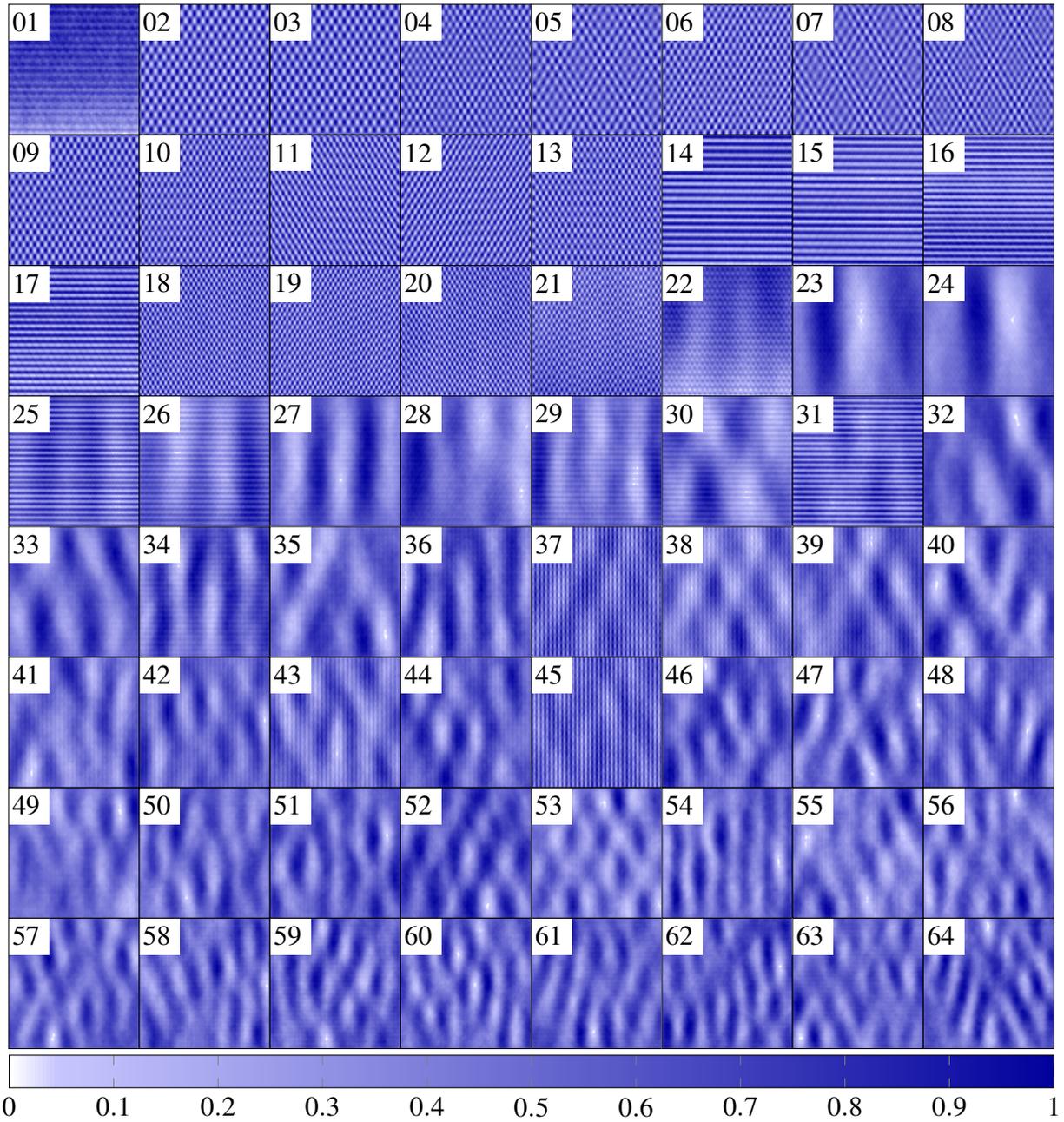

\begin{figure}[t]
\input{Figures/Modes/modes_Uhat.tex}
\caption{First 64 FFT modes ($\mathbf{\hat{U}}$). For interpretability, the FFT modes are displayed in the spatial domain; this is achieved by taking the magnitude of the inverse FFT with phase zero for each FFT mode. The FFT modes were obtained by using SVD on the complete, FFT-transformed dataset $|\mathbf{\hat{X}_{\text{c}}}|$, which consists of \num{26880} images.}
\label{fig:Modes_Uhat}
\end{figure}
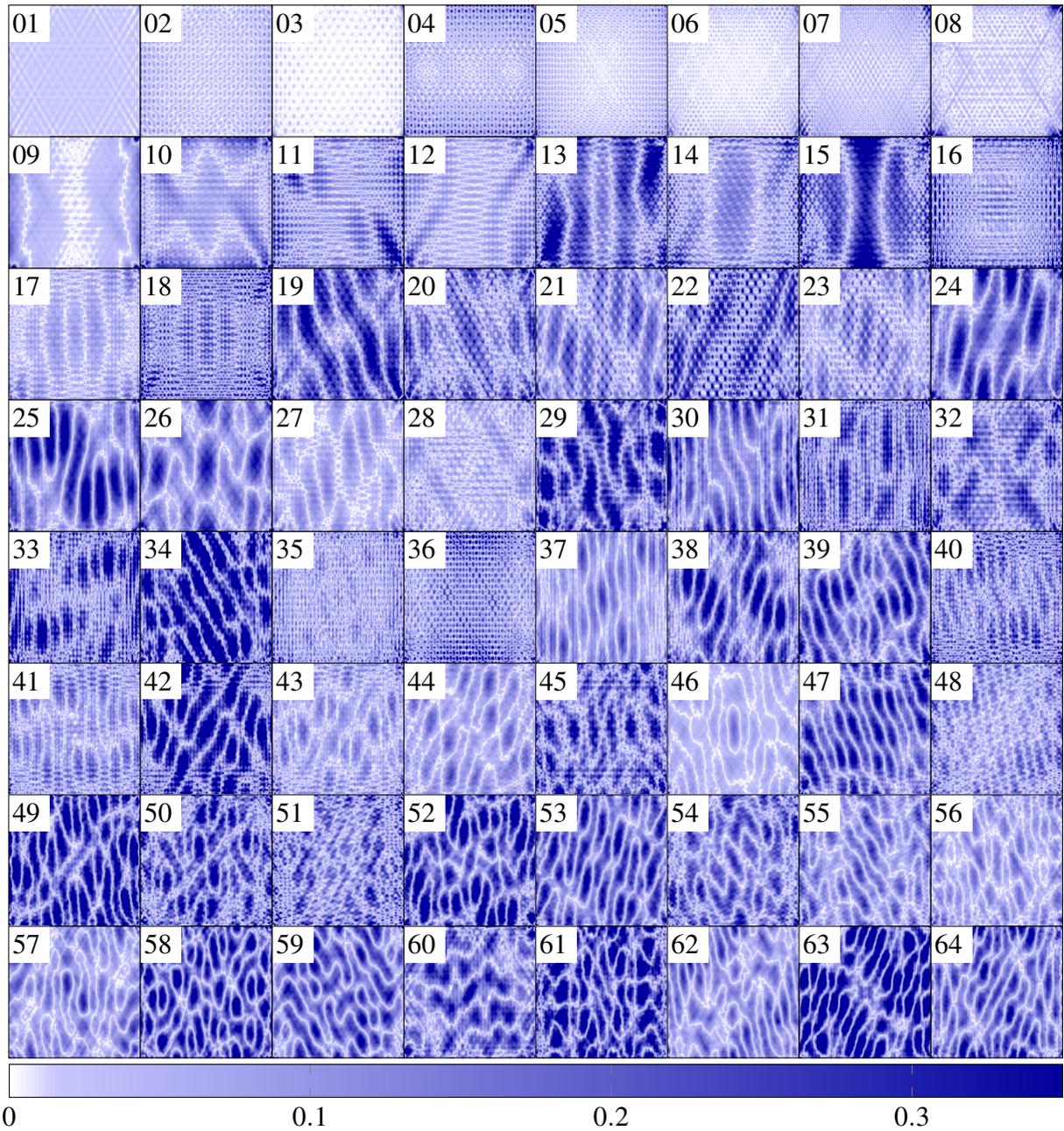

\begin{figure}[t]
    \begin{subfigure}[b]{1\textwidth}
    \input{Figures/Modes_k50/rSVDmodes50_Uhat}
    \caption{$k=50$}
    \end{subfigure}
    \par\bigskip
    \begin{subfigure}[b]{1\textwidth}
    \input{Figures/Modes_k1000/rSVDmodes1000_Uhat}
    \caption{$k=1000$}
    \end{subfigure}  
    \caption{First 50 FFT modes from $\mathbf{\hat{U}}$. For interpretability, the FFT modes are displayed in the spatial domain; this is achieved by taking the magnitude of the inverse FFT with phase zero for each FFT mode. The FFT modes were obtained by using rSVD with different target ranks $k$ on an unbalanced, FFT-transformed, train dataset $|\mathbf{\hat{X}_{\text{train}}}|$. It is $k=50$ (a) and $k=1000$ (b). $|\mathbf{\hat{X}_{\text{train}}}|$ consists of \num{21504} images, which equals \SI{80}{\percent} of the complete dataset.}
    \label{fig:target_rank}
\end{figure}
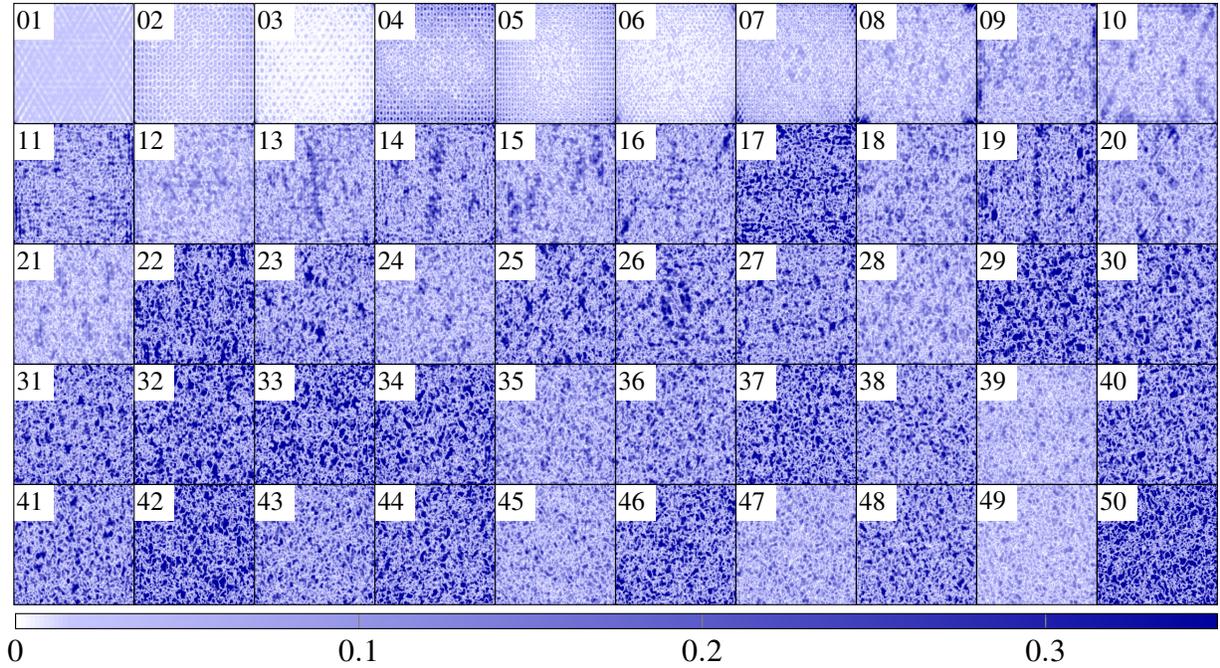
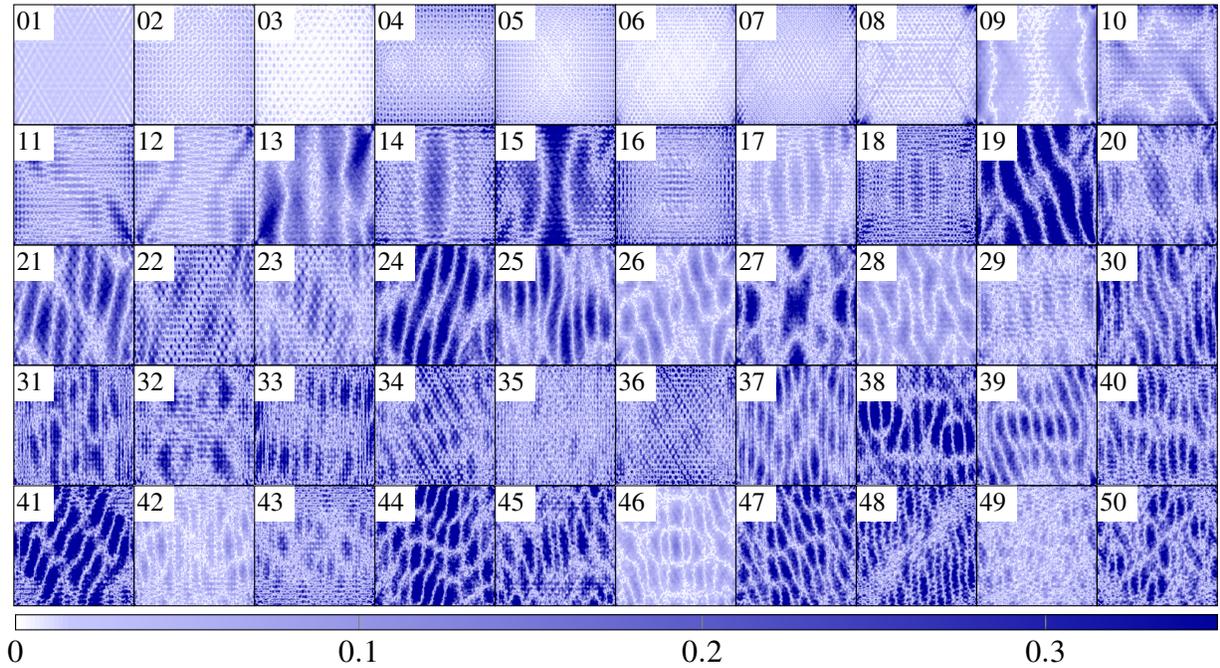

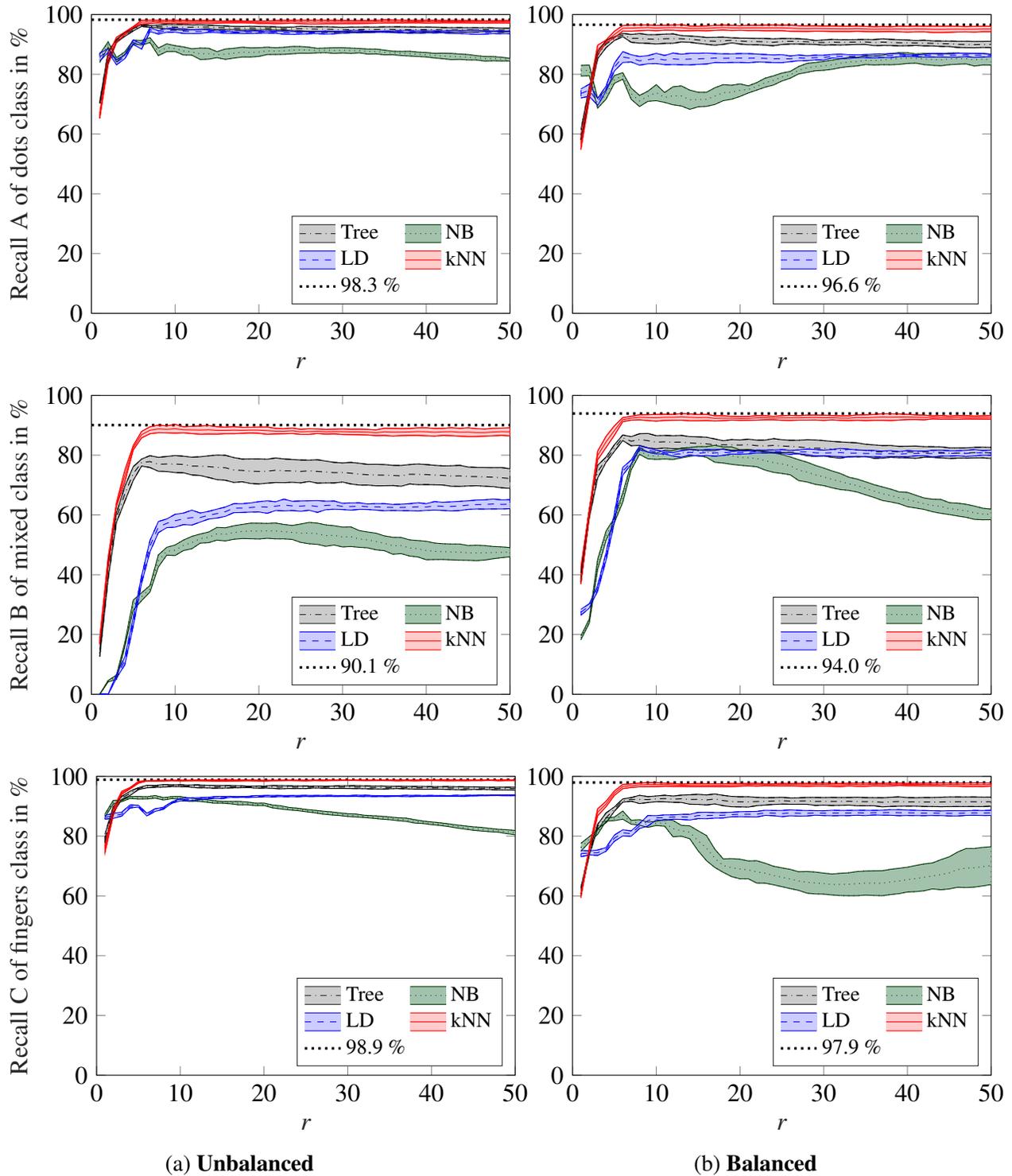
\begin{figure}[t]
\centering
\begin{subfigure}[b]{0.48\textwidth}
    \centering    \input{Figures/balancing0_norm0_FFTtransformed_50rSVDmodes_recall_dots_class}
\end{subfigure}
\hfill
\begin{subfigure}[b]{0.48\textwidth}
    \centering    
    \input{Figures/balancing1_norm0_FFTtransformed_50rSVDmodes_recall_dots_class}
\end{subfigure}
\begin{subfigure}[b]{0.48\textwidth}
    \centering    \input{Figures/balancing0_norm0_FFTtransformed_50rSVDmodes_recall_mixed_class}
\end{subfigure}
\hfill
\begin{subfigure}[b]{0.48\textwidth}
    \centering    
    \input{Figures/balancing1_norm0_FFTtransformed_50rSVDmodes_recall_mixed_class}
\end{subfigure}
\begin{subfigure}[b]{0.48\textwidth}
    \centering    \input{Figures/balancing0_norm0_FFTtransformed_50rSVDmodes_recall_fingers_class}
    \caption{\textbf{Unbalanced}}
\end{subfigure}
\hfill
\begin{subfigure}[b]{0.48\textwidth}
    \centering    
    \input{Figures/balancing1_norm0_FFTtransformed_50rSVDmodes_recall_fingers_class}
    \caption{\textbf{Balanced}}
\end{subfigure}
\caption{Recall A, B and C over $r$. Comparison of the classifier results for an unbalanced (a) and a balanced dataset (b). Unbalanced dataset means \SI{34.8}{\percent} dots, \SI{13.9}{\percent} mixed, \SI{51.3}{\percent} fingers. Balanced data set means \SI{33.3}{\percent} dots, \SI{33.3}{\percent} mixed, \SI{33.3}{\percent} fingers. In all figures, a non-normalized, FFT-transformed dataset was used and the target rank of the rSVD was chosen as $k=50$.}
\label{fig:Recall_Appendix}
\end{figure}

%% file: Figures/Dots/dots.tex
\begin{tikzpicture}

\begin{axis}[%
width=0.787in,
height=0.787in,
at={(0in,0.788in)},
scale only axis,
axis on top,
xmin=0,
xmax=260,
xtick={\empty},
tick align=outside,
y dir=reverse,
ymin=0,
ymax=260,
ytick={\empty},
ticks=none
]
\addplot [forget plot] graphics [xmin=0.5, xmax=260.5, ymin=0.5, ymax=260.5] {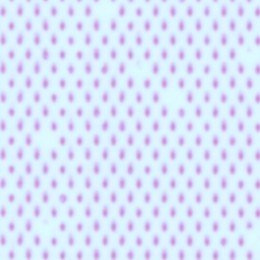};
\node[fill=white, align=center]
at (axis cs:35,35) {01};
\end{axis}

\begin{axis}[%
width=0.787in,
height=0.787in,
at={(0.788in,0.788in)},
scale only axis,
axis on top,
xmin=0,
xmax=260,
xtick={\empty},
tick align=outside,
y dir=reverse,
ymin=0,
ymax=260,
ytick={\empty},
ticks=none
]
\addplot [forget plot] graphics [xmin=0.5, xmax=260.5, ymin=0.5, ymax=260.5] {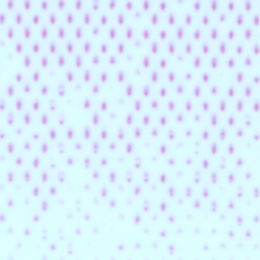};
\node[fill=white, align=center]
at (axis cs:35,35) {02};
\end{axis}

\begin{axis}[%
width=0.787in,
height=0.787in,
at={(1.575in,0.788in)},
scale only axis,
axis on top,
xmin=0,
xmax=260,
xtick={\empty},
tick align=outside,
y dir=reverse,
ymin=0,
ymax=260,
ytick={\empty},
ticks=none
]
\addplot [forget plot] graphics [xmin=0.5, xmax=260.5, ymin=0.5, ymax=260.5] {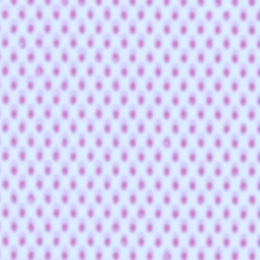};
\node[fill=white, align=center]
at (axis cs:35,35) {03};
\end{axis}

\begin{axis}[%
width=0.787in,
height=0.787in,
at={(2.363in,0.788in)},
scale only axis,
axis on top,
xmin=0,
xmax=260,
xtick={\empty},
tick align=outside,
y dir=reverse,
ymin=0,
ymax=260,
ytick={\empty},
ticks=none
]
\addplot [forget plot] graphics [xmin=0.5, xmax=260.5, ymin=0.5, ymax=260.5] {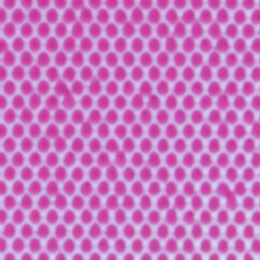};
\node[fill=white, align=center]
at (axis cs:35,35) {04};
\end{axis}

\begin{axis}[%
width=0.787in,
height=0.787in,
at={(3.15in,0.788in)},
scale only axis,
axis on top,
xmin=0,
xmax=260,
xtick={\empty},
tick align=outside,
y dir=reverse,
ymin=0,
ymax=260,
ytick={\empty},
ticks=none
]
\addplot [forget plot] graphics [xmin=0.5, xmax=260.5, ymin=0.5, ymax=260.5] {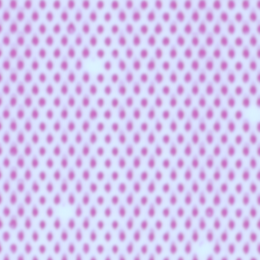};
\node[fill=white, align=center]
at (axis cs:35,35) {05};
\end{axis}

\begin{axis}[%
width=0.787in,
height=0.787in,
at={(3.937in,0.788in)},
scale only axis,
axis on top,
xmin=0,
xmax=260,
xtick={\empty},
tick align=outside,
y dir=reverse,
ymin=0,
ymax=260,
ytick={\empty},
ticks=none
]
\addplot [forget plot] graphics [xmin=0.5, xmax=260.5, ymin=0.5, ymax=260.5] {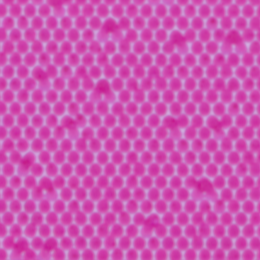};
\node[fill=white, align=center]
at (axis cs:35,35) {06};
\end{axis}

\begin{axis}[%
width=0.787in,
height=0.787in,
at={(4.725in,0.788in)},
scale only axis,
axis on top,
xmin=0,
xmax=260,
xtick={\empty},
tick align=outside,
y dir=reverse,
ymin=0,
ymax=260,
ytick={\empty},
ticks=none
]
\addplot [forget plot] graphics [xmin=0.5, xmax=260.5, ymin=0.5, ymax=260.5] {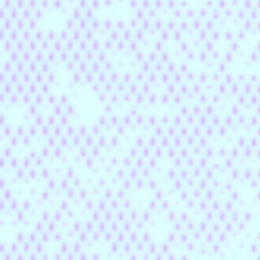};
\node[fill=white, align=center]
at (axis cs:35,35) {07};
\end{axis}

\begin{axis}[%
width=0.787in,
height=0.787in,
at={(5.512in,0.788in)},
scale only axis,
axis on top,
xmin=0,
xmax=260,
xtick={\empty},
tick align=outside,
y dir=reverse,
ymin=0,
ymax=260,
ytick={\empty},
ticks=none
]
\addplot [forget plot] graphics [xmin=0.5, xmax=260.5, ymin=0.5, ymax=260.5] {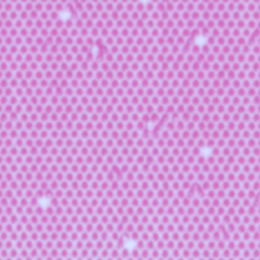};
\node[fill=white, align=center]
at (axis cs:35,35) {08};
\end{axis}

\begin{axis}[%
width=0.787in,
height=0.787in,
at={(0in,0in)},
scale only axis,
axis on top,
xmin=0,
xmax=260,
xtick={\empty},
tick align=outside,
y dir=reverse,
ymin=0,
ymax=260,
ytick={\empty},
ticks=none
]
\addplot [forget plot] graphics [xmin=0.5, xmax=260.5, ymin=0.5, ymax=260.5] {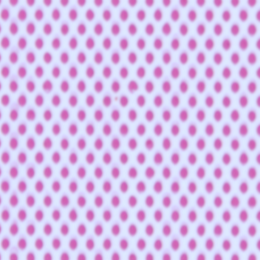};
\node[fill=white, align=center]
at (axis cs:35,35) {09};
\end{axis}

\begin{axis}[%
width=0.787in,
height=0.787in,
at={(0.788in,0in)},
scale only axis,
axis on top,
xmin=0,
xmax=260,
xtick={\empty},
tick align=outside,
y dir=reverse,
ymin=0,
ymax=260,
ytick={\empty},
ticks=none
]
\addplot [forget plot] graphics [xmin=0.5, xmax=260.5, ymin=0.5, ymax=260.5] {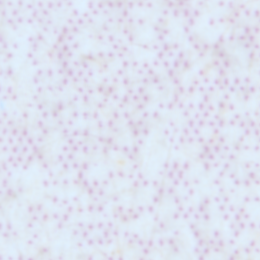};
\node[fill=white, align=center]
at (axis cs:35,35) {10};
\end{axis}

\begin{axis}[%
width=0.787in,
height=0.787in,
at={(1.575in,0in)},
scale only axis,
axis on top,
xmin=0,
xmax=260,
xtick={\empty},
tick align=outside,
y dir=reverse,
ymin=0,
ymax=260,
ytick={\empty},
ticks=none
]
\addplot [forget plot] graphics [xmin=0.5, xmax=260.5, ymin=0.5, ymax=260.5] {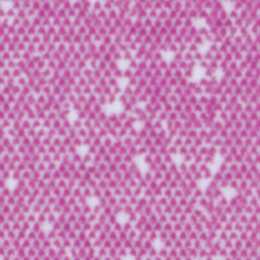};
\node[fill=white, align=center]
at (axis cs:35,35) {11};
\end{axis}

\begin{axis}[%
width=0.787in,
height=0.787in,
at={(2.363in,0in)},
scale only axis,
axis on top,
xmin=0,
xmax=260,
xtick={\empty},
tick align=outside,
y dir=reverse,
ymin=0,
ymax=260,
ytick={\empty},
ticks=none
]
\addplot [forget plot] graphics [xmin=0.5, xmax=260.5, ymin=0.5, ymax=260.5] {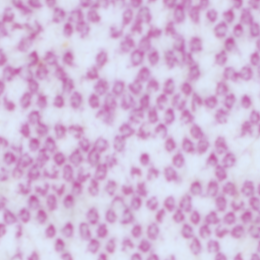};
\node[fill=white, align=center]
at (axis cs:35,35) {12};
\end{axis}

\begin{axis}[%
width=0.787in,
height=0.787in,
at={(3.15in,0in)},
scale only axis,
axis on top,
xmin=0,
xmax=260,
xtick={\empty},
tick align=outside,
y dir=reverse,
ymin=0,
ymax=260,
ytick={\empty},
ticks=none
]
\addplot [forget plot] graphics [xmin=0.5, xmax=260.5, ymin=0.5, ymax=260.5] {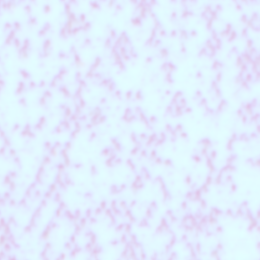};
\node[fill=white, align=center]
at (axis cs:35,35) {13};
\end{axis}

\begin{axis}[%
width=0.787in,
height=0.787in,
at={(3.937in,0in)},
scale only axis,
axis on top,
xmin=0,
xmax=260,
xtick={\empty},
tick align=outside,
y dir=reverse,
ymin=0,
ymax=260,
ytick={\empty},
ticks=none
]
\addplot [forget plot] graphics [xmin=0.5, xmax=260.5, ymin=0.5, ymax=260.5] {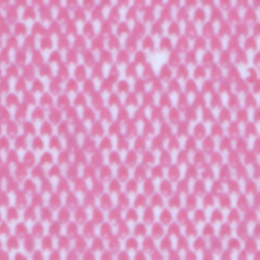};
\node[fill=white, align=center]
at (axis cs:35,35) {14};
\end{axis}

\begin{axis}[%
width=0.787in,
height=0.787in,
at={(4.725in,0in)},
scale only axis,
axis on top,
xmin=0,
xmax=260,
xtick={\empty},
tick align=outside,
y dir=reverse,
ymin=0,
ymax=260,
ytick={\empty},
ticks=none
]
\addplot [forget plot] graphics [xmin=0.5, xmax=260.5, ymin=0.5, ymax=260.5] {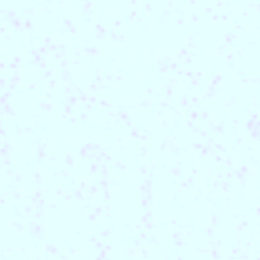};
\node[fill=white, align=center]
at (axis cs:35,35) {15};
\end{axis}

\begin{axis}[%
width=0.787in,
height=0.787in,
at={(5.512in,0in)},
scale only axis,
axis on top,
xmin=0,
xmax=260,
xtick={\empty},
tick align=outside,
y dir=reverse,
ymin=0,
ymax=260,
ytick={\empty},
ticks=none
]
\addplot [forget plot] graphics [xmin=0.5, xmax=260.5, ymin=0.5, ymax=260.5] {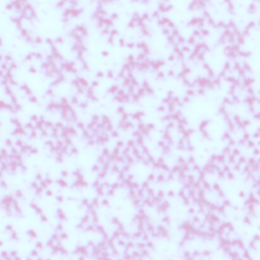};
\node[fill=white, align=center]
at (axis cs:35,35) {16};
\end{axis}
\end{tikzpicture}%

%% file: Figures/Mixed/mixed.tex
\begin{tikzpicture}

\begin{axis}[%
width=0.787in,
height=0.787in,
at={(0in,0.788in)},
scale only axis,
axis on top,
xmin=0,
xmax=260,
xtick={\empty},
tick align=outside,
y dir=reverse,
ymin=0,
ymax=260,
ytick={\empty},
ticks=none
]
\addplot [forget plot] graphics [xmin=0.5, xmax=260.5, ymin=0.5, ymax=260.5] {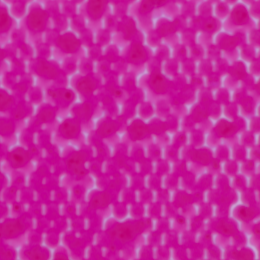};
\node[fill=white, align=center]
at (axis cs:35,35) {01};
\end{axis}

\begin{axis}[%
width=0.787in,
height=0.787in,
at={(0.788in,0.788in)},
scale only axis,
axis on top,
xmin=0,
xmax=260,
xtick={\empty},
tick align=outside,
y dir=reverse,
ymin=0,
ymax=260,
ytick={\empty},
ticks=none
]
\addplot [forget plot] graphics [xmin=0.5, xmax=260.5, ymin=0.5, ymax=260.5] {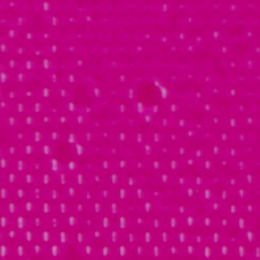};
\node[fill=white, align=center]
at (axis cs:35,35) {02};
\end{axis}

\begin{axis}[%
width=0.787in,
height=0.787in,
at={(1.575in,0.788in)},
scale only axis,
axis on top,
xmin=0,
xmax=260,
xtick={\empty},
tick align=outside,
y dir=reverse,
ymin=0,
ymax=260,
ytick={\empty},
ticks=none
]
\addplot [forget plot] graphics [xmin=0.5, xmax=260.5, ymin=0.5, ymax=260.5] {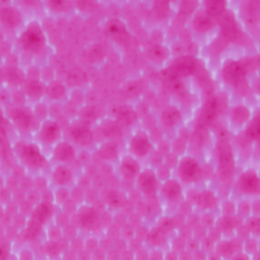};
\node[fill=white, align=center]
at (axis cs:35,35) {03};
\end{axis}

\begin{axis}[%
width=0.787in,
height=0.787in,
at={(2.363in,0.788in)},
scale only axis,
axis on top,
xmin=0,
xmax=260,
xtick={\empty},
tick align=outside,
y dir=reverse,
ymin=0,
ymax=260,
ytick={\empty},
ticks=none
]
\addplot [forget plot] graphics [xmin=0.5, xmax=260.5, ymin=0.5, ymax=260.5] {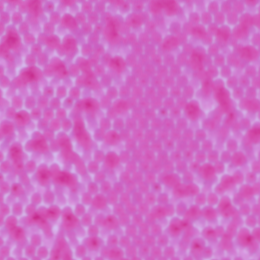};
\node[fill=white, align=center]
at (axis cs:35,35) {04};
\end{axis}

\begin{axis}[%
width=0.787in,
height=0.787in,
at={(3.15in,0.788in)},
scale only axis,
axis on top,
xmin=0,
xmax=260,
xtick={\empty},
tick align=outside,
y dir=reverse,
ymin=0,
ymax=260,
ytick={\empty},
ticks=none
]
\addplot [forget plot] graphics [xmin=0.5, xmax=260.5, ymin=0.5, ymax=260.5] {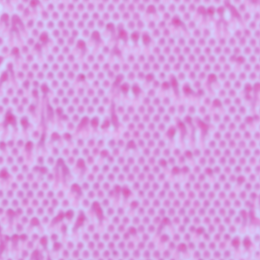};
\node[fill=white, align=center]
at (axis cs:35,35) {05};
\end{axis}

\begin{axis}[%
width=0.787in,
height=0.787in,
at={(3.937in,0.788in)},
scale only axis,
axis on top,
xmin=0,
xmax=260,
xtick={\empty},
tick align=outside,
y dir=reverse,
ymin=0,
ymax=260,
ytick={\empty},
ticks=none
]
\addplot [forget plot] graphics [xmin=0.5, xmax=260.5, ymin=0.5, ymax=260.5] {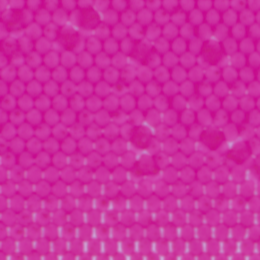};
\node[fill=white, align=center]
at (axis cs:35,35) {06};
\end{axis}

\begin{axis}[%
width=0.787in,
height=0.787in,
at={(4.725in,0.788in)},
scale only axis,
axis on top,
xmin=0,
xmax=260,
xtick={\empty},
tick align=outside,
y dir=reverse,
ymin=0,
ymax=260,
ytick={\empty},
ticks=none
]
\addplot [forget plot] graphics [xmin=0.5, xmax=260.5, ymin=0.5, ymax=260.5] {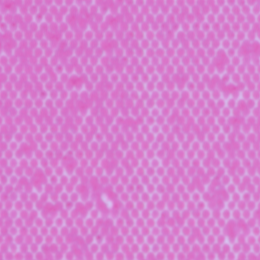};
\node[fill=white, align=center]
at (axis cs:35,35) {07};
\end{axis}

\begin{axis}[%
width=0.787in,
height=0.787in,
at={(5.512in,0.788in)},
scale only axis,
axis on top,
xmin=0,
xmax=260,
xtick={\empty},
tick align=outside,
y dir=reverse,
ymin=0,
ymax=260,
ytick={\empty},
ticks=none
]
\addplot [forget plot] graphics [xmin=0.5, xmax=260.5, ymin=0.5, ymax=260.5] {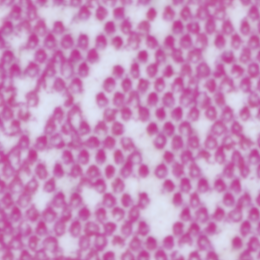};
\node[fill=white, align=center]
at (axis cs:35,35) {08};
\end{axis}

\begin{axis}[%
width=0.787in,
height=0.787in,
at={(0in,0in)},
scale only axis,
axis on top,
xmin=0,
xmax=260,
xtick={\empty},
tick align=outside,
y dir=reverse,
ymin=0,
ymax=260,
ytick={\empty},
ticks=none
]
\addplot [forget plot] graphics [xmin=0.5, xmax=260.5, ymin=0.5, ymax=260.5] {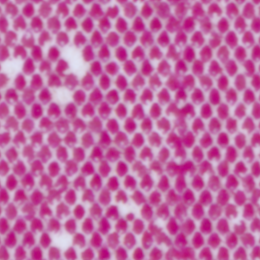};
\node[fill=white, align=center]
at (axis cs:35,35) {09};
\end{axis}

\begin{axis}[%
width=0.787in,
height=0.787in,
at={(0.788in,0in)},
scale only axis,
axis on top,
xmin=0,
xmax=260,
xtick={\empty},
tick align=outside,
y dir=reverse,
ymin=0,
ymax=260,
ytick={\empty},
ticks=none
]
\addplot [forget plot] graphics [xmin=0.5, xmax=260.5, ymin=0.5, ymax=260.5] {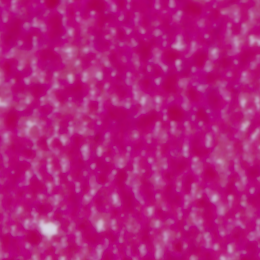};
\node[fill=white, align=center]
at (axis cs:35,35) {10};
\end{axis}

\begin{axis}[%
width=0.787in,
height=0.787in,
at={(1.575in,0in)},
scale only axis,
axis on top,
xmin=0,
xmax=260,
xtick={\empty},
tick align=outside,
y dir=reverse,
ymin=0,
ymax=260,
ytick={\empty},
ticks=none
]
\addplot [forget plot] graphics [xmin=0.5, xmax=260.5, ymin=0.5, ymax=260.5] {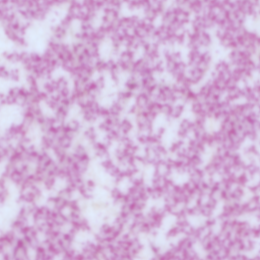};
\node[fill=white, align=center]
at (axis cs:35,35) {11};
\end{axis}

\begin{axis}[%
width=0.787in,
height=0.787in,
at={(2.363in,0in)},
scale only axis,
axis on top,
xmin=0,
xmax=260,
xtick={\empty},
tick align=outside,
y dir=reverse,
ymin=0,
ymax=260,
ytick={\empty},
ticks=none
]
\addplot [forget plot] graphics [xmin=0.5, xmax=260.5, ymin=0.5, ymax=260.5] {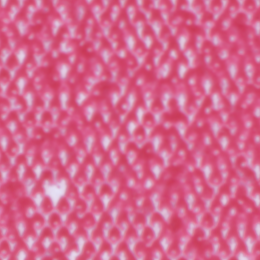};
\node[fill=white, align=center]
at (axis cs:35,35) {12};
\end{axis}

\begin{axis}[%
width=0.787in,
height=0.787in,
at={(3.15in,0in)},
scale only axis,
axis on top,
xmin=0,
xmax=260,
xtick={\empty},
tick align=outside,
y dir=reverse,
ymin=0,
ymax=260,
ytick={\empty},
ticks=none
]
\addplot [forget plot] graphics [xmin=0.5, xmax=260.5, ymin=0.5, ymax=260.5] {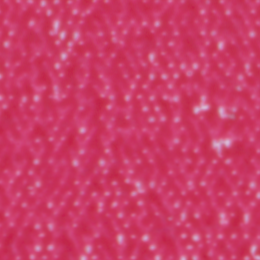};
\node[fill=white, align=center]
at (axis cs:35,35) {13};
\end{axis}

\begin{axis}[%
width=0.787in,
height=0.787in,
at={(3.937in,0in)},
scale only axis,
axis on top,
xmin=0,
xmax=260,
xtick={\empty},
tick align=outside,
y dir=reverse,
ymin=0,
ymax=260,
ytick={\empty},
ticks=none
]
\addplot [forget plot] graphics [xmin=0.5, xmax=260.5, ymin=0.5, ymax=260.5] {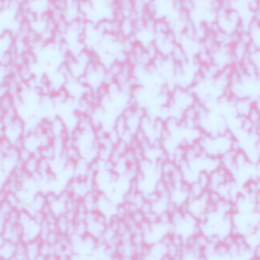};
\node[fill=white, align=center]
at (axis cs:35,35) {14};
\end{axis}

\begin{axis}[%
width=0.787in,
height=0.787in,
at={(4.725in,0in)},
scale only axis,
axis on top,
xmin=0,
xmax=260,
xtick={\empty},
tick align=outside,
y dir=reverse,
ymin=0,
ymax=260,
ytick={\empty},
ticks=none
]
\addplot [forget plot] graphics [xmin=0.5, xmax=260.5, ymin=0.5, ymax=260.5] {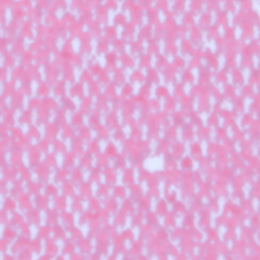};
\node[fill=white, align=center]
at (axis cs:35,35) {15};
\end{axis}

\begin{axis}[%
width=0.787in,
height=0.787in,
at={(5.512in,0in)},
scale only axis,
axis on top,
xmin=0,
xmax=260,
xtick={\empty},
tick align=outside,
y dir=reverse,
ymin=0,
ymax=260,
ytick={\empty},
ticks=none
]
\addplot [forget plot] graphics [xmin=0.5, xmax=260.5, ymin=0.5, ymax=260.5] {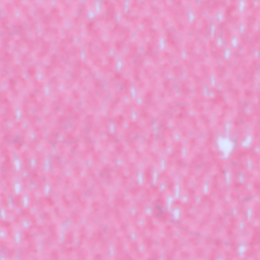};
\node[fill=white, align=center]
at (axis cs:35,35) {16};
\end{axis}
\end{tikzpicture}%

%% file: Figures/Fingers/fingers.tex
\begin{tikzpicture}

\begin{axis}[%
width=0.787in,
height=0.787in,
at={(0in,0.788in)},
scale only axis,
axis on top,
xmin=0,
xmax=260,
xtick={\empty},
tick align=outside,
y dir=reverse,
ymin=0,
ymax=260,
ytick={\empty},
ticks=none
]
\addplot [forget plot] graphics [xmin=0.5, xmax=260.5, ymin=0.5, ymax=260.5] {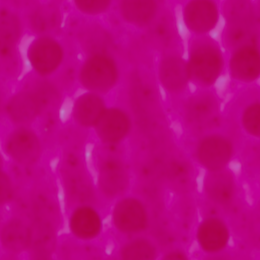};
\node[fill=white, align=center]
at (axis cs:35,35) {01};
\end{axis}

\begin{axis}[%
width=0.787in,
height=0.787in,
at={(0.788in,0.788in)},
scale only axis,
axis on top,
xmin=0,
xmax=260,
xtick={\empty},
tick align=outside,
y dir=reverse,
ymin=0,
ymax=260,
ytick={\empty},
ticks=none
]
\addplot [forget plot] graphics [xmin=0.5, xmax=260.5, ymin=0.5, ymax=260.5] {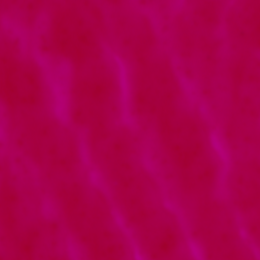};
\node[fill=white, align=center]
at (axis cs:35,35) {02};
\end{axis}

\begin{axis}[%
width=0.787in,
height=0.787in,
at={(1.575in,0.788in)},
scale only axis,
axis on top,
xmin=0,
xmax=260,
xtick={\empty},
tick align=outside,
y dir=reverse,
ymin=0,
ymax=260,
ytick={\empty},
ticks=none
]
\addplot [forget plot] graphics [xmin=0.5, xmax=260.5, ymin=0.5, ymax=260.5] {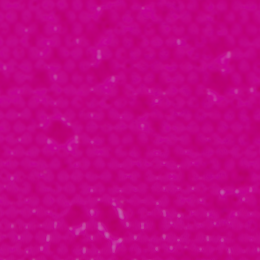};
\node[fill=white, align=center]
at (axis cs:35,35) {03};
\end{axis}

\begin{axis}[%
width=0.787in,
height=0.787in,
at={(2.363in,0.788in)},
scale only axis,
axis on top,
xmin=0,
xmax=260,
xtick={\empty},
tick align=outside,
y dir=reverse,
ymin=0,
ymax=260,
ytick={\empty},
ticks=none
]
\addplot [forget plot] graphics [xmin=0.5, xmax=260.5, ymin=0.5, ymax=260.5] {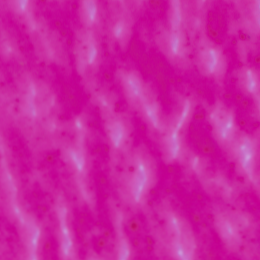};
\node[fill=white, align=center]
at (axis cs:35,35) {04};
\end{axis}

\begin{axis}[%
width=0.787in,
height=0.787in,
at={(3.15in,0.788in)},
scale only axis,
axis on top,
xmin=0,
xmax=260,
xtick={\empty},
tick align=outside,
y dir=reverse,
ymin=0,
ymax=260,
ytick={\empty},
ticks=none
]
\addplot [forget plot] graphics [xmin=0.5, xmax=260.5, ymin=0.5, ymax=260.5] {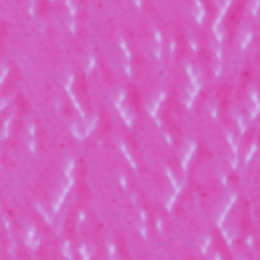};
\node[fill=white, align=center]
at (axis cs:35,35) {05};
\end{axis}

\begin{axis}[%
width=0.787in,
height=0.787in,
at={(3.937in,0.788in)},
scale only axis,
axis on top,
xmin=0,
xmax=260,
xtick={\empty},
tick align=outside,
y dir=reverse,
ymin=0,
ymax=260,
ytick={\empty},
ticks=none
]
\addplot [forget plot] graphics [xmin=0.5, xmax=260.5, ymin=0.5, ymax=260.5] {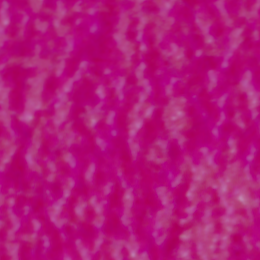};
\node[fill=white, align=center]
at (axis cs:35,35) {06};
\end{axis}

\begin{axis}[%
width=0.787in,
height=0.787in,
at={(4.725in,0.788in)},
scale only axis,
axis on top,
xmin=0,
xmax=260,
xtick={\empty},
tick align=outside,
y dir=reverse,
ymin=0,
ymax=260,
ytick={\empty},
ticks=none
]
\addplot [forget plot] graphics [xmin=0.5, xmax=260.5, ymin=0.5, ymax=260.5] {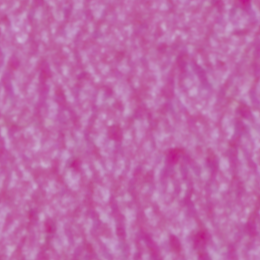};
\node[fill=white, align=center]
at (axis cs:35,35) {07};
\end{axis}

\begin{axis}[%
width=0.787in,
height=0.787in,
at={(5.512in,0.788in)},
scale only axis,
axis on top,
xmin=0,
xmax=260,
xtick={\empty},
tick align=outside,
y dir=reverse,
ymin=0,
ymax=260,
ytick={\empty},
ticks=none
]
\addplot [forget plot] graphics [xmin=0.5, xmax=260.5, ymin=0.5, ymax=260.5] {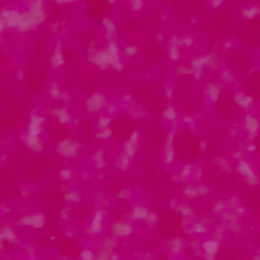};
\node[fill=white, align=center]
at (axis cs:35,35) {08};
\end{axis}

\begin{axis}[%
width=0.787in,
height=0.787in,
at={(0in,0in)},
scale only axis,
axis on top,
xmin=0,
xmax=260,
xtick={\empty},
tick align=outside,
y dir=reverse,
ymin=0,
ymax=260,
ytick={\empty},
ticks=none
]
\addplot [forget plot] graphics [xmin=0.5, xmax=260.5, ymin=0.5, ymax=260.5] {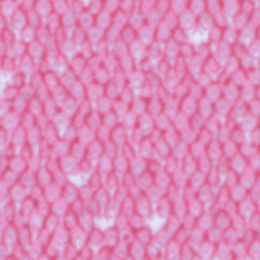};
\node[fill=white, align=center]
at (axis cs:35,35) {09};
\end{axis}

\begin{axis}[%
width=0.787in,
height=0.787in,
at={(0.788in,0in)},
scale only axis,
axis on top,
xmin=0,
xmax=260,
xtick={\empty},
tick align=outside,
y dir=reverse,
ymin=0,
ymax=260,
ytick={\empty},
ticks=none
]
\addplot [forget plot] graphics [xmin=0.5, xmax=260.5, ymin=0.5, ymax=260.5] {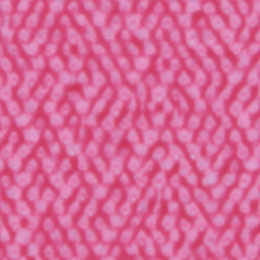};
\node[fill=white, align=center]
at (axis cs:35,35) {10};
\end{axis}

\begin{axis}[%
width=0.787in,
height=0.787in,
at={(1.575in,0in)},
scale only axis,
axis on top,
xmin=0,
xmax=260,
xtick={\empty},
tick align=outside,
y dir=reverse,
ymin=0,
ymax=260,
ytick={\empty},
ticks=none
]
\addplot [forget plot] graphics [xmin=0.5, xmax=260.5, ymin=0.5, ymax=260.5] {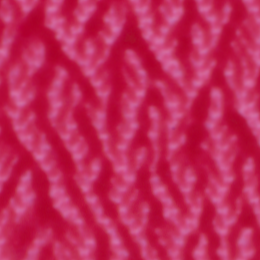};
\node[fill=white, align=center]
at (axis cs:35,35) {11};
\end{axis}

\begin{axis}[%
width=0.787in,
height=0.787in,
at={(2.363in,0in)},
scale only axis,
axis on top,
xmin=0,
xmax=260,
xtick={\empty},
tick align=outside,
y dir=reverse,
ymin=0,
ymax=260,
ytick={\empty},
ticks=none
]
\addplot [forget plot] graphics [xmin=0.5, xmax=260.5, ymin=0.5, ymax=260.5] {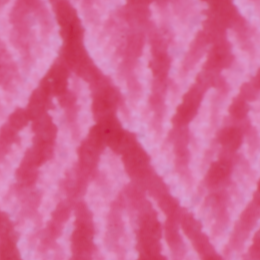};
\node[fill=white, align=center]
at (axis cs:35,35) {12};
\end{axis}

\begin{axis}[%
width=0.787in,
height=0.787in,
at={(3.15in,0in)},
scale only axis,
axis on top,
xmin=0,
xmax=260,
xtick={\empty},
tick align=outside,
y dir=reverse,
ymin=0,
ymax=260,
ytick={\empty},
ticks=none
]
\addplot [forget plot] graphics [xmin=0.5, xmax=260.5, ymin=0.5, ymax=260.5] {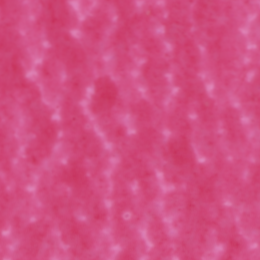};
\node[fill=white, align=center]
at (axis cs:35,35) {13};
\end{axis}

\begin{axis}[%
width=0.787in,
height=0.787in,
at={(3.937in,0in)},
scale only axis,
axis on top,
xmin=0,
xmax=260,
xtick={\empty},
tick align=outside,
y dir=reverse,
ymin=0,
ymax=260,
ytick={\empty},
ticks=none
]
\addplot [forget plot] graphics [xmin=0.5, xmax=260.5, ymin=0.5, ymax=260.5] {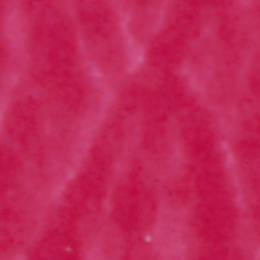};
\node[fill=white, align=center]
at (axis cs:35,35) {14};
\end{axis}

\begin{axis}[%
width=0.787in,
height=0.787in,
at={(4.725in,0in)},
scale only axis,
axis on top,
xmin=0,
xmax=260,
xtick={\empty},
tick align=outside,
y dir=reverse,
ymin=0,
ymax=260,
ytick={\empty},
ticks=none
]
\addplot [forget plot] graphics [xmin=0.5, xmax=260.5, ymin=0.5, ymax=260.5] {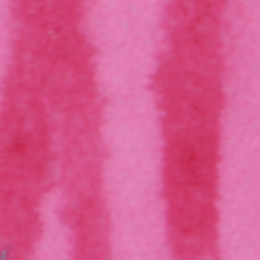};
\node[fill=white, align=center]
at (axis cs:35,35) {15};
\end{axis}

\begin{axis}[%
width=0.787in,
height=0.787in,
at={(5.512in,0in)},
scale only axis,
axis on top,
xmin=0,
xmax=260,
xtick={\empty},
tick align=outside,
y dir=reverse,
ymin=0,
ymax=260,
ytick={\empty},
ticks=none
]
\addplot [forget plot] graphics [xmin=0.5, xmax=260.5, ymin=0.5, ymax=260.5] {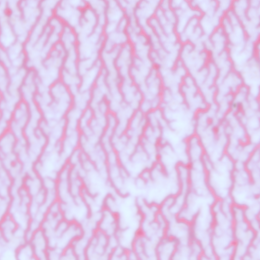};
\node[fill=white, align=center]
at (axis cs:35,35) {16};
\end{axis}
\end{tikzpicture}%

%% file: Figures/Modes/modes_U.tex
\begin{tikzpicture}

\begin{axis}[%
width=0.787in,
height=0.787in,
at={(0in,5.512in)},
scale only axis,
point meta min=0,
point meta max=1,
axis on top,
xmin=0,
xmax=260,
xtick={\empty},
y dir=reverse,
ymin=0,
ymax=260,
ytick={\empty},
axis background/.style={fill=white}
]
\addplot [forget plot] graphics [xmin=0.5, xmax=260.5, ymin=0.5, ymax=260.5] {modes_U-1.png};
\node[fill=white, align=center]
at (axis cs:35,35) {01};
\end{axis}

\begin{axis}[%
width=0.787in,
height=0.787in,
at={(0.787in,5.512in)},
scale only axis,
point meta min=0,
point meta max=1,
axis on top,
xmin=0,
xmax=260,
xtick={\empty},
y dir=reverse,
ymin=0,
ymax=260,
ytick={\empty},
axis background/.style={fill=white}
]
\addplot [forget plot] graphics [xmin=0.5, xmax=260.5, ymin=0.5, ymax=260.5] {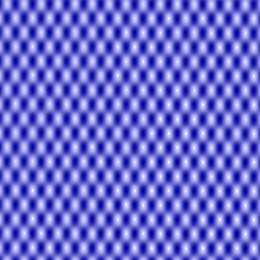};
\node[fill=white, align=center]
at (axis cs:35,35) {02};
\end{axis}

\begin{axis}[%
width=0.787in,
height=0.787in,
at={(1.575in,5.512in)},
scale only axis,
point meta min=0,
point meta max=1,
axis on top,
xmin=0,
xmax=260,
xtick={\empty},
y dir=reverse,
ymin=0,
ymax=260,
ytick={\empty},
axis background/.style={fill=white}
]
\addplot [forget plot] graphics [xmin=0.5, xmax=260.5, ymin=0.5, ymax=260.5] {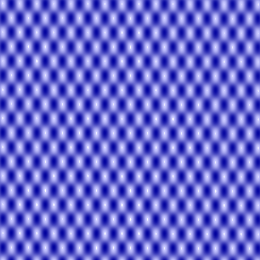};
\node[fill=white, align=center]
at (axis cs:35,35) {03};
\end{axis}

\begin{axis}[%
width=0.787in,
height=0.787in,
at={(2.362in,5.512in)},
scale only axis,
point meta min=0,
point meta max=1,
axis on top,
xmin=0,
xmax=260,
xtick={\empty},
y dir=reverse,
ymin=0,
ymax=260,
ytick={\empty},
axis background/.style={fill=white}
]
\addplot [forget plot] graphics [xmin=0.5, xmax=260.5, ymin=0.5, ymax=260.5] {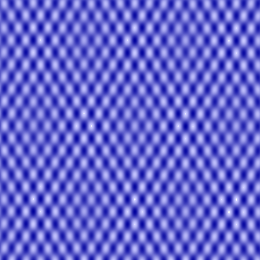};
\node[fill=white, align=center]
at (axis cs:35,35) {04};
\end{axis}

\begin{axis}[%
width=0.787in,
height=0.787in,
at={(3.15in,5.512in)},
scale only axis,
point meta min=0,
point meta max=1,
axis on top,
xmin=0,
xmax=260,
xtick={\empty},
y dir=reverse,
ymin=0,
ymax=260,
ytick={\empty},
axis background/.style={fill=white}
]
\addplot [forget plot] graphics [xmin=0.5, xmax=260.5, ymin=0.5, ymax=260.5] {modes_U-5.png};
\node[fill=white, align=center]
at (axis cs:35,35) {05};
\end{axis}

\begin{axis}[%
width=0.787in,
height=0.787in,
at={(3.937in,5.512in)},
scale only axis,
point meta min=0,
point meta max=1,
axis on top,
xmin=0,
xmax=260,
xtick={\empty},
y dir=reverse,
ymin=0,
ymax=260,
ytick={\empty},
axis background/.style={fill=white}
]
\addplot [forget plot] graphics [xmin=0.5, xmax=260.5, ymin=0.5, ymax=260.5] {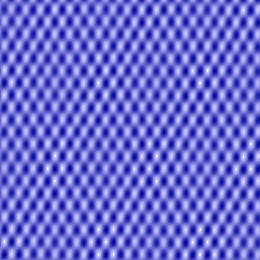};
\node[fill=white, align=center]
at (axis cs:35,35) {06};
\end{axis}

\begin{axis}[%
width=0.787in,
height=0.787in,
at={(4.724in,5.512in)},
scale only axis,
point meta min=0,
point meta max=1,
axis on top,
xmin=0,
xmax=260,
xtick={\empty},
y dir=reverse,
ymin=0,
ymax=260,
ytick={\empty},
axis background/.style={fill=white}
]
\addplot [forget plot] graphics [xmin=0.5, xmax=260.5, ymin=0.5, ymax=260.5] {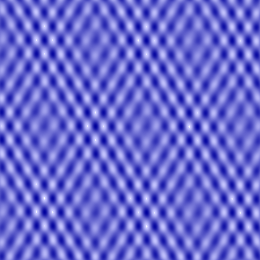};
\node[fill=white, align=center]
at (axis cs:35,35) {07};
\end{axis}

\begin{axis}[%
width=0.787in,
height=0.787in,
at={(5.512in,5.512in)},
scale only axis,
point meta min=0,
point meta max=1,
axis on top,
xmin=0,
xmax=260,
xtick={\empty},
y dir=reverse,
ymin=0,
ymax=260,
ytick={\empty},
axis background/.style={fill=white}
]
\addplot [forget plot] graphics [xmin=0.5, xmax=260.5, ymin=0.5, ymax=260.5] {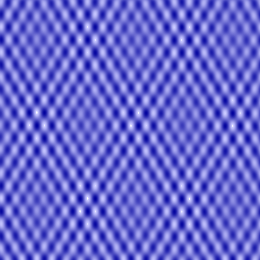};
\node[fill=white, align=center]
at (axis cs:35,35) {08};
\end{axis}

\begin{axis}[%
width=0.787in,
height=0.787in,
at={(0in,4.724in)},
scale only axis,
point meta min=0,
point meta max=1,
axis on top,
xmin=0,
xmax=260,
xtick={\empty},
y dir=reverse,
ymin=0,
ymax=260,
ytick={\empty},
axis background/.style={fill=white}
]
\addplot [forget plot] graphics [xmin=0.5, xmax=260.5, ymin=0.5, ymax=260.5] {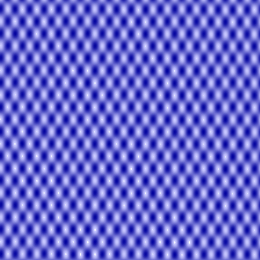};
\node[fill=white, align=center]
at (axis cs:35,35) {09};
\end{axis}

\begin{axis}[%
width=0.787in,
height=0.787in,
at={(0.787in,4.724in)},
scale only axis,
point meta min=0,
point meta max=1,
axis on top,
xmin=0,
xmax=260,
xtick={\empty},
y dir=reverse,
ymin=0,
ymax=260,
ytick={\empty},
axis background/.style={fill=white}
]
\addplot [forget plot] graphics [xmin=0.5, xmax=260.5, ymin=0.5, ymax=260.5] {modes_U-10.png};
\node[fill=white, align=center]
at (axis cs:35,35) {10};
\end{axis}

\begin{axis}[%
width=0.787in,
height=0.787in,
at={(1.575in,4.724in)},
scale only axis,
point meta min=0,
point meta max=1,
axis on top,
xmin=0,
xmax=260,
xtick={\empty},
y dir=reverse,
ymin=0,
ymax=260,
ytick={\empty},
axis background/.style={fill=white}
]
\addplot [forget plot] graphics [xmin=0.5, xmax=260.5, ymin=0.5, ymax=260.5] {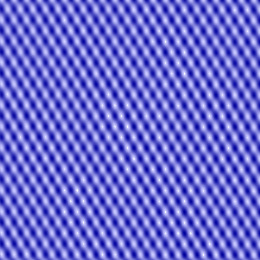};
\node[fill=white, align=center]
at (axis cs:35,35) {11};
\end{axis}

\begin{axis}[%
width=0.787in,
height=0.787in,
at={(2.362in,4.724in)},
scale only axis,
point meta min=0,
point meta max=1,
axis on top,
xmin=0,
xmax=260,
xtick={\empty},
y dir=reverse,
ymin=0,
ymax=260,
ytick={\empty},
axis background/.style={fill=white}
]
\addplot [forget plot] graphics [xmin=0.5, xmax=260.5, ymin=0.5, ymax=260.5] {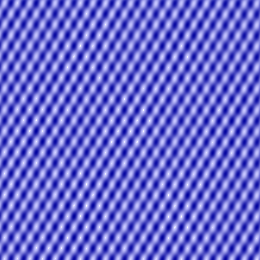};
\node[fill=white, align=center]
at (axis cs:35,35) {12};
\end{axis}

\begin{axis}[%
width=0.787in,
height=0.787in,
at={(3.15in,4.724in)},
scale only axis,
point meta min=0,
point meta max=1,
axis on top,
xmin=0,
xmax=260,
xtick={\empty},
y dir=reverse,
ymin=0,
ymax=260,
ytick={\empty},
axis background/.style={fill=white}
]
\addplot [forget plot] graphics [xmin=0.5, xmax=260.5, ymin=0.5, ymax=260.5] {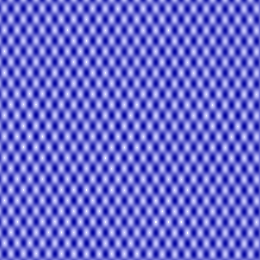};
\node[fill=white, align=center]
at (axis cs:35,35) {13};
\end{axis}

\begin{axis}[%
width=0.787in,
height=0.787in,
at={(3.937in,4.724in)},
scale only axis,
point meta min=0,
point meta max=1,
axis on top,
xmin=0,
xmax=260,
xtick={\empty},
y dir=reverse,
ymin=0,
ymax=260,
ytick={\empty},
axis background/.style={fill=white}
]
\addplot [forget plot] graphics [xmin=0.5, xmax=260.5, ymin=0.5, ymax=260.5] {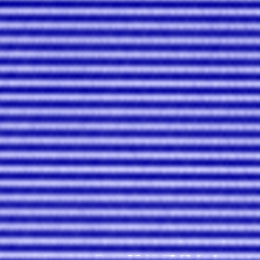};
\node[fill=white, align=center]
at (axis cs:35,35) {14};
\end{axis}

\begin{axis}[%
width=0.787in,
height=0.787in,
at={(4.724in,4.724in)},
scale only axis,
point meta min=0,
point meta max=1,
axis on top,
xmin=0,
xmax=260,
xtick={\empty},
y dir=reverse,
ymin=0,
ymax=260,
ytick={\empty},
axis background/.style={fill=white}
]
\addplot [forget plot] graphics [xmin=0.5, xmax=260.5, ymin=0.5, ymax=260.5] {modes_U-15.png};
\node[fill=white, align=center]
at (axis cs:35,35) {15};
\end{axis}

\begin{axis}[%
width=0.787in,
height=0.787in,
at={(5.512in,4.724in)},
scale only axis,
point meta min=0,
point meta max=1,
axis on top,
xmin=0,
xmax=260,
xtick={\empty},
y dir=reverse,
ymin=0,
ymax=260,
ytick={\empty},
axis background/.style={fill=white}
]
\addplot [forget plot] graphics [xmin=0.5, xmax=260.5, ymin=0.5, ymax=260.5] {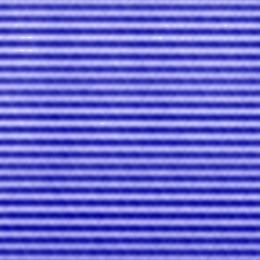};
\node[fill=white, align=center]
at (axis cs:35,35) {16};
\end{axis}

\begin{axis}[%
width=0.787in,
height=0.787in,
at={(0in,3.937in)},
scale only axis,
point meta min=0,
point meta max=1,
axis on top,
xmin=0,
xmax=260,
xtick={\empty},
y dir=reverse,
ymin=0,
ymax=260,
ytick={\empty},
axis background/.style={fill=white}
]
\addplot [forget plot] graphics [xmin=0.5, xmax=260.5, ymin=0.5, ymax=260.5] {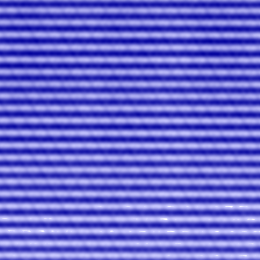};
\node[fill=white, align=center]
at (axis cs:35,35) {17};
\end{axis}

\begin{axis}[%
width=0.787in,
height=0.787in,
at={(0.787in,3.937in)},
scale only axis,
point meta min=0,
point meta max=1,
axis on top,
xmin=0,
xmax=260,
xtick={\empty},
y dir=reverse,
ymin=0,
ymax=260,
ytick={\empty},
axis background/.style={fill=white}
]
\addplot [forget plot] graphics [xmin=0.5, xmax=260.5, ymin=0.5, ymax=260.5] {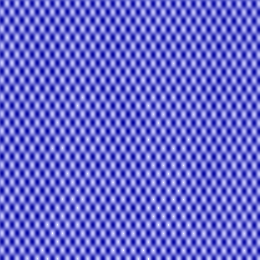};
\node[fill=white, align=center]
at (axis cs:35,35) {18};
\end{axis}

\begin{axis}[%
width=0.787in,
height=0.787in,
at={(1.575in,3.937in)},
scale only axis,
point meta min=0,
point meta max=1,
axis on top,
xmin=0,
xmax=260,
xtick={\empty},
y dir=reverse,
ymin=0,
ymax=260,
ytick={\empty},
axis background/.style={fill=white}
]
\addplot [forget plot] graphics [xmin=0.5, xmax=260.5, ymin=0.5, ymax=260.5] {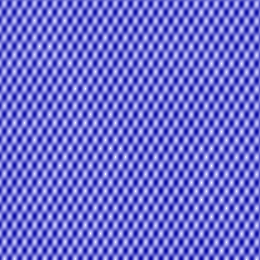};
\node[fill=white, align=center]
at (axis cs:35,35) {19};
\end{axis}

\begin{axis}[%
width=0.787in,
height=0.787in,
at={(2.362in,3.937in)},
scale only axis,
point meta min=0,
point meta max=1,
axis on top,
xmin=0,
xmax=260,
xtick={\empty},
y dir=reverse,
ymin=0,
ymax=260,
ytick={\empty},
axis background/.style={fill=white}
]
\addplot [forget plot] graphics [xmin=0.5, xmax=260.5, ymin=0.5, ymax=260.5] {modes_U-20.png};
\node[fill=white, align=center]
at (axis cs:35,35) {20};
\end{axis}

\begin{axis}[%
width=0.787in,
height=0.787in,
at={(3.15in,3.937in)},
scale only axis,
point meta min=0,
point meta max=1,
axis on top,
xmin=0,
xmax=260,
xtick={\empty},
y dir=reverse,
ymin=0,
ymax=260,
ytick={\empty},
axis background/.style={fill=white}
]
\addplot [forget plot] graphics [xmin=0.5, xmax=260.5, ymin=0.5, ymax=260.5] {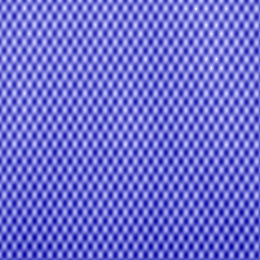};
\node[fill=white, align=center]
at (axis cs:35,35) {21};
\end{axis}

\begin{axis}[%
width=0.787in,
height=0.787in,
at={(3.937in,3.937in)},
scale only axis,
point meta min=0,
point meta max=1,
axis on top,
xmin=0,
xmax=260,
xtick={\empty},
y dir=reverse,
ymin=0,
ymax=260,
ytick={\empty},
axis background/.style={fill=white}
]
\addplot [forget plot] graphics [xmin=0.5, xmax=260.5, ymin=0.5, ymax=260.5] {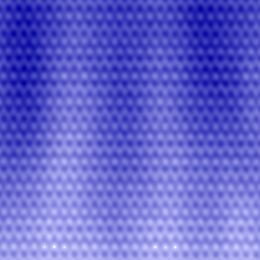};
\node[fill=white, align=center]
at (axis cs:35,35) {22};
\end{axis}

\begin{axis}[%
width=0.787in,
height=0.787in,
at={(4.724in,3.937in)},
scale only axis,
point meta min=0,
point meta max=1,
axis on top,
xmin=0,
xmax=260,
xtick={\empty},
y dir=reverse,
ymin=0,
ymax=260,
ytick={\empty},
axis background/.style={fill=white}
]
\addplot [forget plot] graphics [xmin=0.5, xmax=260.5, ymin=0.5, ymax=260.5] {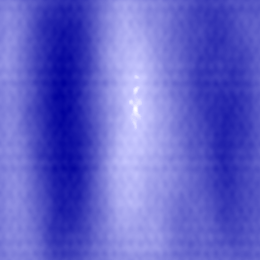};
\node[fill=white, align=center]
at (axis cs:35,35) {23};
\end{axis}

\begin{axis}[%
width=0.787in,
height=0.787in,
at={(5.512in,3.937in)},
scale only axis,
point meta min=0,
point meta max=1,
axis on top,
xmin=0,
xmax=260,
xtick={\empty},
y dir=reverse,
ymin=0,
ymax=260,
ytick={\empty},
axis background/.style={fill=white}
]
\addplot [forget plot] graphics [xmin=0.5, xmax=260.5, ymin=0.5, ymax=260.5] {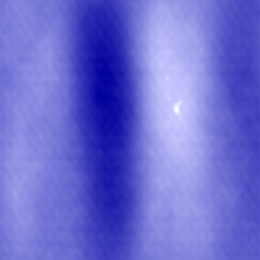};
\node[fill=white, align=center]
at (axis cs:35,35) {24};
\end{axis}

\begin{axis}[%
width=0.787in,
height=0.787in,
at={(0in,3.15in)},
scale only axis,
point meta min=0,
point meta max=1,
axis on top,
xmin=0,
xmax=260,
xtick={\empty},
y dir=reverse,
ymin=0,
ymax=260,
ytick={\empty},
axis background/.style={fill=white}
]
\addplot [forget plot] graphics [xmin=0.5, xmax=260.5, ymin=0.5, ymax=260.5] {modes_U-25.png};
\node[fill=white, align=center]
at (axis cs:35,35) {25};
\end{axis}

\begin{axis}[%
width=0.787in,
height=0.787in,
at={(0.787in,3.15in)},
scale only axis,
point meta min=0,
point meta max=1,
axis on top,
xmin=0,
xmax=260,
xtick={\empty},
y dir=reverse,
ymin=0,
ymax=260,
ytick={\empty},
axis background/.style={fill=white}
]
\addplot [forget plot] graphics [xmin=0.5, xmax=260.5, ymin=0.5, ymax=260.5] {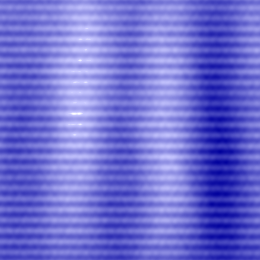};
\node[fill=white, align=center]
at (axis cs:35,35) {26};
\end{axis}

\begin{axis}[%
width=0.787in,
height=0.787in,
at={(1.575in,3.15in)},
scale only axis,
point meta min=0,
point meta max=1,
axis on top,
xmin=0,
xmax=260,
xtick={\empty},
y dir=reverse,
ymin=0,
ymax=260,
ytick={\empty},
axis background/.style={fill=white}
]
\addplot [forget plot] graphics [xmin=0.5, xmax=260.5, ymin=0.5, ymax=260.5] {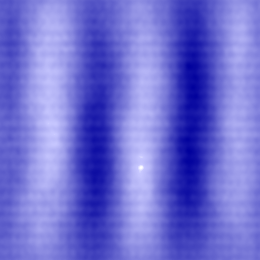};
\node[fill=white, align=center]
at (axis cs:35,35) {27};
\end{axis}

\begin{axis}[%
width=0.787in,
height=0.787in,
at={(2.362in,3.15in)},
scale only axis,
point meta min=0,
point meta max=1,
axis on top,
xmin=0,
xmax=260,
xtick={\empty},
y dir=reverse,
ymin=0,
ymax=260,
ytick={\empty},
axis background/.style={fill=white}
]
\addplot [forget plot] graphics [xmin=0.5, xmax=260.5, ymin=0.5, ymax=260.5] {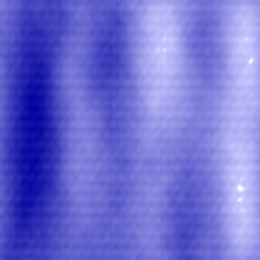};
\node[fill=white, align=center]
at (axis cs:35,35) {28};
\end{axis}

\begin{axis}[%
width=0.787in,
height=0.787in,
at={(3.15in,3.15in)},
scale only axis,
point meta min=0,
point meta max=1,
axis on top,
xmin=0,
xmax=260,
xtick={\empty},
y dir=reverse,
ymin=0,
ymax=260,
ytick={\empty},
axis background/.style={fill=white}
]
\addplot [forget plot] graphics [xmin=0.5, xmax=260.5, ymin=0.5, ymax=260.5] {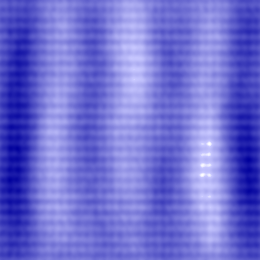};
\node[fill=white, align=center]
at (axis cs:35,35) {29};
\end{axis}

\begin{axis}[%
width=0.787in,
height=0.787in,
at={(3.937in,3.15in)},
scale only axis,
point meta min=0,
point meta max=1,
axis on top,
xmin=0,
xmax=260,
xtick={\empty},
y dir=reverse,
ymin=0,
ymax=260,
ytick={\empty},
axis background/.style={fill=white}
]
\addplot [forget plot] graphics [xmin=0.5, xmax=260.5, ymin=0.5, ymax=260.5] {modes_U-30.png};
\node[fill=white, align=center]
at (axis cs:35,35) {30};
\end{axis}

\begin{axis}[%
width=0.787in,
height=0.787in,
at={(4.724in,3.15in)},
scale only axis,
point meta min=0,
point meta max=1,
axis on top,
xmin=0,
xmax=260,
xtick={\empty},
y dir=reverse,
ymin=0,
ymax=260,
ytick={\empty},
axis background/.style={fill=white}
]
\addplot [forget plot] graphics [xmin=0.5, xmax=260.5, ymin=0.5, ymax=260.5] {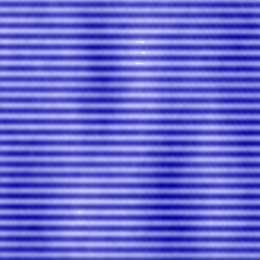};
\node[fill=white, align=center]
at (axis cs:35,35) {31};
\end{axis}

\begin{axis}[%
width=0.787in,
height=0.787in,
at={(5.512in,3.15in)},
scale only axis,
point meta min=0,
point meta max=1,
axis on top,
xmin=0,
xmax=260,
xtick={\empty},
y dir=reverse,
ymin=0,
ymax=260,
ytick={\empty},
axis background/.style={fill=white}
]
\addplot [forget plot] graphics [xmin=0.5, xmax=260.5, ymin=0.5, ymax=260.5] {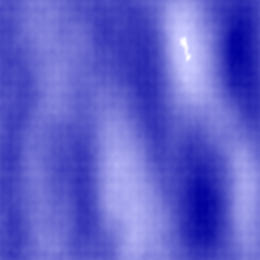};
\node[fill=white, align=center]
at (axis cs:35,35) {32};
\end{axis}

\begin{axis}[%
width=0.787in,
height=0.787in,
at={(0in,2.362in)},
scale only axis,
point meta min=0,
point meta max=1,
axis on top,
xmin=0,
xmax=260,
xtick={\empty},
y dir=reverse,
ymin=0,
ymax=260,
ytick={\empty},
axis background/.style={fill=white}
]
\addplot [forget plot] graphics [xmin=0.5, xmax=260.5, ymin=0.5, ymax=260.5] {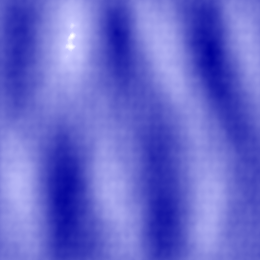};
\node[fill=white, align=center]
at (axis cs:35,35) {33};
\end{axis}

\begin{axis}[%
width=0.787in,
height=0.787in,
at={(0.787in,2.362in)},
scale only axis,
point meta min=0,
point meta max=1,
axis on top,
xmin=0,
xmax=260,
xtick={\empty},
y dir=reverse,
ymin=0,
ymax=260,
ytick={\empty},
axis background/.style={fill=white}
]
\addplot [forget plot] graphics [xmin=0.5, xmax=260.5, ymin=0.5, ymax=260.5] {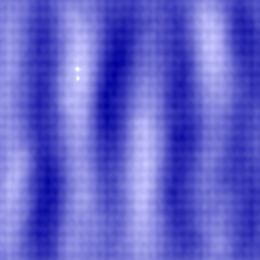};
\node[fill=white, align=center]
at (axis cs:35,35) {34};
\end{axis}

\begin{axis}[%
width=0.787in,
height=0.787in,
at={(1.575in,2.362in)},
scale only axis,
point meta min=0,
point meta max=1,
axis on top,
xmin=0,
xmax=260,
xtick={\empty},
y dir=reverse,
ymin=0,
ymax=260,
ytick={\empty},
axis background/.style={fill=white}
]
\addplot [forget plot] graphics [xmin=0.5, xmax=260.5, ymin=0.5, ymax=260.5] {modes_U-35.png};
\node[fill=white, align=center]
at (axis cs:35,35) {35};
\end{axis}

\begin{axis}[%
width=0.787in,
height=0.787in,
at={(2.362in,2.362in)},
scale only axis,
point meta min=0,
point meta max=1,
axis on top,
xmin=0,
xmax=260,
xtick={\empty},
y dir=reverse,
ymin=0,
ymax=260,
ytick={\empty},
axis background/.style={fill=white}
]
\addplot [forget plot] graphics [xmin=0.5, xmax=260.5, ymin=0.5, ymax=260.5] {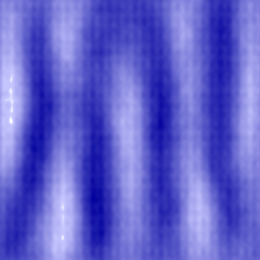};
\node[fill=white, align=center]
at (axis cs:35,35) {36};
\end{axis}

\begin{axis}[%
width=0.787in,
height=0.787in,
at={(3.15in,2.362in)},
scale only axis,
point meta min=0,
point meta max=1,
axis on top,
xmin=0,
xmax=260,
xtick={\empty},
y dir=reverse,
ymin=0,
ymax=260,
ytick={\empty},
axis background/.style={fill=white}
]
\addplot [forget plot] graphics [xmin=0.5, xmax=260.5, ymin=0.5, ymax=260.5] {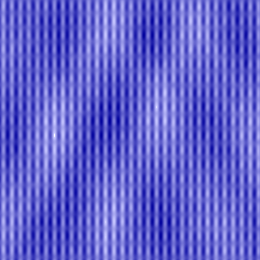};
\node[fill=white, align=center]
at (axis cs:35,35) {37};
\end{axis}

\begin{axis}[%
width=0.787in,
height=0.787in,
at={(3.937in,2.362in)},
scale only axis,
point meta min=0,
point meta max=1,
axis on top,
xmin=0,
xmax=260,
xtick={\empty},
y dir=reverse,
ymin=0,
ymax=260,
ytick={\empty},
axis background/.style={fill=white}
]
\addplot [forget plot] graphics [xmin=0.5, xmax=260.5, ymin=0.5, ymax=260.5] {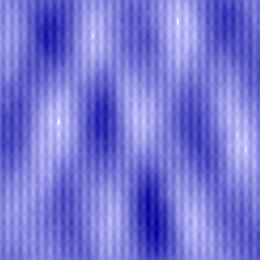};
\node[fill=white, align=center]
at (axis cs:35,35) {38};
\end{axis}

\begin{axis}[%
width=0.787in,
height=0.787in,
at={(4.724in,2.362in)},
scale only axis,
point meta min=0,
point meta max=1,
axis on top,
xmin=0,
xmax=260,
xtick={\empty},
y dir=reverse,
ymin=0,
ymax=260,
ytick={\empty},
axis background/.style={fill=white}
]
\addplot [forget plot] graphics [xmin=0.5, xmax=260.5, ymin=0.5, ymax=260.5] {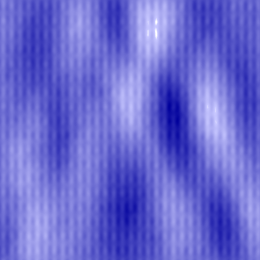};
\node[fill=white, align=center]
at (axis cs:35,35) {39};
\end{axis}

\begin{axis}[%
width=0.787in,
height=0.787in,
at={(5.512in,2.362in)},
scale only axis,
point meta min=0,
point meta max=1,
axis on top,
xmin=0,
xmax=260,
xtick={\empty},
y dir=reverse,
ymin=0,
ymax=260,
ytick={\empty},
axis background/.style={fill=white}
]
\addplot [forget plot] graphics [xmin=0.5, xmax=260.5, ymin=0.5, ymax=260.5] {modes_U-40.png};
\node[fill=white, align=center]
at (axis cs:35,35) {40};
\end{axis}

\begin{axis}[%
width=0.787in,
height=0.787in,
at={(0in,1.575in)},
scale only axis,
point meta min=0,
point meta max=1,
axis on top,
xmin=0,
xmax=260,
xtick={\empty},
y dir=reverse,
ymin=0,
ymax=260,
ytick={\empty},
axis background/.style={fill=white}
]
\addplot [forget plot] graphics [xmin=0.5, xmax=260.5, ymin=0.5, ymax=260.5] {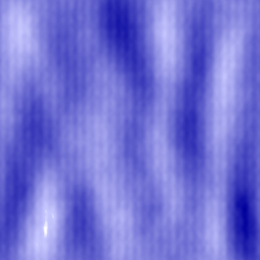};
\node[fill=white, align=center]
at (axis cs:35,35) {41};
\end{axis}

\begin{axis}[%
width=0.787in,
height=0.787in,
at={(0.787in,1.575in)},
scale only axis,
point meta min=0,
point meta max=1,
axis on top,
xmin=0,
xmax=260,
xtick={\empty},
y dir=reverse,
ymin=0,
ymax=260,
ytick={\empty},
axis background/.style={fill=white}
]
\addplot [forget plot] graphics [xmin=0.5, xmax=260.5, ymin=0.5, ymax=260.5] {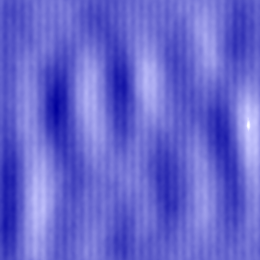};
\node[fill=white, align=center]
at (axis cs:35,35) {42};
\end{axis}

\begin{axis}[%
width=0.787in,
height=0.787in,
at={(1.575in,1.575in)},
scale only axis,
point meta min=0,
point meta max=1,
axis on top,
xmin=0,
xmax=260,
xtick={\empty},
y dir=reverse,
ymin=0,
ymax=260,
ytick={\empty},
axis background/.style={fill=white}
]
\addplot [forget plot] graphics [xmin=0.5, xmax=260.5, ymin=0.5, ymax=260.5] {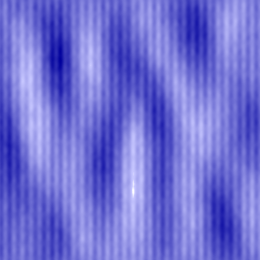};
\node[fill=white, align=center]
at (axis cs:35,35) {43};
\end{axis}

\begin{axis}[%
width=0.787in,
height=0.787in,
at={(2.362in,1.575in)},
scale only axis,
point meta min=0,
point meta max=1,
axis on top,
xmin=0,
xmax=260,
xtick={\empty},
y dir=reverse,
ymin=0,
ymax=260,
ytick={\empty},
axis background/.style={fill=white}
]
\addplot [forget plot] graphics [xmin=0.5, xmax=260.5, ymin=0.5, ymax=260.5] {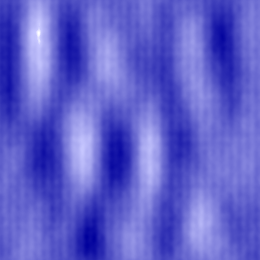};
\node[fill=white, align=center]
at (axis cs:35,35) {44};
\end{axis}

\begin{axis}[%
width=0.787in,
height=0.787in,
at={(3.15in,1.575in)},
scale only axis,
point meta min=0,
point meta max=1,
axis on top,
xmin=0,
xmax=260,
xtick={\empty},
y dir=reverse,
ymin=0,
ymax=260,
ytick={\empty},
axis background/.style={fill=white}
]
\addplot [forget plot] graphics [xmin=0.5, xmax=260.5, ymin=0.5, ymax=260.5] {modes_U-45.png};
\node[fill=white, align=center]
at (axis cs:35,35) {45};
\end{axis}

\begin{axis}[%
width=0.787in,
height=0.787in,
at={(3.937in,1.575in)},
scale only axis,
point meta min=0,
point meta max=1,
axis on top,
xmin=0,
xmax=260,
xtick={\empty},
y dir=reverse,
ymin=0,
ymax=260,
ytick={\empty},
axis background/.style={fill=white}
]
\addplot [forget plot] graphics [xmin=0.5, xmax=260.5, ymin=0.5, ymax=260.5] {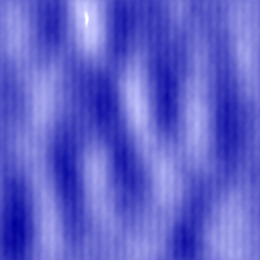};
\node[fill=white, align=center]
at (axis cs:35,35) {46};
\end{axis}

\begin{axis}[%
width=0.787in,
height=0.787in,
at={(4.724in,1.575in)},
scale only axis,
point meta min=0,
point meta max=1,
axis on top,
xmin=0,
xmax=260,
xtick={\empty},
y dir=reverse,
ymin=0,
ymax=260,
ytick={\empty},
axis background/.style={fill=white}
]
\addplot [forget plot] graphics [xmin=0.5, xmax=260.5, ymin=0.5, ymax=260.5] {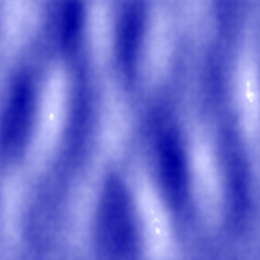};
\node[fill=white, align=center]
at (axis cs:35,35) {47};
\end{axis}

\begin{axis}[%
width=0.787in,
height=0.787in,
at={(5.512in,1.575in)},
scale only axis,
point meta min=0,
point meta max=1,
axis on top,
xmin=0,
xmax=260,
xtick={\empty},
y dir=reverse,
ymin=0,
ymax=260,
ytick={\empty},
axis background/.style={fill=white}
]
\addplot [forget plot] graphics [xmin=0.5, xmax=260.5, ymin=0.5, ymax=260.5] {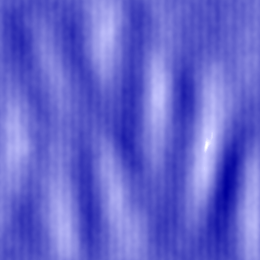};
\node[fill=white, align=center]
at (axis cs:35,35) {48};
\end{axis}

\begin{axis}[%
width=0.787in,
height=0.787in,
at={(0in,0.787in)},
scale only axis,
point meta min=0,
point meta max=1,
axis on top,
xmin=0,
xmax=260,
xtick={\empty},
y dir=reverse,
ymin=0,
ymax=260,
ytick={\empty},
axis background/.style={fill=white}
]
\addplot [forget plot] graphics [xmin=0.5, xmax=260.5, ymin=0.5, ymax=260.5] {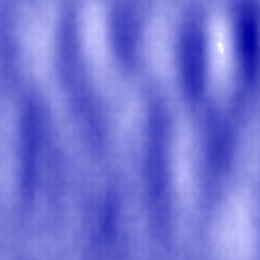};
\node[fill=white, align=center]
at (axis cs:35,35) {49};
\end{axis}

\begin{axis}[%
width=0.787in,
height=0.787in,
at={(0.787in,0.787in)},
scale only axis,
point meta min=0,
point meta max=1,
axis on top,
xmin=0,
xmax=260,
xtick={\empty},
y dir=reverse,
ymin=0,
ymax=260,
ytick={\empty},
axis background/.style={fill=white}
]
\addplot [forget plot] graphics [xmin=0.5, xmax=260.5, ymin=0.5, ymax=260.5] {modes_U-50.png};
\node[fill=white, align=center]
at (axis cs:35,35) {50};
\end{axis}

\begin{axis}[%
width=0.787in,
height=0.787in,
at={(1.575in,0.787in)},
scale only axis,
point meta min=0,
point meta max=1,
axis on top,
xmin=0,
xmax=260,
xtick={\empty},
y dir=reverse,
ymin=0,
ymax=260,
ytick={\empty},
axis background/.style={fill=white}
]
\addplot [forget plot] graphics [xmin=0.5, xmax=260.5, ymin=0.5, ymax=260.5] {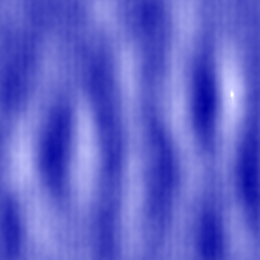};
\node[fill=white, align=center]
at (axis cs:35,35) {51};
\end{axis}

\begin{axis}[%
width=0.787in,
height=0.787in,
at={(2.362in,0.787in)},
scale only axis,
point meta min=0,
point meta max=1,
axis on top,
xmin=0,
xmax=260,
xtick={\empty},
y dir=reverse,
ymin=0,
ymax=260,
ytick={\empty},
axis background/.style={fill=white}
]
\addplot [forget plot] graphics [xmin=0.5, xmax=260.5, ymin=0.5, ymax=260.5] {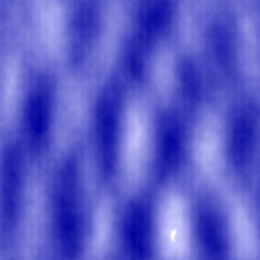};
\node[fill=white, align=center]
at (axis cs:35,35) {52};
\end{axis}

\begin{axis}[%
width=0.787in,
height=0.787in,
at={(3.15in,0.787in)},
scale only axis,
point meta min=0,
point meta max=1,
axis on top,
xmin=0,
xmax=260,
xtick={\empty},
y dir=reverse,
ymin=0,
ymax=260,
ytick={\empty},
axis background/.style={fill=white}
]
\addplot [forget plot] graphics [xmin=0.5, xmax=260.5, ymin=0.5, ymax=260.5] {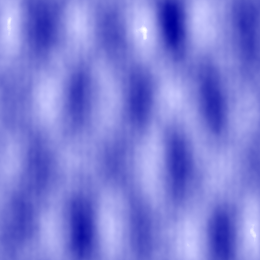};
\node[fill=white, align=center]
at (axis cs:35,35) {53};
\end{axis}

\begin{axis}[%
width=0.787in,
height=0.787in,
at={(3.937in,0.787in)},
scale only axis,
point meta min=0,
point meta max=1,
axis on top,
xmin=0,
xmax=260,
xtick={\empty},
y dir=reverse,
ymin=0,
ymax=260,
ytick={\empty},
axis background/.style={fill=white}
]
\addplot [forget plot] graphics [xmin=0.5, xmax=260.5, ymin=0.5, ymax=260.5] {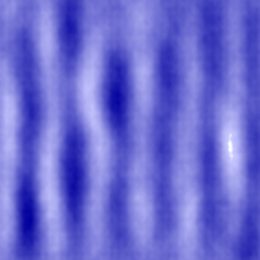};
\node[fill=white, align=center]
at (axis cs:35,35) {54};
\end{axis}

\begin{axis}[%
width=0.787in,
height=0.787in,
at={(4.724in,0.787in)},
scale only axis,
point meta min=0,
point meta max=1,
axis on top,
xmin=0,
xmax=260,
xtick={\empty},
y dir=reverse,
ymin=0,
ymax=260,
ytick={\empty},
axis background/.style={fill=white}
]
\addplot [forget plot] graphics [xmin=0.5, xmax=260.5, ymin=0.5, ymax=260.5] {modes_U-55.png};
\node[fill=white, align=center]
at (axis cs:35,35) {55};
\end{axis}

\begin{axis}[%
width=0.787in,
height=0.787in,
at={(5.512in,0.787in)},
scale only axis,
point meta min=0,
point meta max=1,
axis on top,
xmin=0,
xmax=260,
xtick={\empty},
y dir=reverse,
ymin=0,
ymax=260,
ytick={\empty},
axis background/.style={fill=white}
]
\addplot [forget plot] graphics [xmin=0.5, xmax=260.5, ymin=0.5, ymax=260.5] {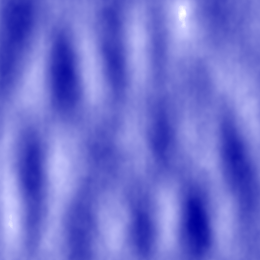};
\node[fill=white, align=center]
at (axis cs:35,35) {56};
\end{axis}

\begin{axis}[%
width=0.787in,
height=0.787in,
at={(0in,0in)},
scale only axis,
point meta min=0,
point meta max=1,
axis on top,
xmin=0,
xmax=260,
xtick={\empty},
y dir=reverse,
ymin=0,
ymax=260,
ytick={\empty},
axis background/.style={fill=white}
]
\addplot [forget plot] graphics [xmin=0.5, xmax=260.5, ymin=0.5, ymax=260.5] {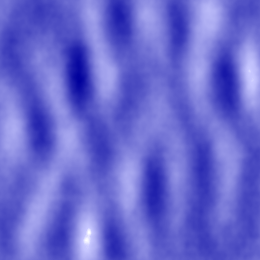};
\node[fill=white, align=center]
at (axis cs:35,35) {57};
\end{axis}

\begin{axis}[%
width=0.787in,
height=0.787in,
at={(0.787in,0in)},
scale only axis,
point meta min=0,
point meta max=1,
axis on top,
xmin=0,
xmax=260,
xtick={\empty},
y dir=reverse,
ymin=0,
ymax=260,
ytick={\empty},
axis background/.style={fill=white}
]
\addplot [forget plot] graphics [xmin=0.5, xmax=260.5, ymin=0.5, ymax=260.5] {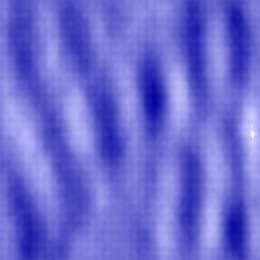};
\node[fill=white, align=center]
at (axis cs:35,35) {58};
\end{axis}

\begin{axis}[%
width=0.787in,
height=0.787in,
at={(1.575in,0in)},
scale only axis,
point meta min=0,
point meta max=1,
axis on top,
xmin=0,
xmax=260,
xtick={\empty},
y dir=reverse,
ymin=0,
ymax=260,
ytick={\empty},
axis background/.style={fill=white}
]
\addplot [forget plot] graphics [xmin=0.5, xmax=260.5, ymin=0.5, ymax=260.5] {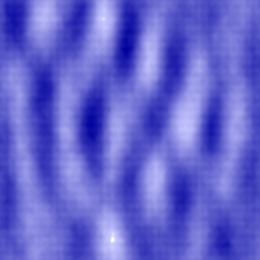};
\node[fill=white, align=center]
at (axis cs:35,35) {59};
\end{axis}

\begin{axis}[%
width=0.787in,
height=0.787in,
at={(2.362in,0in)},
scale only axis,
point meta min=0,
point meta max=1,
axis on top,
xmin=0,
xmax=260,
xtick={\empty},
y dir=reverse,
ymin=0,
ymax=260,
ytick={\empty},
axis background/.style={fill=white}
]
\addplot [forget plot] graphics [xmin=0.5, xmax=260.5, ymin=0.5, ymax=260.5] {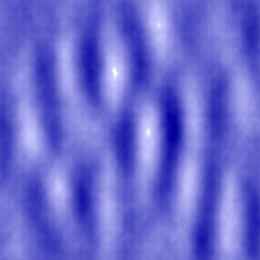};
\node[fill=white, align=center]
at (axis cs:35,35) {60};
\end{axis}

\begin{axis}[%
width=0.787in,
height=0.787in,
at={(3.15in,0in)},
scale only axis,
point meta min=0,
point meta max=1,
axis on top,
xmin=0,
xmax=260,
xtick={\empty},
y dir=reverse,
ymin=0,
ymax=260,
ytick={\empty},
axis background/.style={fill=white}
]
\addplot [forget plot] graphics [xmin=0.5, xmax=260.5, ymin=0.5, ymax=260.5] {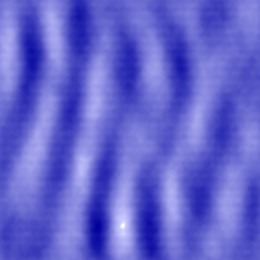};
\node[fill=white, align=center]
at (axis cs:35,35) {61};
\end{axis}

\begin{axis}[%
width=0.787in,
height=0.787in,
at={(3.937in,0in)},
scale only axis,
point meta min=0,
point meta max=1,
axis on top,
xmin=0,
xmax=260,
xtick={\empty},
y dir=reverse,
ymin=0,
ymax=260,
ytick={\empty},
axis background/.style={fill=white}
]
\addplot [forget plot] graphics [xmin=0.5, xmax=260.5, ymin=0.5, ymax=260.5] {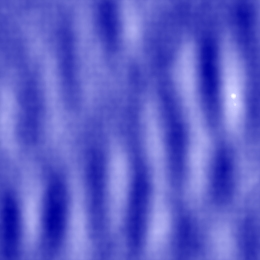};
\node[fill=white, align=center]
at (axis cs:35,35) {62};
\end{axis}

\begin{axis}[%
width=0.787in,
height=0.787in,
at={(4.724in,0in)},
scale only axis,
point meta min=0,
point meta max=1,
axis on top,
xmin=0,
xmax=260,
xtick={\empty},
y dir=reverse,
ymin=0,
ymax=260,
ytick={\empty},
axis background/.style={fill=white}
]
\addplot [forget plot] graphics [xmin=0.5, xmax=260.5, ymin=0.5, ymax=260.5] {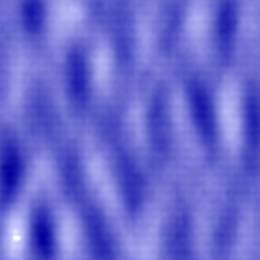};
\node[fill=white, align=center]
at (axis cs:35,35) {63};
\end{axis}

\begin{axis}[%
width=0.787in,
height=0.787in,
at={(5.512in,0in)},
scale only axis,
point meta min=0,
point meta max=1,
axis on top,
xmin=0,
xmax=260,
xtick={\empty},
y dir=reverse,
ymin=0,
ymax=260,
ytick={\empty},
axis background/.style={fill=white},
colormap={mymap}{[1pt] rgb(0pt)=(1,1,1); rgb(11pt)=(0.780392,0.780392,1); rgb(255pt)=(0,0,0.611765)},
colorbar horizontal,
colorbar style={
at={(1,-0.3)},anchor=south east,
width=8*
\pgfkeysvalueof{/pgfplots/parent axis width},
xticklabel pos=lower,
},
]
\addplot [forget plot] graphics [xmin=0.5, xmax=260.5, ymin=0.5, ymax=260.5] {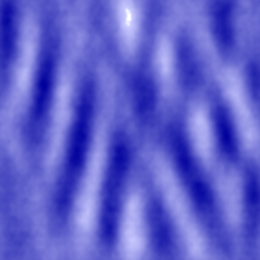};
\node[fill=white, align=center]
at (axis cs:35,35) {64};
\end{axis}
\end{tikzpicture}%

%% file: Figures/Modes/modes_Uhat.tex
\begin{tikzpicture}

\begin{axis}[%
width=0.787in,
height=0.787in,
at={(0in,5.512in)},
scale only axis,
point meta min=0,
point meta max=0.35,
axis on top,
xmin=0,
xmax=260,
xtick={\empty},
y dir=reverse,
ymin=0,
ymax=260,
ytick={\empty},
axis background/.style={fill=white}
]
\addplot [forget plot] graphics [xmin=0.5, xmax=260.5, ymin=0.5, ymax=260.5] {modes_Uhat-1.png};
\node[fill=white, align=center]
at (axis cs:35,35) {01};
\end{axis}

\begin{axis}[%
width=0.787in,
height=0.787in,
at={(0.787in,5.512in)},
scale only axis,
point meta min=0,
point meta max=0.35,
axis on top,
xmin=0,
xmax=260,
xtick={\empty},
y dir=reverse,
ymin=0,
ymax=260,
ytick={\empty},
axis background/.style={fill=white}
]
\addplot [forget plot] graphics [xmin=0.5, xmax=260.5, ymin=0.5, ymax=260.5] {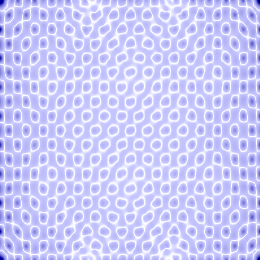};
\node[fill=white, align=center]
at (axis cs:35,35) {02};
\end{axis}

\begin{axis}[%
width=0.787in,
height=0.787in,
at={(1.575in,5.512in)},
scale only axis,
point meta min=0,
point meta max=0.35,
axis on top,
xmin=0,
xmax=260,
xtick={\empty},
y dir=reverse,
ymin=0,
ymax=260,
ytick={\empty},
axis background/.style={fill=white}
]
\addplot [forget plot] graphics [xmin=0.5, xmax=260.5, ymin=0.5, ymax=260.5] {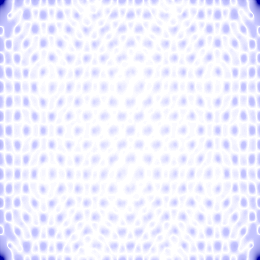};
\node[fill=white, align=center]
at (axis cs:35,35) {03};
\end{axis}

\begin{axis}[%
width=0.787in,
height=0.787in,
at={(2.362in,5.512in)},
scale only axis,
point meta min=0,
point meta max=0.35,
axis on top,
xmin=0,
xmax=260,
xtick={\empty},
y dir=reverse,
ymin=0,
ymax=260,
ytick={\empty},
axis background/.style={fill=white}
]
\addplot [forget plot] graphics [xmin=0.5, xmax=260.5, ymin=0.5, ymax=260.5] {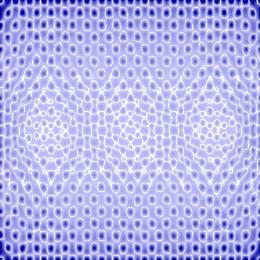};
\node[fill=white, align=center]
at (axis cs:35,35) {04};
\end{axis}

\begin{axis}[%
width=0.787in,
height=0.787in,
at={(3.15in,5.512in)},
scale only axis,
point meta min=0,
point meta max=0.35,
axis on top,
xmin=0,
xmax=260,
xtick={\empty},
y dir=reverse,
ymin=0,
ymax=260,
ytick={\empty},
axis background/.style={fill=white}
]
\addplot [forget plot] graphics [xmin=0.5, xmax=260.5, ymin=0.5, ymax=260.5] {modes_Uhat-5.png};
\node[fill=white, align=center]
at (axis cs:35,35) {05};
\end{axis}

\begin{axis}[%
width=0.787in,
height=0.787in,
at={(3.937in,5.512in)},
scale only axis,
point meta min=0,
point meta max=0.35,
axis on top,
xmin=0,
xmax=260,
xtick={\empty},
y dir=reverse,
ymin=0,
ymax=260,
ytick={\empty},
axis background/.style={fill=white}
]
\addplot [forget plot] graphics [xmin=0.5, xmax=260.5, ymin=0.5, ymax=260.5] {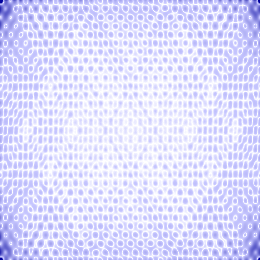};
\node[fill=white, align=center]
at (axis cs:35,35) {06};
\end{axis}

\begin{axis}[%
width=0.787in,
height=0.787in,
at={(4.724in,5.512in)},
scale only axis,
point meta min=0,
point meta max=0.35,
axis on top,
xmin=0,
xmax=260,
xtick={\empty},
y dir=reverse,
ymin=0,
ymax=260,
ytick={\empty},
axis background/.style={fill=white}
]
\addplot [forget plot] graphics [xmin=0.5, xmax=260.5, ymin=0.5, ymax=260.5] {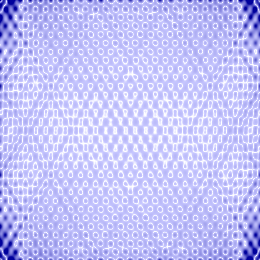};
\node[fill=white, align=center]
at (axis cs:35,35) {07};
\end{axis}

\begin{axis}[%
width=0.787in,
height=0.787in,
at={(5.512in,5.512in)},
scale only axis,
point meta min=0,
point meta max=0.35,
axis on top,
xmin=0,
xmax=260,
xtick={\empty},
y dir=reverse,
ymin=0,
ymax=260,
ytick={\empty},
axis background/.style={fill=white}
]
\addplot [forget plot] graphics [xmin=0.5, xmax=260.5, ymin=0.5, ymax=260.5] {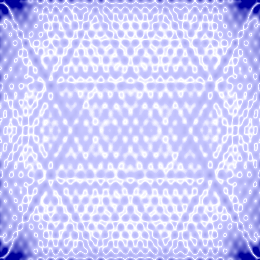};
\node[fill=white, align=center]
at (axis cs:35,35) {08};
\end{axis}

\begin{axis}[%
width=0.787in,
height=0.787in,
at={(0in,4.724in)},
scale only axis,
point meta min=0,
point meta max=0.35,
axis on top,
xmin=0,
xmax=260,
xtick={\empty},
y dir=reverse,
ymin=0,
ymax=260,
ytick={\empty},
axis background/.style={fill=white}
]
\addplot [forget plot] graphics [xmin=0.5, xmax=260.5, ymin=0.5, ymax=260.5] {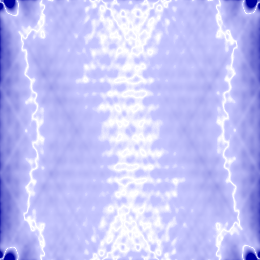};
\node[fill=white, align=center]
at (axis cs:35,35) {09};
\end{axis}

\begin{axis}[%
width=0.787in,
height=0.787in,
at={(0.787in,4.724in)},
scale only axis,
point meta min=0,
point meta max=0.35,
axis on top,
xmin=0,
xmax=260,
xtick={\empty},
y dir=reverse,
ymin=0,
ymax=260,
ytick={\empty},
axis background/.style={fill=white}
]
\addplot [forget plot] graphics [xmin=0.5, xmax=260.5, ymin=0.5, ymax=260.5] {modes_Uhat-10.png};
\node[fill=white, align=center]
at (axis cs:35,35) {10};
\end{axis}

\begin{axis}[%
width=0.787in,
height=0.787in,
at={(1.575in,4.724in)},
scale only axis,
point meta min=0,
point meta max=0.35,
axis on top,
xmin=0,
xmax=260,
xtick={\empty},
y dir=reverse,
ymin=0,
ymax=260,
ytick={\empty},
axis background/.style={fill=white}
]
\addplot [forget plot] graphics [xmin=0.5, xmax=260.5, ymin=0.5, ymax=260.5] {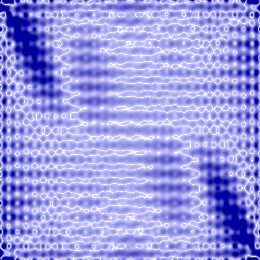};
\node[fill=white, align=center]
at (axis cs:35,35) {11};
\end{axis}

\begin{axis}[%
width=0.787in,
height=0.787in,
at={(2.362in,4.724in)},
scale only axis,
point meta min=0,
point meta max=0.35,
axis on top,
xmin=0,
xmax=260,
xtick={\empty},
y dir=reverse,
ymin=0,
ymax=260,
ytick={\empty},
axis background/.style={fill=white}
]
\addplot [forget plot] graphics [xmin=0.5, xmax=260.5, ymin=0.5, ymax=260.5] {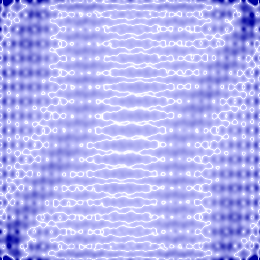};
\node[fill=white, align=center]
at (axis cs:35,35) {12};
\end{axis}

\begin{axis}[%
width=0.787in,
height=0.787in,
at={(3.15in,4.724in)},
scale only axis,
point meta min=0,
point meta max=0.35,
axis on top,
xmin=0,
xmax=260,
xtick={\empty},
y dir=reverse,
ymin=0,
ymax=260,
ytick={\empty},
axis background/.style={fill=white}
]
\addplot [forget plot] graphics [xmin=0.5, xmax=260.5, ymin=0.5, ymax=260.5] {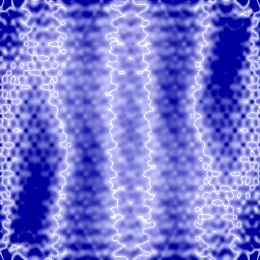};
\node[fill=white, align=center]
at (axis cs:35,35) {13};
\end{axis}

\begin{axis}[%
width=0.787in,
height=0.787in,
at={(3.937in,4.724in)},
scale only axis,
point meta min=0,
point meta max=0.35,
axis on top,
xmin=0,
xmax=260,
xtick={\empty},
y dir=reverse,
ymin=0,
ymax=260,
ytick={\empty},
axis background/.style={fill=white}
]
\addplot [forget plot] graphics [xmin=0.5, xmax=260.5, ymin=0.5, ymax=260.5] {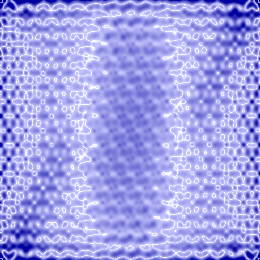};
\node[fill=white, align=center]
at (axis cs:35,35) {14};
\end{axis}

\begin{axis}[%
width=0.787in,
height=0.787in,
at={(4.724in,4.724in)},
scale only axis,
point meta min=0,
point meta max=0.35,
axis on top,
xmin=0,
xmax=260,
xtick={\empty},
y dir=reverse,
ymin=0,
ymax=260,
ytick={\empty},
axis background/.style={fill=white}
]
\addplot [forget plot] graphics [xmin=0.5, xmax=260.5, ymin=0.5, ymax=260.5] {modes_Uhat-15.png};
\node[fill=white, align=center]
at (axis cs:35,35) {15};
\end{axis}

\begin{axis}[%
width=0.787in,
height=0.787in,
at={(5.512in,4.724in)},
scale only axis,
point meta min=0,
point meta max=0.35,
axis on top,
xmin=0,
xmax=260,
xtick={\empty},
y dir=reverse,
ymin=0,
ymax=260,
ytick={\empty},
axis background/.style={fill=white}
]
\addplot [forget plot] graphics [xmin=0.5, xmax=260.5, ymin=0.5, ymax=260.5] {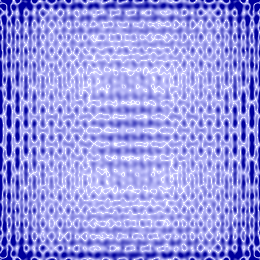};
\node[fill=white, align=center]
at (axis cs:35,35) {16};
\end{axis}

\begin{axis}[%
width=0.787in,
height=0.787in,
at={(0in,3.937in)},
scale only axis,
point meta min=0,
point meta max=0.35,
axis on top,
xmin=0,
xmax=260,
xtick={\empty},
y dir=reverse,
ymin=0,
ymax=260,
ytick={\empty},
axis background/.style={fill=white}
]
\addplot [forget plot] graphics [xmin=0.5, xmax=260.5, ymin=0.5, ymax=260.5] {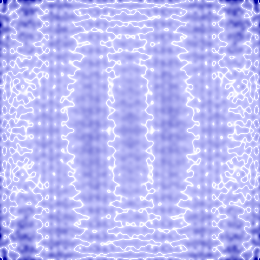};
\node[fill=white, align=center]
at (axis cs:35,35) {17};
\end{axis}

\begin{axis}[%
width=0.787in,
height=0.787in,
at={(0.787in,3.937in)},
scale only axis,
point meta min=0,
point meta max=0.35,
axis on top,
xmin=0,
xmax=260,
xtick={\empty},
y dir=reverse,
ymin=0,
ymax=260,
ytick={\empty},
axis background/.style={fill=white}
]
\addplot [forget plot] graphics [xmin=0.5, xmax=260.5, ymin=0.5, ymax=260.5] {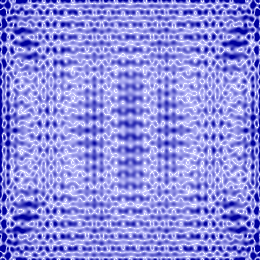};
\node[fill=white, align=center]
at (axis cs:35,35) {18};
\end{axis}

\begin{axis}[%
width=0.787in,
height=0.787in,
at={(1.575in,3.937in)},
scale only axis,
point meta min=0,
point meta max=0.35,
axis on top,
xmin=0,
xmax=260,
xtick={\empty},
y dir=reverse,
ymin=0,
ymax=260,
ytick={\empty},
axis background/.style={fill=white}
]
\addplot [forget plot] graphics [xmin=0.5, xmax=260.5, ymin=0.5, ymax=260.5] {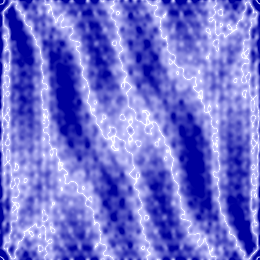};
\node[fill=white, align=center]
at (axis cs:35,35) {19};
\end{axis}

\begin{axis}[%
width=0.787in,
height=0.787in,
at={(2.362in,3.937in)},
scale only axis,
point meta min=0,
point meta max=0.35,
axis on top,
xmin=0,
xmax=260,
xtick={\empty},
y dir=reverse,
ymin=0,
ymax=260,
ytick={\empty},
axis background/.style={fill=white}
]
\addplot [forget plot] graphics [xmin=0.5, xmax=260.5, ymin=0.5, ymax=260.5] {modes_Uhat-20.png};
\node[fill=white, align=center]
at (axis cs:35,35) {20};
\end{axis}

\begin{axis}[%
width=0.787in,
height=0.787in,
at={(3.15in,3.937in)},
scale only axis,
point meta min=0,
point meta max=0.35,
axis on top,
xmin=0,
xmax=260,
xtick={\empty},
y dir=reverse,
ymin=0,
ymax=260,
ytick={\empty},
axis background/.style={fill=white}
]
\addplot [forget plot] graphics [xmin=0.5, xmax=260.5, ymin=0.5, ymax=260.5] {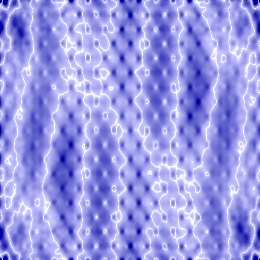};
\node[fill=white, align=center]
at (axis cs:35,35) {21};
\end{axis}

\begin{axis}[%
width=0.787in,
height=0.787in,
at={(3.937in,3.937in)},
scale only axis,
point meta min=0,
point meta max=0.35,
axis on top,
xmin=0,
xmax=260,
xtick={\empty},
y dir=reverse,
ymin=0,
ymax=260,
ytick={\empty},
axis background/.style={fill=white}
]
\addplot [forget plot] graphics [xmin=0.5, xmax=260.5, ymin=0.5, ymax=260.5] {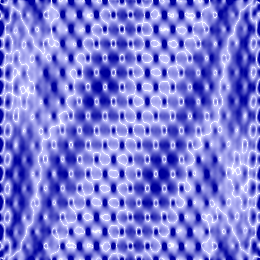};
\node[fill=white, align=center]
at (axis cs:35,35) {22};
\end{axis}

\begin{axis}[%
width=0.787in,
height=0.787in,
at={(4.724in,3.937in)},
scale only axis,
point meta min=0,
point meta max=0.35,
axis on top,
xmin=0,
xmax=260,
xtick={\empty},
y dir=reverse,
ymin=0,
ymax=260,
ytick={\empty},
axis background/.style={fill=white}
]
\addplot [forget plot] graphics [xmin=0.5, xmax=260.5, ymin=0.5, ymax=260.5] {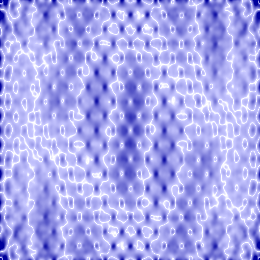};
\node[fill=white, align=center]
at (axis cs:35,35) {23};
\end{axis}

\begin{axis}[%
width=0.787in,
height=0.787in,
at={(5.512in,3.937in)},
scale only axis,
point meta min=0,
point meta max=0.35,
axis on top,
xmin=0,
xmax=260,
xtick={\empty},
y dir=reverse,
ymin=0,
ymax=260,
ytick={\empty},
axis background/.style={fill=white}
]
\addplot [forget plot] graphics [xmin=0.5, xmax=260.5, ymin=0.5, ymax=260.5] {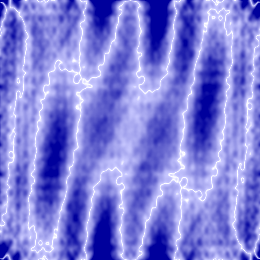};
\node[fill=white, align=center]
at (axis cs:35,35) {24};
\end{axis}

\begin{axis}[%
width=0.787in,
height=0.787in,
at={(0in,3.15in)},
scale only axis,
point meta min=0,
point meta max=0.35,
axis on top,
xmin=0,
xmax=260,
xtick={\empty},
y dir=reverse,
ymin=0,
ymax=260,
ytick={\empty},
axis background/.style={fill=white}
]
\addplot [forget plot] graphics [xmin=0.5, xmax=260.5, ymin=0.5, ymax=260.5] {modes_Uhat-25.png};
\node[fill=white, align=center]
at (axis cs:35,35) {25};
\end{axis}

\begin{axis}[%
width=0.787in,
height=0.787in,
at={(0.787in,3.15in)},
scale only axis,
point meta min=0,
point meta max=0.35,
axis on top,
xmin=0,
xmax=260,
xtick={\empty},
y dir=reverse,
ymin=0,
ymax=260,
ytick={\empty},
axis background/.style={fill=white}
]
\addplot [forget plot] graphics [xmin=0.5, xmax=260.5, ymin=0.5, ymax=260.5] {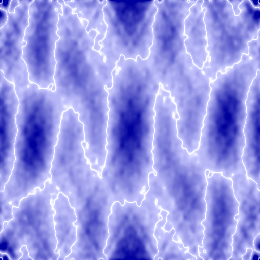};
\node[fill=white, align=center]
at (axis cs:35,35) {26};
\end{axis}

\begin{axis}[%
width=0.787in,
height=0.787in,
at={(1.575in,3.15in)},
scale only axis,
point meta min=0,
point meta max=0.35,
axis on top,
xmin=0,
xmax=260,
xtick={\empty},
y dir=reverse,
ymin=0,
ymax=260,
ytick={\empty},
axis background/.style={fill=white}
]
\addplot [forget plot] graphics [xmin=0.5, xmax=260.5, ymin=0.5, ymax=260.5] {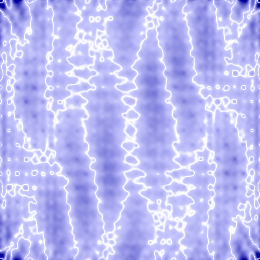};
\node[fill=white, align=center]
at (axis cs:35,35) {27};
\end{axis}

\begin{axis}[%
width=0.787in,
height=0.787in,
at={(2.362in,3.15in)},
scale only axis,
point meta min=0,
point meta max=0.35,
axis on top,
xmin=0,
xmax=260,
xtick={\empty},
y dir=reverse,
ymin=0,
ymax=260,
ytick={\empty},
axis background/.style={fill=white}
]
\addplot [forget plot] graphics [xmin=0.5, xmax=260.5, ymin=0.5, ymax=260.5] {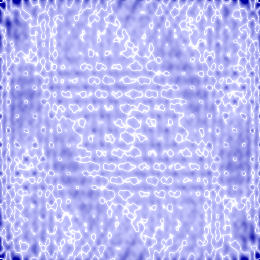};
\node[fill=white, align=center]
at (axis cs:35,35) {28};
\end{axis}

\begin{axis}[%
width=0.787in,
height=0.787in,
at={(3.15in,3.15in)},
scale only axis,
point meta min=0,
point meta max=0.35,
axis on top,
xmin=0,
xmax=260,
xtick={\empty},
y dir=reverse,
ymin=0,
ymax=260,
ytick={\empty},
axis background/.style={fill=white}
]
\addplot [forget plot] graphics [xmin=0.5, xmax=260.5, ymin=0.5, ymax=260.5] {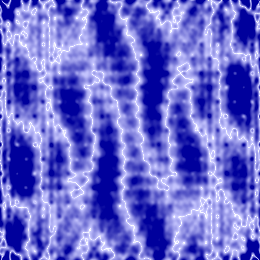};
\node[fill=white, align=center]
at (axis cs:35,35) {29};
\end{axis}

\begin{axis}[%
width=0.787in,
height=0.787in,
at={(3.937in,3.15in)},
scale only axis,
point meta min=0,
point meta max=0.35,
axis on top,
xmin=0,
xmax=260,
xtick={\empty},
y dir=reverse,
ymin=0,
ymax=260,
ytick={\empty},
axis background/.style={fill=white}
]
\addplot [forget plot] graphics [xmin=0.5, xmax=260.5, ymin=0.5, ymax=260.5] {modes_Uhat-30.png};
\node[fill=white, align=center]
at (axis cs:35,35) {30};
\end{axis}

\begin{axis}[%
width=0.787in,
height=0.787in,
at={(4.724in,3.15in)},
scale only axis,
point meta min=0,
point meta max=0.35,
axis on top,
xmin=0,
xmax=260,
xtick={\empty},
y dir=reverse,
ymin=0,
ymax=260,
ytick={\empty},
axis background/.style={fill=white}
]
\addplot [forget plot] graphics [xmin=0.5, xmax=260.5, ymin=0.5, ymax=260.5] {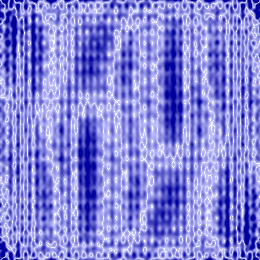};
\node[fill=white, align=center]
at (axis cs:35,35) {31};
\end{axis}

\begin{axis}[%
width=0.787in,
height=0.787in,
at={(5.512in,3.15in)},
scale only axis,
point meta min=0,
point meta max=0.35,
axis on top,
xmin=0,
xmax=260,
xtick={\empty},
y dir=reverse,
ymin=0,
ymax=260,
ytick={\empty},
axis background/.style={fill=white}
]
\addplot [forget plot] graphics [xmin=0.5, xmax=260.5, ymin=0.5, ymax=260.5] {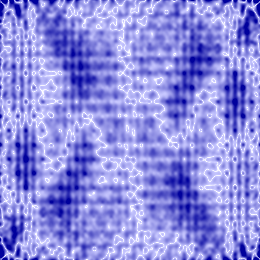};
\node[fill=white, align=center]
at (axis cs:35,35) {32};
\end{axis}

\begin{axis}[%
width=0.787in,
height=0.787in,
at={(0in,2.362in)},
scale only axis,
point meta min=0,
point meta max=0.35,
axis on top,
xmin=0,
xmax=260,
xtick={\empty},
y dir=reverse,
ymin=0,
ymax=260,
ytick={\empty},
axis background/.style={fill=white}
]
\addplot [forget plot] graphics [xmin=0.5, xmax=260.5, ymin=0.5, ymax=260.5] {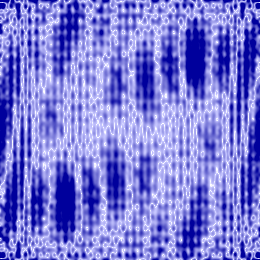};
\node[fill=white, align=center]
at (axis cs:35,35) {33};
\end{axis}

\begin{axis}[%
width=0.787in,
height=0.787in,
at={(0.787in,2.362in)},
scale only axis,
point meta min=0,
point meta max=0.35,
axis on top,
xmin=0,
xmax=260,
xtick={\empty},
y dir=reverse,
ymin=0,
ymax=260,
ytick={\empty},
axis background/.style={fill=white}
]
\addplot [forget plot] graphics [xmin=0.5, xmax=260.5, ymin=0.5, ymax=260.5] {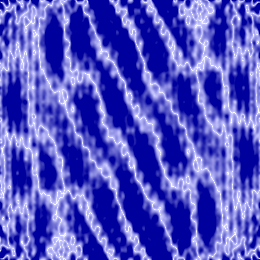};
\node[fill=white, align=center]
at (axis cs:35,35) {34};
\end{axis}

\begin{axis}[%
width=0.787in,
height=0.787in,
at={(1.575in,2.362in)},
scale only axis,
point meta min=0,
point meta max=0.35,
axis on top,
xmin=0,
xmax=260,
xtick={\empty},
y dir=reverse,
ymin=0,
ymax=260,
ytick={\empty},
axis background/.style={fill=white}
]
\addplot [forget plot] graphics [xmin=0.5, xmax=260.5, ymin=0.5, ymax=260.5] {modes_Uhat-35.png};
\node[fill=white, align=center]
at (axis cs:35,35) {35};
\end{axis}

\begin{axis}[%
width=0.787in,
height=0.787in,
at={(2.362in,2.362in)},
scale only axis,
point meta min=0,
point meta max=0.35,
axis on top,
xmin=0,
xmax=260,
xtick={\empty},
y dir=reverse,
ymin=0,
ymax=260,
ytick={\empty},
axis background/.style={fill=white}
]
\addplot [forget plot] graphics [xmin=0.5, xmax=260.5, ymin=0.5, ymax=260.5] {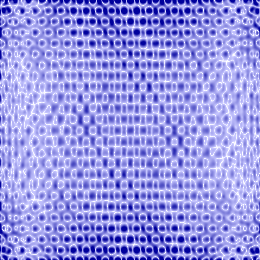};
\node[fill=white, align=center]
at (axis cs:35,35) {36};
\end{axis}

\begin{axis}[%
width=0.787in,
height=0.787in,
at={(3.15in,2.362in)},
scale only axis,
point meta min=0,
point meta max=0.35,
axis on top,
xmin=0,
xmax=260,
xtick={\empty},
y dir=reverse,
ymin=0,
ymax=260,
ytick={\empty},
axis background/.style={fill=white}
]
\addplot [forget plot] graphics [xmin=0.5, xmax=260.5, ymin=0.5, ymax=260.5] {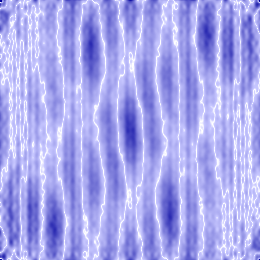};
\node[fill=white, align=center]
at (axis cs:35,35) {37};
\end{axis}

\begin{axis}[%
width=0.787in,
height=0.787in,
at={(3.937in,2.362in)},
scale only axis,
point meta min=0,
point meta max=0.35,
axis on top,
xmin=0,
xmax=260,
xtick={\empty},
y dir=reverse,
ymin=0,
ymax=260,
ytick={\empty},
axis background/.style={fill=white}
]
\addplot [forget plot] graphics [xmin=0.5, xmax=260.5, ymin=0.5, ymax=260.5] {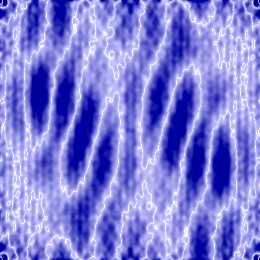};
\node[fill=white, align=center]
at (axis cs:35,35) {38};
\end{axis}

\begin{axis}[%
width=0.787in,
height=0.787in,
at={(4.724in,2.362in)},
scale only axis,
point meta min=0,
point meta max=0.35,
axis on top,
xmin=0,
xmax=260,
xtick={\empty},
y dir=reverse,
ymin=0,
ymax=260,
ytick={\empty},
axis background/.style={fill=white}
]
\addplot [forget plot] graphics [xmin=0.5, xmax=260.5, ymin=0.5, ymax=260.5] {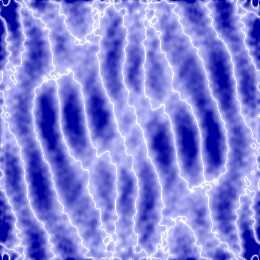};
\node[fill=white, align=center]
at (axis cs:35,35) {39};
\end{axis}

\begin{axis}[%
width=0.787in,
height=0.787in,
at={(5.512in,2.362in)},
scale only axis,
point meta min=0,
point meta max=0.35,
axis on top,
xmin=0,
xmax=260,
xtick={\empty},
y dir=reverse,
ymin=0,
ymax=260,
ytick={\empty},
axis background/.style={fill=white}
]
\addplot [forget plot] graphics [xmin=0.5, xmax=260.5, ymin=0.5, ymax=260.5] {modes_Uhat-40.png};
\node[fill=white, align=center]
at (axis cs:35,35) {40};
\end{axis}

\begin{axis}[%
width=0.787in,
height=0.787in,
at={(0in,1.575in)},
scale only axis,
point meta min=0,
point meta max=0.35,
axis on top,
xmin=0,
xmax=260,
xtick={\empty},
y dir=reverse,
ymin=0,
ymax=260,
ytick={\empty},
axis background/.style={fill=white}
]
\addplot [forget plot] graphics [xmin=0.5, xmax=260.5, ymin=0.5, ymax=260.5] {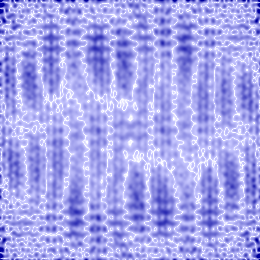};
\node[fill=white, align=center]
at (axis cs:35,35) {41};
\end{axis}

\begin{axis}[%
width=0.787in,
height=0.787in,
at={(0.787in,1.575in)},
scale only axis,
point meta min=0,
point meta max=0.35,
axis on top,
xmin=0,
xmax=260,
xtick={\empty},
y dir=reverse,
ymin=0,
ymax=260,
ytick={\empty},
axis background/.style={fill=white}
]
\addplot [forget plot] graphics [xmin=0.5, xmax=260.5, ymin=0.5, ymax=260.5] {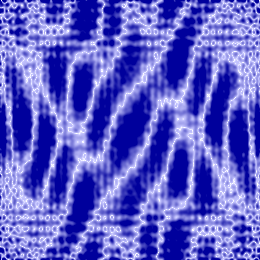};
\node[fill=white, align=center]
at (axis cs:35,35) {42};
\end{axis}

\begin{axis}[%
width=0.787in,
height=0.787in,
at={(1.575in,1.575in)},
scale only axis,
point meta min=0,
point meta max=0.35,
axis on top,
xmin=0,
xmax=260,
xtick={\empty},
y dir=reverse,
ymin=0,
ymax=260,
ytick={\empty},
axis background/.style={fill=white}
]
\addplot [forget plot] graphics [xmin=0.5, xmax=260.5, ymin=0.5, ymax=260.5] {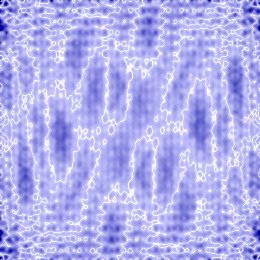};
\node[fill=white, align=center]
at (axis cs:35,35) {43};
\end{axis}

\begin{axis}[%
width=0.787in,
height=0.787in,
at={(2.362in,1.575in)},
scale only axis,
point meta min=0,
point meta max=0.35,
axis on top,
xmin=0,
xmax=260,
xtick={\empty},
y dir=reverse,
ymin=0,
ymax=260,
ytick={\empty},
axis background/.style={fill=white}
]
\addplot [forget plot] graphics [xmin=0.5, xmax=260.5, ymin=0.5, ymax=260.5] {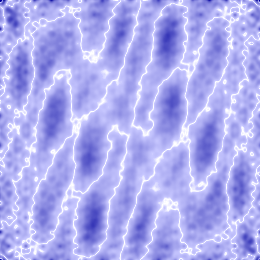};
\node[fill=white, align=center]
at (axis cs:35,35) {44};
\end{axis}

\begin{axis}[%
width=0.787in,
height=0.787in,
at={(3.15in,1.575in)},
scale only axis,
point meta min=0,
point meta max=0.35,
axis on top,
xmin=0,
xmax=260,
xtick={\empty},
y dir=reverse,
ymin=0,
ymax=260,
ytick={\empty},
axis background/.style={fill=white}
]
\addplot [forget plot] graphics [xmin=0.5, xmax=260.5, ymin=0.5, ymax=260.5] {modes_Uhat-45.png};
\node[fill=white, align=center]
at (axis cs:35,35) {45};
\end{axis}

\begin{axis}[%
width=0.787in,
height=0.787in,
at={(3.937in,1.575in)},
scale only axis,
point meta min=0,
point meta max=0.35,
axis on top,
xmin=0,
xmax=260,
xtick={\empty},
y dir=reverse,
ymin=0,
ymax=260,
ytick={\empty},
axis background/.style={fill=white}
]
\addplot [forget plot] graphics [xmin=0.5, xmax=260.5, ymin=0.5, ymax=260.5] {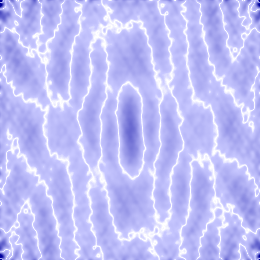};
\node[fill=white, align=center]
at (axis cs:35,35) {46};
\end{axis}

\begin{axis}[%
width=0.787in,
height=0.787in,
at={(4.724in,1.575in)},
scale only axis,
point meta min=0,
point meta max=0.35,
axis on top,
xmin=0,
xmax=260,
xtick={\empty},
y dir=reverse,
ymin=0,
ymax=260,
ytick={\empty},
axis background/.style={fill=white}
]
\addplot [forget plot] graphics [xmin=0.5, xmax=260.5, ymin=0.5, ymax=260.5] {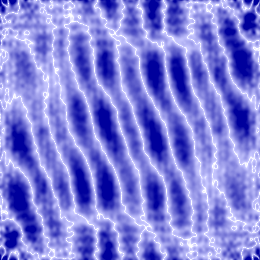};
\node[fill=white, align=center]
at (axis cs:35,35) {47};
\end{axis}

\begin{axis}[%
width=0.787in,
height=0.787in,
at={(5.512in,1.575in)},
scale only axis,
point meta min=0,
point meta max=0.35,
axis on top,
xmin=0,
xmax=260,
xtick={\empty},
y dir=reverse,
ymin=0,
ymax=260,
ytick={\empty},
axis background/.style={fill=white}
]
\addplot [forget plot] graphics [xmin=0.5, xmax=260.5, ymin=0.5, ymax=260.5] {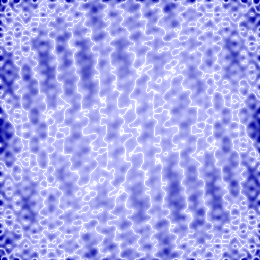};
\node[fill=white, align=center]
at (axis cs:35,35) {48};
\end{axis}

\begin{axis}[%
width=0.787in,
height=0.787in,
at={(0in,0.787in)},
scale only axis,
point meta min=0,
point meta max=0.35,
axis on top,
xmin=0,
xmax=260,
xtick={\empty},
y dir=reverse,
ymin=0,
ymax=260,
ytick={\empty},
axis background/.style={fill=white}
]
\addplot [forget plot] graphics [xmin=0.5, xmax=260.5, ymin=0.5, ymax=260.5] {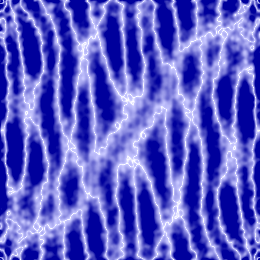};
\node[fill=white, align=center]
at (axis cs:35,35) {49};
\end{axis}

\begin{axis}[%
width=0.787in,
height=0.787in,
at={(0.787in,0.787in)},
scale only axis,
point meta min=0,
point meta max=0.35,
axis on top,
xmin=0,
xmax=260,
xtick={\empty},
y dir=reverse,
ymin=0,
ymax=260,
ytick={\empty},
axis background/.style={fill=white}
]
\addplot [forget plot] graphics [xmin=0.5, xmax=260.5, ymin=0.5, ymax=260.5] {modes_Uhat-50.png};
\node[fill=white, align=center]
at (axis cs:35,35) {50};
\end{axis}

\begin{axis}[%
width=0.787in,
height=0.787in,
at={(1.575in,0.787in)},
scale only axis,
point meta min=0,
point meta max=0.35,
axis on top,
xmin=0,
xmax=260,
xtick={\empty},
y dir=reverse,
ymin=0,
ymax=260,
ytick={\empty},
axis background/.style={fill=white}
]
\addplot [forget plot] graphics [xmin=0.5, xmax=260.5, ymin=0.5, ymax=260.5] {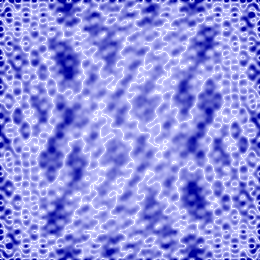};
\node[fill=white, align=center]
at (axis cs:35,35) {51};
\end{axis}

\begin{axis}[%
width=0.787in,
height=0.787in,
at={(2.362in,0.787in)},
scale only axis,
point meta min=0,
point meta max=0.35,
axis on top,
xmin=0,
xmax=260,
xtick={\empty},
y dir=reverse,
ymin=0,
ymax=260,
ytick={\empty},
axis background/.style={fill=white}
]
\addplot [forget plot] graphics [xmin=0.5, xmax=260.5, ymin=0.5, ymax=260.5] {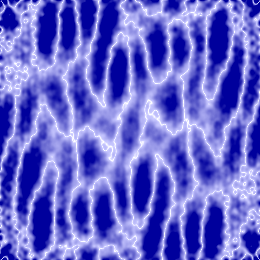};
\node[fill=white, align=center]
at (axis cs:35,35) {52};
\end{axis}

\begin{axis}[%
width=0.787in,
height=0.787in,
at={(3.15in,0.787in)},
scale only axis,
point meta min=0,
point meta max=0.35,
axis on top,
xmin=0,
xmax=260,
xtick={\empty},
y dir=reverse,
ymin=0,
ymax=260,
ytick={\empty},
axis background/.style={fill=white}
]
\addplot [forget plot] graphics [xmin=0.5, xmax=260.5, ymin=0.5, ymax=260.5] {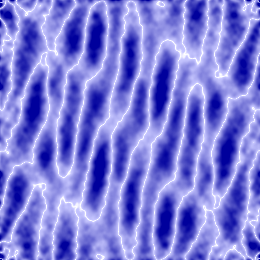};
\node[fill=white, align=center]
at (axis cs:35,35) {53};
\end{axis}

\begin{axis}[%
width=0.787in,
height=0.787in,
at={(3.937in,0.787in)},
scale only axis,
point meta min=0,
point meta max=0.35,
axis on top,
xmin=0,
xmax=260,
xtick={\empty},
y dir=reverse,
ymin=0,
ymax=260,
ytick={\empty},
axis background/.style={fill=white}
]
\addplot [forget plot] graphics [xmin=0.5, xmax=260.5, ymin=0.5, ymax=260.5] {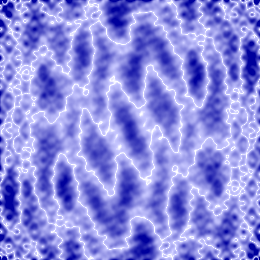};
\node[fill=white, align=center]
at (axis cs:35,35) {54};
\end{axis}

\begin{axis}[%
width=0.787in,
height=0.787in,
at={(4.724in,0.787in)},
scale only axis,
point meta min=0,
point meta max=0.35,
axis on top,
xmin=0,
xmax=260,
xtick={\empty},
y dir=reverse,
ymin=0,
ymax=260,
ytick={\empty},
axis background/.style={fill=white}
]
\addplot [forget plot] graphics [xmin=0.5, xmax=260.5, ymin=0.5, ymax=260.5] {modes_Uhat-55.png};
\node[fill=white, align=center]
at (axis cs:35,35) {55};
\end{axis}

\begin{axis}[%
width=0.787in,
height=0.787in,
at={(5.512in,0.787in)},
scale only axis,
point meta min=0,
point meta max=0.35,
axis on top,
xmin=0,
xmax=260,
xtick={\empty},
y dir=reverse,
ymin=0,
ymax=260,
ytick={\empty},
axis background/.style={fill=white}
]
\addplot [forget plot] graphics [xmin=0.5, xmax=260.5, ymin=0.5, ymax=260.5] {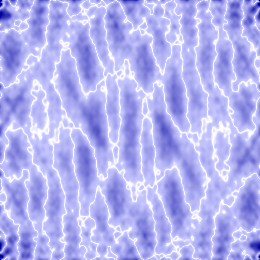};
\node[fill=white, align=center]
at (axis cs:35,35) {56};
\end{axis}

\begin{axis}[%
width=0.787in,
height=0.787in,
at={(0in,0in)},
scale only axis,
point meta min=0,
point meta max=0.35,
axis on top,
xmin=0,
xmax=260,
xtick={\empty},
y dir=reverse,
ymin=0,
ymax=260,
ytick={\empty},
axis background/.style={fill=white}
]
\addplot [forget plot] graphics [xmin=0.5, xmax=260.5, ymin=0.5, ymax=260.5] {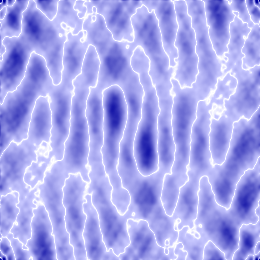};
\node[fill=white, align=center]
at (axis cs:35,35) {57};
\end{axis}

\begin{axis}[%
width=0.787in,
height=0.787in,
at={(0.787in,0in)},
scale only axis,
point meta min=0,
point meta max=0.35,
axis on top,
xmin=0,
xmax=260,
xtick={\empty},
y dir=reverse,
ymin=0,
ymax=260,
ytick={\empty},
axis background/.style={fill=white}
]
\addplot [forget plot] graphics [xmin=0.5, xmax=260.5, ymin=0.5, ymax=260.5] {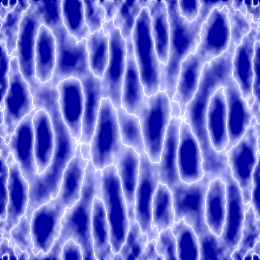};
\node[fill=white, align=center]
at (axis cs:35,35) {58};
\end{axis}

\begin{axis}[%
width=0.787in,
height=0.787in,
at={(1.575in,0in)},
scale only axis,
point meta min=0,
point meta max=0.35,
axis on top,
xmin=0,
xmax=260,
xtick={\empty},
y dir=reverse,
ymin=0,
ymax=260,
ytick={\empty},
axis background/.style={fill=white}
]
\addplot [forget plot] graphics [xmin=0.5, xmax=260.5, ymin=0.5, ymax=260.5] {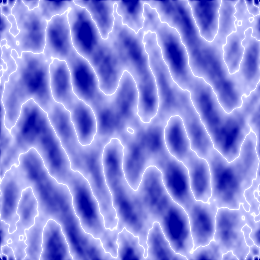};
\node[fill=white, align=center]
at (axis cs:35,35) {59};
\end{axis}

\begin{axis}[%
width=0.787in,
height=0.787in,
at={(2.362in,0in)},
scale only axis,
point meta min=0,
point meta max=0.35,
axis on top,
xmin=0,
xmax=260,
xtick={\empty},
y dir=reverse,
ymin=0,
ymax=260,
ytick={\empty},
axis background/.style={fill=white}
]
\addplot [forget plot] graphics [xmin=0.5, xmax=260.5, ymin=0.5, ymax=260.5] {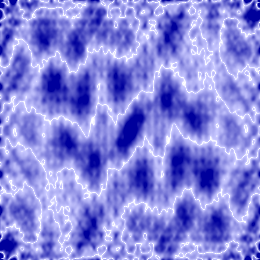};
\node[fill=white, align=center]
at (axis cs:35,35) {60};
\end{axis}

\begin{axis}[%
width=0.787in,
height=0.787in,
at={(3.15in,0in)},
scale only axis,
point meta min=0,
point meta max=0.35,
axis on top,
xmin=0,
xmax=260,
xtick={\empty},
y dir=reverse,
ymin=0,
ymax=260,
ytick={\empty},
axis background/.style={fill=white}
]
\addplot [forget plot] graphics [xmin=0.5, xmax=260.5, ymin=0.5, ymax=260.5] {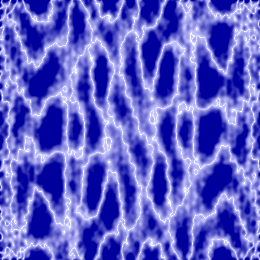};
\node[fill=white, align=center]
at (axis cs:35,35) {61};
\end{axis}

\begin{axis}[%
width=0.787in,
height=0.787in,
at={(3.937in,0in)},
scale only axis,
point meta min=0,
point meta max=0.35,
axis on top,
xmin=0,
xmax=260,
xtick={\empty},
y dir=reverse,
ymin=0,
ymax=260,
ytick={\empty},
axis background/.style={fill=white}
]
\addplot [forget plot] graphics [xmin=0.5, xmax=260.5, ymin=0.5, ymax=260.5] {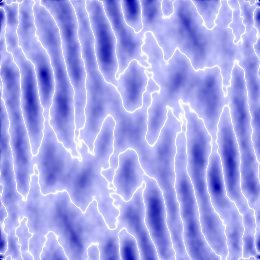};
\node[fill=white, align=center]
at (axis cs:35,35) {62};
\end{axis}

\begin{axis}[%
width=0.787in,
height=0.787in,
at={(4.724in,0in)},
scale only axis,
point meta min=0,
point meta max=0.35,
axis on top,
xmin=0,
xmax=260,
xtick={\empty},
y dir=reverse,
ymin=0,
ymax=260,
ytick={\empty},
axis background/.style={fill=white}
]
\addplot [forget plot] graphics [xmin=0.5, xmax=260.5, ymin=0.5, ymax=260.5] {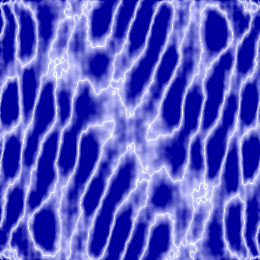};
\node[fill=white, align=center]
at (axis cs:35,35) {63};
\end{axis}

\begin{axis}[%
width=0.787in,
height=0.787in,
at={(5.512in,0in)},
scale only axis,
point meta min=0,
point meta max=0.35,
axis on top,
xmin=0,
xmax=260,
xtick={\empty},
y dir=reverse,
ymin=0,
ymax=260,
ytick={\empty},
axis background/.style={fill=white},
colormap={mymap}{[1pt] rgb(0pt)=(1,1,1); rgb(11pt)=(0.780392,0.780392,1); rgb(255pt)=(0,0,0.611765)},
colorbar horizontal,
colorbar style={
at={(1,-0.3)},anchor=south east,
width=8*
\pgfkeysvalueof{/pgfplots/parent axis width},
xticklabel pos=lower,
xtick={
0,
0.1,
0.2,
0.3
},
},
]
\addplot [forget plot] graphics [xmin=0.5, xmax=260.5, ymin=0.5, ymax=260.5] {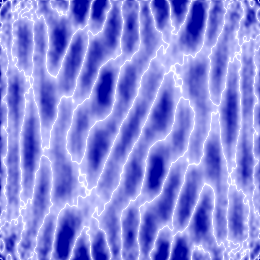};
\node[fill=white, align=center]
at (axis cs:35,35) {64};
\end{axis}

\begin{axis}[%
width=6.299in,
height=6.299in,
at={(0in,0in)},
scale only axis,
xmin=0,
xmax=1,
ymin=0,
ymax=1,
axis line style={draw=none},
ticks=none,
axis x line*=bottom,
axis y line*=left
]
\end{axis}
\end{tikzpicture}%

%% file: Figures/Modes_k50/rSVDmodes50_Uhat.tex
\begin{tikzpicture}

\begin{axis}[%
width=0.629in,
height=0.629in,
at={(0in,2.52in)},
scale only axis,
point meta min=0,
point meta max=0.35,
axis on top,
xmin=0,
xmax=260,
xtick={\empty},
y dir=reverse,
ymin=0,
ymax=260,
ytick={\empty},
axis background/.style={fill=white}
]
\addplot [forget plot] graphics [xmin=0.5, xmax=260.5, ymin=0.5, ymax=260.5] {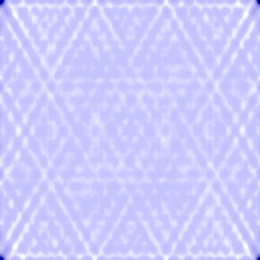};
\node[fill=white, align=center]
at (axis cs:35,35) {\footnotesize 01};
\end{axis}

\begin{axis}[%
width=0.629in,
height=0.629in,
at={(0.63in,2.52in)},
scale only axis,
point meta min=0,
point meta max=0.35,
axis on top,
xmin=0,
xmax=260,
xtick={\empty},
y dir=reverse,
ymin=0,
ymax=260,
ytick={\empty},
axis background/.style={fill=white}
]
\addplot [forget plot] graphics [xmin=0.5, xmax=260.5, ymin=0.5, ymax=260.5] {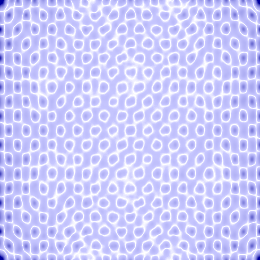};
\node[fill=white, align=center]
at (axis cs:35,35) {\footnotesize 02};
\end{axis}

\begin{axis}[%
width=0.629in,
height=0.629in,
at={(1.26in,2.52in)},
scale only axis,
point meta min=0,
point meta max=0.35,
axis on top,
xmin=0,
xmax=260,
xtick={\empty},
y dir=reverse,
ymin=0,
ymax=260,
ytick={\empty},
axis background/.style={fill=white}
]
\addplot [forget plot] graphics [xmin=0.5, xmax=260.5, ymin=0.5, ymax=260.5] {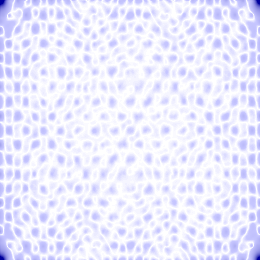};
\node[fill=white, align=center]
at (axis cs:35,35) {\footnotesize 03};
\end{axis}

\begin{axis}[%
width=0.629in,
height=0.629in,
at={(1.89in,2.52in)},
scale only axis,
point meta min=0,
point meta max=0.35,
axis on top,
xmin=0,
xmax=260,
xtick={\empty},
y dir=reverse,
ymin=0,
ymax=260,
ytick={\empty},
axis background/.style={fill=white}
]
\addplot [forget plot] graphics [xmin=0.5, xmax=260.5, ymin=0.5, ymax=260.5] {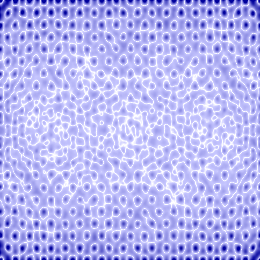};
\node[fill=white, align=center]
at (axis cs:35,35) {\footnotesize 04};
\end{axis}

\begin{axis}[%
width=0.629in,
height=0.629in,
at={(2.52in,2.52in)},
scale only axis,
point meta min=0,
point meta max=0.35,
axis on top,
xmin=0,
xmax=260,
xtick={\empty},
y dir=reverse,
ymin=0,
ymax=260,
ytick={\empty},
axis background/.style={fill=white}
]
\addplot [forget plot] graphics [xmin=0.5, xmax=260.5, ymin=0.5, ymax=260.5] {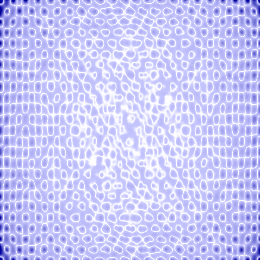};
\node[fill=white, align=center]
at (axis cs:35,35) {\footnotesize 05};
\end{axis}

\begin{axis}[%
width=0.629in,
height=0.629in,
at={(3.15in,2.52in)},
scale only axis,
point meta min=0,
point meta max=0.35,
axis on top,
xmin=0,
xmax=260,
xtick={\empty},
y dir=reverse,
ymin=0,
ymax=260,
ytick={\empty},
axis background/.style={fill=white}
]
\addplot [forget plot] graphics [xmin=0.5, xmax=260.5, ymin=0.5, ymax=260.5] {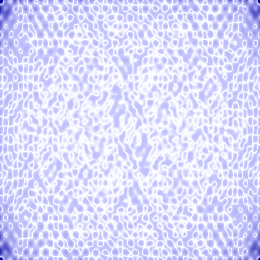};
\node[fill=white, align=center]
at (axis cs:35,35) {\footnotesize 06};
\end{axis}

\begin{axis}[%
width=0.629in,
height=0.629in,
at={(3.78in,2.52in)},
scale only axis,
point meta min=0,
point meta max=0.35,
axis on top,
xmin=0,
xmax=260,
xtick={\empty},
y dir=reverse,
ymin=0,
ymax=260,
ytick={\empty},
axis background/.style={fill=white}
]
\addplot [forget plot] graphics [xmin=0.5, xmax=260.5, ymin=0.5, ymax=260.5] {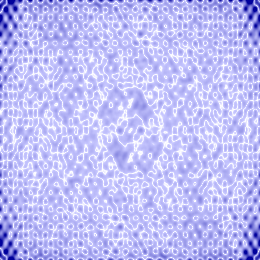};
\node[fill=white, align=center]
at (axis cs:35,35) {\footnotesize 07};
\end{axis}

\begin{axis}[%
width=0.629in,
height=0.629in,
at={(4.41in,2.52in)},
scale only axis,
point meta min=0,
point meta max=0.35,
axis on top,
xmin=0,
xmax=260,
xtick={\empty},
y dir=reverse,
ymin=0,
ymax=260,
ytick={\empty},
axis background/.style={fill=white}
]
\addplot [forget plot] graphics [xmin=0.5, xmax=260.5, ymin=0.5, ymax=260.5] {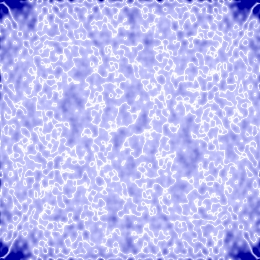};
\node[fill=white, align=center]
at (axis cs:35,35) {\footnotesize 08};
\end{axis}

\begin{axis}[%
width=0.629in,
height=0.629in,
at={(5.04in,2.52in)},
scale only axis,
point meta min=0,
point meta max=0.35,
axis on top,
xmin=0,
xmax=260,
xtick={\empty},
y dir=reverse,
ymin=0,
ymax=260,
ytick={\empty},
axis background/.style={fill=white}
]
\addplot [forget plot] graphics [xmin=0.5, xmax=260.5, ymin=0.5, ymax=260.5] {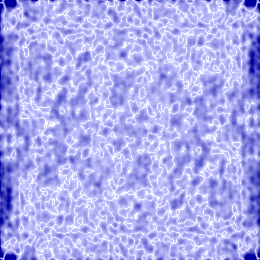};
\node[fill=white, align=center]
at (axis cs:35,35) {\footnotesize 09};
\end{axis}

\begin{axis}[%
width=0.629in,
height=0.629in,
at={(5.67in,2.52in)},
scale only axis,
point meta min=0,
point meta max=0.35,
axis on top,
xmin=0,
xmax=260,
xtick={\empty},
y dir=reverse,
ymin=0,
ymax=260,
ytick={\empty},
axis background/.style={fill=white}
]
\addplot [forget plot] graphics [xmin=0.5, xmax=260.5, ymin=0.5, ymax=260.5] {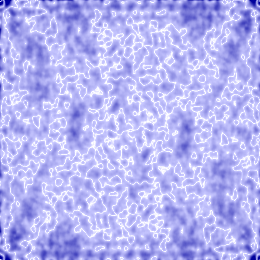};
\node[fill=white, align=center]
at (axis cs:35,35) {\footnotesize 10};
\end{axis}

\begin{axis}[%
width=0.629in,
height=0.629in,
at={(0in,1.89in)},
scale only axis,
point meta min=0,
point meta max=0.35,
axis on top,
xmin=0,
xmax=260,
xtick={\empty},
y dir=reverse,
ymin=0,
ymax=260,
ytick={\empty},
axis background/.style={fill=white}
]
\addplot [forget plot] graphics [xmin=0.5, xmax=260.5, ymin=0.5, ymax=260.5] {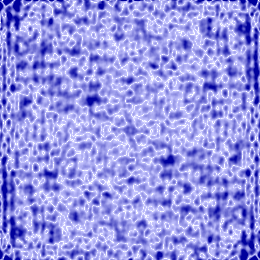};
\node[fill=white, align=center]
at (axis cs:35,35) {\footnotesize 11};
\end{axis}

\begin{axis}[%
width=0.629in,
height=0.629in,
at={(0.63in,1.89in)},
scale only axis,
point meta min=0,
point meta max=0.35,
axis on top,
xmin=0,
xmax=260,
xtick={\empty},
y dir=reverse,
ymin=0,
ymax=260,
ytick={\empty},
axis background/.style={fill=white}
]
\addplot [forget plot] graphics [xmin=0.5, xmax=260.5, ymin=0.5, ymax=260.5] {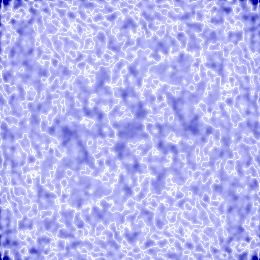};
\node[fill=white, align=center]
at (axis cs:35,35) {\footnotesize 12};
\end{axis}

\begin{axis}[%
width=0.629in,
height=0.629in,
at={(1.26in,1.89in)},
scale only axis,
point meta min=0,
point meta max=0.35,
axis on top,
xmin=0,
xmax=260,
xtick={\empty},
y dir=reverse,
ymin=0,
ymax=260,
ytick={\empty},
axis background/.style={fill=white}
]
\addplot [forget plot] graphics [xmin=0.5, xmax=260.5, ymin=0.5, ymax=260.5] {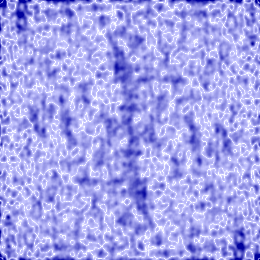};
\node[fill=white, align=center]
at (axis cs:35,35) {\footnotesize 13};
\end{axis}

\begin{axis}[%
width=0.629in,
height=0.629in,
at={(1.89in,1.89in)},
scale only axis,
point meta min=0,
point meta max=0.35,
axis on top,
xmin=0,
xmax=260,
xtick={\empty},
y dir=reverse,
ymin=0,
ymax=260,
ytick={\empty},
axis background/.style={fill=white}
]
\addplot [forget plot] graphics [xmin=0.5, xmax=260.5, ymin=0.5, ymax=260.5] {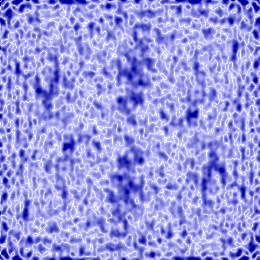};
\node[fill=white, align=center]
at (axis cs:35,35) {\footnotesize 14};
\end{axis}

\begin{axis}[%
width=0.629in,
height=0.629in,
at={(2.52in,1.89in)},
scale only axis,
point meta min=0,
point meta max=0.35,
axis on top,
xmin=0,
xmax=260,
xtick={\empty},
y dir=reverse,
ymin=0,
ymax=260,
ytick={\empty},
axis background/.style={fill=white}
]
\addplot [forget plot] graphics [xmin=0.5, xmax=260.5, ymin=0.5, ymax=260.5] {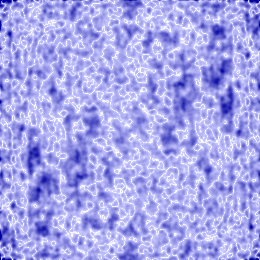};
\node[fill=white, align=center]
at (axis cs:35,35) {\footnotesize 15};
\end{axis}

\begin{axis}[%
width=0.629in,
height=0.629in,
at={(3.15in,1.89in)},
scale only axis,
point meta min=0,
point meta max=0.35,
axis on top,
xmin=0,
xmax=260,
xtick={\empty},
y dir=reverse,
ymin=0,
ymax=260,
ytick={\empty},
axis background/.style={fill=white}
]
\addplot [forget plot] graphics [xmin=0.5, xmax=260.5, ymin=0.5, ymax=260.5] {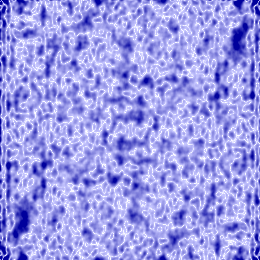};
\node[fill=white, align=center]
at (axis cs:35,35) {\footnotesize 16};
\end{axis}

\begin{axis}[%
width=0.629in,
height=0.629in,
at={(3.78in,1.89in)},
scale only axis,
point meta min=0,
point meta max=0.35,
axis on top,
xmin=0,
xmax=260,
xtick={\empty},
y dir=reverse,
ymin=0,
ymax=260,
ytick={\empty},
axis background/.style={fill=white}
]
\addplot [forget plot] graphics [xmin=0.5, xmax=260.5, ymin=0.5, ymax=260.5] {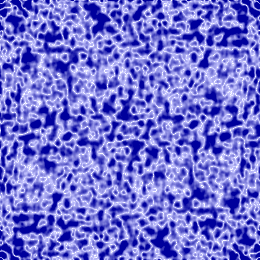};
\node[fill=white, align=center]
at (axis cs:35,35) {\footnotesize 17};
\end{axis}

\begin{axis}[%
width=0.629in,
height=0.629in,
at={(4.41in,1.89in)},
scale only axis,
point meta min=0,
point meta max=0.35,
axis on top,
xmin=0,
xmax=260,
xtick={\empty},
y dir=reverse,
ymin=0,
ymax=260,
ytick={\empty},
axis background/.style={fill=white}
]
\addplot [forget plot] graphics [xmin=0.5, xmax=260.5, ymin=0.5, ymax=260.5] {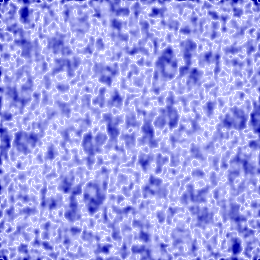};
\node[fill=white, align=center]
at (axis cs:35,35) {\footnotesize 18};
\end{axis}

\begin{axis}[%
width=0.629in,
height=0.629in,
at={(5.04in,1.89in)},
scale only axis,
point meta min=0,
point meta max=0.35,
axis on top,
xmin=0,
xmax=260,
xtick={\empty},
y dir=reverse,
ymin=0,
ymax=260,
ytick={\empty},
axis background/.style={fill=white}
]
\addplot [forget plot] graphics [xmin=0.5, xmax=260.5, ymin=0.5, ymax=260.5] {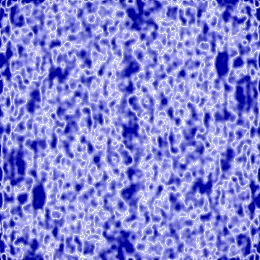};
\node[fill=white, align=center]
at (axis cs:35,35) {\footnotesize 19};
\end{axis}

\begin{axis}[%
width=0.629in,
height=0.629in,
at={(5.67in,1.89in)},
scale only axis,
point meta min=0,
point meta max=0.35,
axis on top,
xmin=0,
xmax=260,
xtick={\empty},
y dir=reverse,
ymin=0,
ymax=260,
ytick={\empty},
axis background/.style={fill=white}
]
\addplot [forget plot] graphics [xmin=0.5, xmax=260.5, ymin=0.5, ymax=260.5] {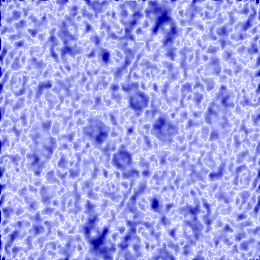};
\node[fill=white, align=center]
at (axis cs:35,35) {\footnotesize 20};
\end{axis}

\begin{axis}[%
width=0.629in,
height=0.629in,
at={(0in,1.26in)},
scale only axis,
point meta min=0,
point meta max=0.35,
axis on top,
xmin=0,
xmax=260,
xtick={\empty},
y dir=reverse,
ymin=0,
ymax=260,
ytick={\empty},
axis background/.style={fill=white}
]
\addplot [forget plot] graphics [xmin=0.5, xmax=260.5, ymin=0.5, ymax=260.5] {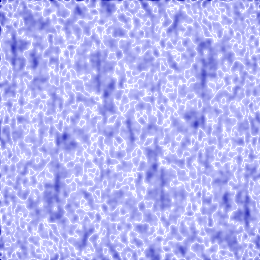};
\node[fill=white, align=center]
at (axis cs:35,35) {\footnotesize 21};
\end{axis}

\begin{axis}[%
width=0.629in,
height=0.629in,
at={(0.63in,1.26in)},
scale only axis,
point meta min=0,
point meta max=0.35,
axis on top,
xmin=0,
xmax=260,
xtick={\empty},
y dir=reverse,
ymin=0,
ymax=260,
ytick={\empty},
axis background/.style={fill=white}
]
\addplot [forget plot] graphics [xmin=0.5, xmax=260.5, ymin=0.5, ymax=260.5] {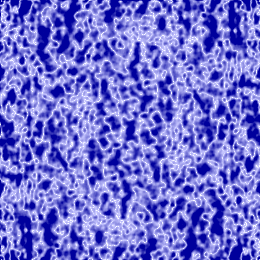};
\node[fill=white, align=center]
at (axis cs:35,35) {\footnotesize 22};
\end{axis}

\begin{axis}[%
width=0.629in,
height=0.629in,
at={(1.26in,1.26in)},
scale only axis,
point meta min=0,
point meta max=0.35,
axis on top,
xmin=0,
xmax=260,
xtick={\empty},
y dir=reverse,
ymin=0,
ymax=260,
ytick={\empty},
axis background/.style={fill=white}
]
\addplot [forget plot] graphics [xmin=0.5, xmax=260.5, ymin=0.5, ymax=260.5] {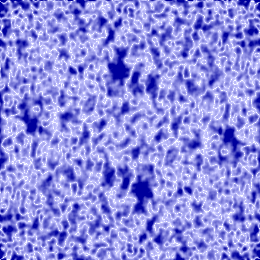};
\node[fill=white, align=center]
at (axis cs:35,35) {\footnotesize 23};
\end{axis}

\begin{axis}[%
width=0.629in,
height=0.629in,
at={(1.89in,1.26in)},
scale only axis,
point meta min=0,
point meta max=0.35,
axis on top,
xmin=0,
xmax=260,
xtick={\empty},
y dir=reverse,
ymin=0,
ymax=260,
ytick={\empty},
axis background/.style={fill=white}
]
\addplot [forget plot] graphics [xmin=0.5, xmax=260.5, ymin=0.5, ymax=260.5] {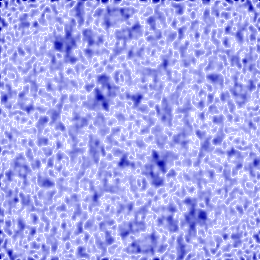};
\node[fill=white, align=center]
at (axis cs:35,35) {\footnotesize 24};
\end{axis}

\begin{axis}[%
width=0.629in,
height=0.629in,
at={(2.52in,1.26in)},
scale only axis,
point meta min=0,
point meta max=0.35,
axis on top,
xmin=0,
xmax=260,
xtick={\empty},
y dir=reverse,
ymin=0,
ymax=260,
ytick={\empty},
axis background/.style={fill=white}
]
\addplot [forget plot] graphics [xmin=0.5, xmax=260.5, ymin=0.5, ymax=260.5] {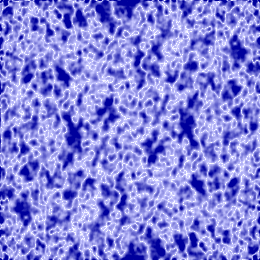};
\node[fill=white, align=center]
at (axis cs:35,35) {\footnotesize 25};
\end{axis}

\begin{axis}[%
width=0.629in,
height=0.629in,
at={(3.15in,1.26in)},
scale only axis,
point meta min=0,
point meta max=0.35,
axis on top,
xmin=0,
xmax=260,
xtick={\empty},
y dir=reverse,
ymin=0,
ymax=260,
ytick={\empty},
axis background/.style={fill=white}
]
\addplot [forget plot] graphics [xmin=0.5, xmax=260.5, ymin=0.5, ymax=260.5] {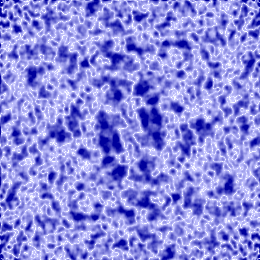};
\node[fill=white, align=center]
at (axis cs:35,35) {\footnotesize 26};
\end{axis}

\begin{axis}[%
width=0.629in,
height=0.629in,
at={(3.78in,1.26in)},
scale only axis,
point meta min=0,
point meta max=0.35,
axis on top,
xmin=0,
xmax=260,
xtick={\empty},
y dir=reverse,
ymin=0,
ymax=260,
ytick={\empty},
axis background/.style={fill=white}
]
\addplot [forget plot] graphics [xmin=0.5, xmax=260.5, ymin=0.5, ymax=260.5] {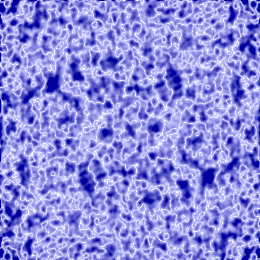};
\node[fill=white, align=center]
at (axis cs:35,35) {\footnotesize 27};
\end{axis}

\begin{axis}[%
width=0.629in,
height=0.629in,
at={(4.41in,1.26in)},
scale only axis,
point meta min=0,
point meta max=0.35,
axis on top,
xmin=0,
xmax=260,
xtick={\empty},
y dir=reverse,
ymin=0,
ymax=260,
ytick={\empty},
axis background/.style={fill=white}
]
\addplot [forget plot] graphics [xmin=0.5, xmax=260.5, ymin=0.5, ymax=260.5] {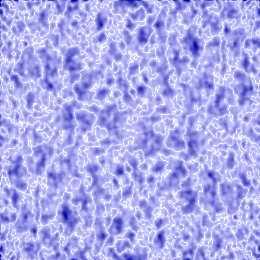};
\node[fill=white, align=center]
at (axis cs:35,35) {\footnotesize 28};
\end{axis}

\begin{axis}[%
width=0.629in,
height=0.629in,
at={(5.04in,1.26in)},
scale only axis,
point meta min=0,
point meta max=0.35,
axis on top,
xmin=0,
xmax=260,
xtick={\empty},
y dir=reverse,
ymin=0,
ymax=260,
ytick={\empty},
axis background/.style={fill=white}
]
\addplot [forget plot] graphics [xmin=0.5, xmax=260.5, ymin=0.5, ymax=260.5] {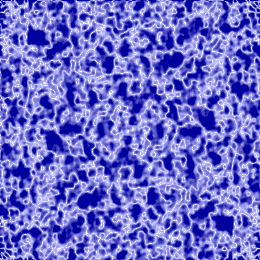};
\node[fill=white, align=center]
at (axis cs:35,35) {\footnotesize 29};
\end{axis}

\begin{axis}[%
width=0.629in,
height=0.629in,
at={(5.67in,1.26in)},
scale only axis,
point meta min=0,
point meta max=0.35,
axis on top,
xmin=0,
xmax=260,
xtick={\empty},
y dir=reverse,
ymin=0,
ymax=260,
ytick={\empty},
axis background/.style={fill=white}
]
\addplot [forget plot] graphics [xmin=0.5, xmax=260.5, ymin=0.5, ymax=260.5] {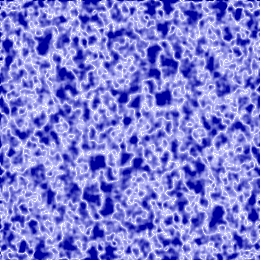};
\node[fill=white, align=center]
at (axis cs:35,35) {\footnotesize 30};
\end{axis}

\begin{axis}[%
width=0.629in,
height=0.629in,
at={(0in,0.63in)},
scale only axis,
point meta min=0,
point meta max=0.35,
axis on top,
xmin=0,
xmax=260,
xtick={\empty},
y dir=reverse,
ymin=0,
ymax=260,
ytick={\empty},
axis background/.style={fill=white}
]
\addplot [forget plot] graphics [xmin=0.5, xmax=260.5, ymin=0.5, ymax=260.5] {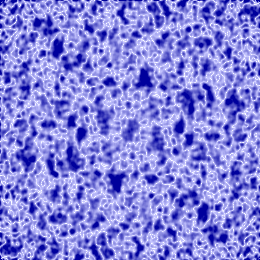};
\node[fill=white, align=center]
at (axis cs:35,35) {\footnotesize 31};
\end{axis}

\begin{axis}[%
width=0.629in,
height=0.629in,
at={(0.63in,0.63in)},
scale only axis,
point meta min=0,
point meta max=0.35,
axis on top,
xmin=0,
xmax=260,
xtick={\empty},
y dir=reverse,
ymin=0,
ymax=260,
ytick={\empty},
axis background/.style={fill=white}
]
\addplot [forget plot] graphics [xmin=0.5, xmax=260.5, ymin=0.5, ymax=260.5] {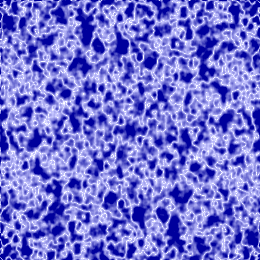};
\node[fill=white, align=center]
at (axis cs:35,35) {\footnotesize 32};
\end{axis}

\begin{axis}[%
width=0.629in,
height=0.629in,
at={(1.26in,0.63in)},
scale only axis,
point meta min=0,
point meta max=0.35,
axis on top,
xmin=0,
xmax=260,
xtick={\empty},
y dir=reverse,
ymin=0,
ymax=260,
ytick={\empty},
axis background/.style={fill=white}
]
\addplot [forget plot] graphics [xmin=0.5, xmax=260.5, ymin=0.5, ymax=260.5] {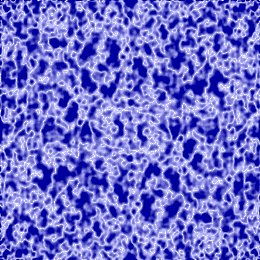};
\node[fill=white, align=center]
at (axis cs:35,35) {\footnotesize 33};
\end{axis}

\begin{axis}[%
width=0.629in,
height=0.629in,
at={(1.89in,0.63in)},
scale only axis,
point meta min=0,
point meta max=0.35,
axis on top,
xmin=0,
xmax=260,
xtick={\empty},
y dir=reverse,
ymin=0,
ymax=260,
ytick={\empty},
axis background/.style={fill=white}
]
\addplot [forget plot] graphics [xmin=0.5, xmax=260.5, ymin=0.5, ymax=260.5] {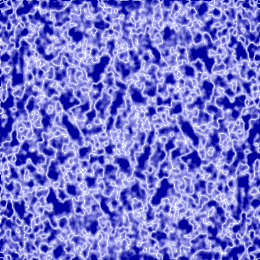};
\node[fill=white, align=center]
at (axis cs:35,35) {\footnotesize 34};
\end{axis}

\begin{axis}[%
width=0.629in,
height=0.629in,
at={(2.52in,0.63in)},
scale only axis,
point meta min=0,
point meta max=0.35,
axis on top,
xmin=0,
xmax=260,
xtick={\empty},
y dir=reverse,
ymin=0,
ymax=260,
ytick={\empty},
axis background/.style={fill=white}
]
\addplot [forget plot] graphics [xmin=0.5, xmax=260.5, ymin=0.5, ymax=260.5] {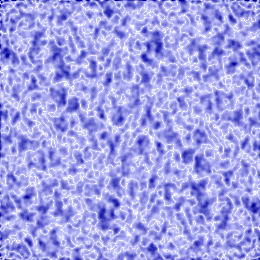};
\node[fill=white, align=center]
at (axis cs:35,35) {\footnotesize 35};
\end{axis}

\begin{axis}[%
width=0.629in,
height=0.629in,
at={(3.15in,0.63in)},
scale only axis,
point meta min=0,
point meta max=0.35,
axis on top,
xmin=0,
xmax=260,
xtick={\empty},
y dir=reverse,
ymin=0,
ymax=260,
ytick={\empty},
axis background/.style={fill=white}
]
\addplot [forget plot] graphics [xmin=0.5, xmax=260.5, ymin=0.5, ymax=260.5] {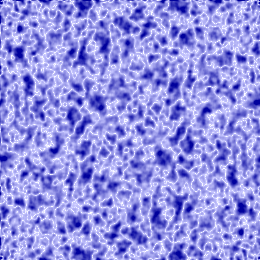};
\node[fill=white, align=center]
at (axis cs:35,35) {\footnotesize 36};
\end{axis}

\begin{axis}[%
width=0.629in,
height=0.629in,
at={(3.78in,0.63in)},
scale only axis,
point meta min=0,
point meta max=0.35,
axis on top,
xmin=0,
xmax=260,
xtick={\empty},
y dir=reverse,
ymin=0,
ymax=260,
ytick={\empty},
axis background/.style={fill=white}
]
\addplot [forget plot] graphics [xmin=0.5, xmax=260.5, ymin=0.5, ymax=260.5] {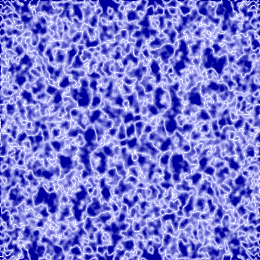};
\node[fill=white, align=center]
at (axis cs:35,35) {\footnotesize 37};
\end{axis}

\begin{axis}[%
width=0.629in,
height=0.629in,
at={(4.41in,0.63in)},
scale only axis,
point meta min=0,
point meta max=0.35,
axis on top,
xmin=0,
xmax=260,
xtick={\empty},
y dir=reverse,
ymin=0,
ymax=260,
ytick={\empty},
axis background/.style={fill=white}
]
\addplot [forget plot] graphics [xmin=0.5, xmax=260.5, ymin=0.5, ymax=260.5] {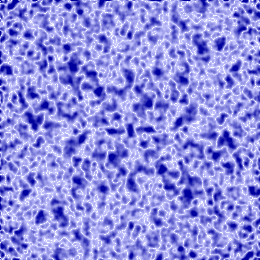};
\node[fill=white, align=center]
at (axis cs:35,35) {\footnotesize 38};
\end{axis}

\begin{axis}[%
width=0.629in,
height=0.629in,
at={(5.04in,0.63in)},
scale only axis,
point meta min=0,
point meta max=0.35,
axis on top,
xmin=0,
xmax=260,
xtick={\empty},
y dir=reverse,
ymin=0,
ymax=260,
ytick={\empty},
axis background/.style={fill=white}
]
\addplot [forget plot] graphics [xmin=0.5, xmax=260.5, ymin=0.5, ymax=260.5] {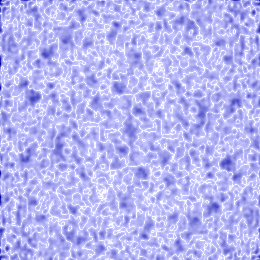};
\node[fill=white, align=center]
at (axis cs:35,35) {\footnotesize 39};
\end{axis}

\begin{axis}[%
width=0.629in,
height=0.629in,
at={(5.67in,0.63in)},
scale only axis,
point meta min=0,
point meta max=0.35,
axis on top,
xmin=0,
xmax=260,
xtick={\empty},
y dir=reverse,
ymin=0,
ymax=260,
ytick={\empty},
axis background/.style={fill=white}
]
\addplot [forget plot] graphics [xmin=0.5, xmax=260.5, ymin=0.5, ymax=260.5] {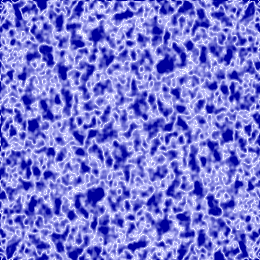};
\node[fill=white, align=center]
at (axis cs:35,35) {\footnotesize 40};
\end{axis}

\begin{axis}[%
width=0.629in,
height=0.629in,
at={(0in,0in)},
scale only axis,
point meta min=0,
point meta max=0.35,
axis on top,
xmin=0,
xmax=260,
xtick={\empty},
y dir=reverse,
ymin=0,
ymax=260,
ytick={\empty},
axis background/.style={fill=white}
]
\addplot [forget plot] graphics [xmin=0.5, xmax=260.5, ymin=0.5, ymax=260.5] {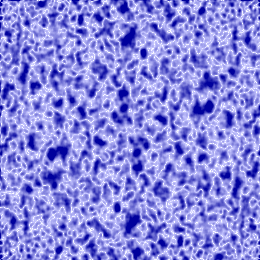};
\node[fill=white, align=center]
at (axis cs:35,35) {\footnotesize 41};
\end{axis}

\begin{axis}[%
width=0.629in,
height=0.629in,
at={(0.63in,0in)},
scale only axis,
point meta min=0,
point meta max=0.35,
axis on top,
xmin=0,
xmax=260,
xtick={\empty},
y dir=reverse,
ymin=0,
ymax=260,
ytick={\empty},
axis background/.style={fill=white}
]
\addplot [forget plot] graphics [xmin=0.5, xmax=260.5, ymin=0.5, ymax=260.5] {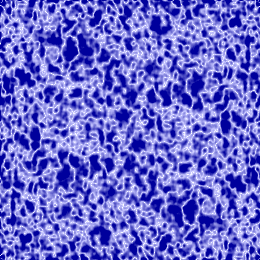};
\node[fill=white, align=center]
at (axis cs:35,35) {\footnotesize 42};
\end{axis}

\begin{axis}[%
width=0.629in,
height=0.629in,
at={(1.26in,0in)},
scale only axis,
point meta min=0,
point meta max=0.35,
axis on top,
xmin=0,
xmax=260,
xtick={\empty},
y dir=reverse,
ymin=0,
ymax=260,
ytick={\empty},
axis background/.style={fill=white}
]
\addplot [forget plot] graphics [xmin=0.5, xmax=260.5, ymin=0.5, ymax=260.5] {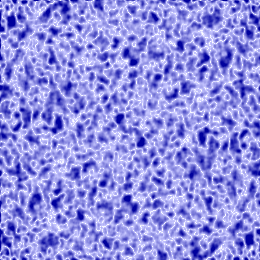};
\node[fill=white, align=center]
at (axis cs:35,35) {\footnotesize 43};
\end{axis}

\begin{axis}[%
width=0.629in,
height=0.629in,
at={(1.89in,0in)},
scale only axis,
point meta min=0,
point meta max=0.35,
axis on top,
xmin=0,
xmax=260,
xtick={\empty},
y dir=reverse,
ymin=0,
ymax=260,
ytick={\empty},
axis background/.style={fill=white}
]
\addplot [forget plot] graphics [xmin=0.5, xmax=260.5, ymin=0.5, ymax=260.5] {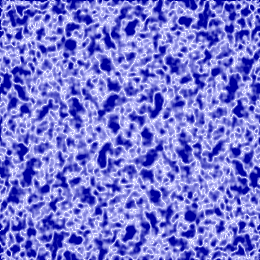};
\node[fill=white, align=center]
at (axis cs:35,35) {\footnotesize 44};
\end{axis}

\begin{axis}[%
width=0.629in,
height=0.629in,
at={(2.52in,0in)},
scale only axis,
point meta min=0,
point meta max=0.35,
axis on top,
xmin=0,
xmax=260,
xtick={\empty},
y dir=reverse,
ymin=0,
ymax=260,
ytick={\empty},
axis background/.style={fill=white}
]
\addplot [forget plot] graphics [xmin=0.5, xmax=260.5, ymin=0.5, ymax=260.5] {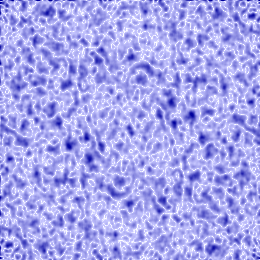};
\node[fill=white, align=center]
at (axis cs:35,35) {\footnotesize 45};
\end{axis}

\begin{axis}[%
width=0.629in,
height=0.629in,
at={(3.15in,0in)},
scale only axis,
point meta min=0,
point meta max=0.35,
axis on top,
xmin=0,
xmax=260,
xtick={\empty},
y dir=reverse,
ymin=0,
ymax=260,
ytick={\empty},
axis background/.style={fill=white}
]
\addplot [forget plot] graphics [xmin=0.5, xmax=260.5, ymin=0.5, ymax=260.5] {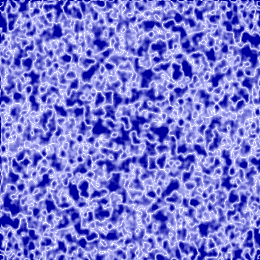};
\node[fill=white, align=center]
at (axis cs:35,35) {\footnotesize 46};
\end{axis}

\begin{axis}[%
width=0.629in,
height=0.629in,
at={(3.78in,0in)},
scale only axis,
point meta min=0,
point meta max=0.35,
axis on top,
xmin=0,
xmax=260,
xtick={\empty},
y dir=reverse,
ymin=0,
ymax=260,
ytick={\empty},
axis background/.style={fill=white}
]
\addplot [forget plot] graphics [xmin=0.5, xmax=260.5, ymin=0.5, ymax=260.5] {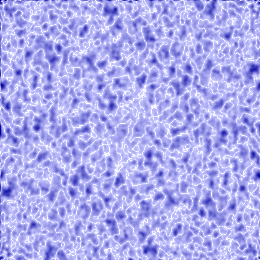};
\node[fill=white, align=center]
at (axis cs:35,35) {\footnotesize 47};
\end{axis}

\begin{axis}[%
width=0.629in,
height=0.629in,
at={(4.41in,0in)},
scale only axis,
point meta min=0,
point meta max=0.35,
axis on top,
xmin=0,
xmax=260,
xtick={\empty},
y dir=reverse,
ymin=0,
ymax=260,
ytick={\empty},
axis background/.style={fill=white}
]
\addplot [forget plot] graphics [xmin=0.5, xmax=260.5, ymin=0.5, ymax=260.5] {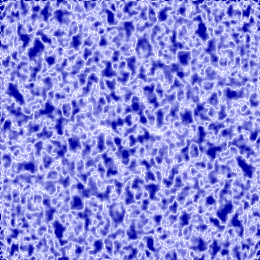};
\node[fill=white, align=center]
at (axis cs:35,35) {\footnotesize 48};
\end{axis}

\begin{axis}[%
width=0.629in,
height=0.629in,
at={(5.04in,0in)},
scale only axis,
point meta min=0,
point meta max=0.35,
axis on top,
xmin=0,
xmax=260,
xtick={\empty},
y dir=reverse,
ymin=0,
ymax=260,
ytick={\empty},
axis background/.style={fill=white}
]
\addplot [forget plot] graphics [xmin=0.5, xmax=260.5, ymin=0.5, ymax=260.5] {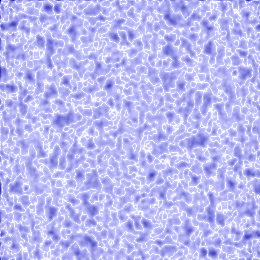};
\node[fill=white, align=center]
at (axis cs:35,35) {\footnotesize 49};
\end{axis}

\begin{axis}[%
width=0.629in,
height=0.629in,
at={(5.67in,0in)},
scale only axis,
point meta min=0,
point meta max=0.35,
axis on top,
xmin=0,
xmax=260,
xtick={\empty},
y dir=reverse,
ymin=0,
ymax=260,
ytick={\empty},
axis background/.style={fill=white},
colormap={mymap}{[1pt] rgb(0pt)=(1,1,1); rgb(11pt)=(0.780392,0.780392,1); rgb(255pt)=(0,0,0.611765)},
colorbar horizontal,
colorbar style={
at={(1,-0.2)},anchor=south east,
width=10*
\pgfkeysvalueof{/pgfplots/parent axis width},
xticklabel pos=lower,
colorbar/width=2mm,
xtick={
0,
0.1,
0.2,
0.3
},
},
]
\addplot [forget plot] graphics [xmin=0.5, xmax=260.5, ymin=0.5, ymax=260.5] {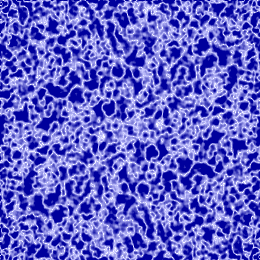};
\node[fill=white, align=center]
at (axis cs:35,35) {\footnotesize 50};
\end{axis}
\end{tikzpicture}%

%% file: Figures/Modes_k1000/rSVDmodes1000_Uhat.tex
\begin{tikzpicture}

\begin{axis}[%
width=0.629in,
height=0.629in,
at={(0in,2.52in)},
scale only axis,
point meta min=0,
point meta max=0.35,
axis on top,
xmin=0,
xmax=260,
xtick={\empty},
y dir=reverse,
ymin=0,
ymax=260,
ytick={\empty},
axis background/.style={fill=white}
]
\addplot [forget plot] graphics [xmin=0.5, xmax=260.5, ymin=0.5, ymax=260.5] {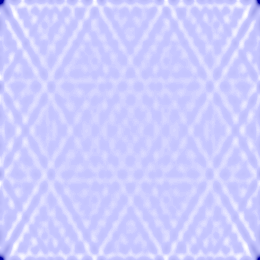};
\node[fill=white, align=center]
at (axis cs:35,35) {\footnotesize 01};
\end{axis}

\begin{axis}[%
width=0.629in,
height=0.629in,
at={(0.63in,2.52in)},
scale only axis,
point meta min=0,
point meta max=0.35,
axis on top,
xmin=0,
xmax=260,
xtick={\empty},
y dir=reverse,
ymin=0,
ymax=260,
ytick={\empty},
axis background/.style={fill=white}
]
\addplot [forget plot] graphics [xmin=0.5, xmax=260.5, ymin=0.5, ymax=260.5] {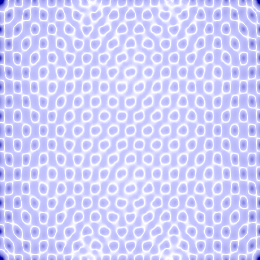};
\node[fill=white, align=center]
at (axis cs:35,35) {\footnotesize 02};
\end{axis}

\begin{axis}[%
width=0.629in,
height=0.629in,
at={(1.26in,2.52in)},
scale only axis,
point meta min=0,
point meta max=0.35,
axis on top,
xmin=0,
xmax=260,
xtick={\empty},
y dir=reverse,
ymin=0,
ymax=260,
ytick={\empty},
axis background/.style={fill=white}
]
\addplot [forget plot] graphics [xmin=0.5, xmax=260.5, ymin=0.5, ymax=260.5] {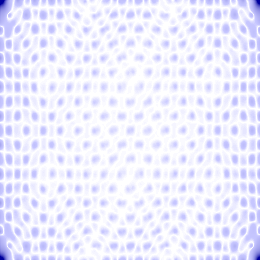};
\node[fill=white, align=center]
at (axis cs:35,35) {\footnotesize 03};
\end{axis}

\begin{axis}[%
width=0.629in,
height=0.629in,
at={(1.89in,2.52in)},
scale only axis,
point meta min=0,
point meta max=0.35,
axis on top,
xmin=0,
xmax=260,
xtick={\empty},
y dir=reverse,
ymin=0,
ymax=260,
ytick={\empty},
axis background/.style={fill=white}
]
\addplot [forget plot] graphics [xmin=0.5, xmax=260.5, ymin=0.5, ymax=260.5] {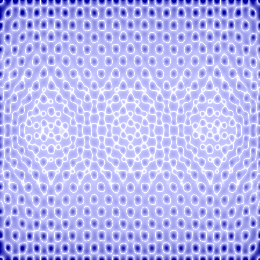};
\node[fill=white, align=center]
at (axis cs:35,35) {\footnotesize 04};
\end{axis}

\begin{axis}[%
width=0.629in,
height=0.629in,
at={(2.52in,2.52in)},
scale only axis,
point meta min=0,
point meta max=0.35,
axis on top,
xmin=0,
xmax=260,
xtick={\empty},
y dir=reverse,
ymin=0,
ymax=260,
ytick={\empty},
axis background/.style={fill=white}
]
\addplot [forget plot] graphics [xmin=0.5, xmax=260.5, ymin=0.5, ymax=260.5] {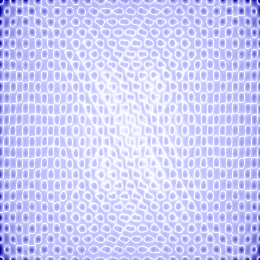};
\node[fill=white, align=center]
at (axis cs:35,35) {\footnotesize 05};
\end{axis}

\begin{axis}[%
width=0.629in,
height=0.629in,
at={(3.15in,2.52in)},
scale only axis,
point meta min=0,
point meta max=0.35,
axis on top,
xmin=0,
xmax=260,
xtick={\empty},
y dir=reverse,
ymin=0,
ymax=260,
ytick={\empty},
axis background/.style={fill=white}
]
\addplot [forget plot] graphics [xmin=0.5, xmax=260.5, ymin=0.5, ymax=260.5] {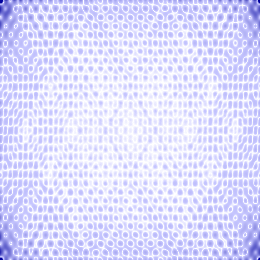};
\node[fill=white, align=center]
at (axis cs:35,35) {\footnotesize 06};
\end{axis}

\begin{axis}[%
width=0.629in,
height=0.629in,
at={(3.78in,2.52in)},
scale only axis,
point meta min=0,
point meta max=0.35,
axis on top,
xmin=0,
xmax=260,
xtick={\empty},
y dir=reverse,
ymin=0,
ymax=260,
ytick={\empty},
axis background/.style={fill=white}
]
\addplot [forget plot] graphics [xmin=0.5, xmax=260.5, ymin=0.5, ymax=260.5] {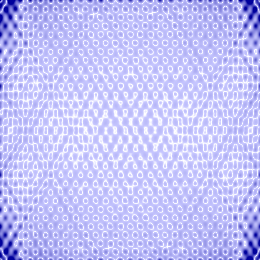};
\node[fill=white, align=center]
at (axis cs:35,35) {\footnotesize 07};
\end{axis}

\begin{axis}[%
width=0.629in,
height=0.629in,
at={(4.41in,2.52in)},
scale only axis,
point meta min=0,
point meta max=0.35,
axis on top,
xmin=0,
xmax=260,
xtick={\empty},
y dir=reverse,
ymin=0,
ymax=260,
ytick={\empty},
axis background/.style={fill=white}
]
\addplot [forget plot] graphics [xmin=0.5, xmax=260.5, ymin=0.5, ymax=260.5] {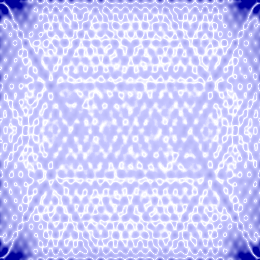};
\node[fill=white, align=center]
at (axis cs:35,35) {\footnotesize 08};
\end{axis}

\begin{axis}[%
width=0.629in,
height=0.629in,
at={(5.04in,2.52in)},
scale only axis,
point meta min=0,
point meta max=0.35,
axis on top,
xmin=0,
xmax=260,
xtick={\empty},
y dir=reverse,
ymin=0,
ymax=260,
ytick={\empty},
axis background/.style={fill=white}
]
\addplot [forget plot] graphics [xmin=0.5, xmax=260.5, ymin=0.5, ymax=260.5] {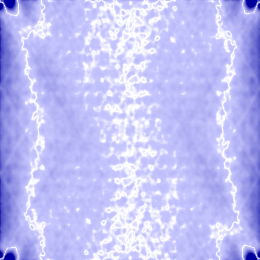};
\node[fill=white, align=center]
at (axis cs:35,35) {\footnotesize 09};
\end{axis}

\begin{axis}[%
width=0.629in,
height=0.629in,
at={(5.67in,2.52in)},
scale only axis,
point meta min=0,
point meta max=0.35,
axis on top,
xmin=0,
xmax=260,
xtick={\empty},
y dir=reverse,
ymin=0,
ymax=260,
ytick={\empty},
axis background/.style={fill=white}
]
\addplot [forget plot] graphics [xmin=0.5, xmax=260.5, ymin=0.5, ymax=260.5] {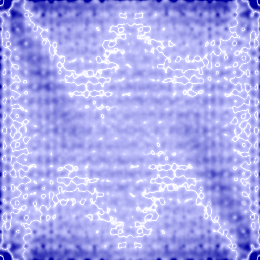};
\node[fill=white, align=center]
at (axis cs:35,35) {\footnotesize 10};
\end{axis}

\begin{axis}[%
width=0.629in,
height=0.629in,
at={(0in,1.89in)},
scale only axis,
point meta min=0,
point meta max=0.35,
axis on top,
xmin=0,
xmax=260,
xtick={\empty},
y dir=reverse,
ymin=0,
ymax=260,
ytick={\empty},
axis background/.style={fill=white}
]
\addplot [forget plot] graphics [xmin=0.5, xmax=260.5, ymin=0.5, ymax=260.5] {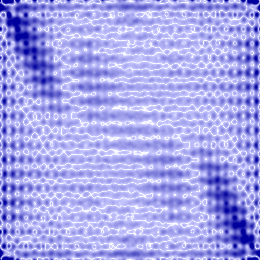};
\node[fill=white, align=center]
at (axis cs:35,35) {\footnotesize 11};
\end{axis}

\begin{axis}[%
width=0.629in,
height=0.629in,
at={(0.63in,1.89in)},
scale only axis,
point meta min=0,
point meta max=0.35,
axis on top,
xmin=0,
xmax=260,
xtick={\empty},
y dir=reverse,
ymin=0,
ymax=260,
ytick={\empty},
axis background/.style={fill=white}
]
\addplot [forget plot] graphics [xmin=0.5, xmax=260.5, ymin=0.5, ymax=260.5] {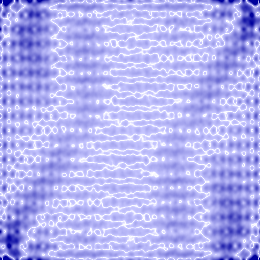};
\node[fill=white, align=center]
at (axis cs:35,35) {\footnotesize 12};
\end{axis}

\begin{axis}[%
width=0.629in,
height=0.629in,
at={(1.26in,1.89in)},
scale only axis,
point meta min=0,
point meta max=0.35,
axis on top,
xmin=0,
xmax=260,
xtick={\empty},
y dir=reverse,
ymin=0,
ymax=260,
ytick={\empty},
axis background/.style={fill=white}
]
\addplot [forget plot] graphics [xmin=0.5, xmax=260.5, ymin=0.5, ymax=260.5] {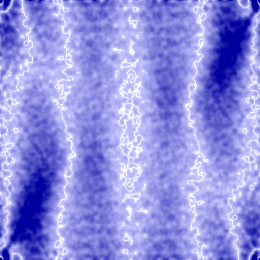};
\node[fill=white, align=center]
at (axis cs:35,35) {\footnotesize 13};
\end{axis}

\begin{axis}[%
width=0.629in,
height=0.629in,
at={(1.89in,1.89in)},
scale only axis,
point meta min=0,
point meta max=0.35,
axis on top,
xmin=0,
xmax=260,
xtick={\empty},
y dir=reverse,
ymin=0,
ymax=260,
ytick={\empty},
axis background/.style={fill=white}
]
\addplot [forget plot] graphics [xmin=0.5, xmax=260.5, ymin=0.5, ymax=260.5] {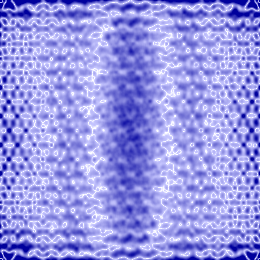};
\node[fill=white, align=center]
at (axis cs:35,35) {\footnotesize 14};
\end{axis}

\begin{axis}[%
width=0.629in,
height=0.629in,
at={(2.52in,1.89in)},
scale only axis,
point meta min=0,
point meta max=0.35,
axis on top,
xmin=0,
xmax=260,
xtick={\empty},
y dir=reverse,
ymin=0,
ymax=260,
ytick={\empty},
axis background/.style={fill=white}
]
\addplot [forget plot] graphics [xmin=0.5, xmax=260.5, ymin=0.5, ymax=260.5] {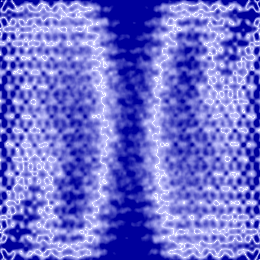};
\node[fill=white, align=center]
at (axis cs:35,35) {\footnotesize 15};
\end{axis}

\begin{axis}[%
width=0.629in,
height=0.629in,
at={(3.15in,1.89in)},
scale only axis,
point meta min=0,
point meta max=0.35,
axis on top,
xmin=0,
xmax=260,
xtick={\empty},
y dir=reverse,
ymin=0,
ymax=260,
ytick={\empty},
axis background/.style={fill=white}
]
\addplot [forget plot] graphics [xmin=0.5, xmax=260.5, ymin=0.5, ymax=260.5] {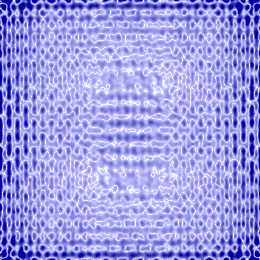};
\node[fill=white, align=center]
at (axis cs:35,35) {\footnotesize 16};
\end{axis}

\begin{axis}[%
width=0.629in,
height=0.629in,
at={(3.78in,1.89in)},
scale only axis,
point meta min=0,
point meta max=0.35,
axis on top,
xmin=0,
xmax=260,
xtick={\empty},
y dir=reverse,
ymin=0,
ymax=260,
ytick={\empty},
axis background/.style={fill=white}
]
\addplot [forget plot] graphics [xmin=0.5, xmax=260.5, ymin=0.5, ymax=260.5] {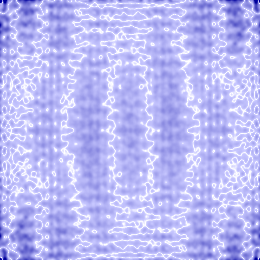};
\node[fill=white, align=center]
at (axis cs:35,35) {\footnotesize 17};
\end{axis}

\begin{axis}[%
width=0.629in,
height=0.629in,
at={(4.41in,1.89in)},
scale only axis,
point meta min=0,
point meta max=0.35,
axis on top,
xmin=0,
xmax=260,
xtick={\empty},
y dir=reverse,
ymin=0,
ymax=260,
ytick={\empty},
axis background/.style={fill=white}
]
\addplot [forget plot] graphics [xmin=0.5, xmax=260.5, ymin=0.5, ymax=260.5] {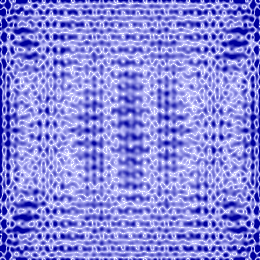};
\node[fill=white, align=center]
at (axis cs:35,35) {\footnotesize 18};
\end{axis}

\begin{axis}[%
width=0.629in,
height=0.629in,
at={(5.04in,1.89in)},
scale only axis,
point meta min=0,
point meta max=0.35,
axis on top,
xmin=0,
xmax=260,
xtick={\empty},
y dir=reverse,
ymin=0,
ymax=260,
ytick={\empty},
axis background/.style={fill=white}
]
\addplot [forget plot] graphics [xmin=0.5, xmax=260.5, ymin=0.5, ymax=260.5] {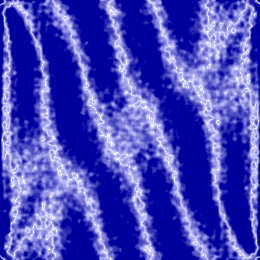};
\node[fill=white, align=center]
at (axis cs:35,35) {\footnotesize 19};
\end{axis}

\begin{axis}[%
width=0.629in,
height=0.629in,
at={(5.67in,1.89in)},
scale only axis,
point meta min=0,
point meta max=0.35,
axis on top,
xmin=0,
xmax=260,
xtick={\empty},
y dir=reverse,
ymin=0,
ymax=260,
ytick={\empty},
axis background/.style={fill=white}
]
\addplot [forget plot] graphics [xmin=0.5, xmax=260.5, ymin=0.5, ymax=260.5] {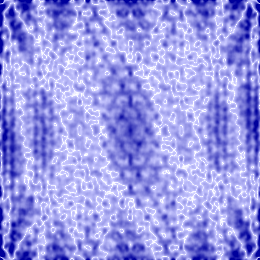};
\node[fill=white, align=center]
at (axis cs:35,35) {\footnotesize 20};
\end{axis}

\begin{axis}[%
width=0.629in,
height=0.629in,
at={(0in,1.26in)},
scale only axis,
point meta min=0,
point meta max=0.35,
axis on top,
xmin=0,
xmax=260,
xtick={\empty},
y dir=reverse,
ymin=0,
ymax=260,
ytick={\empty},
axis background/.style={fill=white}
]
\addplot [forget plot] graphics [xmin=0.5, xmax=260.5, ymin=0.5, ymax=260.5] {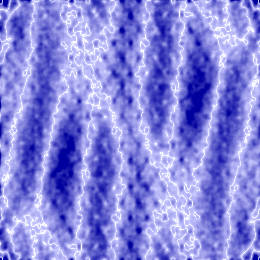};
\node[fill=white, align=center]
at (axis cs:35,35) {\footnotesize 21};
\end{axis}

\begin{axis}[%
width=0.629in,
height=0.629in,
at={(0.63in,1.26in)},
scale only axis,
point meta min=0,
point meta max=0.35,
axis on top,
xmin=0,
xmax=260,
xtick={\empty},
y dir=reverse,
ymin=0,
ymax=260,
ytick={\empty},
axis background/.style={fill=white}
]
\addplot [forget plot] graphics [xmin=0.5, xmax=260.5, ymin=0.5, ymax=260.5] {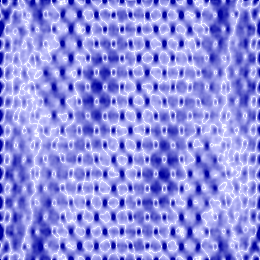};
\node[fill=white, align=center]
at (axis cs:35,35) {\footnotesize 22};
\end{axis}

\begin{axis}[%
width=0.629in,
height=0.629in,
at={(1.26in,1.26in)},
scale only axis,
point meta min=0,
point meta max=0.35,
axis on top,
xmin=0,
xmax=260,
xtick={\empty},
y dir=reverse,
ymin=0,
ymax=260,
ytick={\empty},
axis background/.style={fill=white}
]
\addplot [forget plot] graphics [xmin=0.5, xmax=260.5, ymin=0.5, ymax=260.5] {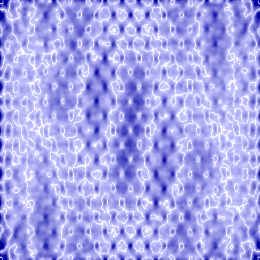};
\node[fill=white, align=center]
at (axis cs:35,35) {\footnotesize 23};
\end{axis}

\begin{axis}[%
width=0.629in,
height=0.629in,
at={(1.89in,1.26in)},
scale only axis,
point meta min=0,
point meta max=0.35,
axis on top,
xmin=0,
xmax=260,
xtick={\empty},
y dir=reverse,
ymin=0,
ymax=260,
ytick={\empty},
axis background/.style={fill=white}
]
\addplot [forget plot] graphics [xmin=0.5, xmax=260.5, ymin=0.5, ymax=260.5] {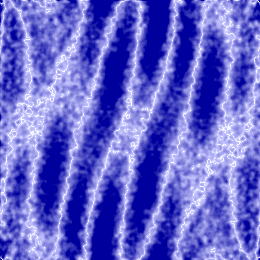};
\node[fill=white, align=center]
at (axis cs:35,35) {\footnotesize 24};
\end{axis}

\begin{axis}[%
width=0.629in,
height=0.629in,
at={(2.52in,1.26in)},
scale only axis,
point meta min=0,
point meta max=0.35,
axis on top,
xmin=0,
xmax=260,
xtick={\empty},
y dir=reverse,
ymin=0,
ymax=260,
ytick={\empty},
axis background/.style={fill=white}
]
\addplot [forget plot] graphics [xmin=0.5, xmax=260.5, ymin=0.5, ymax=260.5] {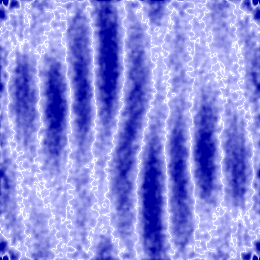};
\node[fill=white, align=center]
at (axis cs:35,35) {\footnotesize 25};
\end{axis}

\begin{axis}[%
width=0.629in,
height=0.629in,
at={(3.15in,1.26in)},
scale only axis,
point meta min=0,
point meta max=0.35,
axis on top,
xmin=0,
xmax=260,
xtick={\empty},
y dir=reverse,
ymin=0,
ymax=260,
ytick={\empty},
axis background/.style={fill=white}
]
\addplot [forget plot] graphics [xmin=0.5, xmax=260.5, ymin=0.5, ymax=260.5] {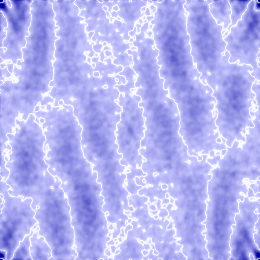};
\node[fill=white, align=center]
at (axis cs:35,35) {\footnotesize 26};
\end{axis}

\begin{axis}[%
width=0.629in,
height=0.629in,
at={(3.78in,1.26in)},
scale only axis,
point meta min=0,
point meta max=0.35,
axis on top,
xmin=0,
xmax=260,
xtick={\empty},
y dir=reverse,
ymin=0,
ymax=260,
ytick={\empty},
axis background/.style={fill=white}
]
\addplot [forget plot] graphics [xmin=0.5, xmax=260.5, ymin=0.5, ymax=260.5] {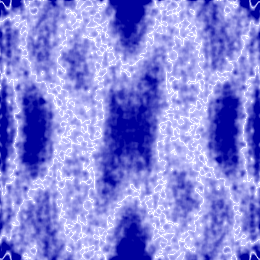};
\node[fill=white, align=center]
at (axis cs:35,35) {\footnotesize 27};
\end{axis}

\begin{axis}[%
width=0.629in,
height=0.629in,
at={(4.41in,1.26in)},
scale only axis,
point meta min=0,
point meta max=0.35,
axis on top,
xmin=0,
xmax=260,
xtick={\empty},
y dir=reverse,
ymin=0,
ymax=260,
ytick={\empty},
axis background/.style={fill=white}
]
\addplot [forget plot] graphics [xmin=0.5, xmax=260.5, ymin=0.5, ymax=260.5] {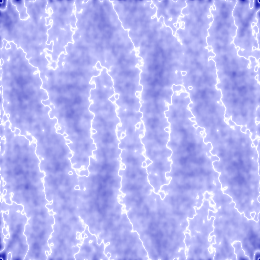};
\node[fill=white, align=center]
at (axis cs:35,35) {\footnotesize 28};
\end{axis}

\begin{axis}[%
width=0.629in,
height=0.629in,
at={(5.04in,1.26in)},
scale only axis,
point meta min=0,
point meta max=0.35,
axis on top,
xmin=0,
xmax=260,
xtick={\empty},
y dir=reverse,
ymin=0,
ymax=260,
ytick={\empty},
axis background/.style={fill=white}
]
\addplot [forget plot] graphics [xmin=0.5, xmax=260.5, ymin=0.5, ymax=260.5] {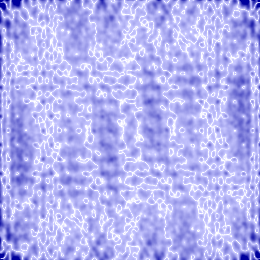};
\node[fill=white, align=center]
at (axis cs:35,35) {\footnotesize 29};
\end{axis}

\begin{axis}[%
width=0.629in,
height=0.629in,
at={(5.67in,1.26in)},
scale only axis,
point meta min=0,
point meta max=0.35,
axis on top,
xmin=0,
xmax=260,
xtick={\empty},
y dir=reverse,
ymin=0,
ymax=260,
ytick={\empty},
axis background/.style={fill=white}
]
\addplot [forget plot] graphics [xmin=0.5, xmax=260.5, ymin=0.5, ymax=260.5] {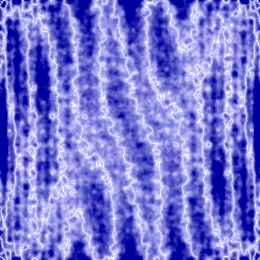};
\node[fill=white, align=center]
at (axis cs:35,35) {\footnotesize 30};
\end{axis}

\begin{axis}[%
width=0.629in,
height=0.629in,
at={(0in,0.63in)},
scale only axis,
point meta min=0,
point meta max=0.35,
axis on top,
xmin=0,
xmax=260,
xtick={\empty},
y dir=reverse,
ymin=0,
ymax=260,
ytick={\empty},
axis background/.style={fill=white}
]
\addplot [forget plot] graphics [xmin=0.5, xmax=260.5, ymin=0.5, ymax=260.5] {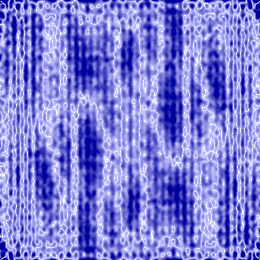};
\node[fill=white, align=center]
at (axis cs:35,35) {\footnotesize 31};
\end{axis}

\begin{axis}[%
width=0.629in,
height=0.629in,
at={(0.63in,0.63in)},
scale only axis,
point meta min=0,
point meta max=0.35,
axis on top,
xmin=0,
xmax=260,
xtick={\empty},
y dir=reverse,
ymin=0,
ymax=260,
ytick={\empty},
axis background/.style={fill=white}
]
\addplot [forget plot] graphics [xmin=0.5, xmax=260.5, ymin=0.5, ymax=260.5] {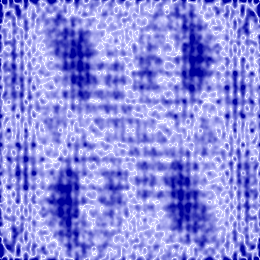};
\node[fill=white, align=center]
at (axis cs:35,35) {\footnotesize 32};
\end{axis}

\begin{axis}[%
width=0.629in,
height=0.629in,
at={(1.26in,0.63in)},
scale only axis,
point meta min=0,
point meta max=0.35,
axis on top,
xmin=0,
xmax=260,
xtick={\empty},
y dir=reverse,
ymin=0,
ymax=260,
ytick={\empty},
axis background/.style={fill=white}
]
\addplot [forget plot] graphics [xmin=0.5, xmax=260.5, ymin=0.5, ymax=260.5] {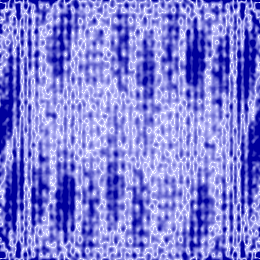};
\node[fill=white, align=center]
at (axis cs:35,35) {\footnotesize 33};
\end{axis}

\begin{axis}[%
width=0.629in,
height=0.629in,
at={(1.89in,0.63in)},
scale only axis,
point meta min=0,
point meta max=0.35,
axis on top,
xmin=0,
xmax=260,
xtick={\empty},
y dir=reverse,
ymin=0,
ymax=260,
ytick={\empty},
axis background/.style={fill=white}
]
\addplot [forget plot] graphics [xmin=0.5, xmax=260.5, ymin=0.5, ymax=260.5] {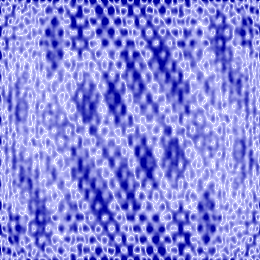};
\node[fill=white, align=center]
at (axis cs:35,35) {\footnotesize 34};
\end{axis}

\begin{axis}[%
width=0.629in,
height=0.629in,
at={(2.52in,0.63in)},
scale only axis,
point meta min=0,
point meta max=0.35,
axis on top,
xmin=0,
xmax=260,
xtick={\empty},
y dir=reverse,
ymin=0,
ymax=260,
ytick={\empty},
axis background/.style={fill=white}
]
\addplot [forget plot] graphics [xmin=0.5, xmax=260.5, ymin=0.5, ymax=260.5] {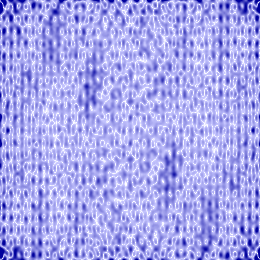};
\node[fill=white, align=center]
at (axis cs:35,35) {\footnotesize 35};
\end{axis}

\begin{axis}[%
width=0.629in,
height=0.629in,
at={(3.15in,0.63in)},
scale only axis,
point meta min=0,
point meta max=0.35,
axis on top,
xmin=0,
xmax=260,
xtick={\empty},
y dir=reverse,
ymin=0,
ymax=260,
ytick={\empty},
axis background/.style={fill=white}
]
\addplot [forget plot] graphics [xmin=0.5, xmax=260.5, ymin=0.5, ymax=260.5] {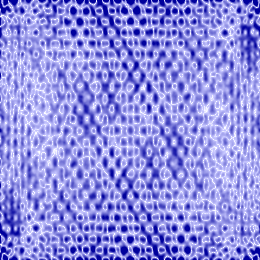};
\node[fill=white, align=center]
at (axis cs:35,35) {\footnotesize 36};
\end{axis}

\begin{axis}[%
width=0.629in,
height=0.629in,
at={(3.78in,0.63in)},
scale only axis,
point meta min=0,
point meta max=0.35,
axis on top,
xmin=0,
xmax=260,
xtick={\empty},
y dir=reverse,
ymin=0,
ymax=260,
ytick={\empty},
axis background/.style={fill=white}
]
\addplot [forget plot] graphics [xmin=0.5, xmax=260.5, ymin=0.5, ymax=260.5] {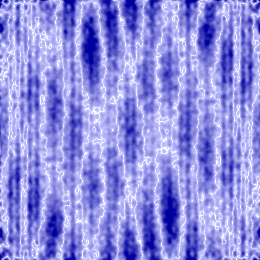};
\node[fill=white, align=center]
at (axis cs:35,35) {\footnotesize 37};
\end{axis}

\begin{axis}[%
width=0.629in,
height=0.629in,
at={(4.41in,0.63in)},
scale only axis,
point meta min=0,
point meta max=0.35,
axis on top,
xmin=0,
xmax=260,
xtick={\empty},
y dir=reverse,
ymin=0,
ymax=260,
ytick={\empty},
axis background/.style={fill=white}
]
\addplot [forget plot] graphics [xmin=0.5, xmax=260.5, ymin=0.5, ymax=260.5] {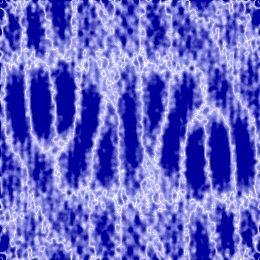};
\node[fill=white, align=center]
at (axis cs:35,35) {\footnotesize 38};
\end{axis}

\begin{axis}[%
width=0.629in,
height=0.629in,
at={(5.04in,0.63in)},
scale only axis,
point meta min=0,
point meta max=0.35,
axis on top,
xmin=0,
xmax=260,
xtick={\empty},
y dir=reverse,
ymin=0,
ymax=260,
ytick={\empty},
axis background/.style={fill=white}
]
\addplot [forget plot] graphics [xmin=0.5, xmax=260.5, ymin=0.5, ymax=260.5] {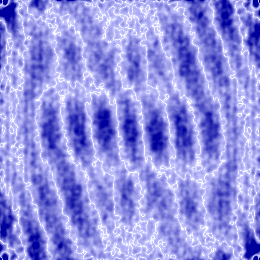};
\node[fill=white, align=center]
at (axis cs:35,35) {\footnotesize 39};
\end{axis}

\begin{axis}[%
width=0.629in,
height=0.629in,
at={(5.67in,0.63in)},
scale only axis,
point meta min=0,
point meta max=0.35,
axis on top,
xmin=0,
xmax=260,
xtick={\empty},
y dir=reverse,
ymin=0,
ymax=260,
ytick={\empty},
axis background/.style={fill=white}
]
\addplot [forget plot] graphics [xmin=0.5, xmax=260.5, ymin=0.5, ymax=260.5] {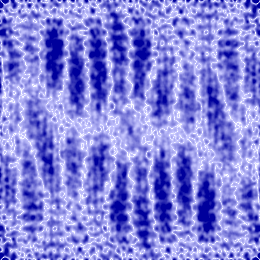};
\node[fill=white, align=center]
at (axis cs:35,35) {\footnotesize 40};
\end{axis}

\begin{axis}[%
width=0.629in,
height=0.629in,
at={(0in,0in)},
scale only axis,
point meta min=0,
point meta max=0.35,
axis on top,
xmin=0,
xmax=260,
xtick={\empty},
y dir=reverse,
ymin=0,
ymax=260,
ytick={\empty},
axis background/.style={fill=white}
]
\addplot [forget plot] graphics [xmin=0.5, xmax=260.5, ymin=0.5, ymax=260.5] {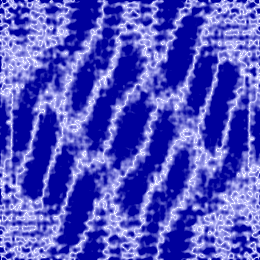};
\node[fill=white, align=center]
at (axis cs:35,35) {\footnotesize 41};
\end{axis}

\begin{axis}[%
width=0.629in,
height=0.629in,
at={(0.63in,0in)},
scale only axis,
point meta min=0,
point meta max=0.35,
axis on top,
xmin=0,
xmax=260,
xtick={\empty},
y dir=reverse,
ymin=0,
ymax=260,
ytick={\empty},
axis background/.style={fill=white}
]
\addplot [forget plot] graphics [xmin=0.5, xmax=260.5, ymin=0.5, ymax=260.5] {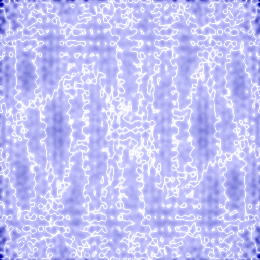};
\node[fill=white, align=center]
at (axis cs:35,35) {\footnotesize 42};
\end{axis}

\begin{axis}[%
width=0.629in,
height=0.629in,
at={(1.26in,0in)},
scale only axis,
point meta min=0,
point meta max=0.35,
axis on top,
xmin=0,
xmax=260,
xtick={\empty},
y dir=reverse,
ymin=0,
ymax=260,
ytick={\empty},
axis background/.style={fill=white}
]
\addplot [forget plot] graphics [xmin=0.5, xmax=260.5, ymin=0.5, ymax=260.5] {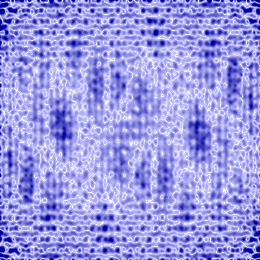};
\node[fill=white, align=center]
at (axis cs:35,35) {\footnotesize 43};
\end{axis}

\begin{axis}[%
width=0.629in,
height=0.629in,
at={(1.89in,0in)},
scale only axis,
point meta min=0,
point meta max=0.35,
axis on top,
xmin=0,
xmax=260,
xtick={\empty},
y dir=reverse,
ymin=0,
ymax=260,
ytick={\empty},
axis background/.style={fill=white}
]
\addplot [forget plot] graphics [xmin=0.5, xmax=260.5, ymin=0.5, ymax=260.5] {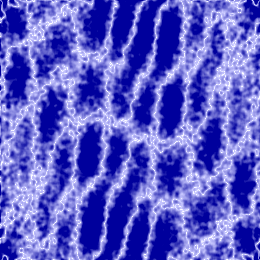};
\node[fill=white, align=center]
at (axis cs:35,35) {\footnotesize 44};
\end{axis}

\begin{axis}[%
width=0.629in,
height=0.629in,
at={(2.52in,0in)},
scale only axis,
point meta min=0,
point meta max=0.35,
axis on top,
xmin=0,
xmax=260,
xtick={\empty},
y dir=reverse,
ymin=0,
ymax=260,
ytick={\empty},
axis background/.style={fill=white}
]
\addplot [forget plot] graphics [xmin=0.5, xmax=260.5, ymin=0.5, ymax=260.5] {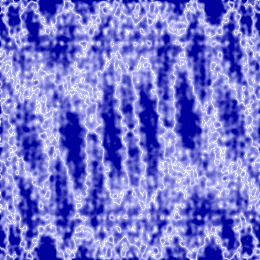};
\node[fill=white, align=center]
at (axis cs:35,35) {\footnotesize 45};
\end{axis}

\begin{axis}[%
width=0.629in,
height=0.629in,
at={(3.15in,0in)},
scale only axis,
point meta min=0,
point meta max=0.35,
axis on top,
xmin=0,
xmax=260,
xtick={\empty},
y dir=reverse,
ymin=0,
ymax=260,
ytick={\empty},
axis background/.style={fill=white}
]
\addplot [forget plot] graphics [xmin=0.5, xmax=260.5, ymin=0.5, ymax=260.5] {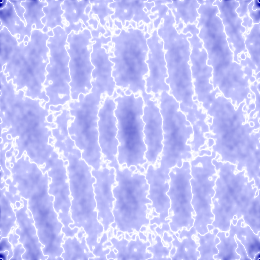};
\node[fill=white, align=center]
at (axis cs:35,35) {\footnotesize 46};
\end{axis}

\begin{axis}[%
width=0.629in,
height=0.629in,
at={(3.78in,0in)},
scale only axis,
point meta min=0,
point meta max=0.35,
axis on top,
xmin=0,
xmax=260,
xtick={\empty},
y dir=reverse,
ymin=0,
ymax=260,
ytick={\empty},
axis background/.style={fill=white}
]
\addplot [forget plot] graphics [xmin=0.5, xmax=260.5, ymin=0.5, ymax=260.5] {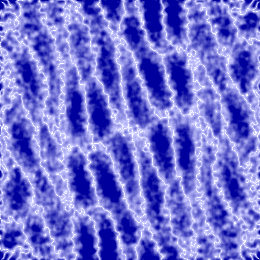};
\node[fill=white, align=center]
at (axis cs:35,35) {\footnotesize 47};
\end{axis}

\begin{axis}[%
width=0.629in,
height=0.629in,
at={(4.41in,0in)},
scale only axis,
point meta min=0,
point meta max=0.35,
axis on top,
xmin=0,
xmax=260,
xtick={\empty},
y dir=reverse,
ymin=0,
ymax=260,
ytick={\empty},
axis background/.style={fill=white}
]
\addplot [forget plot] graphics [xmin=0.5, xmax=260.5, ymin=0.5, ymax=260.5] {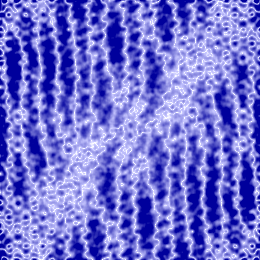};
\node[fill=white, align=center]
at (axis cs:35,35) {\footnotesize 48};
\end{axis}

\begin{axis}[%
width=0.629in,
height=0.629in,
at={(5.04in,0in)},
scale only axis,
point meta min=0,
point meta max=0.35,
axis on top,
xmin=0,
xmax=260,
xtick={\empty},
y dir=reverse,
ymin=0,
ymax=260,
ytick={\empty},
axis background/.style={fill=white}
]
\addplot [forget plot] graphics [xmin=0.5, xmax=260.5, ymin=0.5, ymax=260.5] {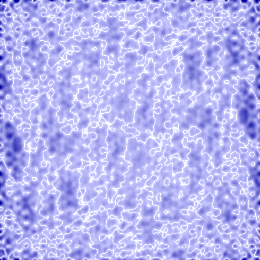};
\node[fill=white, align=center]
at (axis cs:35,35) {\footnotesize 49};
\end{axis}

\begin{axis}[%
width=0.629in,
height=0.629in,
at={(5.67in,0in)},
scale only axis,
point meta min=0,
point meta max=0.35,
axis on top,
xmin=0,
xmax=260,
xtick={\empty},
y dir=reverse,
ymin=0,
ymax=260,
ytick={\empty},
axis background/.style={fill=white},
colormap={mymap}{[1pt] rgb(0pt)=(1,1,1); rgb(11pt)=(0.780392,0.780392,1); rgb(255pt)=(0,0,0.611765)},
colorbar horizontal,
colorbar style={
at={(1,-0.2)},anchor=south east,
width=10*
\pgfkeysvalueof{/pgfplots/parent axis width},
xticklabel pos=lower,
colorbar/width=2mm,
xtick={
0,
0.1,
0.2,
0.3
},
},
]
\addplot [forget plot] graphics [xmin=0.5, xmax=260.5, ymin=0.5, ymax=260.5] {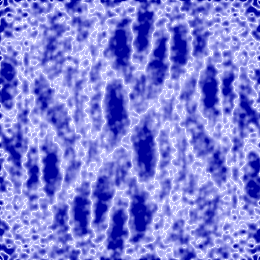};
\node[fill=white, align=center]
at (axis cs:35,35) {\footnotesize 50};
\end{axis}
\end{tikzpicture}%

%% file: Figures/balancing0_norm0_FFTtransformed_50rSVDmodes_recall_dots_class.tex
\definecolor{mycolor1}{rgb}{0.64, 0.76, 0.68}%
\begin{tikzpicture}

\begin{axis}[%
width=7cm,
height=5cm,
scale only axis,
unbounded coords=jump,
xmin=0,
xmax=50,
xlabel style={font=\color{white!15!black}},
xlabel={$r$},
ymin=0,
ymax=100,
ylabel style={font=\color{white!15!black}},
ylabel={Recall A of dots class in \%},
axis background/.style={fill=white},
title style={font=\bfseries},
legend columns = 2,
legend pos = south east,
legend style={legend cell align=left, align=left, draw=white!15!black, font=\footnotesize},
legend entries={Tree,
                NB,
                LD,
                kNN,
                98.3~\%},
]
\addlegendimage{legend image code/.code={
\draw [draw=none, fill=white!80!black] (0cm,-0.15cm) rectangle (0.6cm,0.15cm);
\draw[dashdotted] (0cm,0cm) -- (0.6cm,0cm);
\draw[-] (0cm,0.15cm) -- (0.6cm,0.15cm);
\draw[-] (0cm,-0.15cm) -- (0.6cm,-0.15cm);
}
} 
\addlegendimage{legend image code/.code={
\draw [draw=none, fill=mycolor1] (0cm,-0.15cm) rectangle (0.6cm,0.15cm);
\draw[dotted, black!80!green] (0cm,0cm) -- (0.6cm,0cm);
\draw[-, black!80!green] (0cm,0.15cm) -- (0.6cm,0.15cm);
\draw[-, black!80!green] (0cm,-0.15cm) -- (0.6cm,-0.15cm);
}
}
\addlegendimage{legend image code/.code={
\draw [draw=none, fill=white!80!blue] (0cm,-0.15cm) rectangle (0.6cm,0.15cm);
\draw[dashed, blue] (0cm,0cm) -- (0.6cm,0cm);
\draw[-, blue] (0cm,0.15cm) -- (0.6cm,0.15cm);
\draw[-, blue] (0cm,-0.15cm) -- (0.6cm,-0.15cm);
}
}
\addlegendimage{legend image code/.code={
\draw [draw=none, fill=white!80!red] (0cm,-0.15cm) rectangle (0.6cm,0.15cm);
\draw[-, red] (0cm,0cm) -- (0.6cm,0cm);
\draw[-, red] (0cm,0.15cm) -- (0.6cm,0.15cm);
\draw[-, red] (0cm,-0.15cm) -- (0.6cm,-0.15cm);
}
}
\addlegendimage{mark=none, black, very thick, dotted}

\addplot[area legend, draw=black, fill=white!80!black, dashdotted]
table[row sep=crcr] {%
x	y\\
1	70.210902262794\\
2	84.2088783029252\\
3	90.6780028227622\\
4	92.6912411667143\\
5	94.4957786044265\\
6	96.0177281013134\\
7	95.7393429256889\\
8	95.5111552925381\\
9	95.1790576556566\\
10	94.9638075339863\\
11	94.7653659769567\\
12	94.9286468009622\\
13	95.0288792925194\\
14	95.198455277794\\
15	94.9576501567527\\
16	94.8980095710758\\
17	94.8764631698\\
18	94.7573731329925\\
19	94.8837016262933\\
20	94.7842823494134\\
21	94.9131292535815\\
22	94.7985208179909\\
23	94.8078991800562\\
24	94.556296138469\\
25	94.5259218267462\\
26	94.3718466889383\\
27	94.3002018124022\\
28	94.2552961350501\\
29	94.1734935298058\\
30	94.0591304523422\\
31	94.1565856692322\\
32	94.1427087549708\\
33	94.1966798175035\\
34	94.1625005785276\\
35	94.1523463513104\\
36	94.1161745229541\\
37	93.968256937268\\
38	93.9823228285834\\
39	94.04202545479\\
40	94.1460696445894\\
41	94.0884201112225\\
42	94.0693476658185\\
43	94.0713070298993\\
44	94.1344907420443\\
45	94.1419398873229\\
46	94.2347020344152\\
47	94.3101075856103\\
48	94.3378869587277\\
49	94.1985781149804\\
50	94.2369824467604\\
50	95.4222894375515\\
49	95.4607051775117\\
48	95.5350487310793\\
47	95.584172809204\\
46	95.5741310912893\\
45	95.5814459692717\\
44	95.6102740440979\\
43	95.5879876707729\\
42	95.6326820775887\\
41	95.7845269867646\\
40	95.8123703552281\\
39	95.8095997545501\\
38	95.8479348593892\\
37	95.8833340475317\\
36	95.8849777032382\\
35	95.8701962126097\\
34	95.7959394212899\\
33	95.8258627464165\\
32	95.8157426530268\\
31	95.8018885551257\\
30	95.707036079708\\
29	95.8062683586587\\
28	95.916762037542\\
27	95.8718335438296\\
26	95.9497727250463\\
25	96.0520736191083\\
24	95.7866907968831\\
23	95.7700734494381\\
22	95.8649447133336\\
21	95.9639430423375\\
20	96.0073654937561\\
19	96.0788749796362\\
18	96.0129300051697\\
17	96.0434012097547\\
16	96.1714274580515\\
15	96.3041287892227\\
14	96.4479276446167\\
13	96.6175606707919\\
12	96.5255082864017\\
11	96.6674672217601\\
10	96.6399316111303\\
9	96.7237925640649\\
8	96.5197061163265\\
7	96.2914728504552\\
6	96.7394921358855\\
5	95.291550101749\\
4	94.0409769029304\\
3	91.6529507641978\\
2	87.6107720591704\\
1	70.8424606405423\\
}--cycle;
\addlegendentry{Tree}

\addplot[area legend, draw=black!80!green, fill=mycolor1, dotted]
table[row sep=crcr] {%
x	y\\
1	83.9449157091527\\
2	89.0153190309405\\
3	84.1054844268445\\
4	86.6547577208374\\
5	90.2350701630918\\
6	89.7763402042124\\
7	90.5942067240482\\
8	86.2358631956905\\
9	87.5978949997213\\
10	87.6303278768445\\
11	86.4405210680531\\
12	86.3972247781844\\
13	84.9297685815651\\
14	85.0512689914739\\
15	85.5572980473007\\
16	84.9833418309981\\
17	85.3984263359301\\
18	85.7720665350425\\
19	85.7594161929551\\
20	85.9772953908942\\
21	86.6021299051841\\
22	86.6115123203925\\
23	86.6573246666548\\
24	86.8503537282002\\
25	87.0412610172848\\
26	87.0157150808153\\
27	87.1618849063777\\
28	87.215555788156\\
29	86.9814464908205\\
30	87.1223824367939\\
31	86.9765668915086\\
32	87.1714375078094\\
33	86.7441503869356\\
34	86.8114858822913\\
35	86.5002124002735\\
36	86.5397788446282\\
37	86.2617537247246\\
38	86.1007956408981\\
39	85.948465206717\\
40	85.7873674061234\\
41	85.451307099292\\
42	85.1994045344999\\
43	84.8998292510071\\
44	84.9780090901894\\
45	84.7839316594982\\
46	84.7825691476931\\
47	84.4037045129037\\
48	84.5054441635413\\
49	84.4359590958538\\
50	84.4612081754503\\
50	85.3283907585693\\
49	85.6955544045866\\
48	85.9252260442245\\
47	85.9415070175721\\
46	86.2249556879151\\
45	86.6081857443651\\
44	86.7773219523813\\
43	86.9837069197688\\
42	87.0473110504431\\
41	86.987647728878\\
40	87.2711200517234\\
39	87.4091447339149\\
38	87.7054409830107\\
37	87.6513775469224\\
36	87.7151871361786\\
35	88.0752664010462\\
34	88.4262491832601\\
33	88.4294364817927\\
32	88.6217504396765\\
31	88.7525869409554\\
30	88.8845035407274\\
29	88.8759352862089\\
28	88.7700083007182\\
27	88.6740951247438\\
26	89.0766866148966\\
25	89.2861948216499\\
24	89.3061048988938\\
23	89.0291169394345\\
22	88.88271285883\\
21	89.0630582613389\\
20	88.875938371843\\
19	88.8374301297322\\
18	89.0597426654266\\
17	88.6855652493962\\
16	88.4597496032839\\
15	87.5653946482697\\
14	88.7119816097372\\
13	88.7694479046242\\
12	88.5200545077548\\
11	89.1601310235227\\
10	89.5728997254755\\
9	90.1821301895404\\
8	88.7880825743845\\
7	92.2494418857509\\
6	91.3155253070338\\
5	91.4122596382689\\
4	87.5148180317603\\
3	85.5988725853075\\
2	90.8589285995805\\
1	86.7637033990926\\
}--cycle;
\addlegendentry{NB}

\addplot[area legend, draw=blue, fill=white!80!blue, dashed]
table[row sep=crcr] {%
x	y\\
1	84.4575596578727\\
2	87.3164195223037\\
3	83.1883652895836\\
4	85.2113550495485\\
5	89.9910424835366\\
6	88.4118708985516\\
7	94.2827398105184\\
8	93.456488031927\\
9	94.0620796045544\\
10	94.1000281312945\\
11	93.6679605499018\\
12	93.6209215320153\\
13	93.6456017660314\\
14	94.0474657484179\\
15	93.9520703489479\\
16	93.7174163254741\\
17	93.7208696626469\\
18	93.6274543719507\\
19	93.6385706848278\\
20	93.7585161914845\\
21	93.7702615679558\\
22	93.7399590914641\\
23	93.3982043014271\\
24	93.3328787907275\\
25	93.4317994022464\\
26	93.5581219141407\\
27	93.594987888356\\
28	93.4527123746279\\
29	93.5557539052176\\
30	93.76464801908\\
31	93.8367021802588\\
32	93.8808498842885\\
33	93.8675407262121\\
34	94.0473220172161\\
35	94.1076229877956\\
36	93.9294044428103\\
37	93.8236546558814\\
38	93.755786517492\\
39	93.7441547755374\\
40	93.8167929735676\\
41	93.8250417490555\\
42	93.8547619469042\\
43	93.6758963955854\\
44	93.6240811260952\\
45	93.5084827483943\\
46	93.3393158138247\\
47	93.4161332315065\\
48	93.6076287962186\\
49	93.5655438870897\\
50	93.589440189285\\
50	94.4674501233447\\
49	94.5127253550877\\
48	94.4920307836866\\
47	94.533953698826\\
46	94.8030788088155\\
45	94.7193591433558\\
44	94.6464501756694\\
43	94.6801049916494\\
42	94.5225955535179\\
41	94.5095921168117\\
40	94.6032881614094\\
39	94.5263651180471\\
38	94.6215367583898\\
37	94.5963808463752\\
36	94.5119871726337\\
35	94.4619737558535\\
34	94.4581835705105\\
33	94.4456799855671\\
32	94.6032767738906\\
31	94.6260341401924\\
30	94.3989886349464\\
29	94.3942873923945\\
28	94.4973289229842\\
27	94.3123297747011\\
26	94.1996801402444\\
25	94.2619343125764\\
24	94.2967523599929\\
23	94.4023670202334\\
22	94.4023556739153\\
21	94.3507313087765\\
20	94.4265792493504\\
19	94.5252028673601\\
18	94.3012650370143\\
17	94.2078497463181\\
16	94.531838720364\\
15	94.5321475746721\\
14	94.8640797861922\\
13	94.9455792544075\\
12	94.9061455161409\\
11	95.1582632055798\\
10	95.3884176638785\\
9	95.3195057674207\\
8	94.6003680561625\\
7	95.5901160220278\\
6	90.0949009971685\\
5	91.506760300972\\
4	86.3943348943688\\
3	84.8494961786035\\
2	88.434215913329\\
1	87.1056462232723\\
}--cycle;
\addlegendentry{LD}

\addplot[area legend, draw=red, fill=white!80!red]
table[row sep=crcr] {%
x	y\\
1	65.2476448267889\\
2	83.2568087002571\\
3	90.6933242062155\\
4	93.266478112385\\
5	95.6227692325661\\
6	96.8121692346041\\
7	96.9846288073201\\
8	97.17585459102\\
9	97.0412701387041\\
10	97.1118248483615\\
11	96.9593596369471\\
12	97.0694045922711\\
13	97.0738029293189\\
14	97.0665203133686\\
15	97.0446252021346\\
16	97.1916900378402\\
17	97.1964428200599\\
18	97.2339156284337\\
19	97.1620858180845\\
20	96.9849787864403\\
21	97.0023997219426\\
22	96.937523917841\\
23	97.0075386182372\\
24	97.0777121551162\\
25	96.938548344244\\
26	96.8346501578565\\
27	96.8410763668508\\
28	96.9253465211415\\
29	96.9282558292868\\
30	96.959181598032\\
31	96.9516726527297\\
32	96.9722809851186\\
33	97.0162429435578\\
34	97.0801330995015\\
35	97.1623708747108\\
36	97.1715957586542\\
37	97.1778006256378\\
38	97.207985112667\\
39	97.2171307596789\\
40	97.1203417222844\\
41	97.1394502087112\\
42	97.2188614487738\\
43	97.2067406831769\\
44	97.2176900064216\\
45	97.1836850545039\\
46	97.1462944740028\\
47	97.1174976453458\\
48	97.177263311637\\
49	97.1697325650438\\
50	97.1413157744245\\
50	97.9874711709668\\
49	98.0017666067223\\
48	97.9301332960266\\
47	97.9044402850278\\
46	97.9824696550283\\
45	97.9878141172622\\
44	97.9965442080796\\
43	98.007470714964\\
42	98.0167174707346\\
41	98.07478400579\\
40	98.0725363790293\\
39	98.0184595680096\\
38	98.0703402577566\\
37	98.1432483793407\\
36	98.1921882890594\\
35	98.2655385534656\\
34	98.0273091408826\\
33	97.9843616899886\\
32	97.9642324925054\\
31	97.9634847117069\\
30	98.0414230355144\\
29	98.050969874712\\
28	98.0966370419526\\
27	98.1168160403208\\
26	98.1018747279476\\
25	98.1261816697652\\
24	98.0724651280029\\
23	98.0999378466873\\
22	97.977576405695\\
21	97.8699883752185\\
20	97.8446742679858\\
19	97.7957381400064\\
18	97.8948370924172\\
17	97.9537002385188\\
16	97.8729715270883\\
15	97.8704865295815\\
14	98.0408648861148\\
13	97.9908472274295\\
12	97.9524762972019\\
11	98.0839115902536\\
10	98.0596286906842\\
9	98.1087816544336\\
8	98.102413738503\\
7	98.1441125053507\\
6	98.0815635675644\\
5	95.9807303407679\\
4	94.1495234573806\\
3	92.385515444968\\
2	86.4262720559684\\
1	67.0461295011552\\
}--cycle;
\addlegendentry{kNN}

\addplot[mark=none, black, very thick, dotted, domain=0:50] {98.26};
\addlegendentry{98.3~\%}

\addplot [color=black, forget plot, dashdotted]
  table[row sep=crcr]{%
1	70.5266814516681\\
2	85.9098251810478\\
3	91.16547679348\\
4	93.3661090348223\\
5	94.8936643530877\\
6	96.3786101185994\\
7	96.0154078880721\\
8	96.0154307044323\\
9	95.9514251098608\\
10	95.8018695725583\\
11	95.7164165993584\\
12	95.7270775436819\\
13	95.8232199816557\\
14	95.8231914612054\\
15	95.6308894729877\\
16	95.5347185145637\\
18	95.3851515690811\\
19	95.4812883029648\\
20	95.3958239215847\\
21	95.4385361479595\\
22	95.3317327656623\\
23	95.2889863147471\\
24	95.1714934676761\\
25	95.2889977229272\\
26	95.1608097069923\\
27	95.0860176781159\\
28	95.086029086296\\
29	94.9898809442323\\
30	94.8830832660251\\
31	94.9792371121789\\
32	94.9792257039988\\
33	95.01127128196\\
34	94.9792199999087\\
35	95.01127128196\\
36	95.0005761130961\\
37	94.9257954923999\\
38	94.9151288439863\\
39	94.92581260467\\
40	94.9792199999087\\
41	94.9364735489936\\
42	94.8510148717036\\
43	94.8296473503361\\
44	94.8723823930711\\
45	94.8616929282973\\
47	94.9471401974072\\
48	94.9364678449035\\
49	94.829641646246\\
50	94.829635942156\\
};
\addplot [color=black!80!green, forget plot, dotted]
  table[row sep=crcr]{%
1	85.3543095541227\\
2	89.9371238152605\\
3	84.852178506076\\
4	87.0847878762988\\
5	90.8236649006804\\
6	90.5459327556231\\
7	91.4218243048996\\
8	87.5119728850375\\
9	88.8900125946309\\
10	88.60161380116\\
11	87.8003260457879\\
12	87.4586396429696\\
13	86.8496082430946\\
14	86.8816253006055\\
15	86.5613463477852\\
16	86.721545717141\\
17	87.0419957926632\\
18	87.4159046002345\\
19	87.2984231613436\\
20	87.4266168813686\\
21	87.8325940832615\\
22	87.7471125896113\\
23	87.8432208030446\\
24	88.078229313547\\
25	88.1637279194674\\
26	88.0462008478559\\
27	87.9179900155607\\
28	87.9927820444371\\
29	87.9286908885147\\
30	88.0034429887607\\
31	87.864576916232\\
32	87.8965939737429\\
33	87.5867934343642\\
34	87.6188675327757\\
35	87.2877394006599\\
37	86.9565656358235\\
38	86.9031183119544\\
39	86.6788049703159\\
40	86.5292437289234\\
41	86.219477414085\\
42	86.1233577924715\\
43	85.9417680853879\\
44	85.8776655212854\\
46	85.5037624178041\\
47	85.1726057652379\\
48	85.2153351038829\\
49	85.0657567502202\\
50	84.8947994670098\\
};
\addplot [color=blue, forget plot, dashed]
  table[row sep=crcr]{%
1	85.7816029405725\\
2	87.8753177178164\\
3	84.0189307340936\\
4	85.8028449719587\\
5	90.7489013922543\\
6	89.2533859478601\\
7	94.9364279162731\\
8	94.0284280440447\\
9	94.6907926859875\\
10	94.7442228975865\\
11	94.4131118777408\\
12	94.2635335240781\\
13	94.2955905102195\\
14	94.4557727673051\\
15	94.24210896181\\
16	94.124627522919\\
17	93.9643597044825\\
18	93.9643597044825\\
19	94.0818867760939\\
20	94.0925477204175\\
21	94.0604964383662\\
22	94.0711573826897\\
23	93.9002856608302\\
24	93.8148155753602\\
26	93.8789010271925\\
27	93.9536588315286\\
28	93.975020648806\\
29	93.975020648806\\
30	94.0818183270132\\
31	94.2313681602256\\
32	94.2420633290895\\
33	94.1566103558896\\
34	94.2527527938633\\
35	94.2847983718245\\
36	94.220695807722\\
37	94.2100177511283\\
38	94.1886616379409\\
39	94.1352599467922\\
40	94.2100405674885\\
41	94.1673169329336\\
42	94.1886787502111\\
43	94.1780006936174\\
44	94.1352656508823\\
45	94.113920945875\\
46	94.0711973113201\\
47	93.9750434651663\\
48	94.0498297899526\\
50	94.0284451563149\\
};
\addplot [color=red, forget plot]
  table[row sep=crcr]{%
1	66.1468871639721\\
2	84.8415403781127\\
3	91.5394198255918\\
4	93.7080007848828\\
5	95.801749786667\\
6	97.4468664010842\\
7	97.5643706563354\\
8	97.6391341647615\\
9	97.5750258965689\\
10	97.5857267695228\\
11	97.5216356136004\\
12	97.5109404447365\\
14	97.5536925997417\\
15	97.457555865858\\
16	97.5323307824643\\
17	97.5750715292894\\
18	97.5643763604255\\
19	97.4789119790455\\
20	97.4148265272131\\
22	97.457550161768\\
23	97.5537382324622\\
24	97.5750886415595\\
25	97.5323650070046\\
26	97.4682624429021\\
27	97.4789462035858\\
28	97.510991781547\\
29	97.4896128519994\\
30	97.5003023167732\\
31	97.4575786822183\\
32	97.468256738812\\
33	97.5003023167732\\
34	97.553721120192\\
35	97.7139547140882\\
36	97.6818920238568\\
40	97.5964390506569\\
42	97.6177894597542\\
43	97.6071056990704\\
44	97.6071171072506\\
46	97.5643820645155\\
47	97.5109689651868\\
48	97.5536983038318\\
49	97.5857495858831\\
50	97.5643934726957\\
};
\addplot [color=black, forget plot]
  table[row sep=crcr]{%
1	70.210902262794\\
2	84.2088783029252\\
3	90.6780028227622\\
4	92.6912411667143\\
5	94.4957786044265\\
6	96.0177281013134\\
7	95.7393429256889\\
8	95.5111552925381\\
9	95.1790576556566\\
10	94.9638075339863\\
11	94.7653659769567\\
12	94.9286468009622\\
13	95.0288792925194\\
14	95.198455277794\\
15	94.9576501567527\\
16	94.8980095710758\\
17	94.8764631698\\
18	94.7573731329925\\
19	94.8837016262933\\
20	94.7842823494134\\
21	94.9131292535815\\
22	94.7985208179909\\
23	94.8078991800562\\
24	94.556296138469\\
25	94.5259218267462\\
26	94.3718466889383\\
27	94.3002018124022\\
28	94.2552961350501\\
29	94.1734935298058\\
30	94.0591304523422\\
31	94.1565856692322\\
32	94.1427087549708\\
33	94.1966798175035\\
34	94.1625005785276\\
35	94.1523463513104\\
36	94.1161745229541\\
37	93.968256937268\\
38	93.9823228285834\\
39	94.04202545479\\
40	94.1460696445894\\
41	94.0884201112225\\
42	94.0693476658185\\
43	94.0713070298993\\
44	94.1344907420443\\
45	94.1419398873229\\
46	94.2347020344152\\
47	94.3101075856103\\
48	94.3378869587277\\
49	94.1985781149804\\
50	94.2369824467604\\
nan	nan\\
50	95.4222894375515\\
49	95.4607051775117\\
48	95.5350487310793\\
47	95.584172809204\\
46	95.5741310912893\\
45	95.5814459692717\\
44	95.6102740440979\\
43	95.5879876707729\\
42	95.6326820775887\\
41	95.7845269867646\\
40	95.8123703552281\\
39	95.8095997545501\\
37	95.8833340475317\\
36	95.8849777032382\\
35	95.8701962126097\\
34	95.7959394212899\\
33	95.8258627464165\\
31	95.8018885551257\\
30	95.707036079708\\
29	95.8062683586587\\
28	95.916762037542\\
27	95.8718335438296\\
26	95.9497727250463\\
25	96.0520736191083\\
24	95.7866907968831\\
23	95.7700734494381\\
21	95.9639430423375\\
20	96.0073654937561\\
19	96.0788749796362\\
18	96.0129300051697\\
17	96.0434012097547\\
15	96.3041287892227\\
14	96.4479276446167\\
13	96.6175606707919\\
12	96.5255082864017\\
11	96.6674672217601\\
10	96.6399316111303\\
9	96.7237925640649\\
8	96.5197061163265\\
7	96.2914728504552\\
6	96.7394921358855\\
5	95.291550101749\\
4	94.0409769029304\\
3	91.6529507641978\\
2	87.6107720591704\\
1	70.8424606405423\\
};
\addplot [color=black!80!green, forget plot]
  table[row sep=crcr]{%
1	83.9449157091527\\
2	89.0153190309405\\
3	84.1054844268445\\
4	86.6547577208374\\
5	90.2350701630918\\
6	89.7763402042124\\
7	90.5942067240482\\
8	86.2358631956905\\
9	87.5978949997213\\
10	87.6303278768445\\
11	86.4405210680531\\
12	86.3972247781844\\
13	84.9297685815651\\
14	85.0512689914739\\
15	85.5572980473007\\
16	84.9833418309981\\
17	85.3984263359301\\
18	85.7720665350425\\
19	85.7594161929551\\
20	85.9772953908942\\
21	86.6021299051841\\
22	86.6115123203925\\
23	86.6573246666548\\
25	87.0412610172848\\
26	87.0157150808153\\
27	87.1618849063777\\
28	87.215555788156\\
29	86.9814464908205\\
30	87.1223824367939\\
31	86.9765668915086\\
32	87.1714375078094\\
33	86.7441503869356\\
34	86.8114858822913\\
35	86.5002124002735\\
36	86.5397788446282\\
37	86.2617537247246\\
40	85.7873674061234\\
41	85.451307099292\\
42	85.1994045344999\\
43	84.8998292510071\\
44	84.9780090901894\\
45	84.7839316594982\\
46	84.7825691476931\\
47	84.4037045129037\\
48	84.5054441635413\\
49	84.4359590958538\\
50	84.4612081754503\\
nan	nan\\
50	85.3283907585693\\
49	85.6955544045866\\
48	85.9252260442245\\
47	85.9415070175721\\
46	86.2249556879151\\
45	86.6081857443651\\
44	86.7773219523813\\
43	86.9837069197688\\
42	87.0473110504431\\
41	86.987647728878\\
40	87.2711200517234\\
39	87.4091447339149\\
38	87.7054409830107\\
37	87.6513775469224\\
36	87.7151871361786\\
34	88.4262491832601\\
33	88.4294364817927\\
32	88.6217504396765\\
30	88.8845035407274\\
29	88.8759352862089\\
27	88.6740951247438\\
26	89.0766866148966\\
25	89.2861948216499\\
24	89.3061048988938\\
23	89.0291169394345\\
22	88.88271285883\\
21	89.0630582613389\\
20	88.875938371843\\
19	88.8374301297322\\
18	89.0597426654266\\
17	88.6855652493962\\
16	88.4597496032839\\
15	87.5653946482697\\
14	88.7119816097372\\
13	88.7694479046242\\
12	88.5200545077548\\
11	89.1601310235227\\
10	89.5728997254755\\
9	90.1821301895404\\
8	88.7880825743845\\
7	92.2494418857509\\
6	91.3155253070338\\
5	91.4122596382689\\
4	87.5148180317603\\
3	85.5988725853075\\
2	90.8589285995805\\
1	86.7637033990926\\
};
\addplot [color=blue, forget plot]
  table[row sep=crcr]{%
1	84.4575596578727\\
2	87.3164195223037\\
3	83.1883652895836\\
4	85.2113550495485\\
5	89.9910424835366\\
6	88.4118708985516\\
7	94.2827398105184\\
8	93.456488031927\\
9	94.0620796045544\\
10	94.1000281312945\\
11	93.6679605499018\\
12	93.6209215320153\\
13	93.6456017660314\\
14	94.0474657484179\\
15	93.9520703489479\\
16	93.7174163254741\\
17	93.7208696626469\\
18	93.6274543719507\\
19	93.6385706848278\\
20	93.7585161914845\\
21	93.7702615679558\\
22	93.7399590914641\\
23	93.3982043014271\\
24	93.3328787907275\\
25	93.4317994022464\\
26	93.5581219141407\\
27	93.594987888356\\
28	93.4527123746279\\
29	93.5557539052176\\
30	93.76464801908\\
31	93.8367021802588\\
32	93.8808498842885\\
33	93.8675407262121\\
34	94.0473220172161\\
35	94.1076229877956\\
36	93.9294044428103\\
37	93.8236546558814\\
38	93.755786517492\\
39	93.7441547755374\\
40	93.8167929735676\\
41	93.8250417490555\\
42	93.8547619469042\\
43	93.6758963955854\\
44	93.6240811260952\\
45	93.5084827483943\\
46	93.3393158138247\\
47	93.4161332315065\\
48	93.6076287962186\\
49	93.5655438870897\\
50	93.589440189285\\
nan	nan\\
50	94.4674501233447\\
49	94.5127253550877\\
48	94.4920307836866\\
47	94.533953698826\\
46	94.8030788088155\\
45	94.7193591433558\\
44	94.6464501756694\\
43	94.6801049916494\\
42	94.5225955535179\\
41	94.5095921168117\\
40	94.6032881614094\\
39	94.5263651180471\\
38	94.6215367583898\\
37	94.5963808463752\\
36	94.5119871726337\\
35	94.4619737558535\\
33	94.4456799855671\\
32	94.6032767738906\\
31	94.6260341401924\\
30	94.3989886349464\\
29	94.3942873923945\\
28	94.4973289229842\\
27	94.3123297747011\\
26	94.1996801402444\\
25	94.2619343125764\\
24	94.2967523599929\\
23	94.4023670202334\\
22	94.4023556739153\\
21	94.3507313087765\\
20	94.4265792493504\\
19	94.5252028673601\\
18	94.3012650370143\\
17	94.2078497463181\\
16	94.531838720364\\
15	94.5321475746721\\
14	94.8640797861922\\
13	94.9455792544075\\
12	94.9061455161409\\
11	95.1582632055798\\
10	95.3884176638785\\
9	95.3195057674207\\
8	94.6003680561625\\
7	95.5901160220278\\
6	90.0949009971685\\
5	91.506760300972\\
4	86.3943348943688\\
3	84.8494961786035\\
2	88.434215913329\\
1	87.1056462232723\\
};
\addplot [color=red, forget plot]
  table[row sep=crcr]{%
1	65.2476448267889\\
2	83.2568087002571\\
3	90.6933242062155\\
4	93.266478112385\\
5	95.6227692325661\\
6	96.8121692346041\\
7	96.9846288073201\\
8	97.17585459102\\
9	97.0412701387041\\
10	97.1118248483615\\
11	96.9593596369471\\
12	97.0694045922711\\
13	97.0738029293189\\
14	97.0665203133686\\
15	97.0446252021346\\
16	97.1916900378402\\
17	97.1964428200599\\
18	97.2339156284337\\
19	97.1620858180845\\
20	96.9849787864403\\
21	97.0023997219426\\
22	96.937523917841\\
24	97.0777121551162\\
25	96.938548344244\\
26	96.8346501578565\\
27	96.8410763668508\\
28	96.9253465211415\\
29	96.9282558292868\\
30	96.959181598032\\
31	96.9516726527297\\
32	96.9722809851186\\
33	97.0162429435578\\
34	97.0801330995015\\
35	97.1623708747108\\
37	97.1778006256378\\
38	97.207985112667\\
39	97.2171307596789\\
40	97.1203417222844\\
41	97.1394502087112\\
42	97.2188614487738\\
43	97.2067406831769\\
44	97.2176900064216\\
47	97.1174976453458\\
48	97.177263311637\\
49	97.1697325650438\\
50	97.1413157744245\\
nan	nan\\
50	97.9874711709668\\
49	98.0017666067223\\
48	97.9301332960266\\
47	97.9044402850278\\
46	97.9824696550283\\
43	98.007470714964\\
42	98.0167174707346\\
41	98.07478400579\\
40	98.0725363790293\\
39	98.0184595680096\\
38	98.0703402577566\\
37	98.1432483793407\\
36	98.1921882890594\\
35	98.2655385534656\\
34	98.0273091408826\\
33	97.9843616899886\\
32	97.9642324925054\\
31	97.9634847117069\\
30	98.0414230355144\\
29	98.050969874712\\
28	98.0966370419526\\
27	98.1168160403208\\
26	98.1018747279476\\
25	98.1261816697652\\
24	98.0724651280029\\
23	98.0999378466873\\
22	97.977576405695\\
21	97.8699883752185\\
20	97.8446742679858\\
19	97.7957381400064\\
18	97.8948370924172\\
17	97.9537002385188\\
16	97.8729715270883\\
15	97.8704865295815\\
14	98.0408648861148\\
13	97.9908472274295\\
12	97.9524762972019\\
11	98.0839115902536\\
10	98.0596286906842\\
9	98.1087816544336\\
8	98.102413738503\\
7	98.1441125053507\\
6	98.0815635675644\\
5	95.9807303407679\\
4	94.1495234573806\\
3	92.385515444968\\
2	86.4262720559684\\
1	67.0461295011552\\
};
\end{axis}
\end{tikzpicture}%

%% file: Figures/balancing1_norm0_FFTtransformed_50rSVDmodes_recall_dots_class.tex
\definecolor{mycolor1}{rgb}{0.64, 0.76, 0.68}%
\begin{tikzpicture}

\begin{axis}[%
width=7cm,
height=5cm,
scale only axis,
unbounded coords=jump,
xmin=0,
xmax=50,
xlabel style={font=\color{white!15!black}},
xlabel={$r$},
ymin=0,
ymax=100,
ylabel style={font=\color{white!15!black}},
axis background/.style={fill=white},
title style={font=\bfseries},
legend columns = 2,
legend pos = south east,
legend style={legend cell align=left, align=left, draw=white!15!black, font=\footnotesize},
legend entries={Tree,
                NB,
                LD,
                kNN,
                96.6~\%},
]
\addlegendimage{legend image code/.code={
\draw [draw=none, fill=white!80!black] (0cm,-0.15cm) rectangle (0.6cm,0.15cm);
\draw[dashdotted] (0cm,0cm) -- (0.6cm,0cm);
\draw[-] (0cm,0.15cm) -- (0.6cm,0.15cm);
\draw[-] (0cm,-0.15cm) -- (0.6cm,-0.15cm);
}
} 
\addlegendimage{legend image code/.code={
\draw [draw=none, fill=mycolor1] (0cm,-0.15cm) rectangle (0.6cm,0.15cm);
\draw[dotted, black!80!green] (0cm,0cm) -- (0.6cm,0cm);
\draw[-, black!80!green] (0cm,0.15cm) -- (0.6cm,0.15cm);
\draw[-, black!80!green] (0cm,-0.15cm) -- (0.6cm,-0.15cm);
}
}
\addlegendimage{legend image code/.code={
\draw [draw=none, fill=white!80!blue] (0cm,-0.15cm) rectangle (0.6cm,0.15cm);
\draw[dashed, blue] (0cm,0cm) -- (0.6cm,0cm);
\draw[-, blue] (0cm,0.15cm) -- (0.6cm,0.15cm);
\draw[-, blue] (0cm,-0.15cm) -- (0.6cm,-0.15cm);
}
}
\addlegendimage{legend image code/.code={
\draw [draw=none, fill=white!80!red] (0cm,-0.15cm) rectangle (0.6cm,0.15cm);
\draw[-, red] (0cm,0cm) -- (0.6cm,0cm);
\draw[-, red] (0cm,0.15cm) -- (0.6cm,0.15cm);
\draw[-, red] (0cm,-0.15cm) -- (0.6cm,-0.15cm);
}
}
\addlegendimage{mark=none, black, very thick, dotted}

\addplot[area legend, draw=black, fill=white!80!black, dashdotted]
table[row sep=crcr] {%
x	y\\
1	58.0190937614347\\
2	71.9065497739471\\
3	85.1950735771644\\
4	88.8734538783722\\
5	91.305051645185\\
6	92.5828897671142\\
7	90.7074557586411\\
8	90.4127746818875\\
9	90.350333794024\\
10	90.1220594727888\\
11	90.3298975916805\\
12	90.3508298222377\\
13	90.8039343980253\\
14	90.6864200397489\\
15	90.335626254329\\
16	90.005954320634\\
17	89.7328296448127\\
18	90.0014943519224\\
19	90.4908295565535\\
20	90.108071930801\\
21	89.8741480856555\\
22	89.8799684268456\\
23	89.7316629285117\\
24	89.5294926487013\\
25	89.9556531937088\\
26	90.0883572603382\\
27	90.0980064473033\\
28	90.0402159138697\\
29	89.963942933084\\
30	89.7399833735105\\
31	89.8540583521265\\
32	89.7825058006573\\
33	89.8972561823966\\
34	89.9532028952183\\
35	89.9692251880432\\
36	89.5855759712973\\
37	89.6305574481322\\
38	89.6101277791302\\
39	89.5157753866264\\
40	89.5387266273245\\
41	89.6505376344086\\
42	89.4825507133326\\
43	89.477835679437\\
44	89.3114120261306\\
45	89.2765685506552\\
46	88.9693392735118\\
47	88.9037460064024\\
48	88.8695900973244\\
49	88.9556886630228\\
50	88.9164922576864\\
50	91.0297443014534\\
49	90.990547896117\\
48	90.969119580095\\
47	91.0424905527374\\
46	91.0306607264882\\
45	91.0997755353663\\
44	91.1186955007511\\
43	91.1673256108855\\
42	91.3776643404309\\
41	91.2634408602151\\
40	91.4827787490196\\
39	91.5594934305779\\
38	91.680194801515\\
37	91.8748188959539\\
36	91.5972197276274\\
35	91.6436780377633\\
34	91.6059368897279\\
33	91.5543567208292\\
32	91.5615802208481\\
31	91.6513179919595\\
30	91.4965757662744\\
29	91.272616206701\\
28	91.7339776345174\\
27	91.8912408645246\\
26	91.95465349235\\
25	91.9798306772589\\
24	91.8683568136643\\
23	91.9887671790152\\
22	92.0017520032619\\
21	92.0613357853123\\
20	92.0962291444678\\
19	92.1973424864572\\
18	92.4178604867873\\
17	92.3639445487357\\
16	92.3058736363552\\
15	92.6213629929829\\
14	92.8082036161651\\
13	92.9595064621898\\
12	93.5201379196978\\
11	93.4335432685346\\
10	93.2650373014047\\
9	93.2518167436104\\
8	93.6194833826287\\
7	93.3248023058751\\
6	93.707432813531\\
5	92.5659160967505\\
4	91.3415998850686\\
3	87.4393350249861\\
2	75.7816222690637\\
1	61.389508389103\\
}--cycle;
\addlegendentry{Tree}

\addplot[area legend, draw=black!80!green, fill=mycolor1, dotted]
table[row sep=crcr] {%
x	y\\
1	79.4123736730753\\
2	79.5011224416888\\
3	68.6577879560546\\
4	71.6471018927557\\
5	76.9708966188561\\
6	78.3312045211589\\
7	71.632598209334\\
8	69.1893650105355\\
9	70.9433154852095\\
10	71.1616324264063\\
11	70.577902352507\\
12	69.780329480579\\
13	69.3717888705666\\
14	68.2605382254526\\
15	69.0935659290233\\
16	69.2475409446243\\
17	69.8906643568285\\
18	71.1582671589995\\
19	71.633844523823\\
20	72.5723042931248\\
21	73.7224388784466\\
22	74.6835828673475\\
23	75.6734837923979\\
24	76.3905568749165\\
25	77.7204409034421\\
26	78.6658024867047\\
27	80.0998078697814\\
28	80.611932543118\\
29	81.1653804829444\\
30	81.3246632563704\\
31	81.6488491136195\\
32	82.2178035921034\\
33	82.5723247594502\\
34	82.7771796026294\\
35	83.2631110300093\\
36	83.0982538997668\\
37	83.1975349168064\\
38	83.1686857905454\\
39	83.3058072938756\\
40	83.6185751313495\\
41	83.7140866945246\\
42	83.2296506240244\\
43	82.9754195835195\\
44	82.8364330031318\\
45	83.1982804121552\\
46	83.3568578335561\\
47	83.4763741424803\\
48	82.7455113813838\\
49	83.1027541737461\\
50	83.0727374950456\\
50	86.7122087415136\\
49	87.0047727079744\\
48	86.9856714143151\\
47	86.8999699435412\\
46	87.1270131341858\\
45	87.1243002330062\\
44	86.7872229108466\\
43	86.9170535347601\\
42	86.9854031394165\\
41	86.8235477140776\\
40	87.1878764815537\\
39	87.2318271147266\\
38	86.938841091175\\
37	86.3185941154516\\
36	86.09529448733\\
35	86.1992545613885\\
34	85.6636806124243\\
33	85.7072451330229\\
32	85.1477878057461\\
31	84.7490003487461\\
30	84.3204980339522\\
29	83.942146398776\\
28	83.635379284839\\
27	82.5883641732294\\
26	80.9578534272738\\
25	80.3978386664504\\
24	79.4696581788469\\
23	78.6813549172795\\
22	77.1981375627601\\
21	76.1162707989727\\
20	76.7825344165526\\
19	76.3231447234888\\
18	76.5299048840112\\
17	75.4856797291931\\
16	73.9782655069887\\
15	73.970950200009\\
14	74.6964510218592\\
13	76.1120820971753\\
12	75.8110683688834\\
11	73.8844632388909\\
10	76.5265396166045\\
9	74.3255017190916\\
8	72.9074091830129\\
7	75.4104125433542\\
6	80.5397632207765\\
5	79.5882431660901\\
4	74.7507475696099\\
3	71.9336098934078\\
2	83.0795227196015\\
1	82.9532177247741\\
}--cycle;
\addlegendentry{NB}

\addplot[area legend, draw=blue, fill=white!80!blue,dashed]
table[row sep=crcr] {%
x	y\\
1	72.0677338824936\\
2	72.6285926181568\\
3	69.5941203258328\\
4	74.3402252529254\\
5	81.0997725765006\\
6	83.9270056845153\\
7	84.111410717627\\
8	83.0335332724127\\
9	83.5028180800391\\
10	83.5833734242806\\
11	83.2179272569214\\
12	83.2237540750223\\
13	83.180923987021\\
14	83.1305620794964\\
15	83.3144554798583\\
16	83.4317903695409\\
17	83.9136115368206\\
18	83.855772125612\\
19	83.9243578735692\\
20	83.9315872331001\\
21	83.8397298293519\\
22	84.094110361313\\
23	84.0249980229949\\
24	83.9010736967877\\
25	83.8609246183058\\
26	84.1986019043296\\
27	84.5146367539075\\
28	84.5676554324467\\
29	84.7939571519252\\
30	84.7659435065853\\
31	85.4518187897534\\
32	85.2378662225667\\
33	85.2951148154182\\
34	85.5985698782075\\
35	85.5759630209407\\
36	85.4674344648861\\
37	85.6384786166321\\
38	85.5749613466066\\
39	85.8738282610318\\
40	85.974548539118\\
41	85.8870967741936\\
42	85.7431336765109\\
43	85.7579702015596\\
44	85.7385010577382\\
45	85.7060096410396\\
46	85.7084001320502\\
47	85.9035147947256\\
48	85.6358176421701\\
49	85.7613042099818\\
50	85.844827647785\\
50	86.6820540726451\\
49	86.926867833029\\
48	86.9985909599804\\
47	87.0534744525862\\
46	87.0335353518208\\
45	86.8208720793905\\
44	86.9496709852725\\
43	86.9839652823114\\
42	86.6224577213386\\
41	86.4247311827957\\
40	86.4985697404519\\
39	86.9218706636994\\
38	86.8443934921031\\
37	86.8346396629378\\
36	86.7368666103827\\
35	86.5745746134679\\
34	86.4982043153409\\
33	86.5328421738291\\
32	86.7513810892613\\
31	86.644955403795\\
30	86.3630887514792\\
29	86.3888385469996\\
28	86.1850327395963\\
27	86.4531051815763\\
26	86.1239787408317\\
25	86.5691829085759\\
24	86.4752703892338\\
23	86.6201632673277\\
22	86.7661046924504\\
21	86.9129583426912\\
20	86.9286278206633\\
19	86.8283302984738\\
18	86.8431526055708\\
17	86.9466035169428\\
16	87.1058440390613\\
15	87.008125165303\\
14	87.0307282430843\\
13	86.7115491312586\\
12	87.098826570139\\
11	86.2982017753367\\
10	87.0080244251818\\
9	86.2283647156598\\
8	85.9449613512433\\
7	86.5337505726956\\
6	87.5783706595707\\
5	83.6851736600586\\
4	76.0361188330961\\
3	71.9112560182532\\
2	76.6724826506604\\
1	75.0828037519151\\
}--cycle;
\addlegendentry{LD}

\addplot[area legend, draw=red, fill=white!80!red]
table[row sep=crcr] {%
x	y\\
1	54.8106476910287\\
2	71.5970447501843\\
3	86.6190564504985\\
4	90.602892163914\\
5	92.2087356457476\\
6	94.4470760205676\\
7	94.409809076302\\
8	94.5712001068684\\
9	94.7037082883761\\
10	94.5684680709119\\
11	94.7469405201212\\
12	94.7287423597971\\
13	94.5453124403157\\
14	94.5508897431445\\
15	94.6219156665202\\
16	94.5391176611614\\
17	94.6843856909172\\
18	94.8025576334055\\
19	94.6469747842441\\
20	94.7199853635278\\
21	94.6263073184751\\
22	94.6522193858779\\
23	94.5527672214406\\
24	94.5618983111354\\
25	94.5863575879543\\
26	94.3632723208073\\
27	94.3500505860571\\
28	94.3212603907519\\
29	94.418938907851\\
30	94.3905210079191\\
31	94.3078198734466\\
32	94.3204536939627\\
33	94.2945822506132\\
34	94.3905210079191\\
35	94.285053440161\\
36	94.2488307286685\\
37	94.2769280694765\\
38	94.2921639307472\\
39	94.2204301075269\\
40	94.0993703092234\\
41	94.1920561087462\\
42	94.2012604861446\\
43	93.9852361874004\\
44	93.9814281460952\\
45	94.0456116500979\\
46	94.1542675684679\\
47	94.1860450918122\\
48	94.1086191004428\\
49	94.2398313709138\\
50	94.2398313709138\\
50	96.0827492742474\\
49	96.0827492742474\\
48	96.0526712221379\\
47	95.9752452307685\\
46	96.060786194973\\
45	96.1156786724827\\
44	96.0723352947651\\
43	96.12229069432\\
42	96.0675567181564\\
41	96.1305245364151\\
40	96.0619200133572\\
39	96.1021505376344\\
38	95.9228898326937\\
37	95.9918891348245\\
36	95.9124595939121\\
35	96.0375272050004\\
34	96.1471134006831\\
33	96.0817618354083\\
32	96.109653832919\\
31	96.1760510942954\\
30	96.1471134006831\\
29	96.2262223824716\\
28	96.2163740178503\\
27	96.2413472634053\\
26	96.4969427329562\\
25	96.4351477883897\\
24	96.5133705060689\\
23	96.4687381549035\\
22	96.3692859904662\\
21	96.3951980578689\\
20	96.2477565719561\\
19	96.3745305921\\
18	96.4877649472397\\
17	96.3371196854268\\
16	96.2135705108816\\
15	96.0770090646626\\
14	96.1480349880383\\
13	96.5837198177489\\
12	96.5078167799879\\
11	96.3820917379433\\
10	96.3455104237118\\
9	96.5328508514088\\
8	96.5040687103359\\
7	96.5041694183217\\
6	96.3056121514755\\
5	94.4041675800589\\
4	91.4401185887741\\
3	89.7250295710069\\
2	74.9620950347619\\
1	58.7915028466057\\
}--cycle;
\addlegendentry{kNN}

\addplot[mark=none, black, very thick, dotted, domain=0:50] {96.58};
\addlegendentry{96.6~\%}

\addplot [color=black, forget plot,dashdotted]
  table[row sep=crcr]{%
1	59.7043010752688\\
2	73.8440860215054\\
3	86.3172043010753\\
4	90.1075268817204\\
5	91.9354838709677\\
6	93.1451612903226\\
7	92.0161290322581\\
8	92.0161290322581\\
9	91.8010752688172\\
10	91.6935483870968\\
11	91.8817204301075\\
12	91.9354838709677\\
13	91.8817204301075\\
14	91.747311827957\\
15	91.4784946236559\\
16	91.1559139784946\\
17	91.0483870967742\\
18	91.2096774193548\\
19	91.3440860215054\\
20	91.1021505376344\\
21	90.9677419354839\\
22	90.9408602150538\\
23	90.8602150537634\\
24	90.6989247311828\\
25	90.9677419354839\\
26	91.0215053763441\\
27	90.994623655914\\
28	90.8870967741935\\
29	90.6182795698925\\
30	90.6182795698925\\
31	90.752688172043\\
32	90.6720430107527\\
34	90.7795698924731\\
35	90.8064516129032\\
36	90.5913978494624\\
37	90.752688172043\\
39	90.5376344086022\\
40	90.510752688172\\
41	90.4569892473118\\
42	90.4301075268817\\
44	90.2150537634408\\
45	90.1881720430107\\
46	90\\
47	89.9731182795699\\
48	89.9193548387097\\
49	89.9731182795699\\
50	89.9731182795699\\
};
\addplot [color=black!80!green, forget plot,dotted]
  table[row sep=crcr]{%
1	81.1827956989247\\
2	81.2903225806452\\
3	70.2956989247312\\
4	73.1989247311828\\
5	78.2795698924731\\
6	79.4354838709677\\
7	73.5215053763441\\
8	71.0483870967742\\
9	72.6344086021505\\
10	73.8440860215054\\
11	72.2311827956989\\
12	72.7956989247312\\
13	72.741935483871\\
14	71.4784946236559\\
15	71.5322580645161\\
16	71.6129032258065\\
17	72.6881720430108\\
18	73.8440860215054\\
19	73.9784946236559\\
20	74.6774193548387\\
21	74.9193548387097\\
22	75.9408602150538\\
23	77.1774193548387\\
24	77.9301075268817\\
25	79.0591397849462\\
26	79.8118279569892\\
27	81.3440860215054\\
28	82.1236559139785\\
29	82.5537634408602\\
30	82.8225806451613\\
31	83.1989247311828\\
32	83.6827956989247\\
33	84.1397849462366\\
34	84.2204301075269\\
35	84.7311827956989\\
36	84.5967741935484\\
37	84.758064516129\\
38	85.0537634408602\\
39	85.2688172043011\\
40	85.4032258064516\\
41	85.2688172043011\\
43	84.9462365591398\\
44	84.8118279569892\\
45	85.1612903225807\\
46	85.241935483871\\
47	85.1881720430108\\
48	84.8655913978495\\
49	85.0537634408602\\
50	84.8924731182796\\
};
\addplot [color=blue, forget plot, dashed]
  table[row sep=crcr]{%
1	73.5752688172043\\
2	74.6505376344086\\
3	70.752688172043\\
4	75.1881720430107\\
5	82.3924731182796\\
6	85.752688172043\\
7	85.3225806451613\\
8	84.489247311828\\
9	84.8655913978495\\
10	85.2956989247312\\
11	84.758064516129\\
12	85.1612903225806\\
13	84.9462365591398\\
14	85.0806451612903\\
15	85.1612903225807\\
16	85.2688172043011\\
17	85.4301075268817\\
18	85.3494623655914\\
19	85.3763440860215\\
20	85.4301075268817\\
21	85.3763440860215\\
22	85.4301075268817\\
23	85.3225806451613\\
24	85.1881720430108\\
25	85.2150537634409\\
26	85.1612903225807\\
27	85.4838709677419\\
28	85.3763440860215\\
29	85.5913978494624\\
30	85.5645161290323\\
31	86.0483870967742\\
32	85.994623655914\\
33	85.9139784946236\\
34	86.0483870967742\\
36	86.1021505376344\\
37	86.236559139785\\
38	86.2096774193548\\
39	86.3978494623656\\
40	86.2365591397849\\
41	86.1559139784946\\
42	86.1827956989247\\
43	86.3709677419355\\
44	86.3440860215054\\
45	86.263440860215\\
47	86.4784946236559\\
48	86.3172043010753\\
49	86.3440860215054\\
50	86.2634408602151\\
};
\addplot [color=red, forget plot]
  table[row sep=crcr]{%
1	56.8010752688172\\
2	73.2795698924731\\
3	88.1720430107527\\
4	91.0215053763441\\
5	93.3064516129032\\
6	95.3763440860215\\
9	95.6182795698925\\
10	95.4569892473118\\
11	95.5645161290323\\
12	95.6182795698925\\
13	95.5645161290323\\
14	95.3494623655914\\
15	95.3494623655914\\
16	95.3763440860215\\
18	95.6451612903226\\
19	95.510752688172\\
20	95.4838709677419\\
21	95.510752688172\\
23	95.510752688172\\
24	95.5376344086022\\
25	95.510752688172\\
26	95.4301075268817\\
27	95.2956989247312\\
28	95.2688172043011\\
29	95.3225806451613\\
30	95.2688172043011\\
33	95.1881720430108\\
34	95.2688172043011\\
35	95.1612903225807\\
36	95.0806451612903\\
37	95.1344086021505\\
38	95.1075268817204\\
39	95.1612903225807\\
40	95.0806451612903\\
41	95.1612903225806\\
42	95.1344086021505\\
43	95.0537634408602\\
44	95.0268817204301\\
45	95.0806451612903\\
46	95.1075268817204\\
47	95.0806451612903\\
48	95.0806451612903\\
49	95.1612903225806\\
50	95.1612903225806\\
};
\addplot [color=black, forget plot]
  table[row sep=crcr]{%
1	58.0190937614347\\
2	71.9065497739471\\
3	85.1950735771644\\
4	88.8734538783722\\
5	91.305051645185\\
6	92.5828897671142\\
7	90.7074557586411\\
8	90.4127746818875\\
9	90.350333794024\\
10	90.1220594727888\\
11	90.3298975916805\\
12	90.3508298222377\\
13	90.8039343980253\\
14	90.6864200397489\\
15	90.335626254329\\
16	90.005954320634\\
17	89.7328296448127\\
18	90.0014943519224\\
19	90.4908295565535\\
20	90.108071930801\\
21	89.8741480856555\\
22	89.8799684268456\\
23	89.7316629285117\\
24	89.5294926487013\\
25	89.9556531937088\\
26	90.0883572603382\\
27	90.0980064473033\\
28	90.0402159138697\\
29	89.963942933084\\
30	89.7399833735105\\
31	89.8540583521265\\
32	89.7825058006573\\
33	89.8972561823966\\
34	89.9532028952183\\
35	89.9692251880432\\
36	89.5855759712973\\
37	89.6305574481322\\
38	89.6101277791302\\
39	89.5157753866264\\
40	89.5387266273245\\
41	89.6505376344086\\
42	89.4825507133326\\
43	89.477835679437\\
44	89.3114120261306\\
45	89.2765685506552\\
46	88.9693392735118\\
47	88.9037460064024\\
48	88.8695900973244\\
49	88.9556886630228\\
50	88.9164922576864\\
nan	nan\\
50	91.0297443014534\\
49	90.990547896117\\
48	90.969119580095\\
47	91.0424905527374\\
46	91.0306607264882\\
45	91.0997755353663\\
44	91.1186955007511\\
43	91.1673256108855\\
42	91.3776643404309\\
41	91.2634408602151\\
40	91.4827787490196\\
39	91.5594934305779\\
38	91.680194801515\\
37	91.8748188959539\\
36	91.5972197276274\\
35	91.6436780377633\\
34	91.6059368897279\\
33	91.5543567208292\\
32	91.5615802208481\\
31	91.6513179919595\\
30	91.4965757662744\\
29	91.272616206701\\
28	91.7339776345174\\
27	91.8912408645246\\
26	91.95465349235\\
25	91.9798306772589\\
24	91.8683568136643\\
23	91.9887671790152\\
22	92.0017520032619\\
21	92.0613357853123\\
20	92.0962291444678\\
19	92.1973424864572\\
18	92.4178604867873\\
16	92.3058736363552\\
15	92.6213629929829\\
14	92.8082036161651\\
13	92.9595064621898\\
12	93.5201379196978\\
11	93.4335432685346\\
10	93.2650373014047\\
9	93.2518167436104\\
8	93.6194833826287\\
7	93.3248023058751\\
6	93.707432813531\\
5	92.5659160967505\\
4	91.3415998850686\\
3	87.4393350249861\\
2	75.7816222690637\\
1	61.389508389103\\
};
\addplot [color=black!80!green, forget plot]
  table[row sep=crcr]{%
1	79.4123736730753\\
2	79.5011224416888\\
3	68.6577879560546\\
4	71.6471018927557\\
5	76.9708966188561\\
6	78.3312045211589\\
7	71.632598209334\\
8	69.1893650105355\\
9	70.9433154852095\\
10	71.1616324264063\\
11	70.577902352507\\
12	69.780329480579\\
13	69.3717888705666\\
14	68.2605382254526\\
15	69.0935659290233\\
16	69.2475409446243\\
17	69.8906643568285\\
18	71.1582671589995\\
19	71.633844523823\\
20	72.5723042931248\\
21	73.7224388784466\\
22	74.6835828673475\\
23	75.6734837923979\\
24	76.3905568749165\\
25	77.7204409034421\\
26	78.6658024867047\\
27	80.0998078697814\\
28	80.611932543118\\
29	81.1653804829444\\
30	81.3246632563704\\
31	81.6488491136195\\
32	82.2178035921034\\
33	82.5723247594502\\
34	82.7771796026294\\
35	83.2631110300093\\
36	83.0982538997668\\
37	83.1975349168064\\
38	83.1686857905454\\
39	83.3058072938756\\
40	83.6185751313495\\
41	83.7140866945246\\
42	83.2296506240244\\
43	82.9754195835195\\
44	82.8364330031318\\
45	83.1982804121552\\
46	83.3568578335561\\
47	83.4763741424803\\
48	82.7455113813838\\
49	83.1027541737461\\
50	83.0727374950456\\
nan	nan\\
50	86.7122087415136\\
49	87.0047727079744\\
48	86.9856714143151\\
47	86.8999699435412\\
46	87.1270131341858\\
45	87.1243002330062\\
44	86.7872229108466\\
43	86.9170535347601\\
42	86.9854031394165\\
41	86.8235477140776\\
40	87.1878764815537\\
39	87.2318271147266\\
38	86.938841091175\\
37	86.3185941154516\\
36	86.09529448733\\
35	86.1992545613885\\
34	85.6636806124243\\
33	85.7072451330229\\
32	85.1477878057461\\
31	84.7490003487461\\
30	84.3204980339522\\
29	83.942146398776\\
28	83.635379284839\\
27	82.5883641732294\\
26	80.9578534272738\\
25	80.3978386664504\\
24	79.4696581788469\\
23	78.6813549172795\\
22	77.1981375627601\\
21	76.1162707989727\\
20	76.7825344165526\\
19	76.3231447234888\\
18	76.5299048840112\\
17	75.4856797291931\\
16	73.9782655069887\\
15	73.970950200009\\
14	74.6964510218592\\
13	76.1120820971753\\
12	75.8110683688834\\
11	73.8844632388909\\
10	76.5265396166045\\
9	74.3255017190916\\
8	72.9074091830129\\
7	75.4104125433542\\
6	80.5397632207765\\
5	79.5882431660901\\
4	74.7507475696099\\
3	71.9336098934078\\
2	83.0795227196015\\
1	82.9532177247741\\
};
\addplot [color=blue, forget plot]
  table[row sep=crcr]{%
1	72.0677338824936\\
2	72.6285926181568\\
3	69.5941203258328\\
4	74.3402252529254\\
5	81.0997725765006\\
6	83.9270056845153\\
7	84.111410717627\\
8	83.0335332724127\\
9	83.5028180800391\\
10	83.5833734242806\\
11	83.2179272569214\\
12	83.2237540750223\\
14	83.1305620794964\\
15	83.3144554798583\\
16	83.4317903695409\\
17	83.9136115368206\\
18	83.855772125612\\
19	83.9243578735692\\
20	83.9315872331001\\
21	83.8397298293519\\
22	84.094110361313\\
23	84.0249980229949\\
24	83.9010736967877\\
25	83.8609246183058\\
26	84.1986019043296\\
27	84.5146367539075\\
28	84.5676554324467\\
29	84.7939571519252\\
30	84.7659435065853\\
31	85.4518187897534\\
32	85.2378662225667\\
33	85.2951148154182\\
34	85.5985698782075\\
35	85.5759630209407\\
36	85.4674344648861\\
37	85.6384786166321\\
38	85.5749613466066\\
39	85.8738282610318\\
40	85.974548539118\\
41	85.8870967741936\\
42	85.7431336765109\\
43	85.7579702015596\\
44	85.7385010577382\\
45	85.7060096410396\\
46	85.7084001320502\\
47	85.9035147947256\\
48	85.6358176421701\\
49	85.7613042099818\\
50	85.844827647785\\
nan	nan\\
50	86.6820540726451\\
49	86.926867833029\\
48	86.9985909599804\\
47	87.0534744525862\\
46	87.0335353518208\\
45	86.8208720793905\\
44	86.9496709852725\\
43	86.9839652823114\\
42	86.6224577213386\\
41	86.4247311827957\\
40	86.4985697404519\\
39	86.9218706636994\\
38	86.8443934921031\\
37	86.8346396629378\\
36	86.7368666103827\\
35	86.5745746134679\\
34	86.4982043153409\\
33	86.5328421738291\\
32	86.7513810892613\\
31	86.644955403795\\
30	86.3630887514792\\
29	86.3888385469996\\
28	86.1850327395963\\
27	86.4531051815763\\
26	86.1239787408317\\
25	86.5691829085759\\
24	86.4752703892338\\
21	86.9129583426912\\
20	86.9286278206633\\
19	86.8283302984738\\
18	86.8431526055708\\
17	86.9466035169428\\
16	87.1058440390613\\
15	87.008125165303\\
14	87.0307282430843\\
13	86.7115491312586\\
12	87.098826570139\\
11	86.2982017753367\\
10	87.0080244251818\\
9	86.2283647156598\\
8	85.9449613512433\\
7	86.5337505726956\\
6	87.5783706595707\\
5	83.6851736600586\\
4	76.0361188330961\\
3	71.9112560182532\\
2	76.6724826506604\\
1	75.0828037519151\\
};
\addplot [color=red, forget plot]
  table[row sep=crcr]{%
1	54.8106476910287\\
2	71.5970447501843\\
3	86.6190564504985\\
4	90.602892163914\\
5	92.2087356457476\\
6	94.4470760205676\\
7	94.409809076302\\
8	94.5712001068684\\
9	94.7037082883761\\
10	94.5684680709119\\
11	94.7469405201212\\
12	94.7287423597971\\
13	94.5453124403157\\
14	94.5508897431445\\
15	94.6219156665202\\
16	94.5391176611614\\
17	94.6843856909172\\
18	94.8025576334055\\
19	94.6469747842441\\
20	94.7199853635278\\
21	94.6263073184751\\
22	94.6522193858779\\
23	94.5527672214406\\
24	94.5618983111354\\
25	94.5863575879543\\
26	94.3632723208073\\
27	94.3500505860571\\
28	94.3212603907519\\
29	94.418938907851\\
30	94.3905210079191\\
31	94.3078198734466\\
32	94.3204536939627\\
33	94.2945822506132\\
34	94.3905210079191\\
35	94.285053440161\\
36	94.2488307286685\\
37	94.2769280694765\\
38	94.2921639307472\\
39	94.2204301075269\\
40	94.0993703092234\\
41	94.1920561087462\\
42	94.2012604861446\\
43	93.9852361874004\\
44	93.9814281460952\\
45	94.0456116500979\\
46	94.1542675684679\\
47	94.1860450918122\\
48	94.1086191004428\\
49	94.2398313709138\\
50	94.2398313709138\\
nan	nan\\
50	96.0827492742474\\
49	96.0827492742474\\
48	96.0526712221379\\
47	95.9752452307685\\
46	96.060786194973\\
45	96.1156786724827\\
44	96.0723352947651\\
43	96.12229069432\\
42	96.0675567181564\\
41	96.1305245364151\\
40	96.0619200133572\\
39	96.1021505376344\\
38	95.9228898326937\\
37	95.9918891348245\\
36	95.9124595939121\\
35	96.0375272050004\\
34	96.1471134006831\\
33	96.0817618354083\\
32	96.109653832919\\
31	96.1760510942954\\
30	96.1471134006831\\
29	96.2262223824716\\
28	96.2163740178503\\
27	96.2413472634053\\
26	96.4969427329562\\
25	96.4351477883897\\
24	96.5133705060689\\
23	96.4687381549035\\
22	96.3692859904662\\
21	96.3951980578689\\
20	96.2477565719561\\
19	96.3745305921\\
18	96.4877649472397\\
17	96.3371196854268\\
16	96.2135705108816\\
15	96.0770090646626\\
14	96.1480349880383\\
13	96.5837198177489\\
12	96.5078167799879\\
11	96.3820917379433\\
10	96.3455104237118\\
9	96.5328508514088\\
8	96.5040687103359\\
7	96.5041694183217\\
6	96.3056121514755\\
5	94.4041675800589\\
4	91.4401185887741\\
3	89.7250295710069\\
2	74.9620950347619\\
1	58.7915028466057\\
};
\end{axis}
\end{tikzpicture}%

%% file: Figures/balancing0_norm0_FFTtransformed_50rSVDmodes_recall_fingers_class.tex
\definecolor{mycolor1}{rgb}{0.64, 0.76, 0.68}%
\begin{tikzpicture}

\begin{axis}[%
width=7cm,
height=5cm,
scale only axis,
unbounded coords=jump,
xmin=0,
xmax=50,
xlabel style={font=\color{white!15!black}},
xlabel={$r$},
ymin=0,
ymax=100,
ylabel style={font=\color{white!15!black}},
ylabel={Recall C of fingers class in \%},
axis background/.style={fill=white},
title style={font=\bfseries},
legend columns = 2,
legend pos = south east,
legend style={legend cell align=left, align=left, draw=white!15!black, font=\footnotesize},
legend entries={Tree,
                NB,
                LD,
                kNN,
                98.9~\%},
]
\addlegendimage{legend image code/.code={
\draw [draw=none, fill=white!80!black] (0cm,-0.15cm) rectangle (0.6cm,0.15cm);
\draw[dashdotted] (0cm,0cm) -- (0.6cm,0cm);
\draw[-] (0cm,0.15cm) -- (0.6cm,0.15cm);
\draw[-] (0cm,-0.15cm) -- (0.6cm,-0.15cm);
}
} 
\addlegendimage{legend image code/.code={
\draw [draw=none, fill=mycolor1] (0cm,-0.15cm) rectangle (0.6cm,0.15cm);
\draw[dotted, black!80!green] (0cm,0cm) -- (0.6cm,0cm);
\draw[-, black!80!green] (0cm,0.15cm) -- (0.6cm,0.15cm);
\draw[-, black!80!green] (0cm,-0.15cm) -- (0.6cm,-0.15cm);
}
}
\addlegendimage{legend image code/.code={
\draw [draw=none, fill=white!80!blue] (0cm,-0.15cm) rectangle (0.6cm,0.15cm);
\draw[dashed, blue] (0cm,0cm) -- (0.6cm,0cm);
\draw[-, blue] (0cm,0.15cm) -- (0.6cm,0.15cm);
\draw[-, blue] (0cm,-0.15cm) -- (0.6cm,-0.15cm);
}
}
\addlegendimage{legend image code/.code={
\draw [draw=none, fill=white!80!red] (0cm,-0.15cm) rectangle (0.6cm,0.15cm);
\draw[-, red] (0cm,0cm) -- (0.6cm,0cm);
\draw[-, red] (0cm,0.15cm) -- (0.6cm,0.15cm);
\draw[-, red] (0cm,-0.15cm) -- (0.6cm,-0.15cm);
}
}
\addlegendimage{mark=none, black, very thick, dotted}

\addplot[area legend, draw=black, fill=white!80!black, dashdotted]
table[row sep=crcr] {%
x	y\\
1	77.7010097206438\\
2	87.8775482111069\\
3	92.2103397107741\\
4	93.5996636347119\\
5	95.5723711595146\\
6	96.1522864049922\\
7	96.3364735764658\\
8	96.4026757885762\\
9	96.5916045548038\\
10	96.5495654597671\\
11	96.4438520034418\\
12	96.1768040709561\\
13	96.142553128961\\
14	96.0336365449824\\
15	96.0365542977309\\
16	96.1400672447157\\
17	96.1667957283196\\
18	96.3031107022384\\
19	96.0503737102083\\
20	96.0862726689533\\
21	95.9830101979916\\
22	95.8210409960337\\
23	95.9309479190219\\
24	95.9042609678968\\
25	95.9128817148687\\
26	96.0925043863021\\
27	96.0568679837545\\
28	95.9411127176965\\
29	95.9618906313519\\
30	95.9297741615548\\
31	95.8957948136432\\
32	95.8479983214007\\
33	95.7712311690668\\
34	95.7263252244064\\
35	95.8202595205501\\
36	95.901069674323\\
37	95.8008395111219\\
38	95.7207125757569\\
39	95.6742748972227\\
40	95.6963219583253\\
41	95.6892030741858\\
42	95.6290713975559\\
43	95.5221362562465\\
44	95.5033374139915\\
45	95.6114703130698\\
46	95.6067106267384\\
47	95.5610121759469\\
48	95.4448792019834\\
49	95.4337424037886\\
50	95.3869774272053\\
50	96.3762729648795\\
49	96.373023034709\\
48	96.3763789863283\\
47	96.3472655917702\\
46	96.4320544564056\\
45	96.4997900594049\\
44	96.5209033790778\\
43	96.5456090698154\\
42	96.5981835405318\\
41	96.5670689037907\\
40	96.5744532828855\\
39	96.6254911003261\\
38	96.6225579547844\\
37	96.6584413551629\\
36	96.68874581778\\
35	96.653529865679\\
34	96.7764444047406\\
33	96.7750377363626\\
32	96.8142966899025\\
31	96.8970190533478\\
30	96.8630397054363\\
29	96.8309284923493\\
28	96.7792163733842\\
27	96.7214899303935\\
26	96.7728520804108\\
25	96.7784461064536\\
24	96.8160576097636\\
23	96.818382441817\\
22	96.9137545613103\\
21	96.9548205312114\\
20	96.7645910479455\\
19	96.7569697035676\\
18	96.7362744098649\\
17	96.8435460603443\\
16	96.9282613133346\\
15	96.9302514176798\\
14	96.9766947302612\\
13	97.0418067916731\\
12	97.167065461704\\
11	97.0740461746087\\
10	97.1713468633019\\
9	97.1728070445477\\
8	97.2456834213507\\
7	96.9639545037232\\
6	97.0031826367962\\
5	96.2199593529805\\
4	95.0167413485442\\
3	93.302950435181\\
2	89.906517176059\\
1	80.5933928555647\\
}--cycle;
\addlegendentry{Tree}

\addplot[area legend, draw=black!80!green, fill=mycolor1, dotted]
table[row sep=crcr] {%
x	y\\
1	86.369919588866\\
2	90.7576444751129\\
3	91.5219293884986\\
4	92.5977180024671\\
5	92.4120933850905\\
6	92.4329083484411\\
7	92.7057126208932\\
8	92.0867533610352\\
9	92.5575718755273\\
10	92.304189817332\\
11	91.6230122710067\\
12	91.6645284167859\\
13	91.4104010118256\\
14	91.0347494561662\\
15	91.0354363421779\\
16	90.8176280670333\\
17	90.465643245502\\
18	90.4812611511186\\
19	90.2160628141167\\
20	90.2392797413932\\
21	89.4707839338981\\
22	89.0879932675199\\
23	88.7697637035184\\
24	88.5410719051033\\
25	88.1681648447666\\
26	87.8278326006238\\
27	87.3832479253271\\
28	87.1036580992142\\
29	86.9046544143713\\
30	86.5835036326836\\
31	86.4046519853655\\
32	86.2255159327648\\
33	85.9904261592187\\
34	85.6956838485922\\
35	85.5062952278591\\
36	85.1751373253508\\
37	84.9328127380998\\
38	84.384654814503\\
39	84.2254280660142\\
40	83.8419178880646\\
41	83.597317442393\\
42	83.2187765244889\\
43	82.7541979214219\\
44	82.3547974319281\\
45	81.9749346939399\\
46	81.7144740959679\\
47	81.4814752434394\\
48	81.3250440531357\\
49	80.8562912093323\\
50	80.4769781014053\\
50	81.9071081548998\\
49	82.1658808054674\\
48	82.3061178296182\\
47	82.5847267125298\\
46	82.9317847954287\\
45	83.120893817457\\
44	83.4370825805092\\
43	83.7627611453648\\
42	84.1537541687845\\
41	84.4857627584448\\
40	84.7632061952637\\
39	85.0612442551203\\
38	85.5110967386577\\
37	86.0360278498001\\
36	86.1417184996492\\
35	86.4340905253\\
34	86.6797419777824\\
33	86.8345272334752\\
32	86.9909467166014\\
31	87.4063971462129\\
30	87.5175634531882\\
29	87.8054340797153\\
28	88.128484790783\\
27	88.3709598740012\\
26	88.6803976990008\\
25	88.9490342959715\\
24	89.3591983160807\\
23	89.6089827866419\\
22	90.1608281123612\\
21	90.4306580079727\\
20	91.0687632090825\\
19	91.0339460565817\\
18	91.4212053235555\\
17	91.625323597536\\
16	91.6069401139426\\
15	91.6356715436495\\
14	91.8973540873558\\
13	92.2757408021232\\
12	92.5001369765303\\
11	92.8026067263232\\
10	93.2525260109638\\
9	93.2312434749073\\
8	92.8900369495864\\
7	93.5906223219572\\
6	93.1673961202093\\
5	93.2751991227046\\
4	93.2346702181917\\
3	92.019185049818\\
2	92.0004619632794\\
1	87.3829955854425\\
}--cycle;
\addlegendentry{NB}

\addplot[area legend, draw=blue, fill=white!80!blue, dashed]
table[row sep=crcr] {%
x	y\\
1	85.7021183832874\\
2	85.9471129456235\\
3	86.7044005971061\\
4	89.4192684478947\\
5	89.7166995010979\\
6	86.7565401361972\\
7	88.2740283376715\\
8	89.0100350034701\\
9	91.0305268862402\\
10	91.7729989314609\\
11	91.8404913605125\\
12	92.0829598872456\\
13	92.421971523726\\
14	92.589490490537\\
15	92.5500335019131\\
16	92.9588474041311\\
17	92.9609723067942\\
18	92.9544927510677\\
19	93.0112418787567\\
20	93.1547775570303\\
21	93.117025002663\\
22	93.2004594824271\\
23	93.1729053312008\\
24	93.1737731927983\\
25	93.2570483875708\\
26	93.2377544655535\\
27	93.2841316842504\\
28	93.2712611399834\\
29	93.2603768664902\\
30	93.2751212872816\\
31	93.2936048621562\\
32	93.2313144250924\\
33	93.2297001701083\\
34	93.2125077707941\\
35	93.2405597053028\\
36	93.2995049233876\\
37	93.1895688939404\\
38	93.2680527112941\\
39	93.2824025945883\\
40	93.3011781726378\\
41	93.3168173114537\\
42	93.3305197700892\\
43	93.3046530024067\\
44	93.2811516778258\\
45	93.4716221097034\\
46	93.4963921220288\\
47	93.5508565622345\\
48	93.5669893479133\\
49	93.4784728895994\\
50	93.5478740326192\\
50	93.8361478289869\\
49	93.8040313860772\\
48	93.8025239937484\\
47	93.7606489832005\\
46	93.7425813371046\\
45	93.7093435532032\\
44	93.6533583875913\\
43	93.6878648592372\\
42	93.5604542220746\\
41	93.6031684638887\\
40	93.6333108659388\\
39	93.62308517423\\
38	93.6519435774685\\
37	93.6724353687258\\
36	93.5624520288875\\
35	93.5489019576417\\
34	93.5334493591579\\
33	93.5887732760147\\
32	93.7031956403248\\
31	93.6843939661231\\
30	93.5868619485441\\
29	93.5726103562873\\
28	93.648724635359\\
27	93.6358646045122\\
26	93.6242392836925\\
25	93.6484288678273\\
24	93.615677956726\\
23	93.5875498052752\\
22	93.6035001870415\\
21	93.643430133813\\
20	93.6056828361559\\
19	93.5751846123288\\
18	93.3128882324155\\
17	93.2194364076747\\
16	93.3085598629027\\
15	93.2823757455862\\
14	93.1994089672597\\
13	92.9463438136983\\
12	92.8648344103279\\
11	92.788231145908\\
10	92.551249667823\\
9	91.4956167044765\\
8	89.6877937912635\\
7	89.1622268240893\\
6	87.692473849543\\
5	90.329727762724\\
4	90.4821997775267\\
3	87.9911110398045\\
2	87.6608694738352\\
1	86.9197052167653\\
}--cycle;
\addlegendentry{LD}

\addplot[area legend, draw=red, fill=white!80!red]
table[row sep=crcr] {%
x	y\\
1	74.4447705891167\\
2	86.9411249441942\\
3	93.9842414927358\\
4	95.8049103891504\\
5	97.6939455115177\\
6	98.4325812469221\\
7	98.4022972565978\\
8	98.4480622762415\\
9	98.5172890927368\\
10	98.4616861332457\\
11	98.496027360265\\
12	98.5089165921868\\
13	98.482632884094\\
14	98.4217495740064\\
15	98.373242643204\\
16	98.442830142854\\
17	98.4561393724132\\
18	98.5142904263309\\
19	98.4633500890582\\
20	98.3887756949583\\
21	98.5132032238429\\
22	98.5298057508957\\
23	98.5625305227293\\
24	98.5392782021753\\
25	98.4775396038177\\
26	98.5686810192877\\
27	98.5138211470985\\
28	98.5583859600022\\
29	98.5729227731785\\
30	98.6378950496178\\
31	98.5843252944403\\
32	98.5439060327274\\
33	98.5559578345138\\
34	98.5685137457548\\
35	98.5558693911641\\
36	98.4954645368527\\
37	98.555919482958\\
38	98.4965042397365\\
39	98.544092736638\\
40	98.5631448886904\\
41	98.5326257932399\\
42	98.5028665297963\\
43	98.5306807080476\\
44	98.510992783209\\
45	98.546173798151\\
46	98.5431466695506\\
47	98.6196845965957\\
48	98.4999622230123\\
49	98.5492255897569\\
50	98.5538812942216\\
50	98.8503920431401\\
49	98.8840490173631\\
48	98.8463085748327\\
47	98.857094543517\\
46	98.846628661287\\
45	98.8290930127422\\
44	98.864263514264\\
43	98.8445755894255\\
42	98.8868825174907\\
41	98.8571390241775\\
40	98.8411442188016\\
39	98.8747048907985\\
38	98.8207810584806\\
37	98.8048756049619\\
36	98.7928037214757\\
35	98.775908656867\\
34	98.7777517953802\\
33	98.7613116935729\\
32	98.7733687520695\\
31	98.8489545693899\\
30	98.8389156307556\\
29	98.8313663343136\\
28	98.8168966210215\\
27	98.8324601641668\\
26	98.7776055486877\\
25	98.7092268387116\\
24	98.7199940431049\\
23	98.7547337486473\\
22	98.7149579744401\\
21	98.7025592317345\\
20	98.681954128277\\
19	98.7378985898648\\
18	98.7739725752874\\
17	98.7741158329783\\
16	98.9034143714404\\
15	98.9004802982092\\
14	98.7939813413103\\
13	98.8345998470218\\
12	98.8083266523493\\
11	98.6907390822643\\
10	98.7250908227038\\
9	98.6839963831571\\
8	98.6807121401915\\
7	98.63946809385\\
6	98.5946808402915\\
5	98.3037163038188\\
4	96.2629243009836\\
3	94.9802786051126\\
2	89.2044659507634\\
1	76.9471931101458\\
}--cycle;
\addlegendentry{kNN}

\addplot[mark=none, black, very thick, dotted, domain=0:50] {98.90};
\addlegendentry{98.9~\%}

\addplot [color=black, forget plot, dashdotted]
  table[row sep=crcr]{%
1	79.1472012881042\\
2	88.8920326935829\\
3	92.7566450729776\\
4	94.308202491628\\
5	95.8961652562475\\
6	96.5777345208942\\
7	96.6502140400945\\
8	96.8241796049635\\
9	96.8822057996757\\
10	96.8604561615345\\
11	96.7589490890253\\
14	96.5051656376218\\
15	96.4834028577053\\
16	96.5341642790251\\
17	96.505170894332\\
18	96.5196925560517\\
19	96.4036717068879\\
20	96.4254318584494\\
21	96.4689153646015\\
22	96.367397778672\\
23	96.3746651804195\\
25	96.3456639106612\\
26	96.4326782333564\\
27	96.389178957074\\
28	96.3601645455403\\
29	96.3964095618506\\
31	96.3964069334955\\
33	96.2731344527147\\
35	96.2368946931146\\
36	96.2949077460515\\
38	96.1716352652707\\
40	96.1353876206054\\
42	96.1136274690439\\
43	96.0338726630309\\
44	96.0121203965347\\
45	96.0556301862374\\
46	96.019382541572\\
47	95.9541388838585\\
48	95.9106290941558\\
49	95.9033827192488\\
50	95.8816251960424\\
};
\addplot [color=black!80!green, forget plot, dotted]
  table[row sep=crcr]{%
1	86.8764575871543\\
2	91.3790532191962\\
3	91.7705572191583\\
4	92.9161941103294\\
5	92.8436462538975\\
6	92.8001522343252\\
7	93.1481674714252\\
8	92.4883951553108\\
9	92.8944076752173\\
10	92.7783579141479\\
11	92.2128094986649\\
12	92.0823326966581\\
13	91.8430709069744\\
14	91.466051771761\\
16	91.212284090488\\
17	91.045483421519\\
18	90.951233237337\\
19	90.6250044353492\\
20	90.6540214752379\\
21	89.9507209709354\\
22	89.6244106899406\\
23	89.1893732450802\\
24	88.950135110592\\
25	88.5585995703691\\
26	88.2541151498123\\
27	87.8771038996641\\
29	87.3550442470433\\
30	87.0505335429359\\
31	86.9055245657892\\
32	86.6082313246831\\
33	86.4124766963469\\
35	85.9701928765795\\
36	85.6584279125\\
37	85.48442029395\\
38	84.9478757765804\\
39	84.6433361605672\\
40	84.3025620416641\\
41	84.0415401004189\\
42	83.6862653466367\\
43	83.2584795333934\\
44	82.8959400062187\\
45	82.5479142556985\\
46	82.3231294456983\\
47	82.0331009779846\\
48	81.8155809413769\\
49	81.5110860073999\\
50	81.1920431281525\\
};
\addplot [color=blue, forget plot, dashed]
  table[row sep=crcr]{%
1	86.3109118000263\\
2	86.8039912097293\\
3	87.3477558184553\\
4	89.9507341127107\\
5	90.0232136319109\\
6	87.2245069928701\\
7	88.7181275808804\\
8	89.3489143973668\\
9	91.2630717953584\\
10	92.1621242996419\\
12	92.4738971487867\\
14	92.8944497288983\\
15	92.9162046237496\\
16	93.1337036335169\\
17	93.0902043572345\\
18	93.1336904917416\\
19	93.2932132455428\\
20	93.3802301965931\\
21	93.380227568238\\
22	93.4019798347343\\
23	93.380227568238\\
24	93.3947255747621\\
25	93.452738627699\\
26	93.430996874623\\
27	93.4599981443813\\
28	93.4599928876712\\
29	93.4164936113888\\
30	93.4309916179129\\
31	93.4889994141397\\
32	93.4672550327086\\
33	93.4092367230615\\
34	93.372978564976\\
35	93.3947308314722\\
36	93.4309784761376\\
37	93.4310021313331\\
38	93.4599981443813\\
39	93.4527438844091\\
40	93.4672445192883\\
42	93.4454869960819\\
43	93.496258930822\\
44	93.4672550327086\\
45	93.5904828314533\\
48	93.6847566708309\\
49	93.6412521378383\\
50	93.692010930803\\
};
\addplot [color=red, forget plot]
  table[row sep=crcr]{%
1	75.6959818496313\\
2	88.0727954474788\\
3	94.4822600489242\\
4	96.033917345067\\
5	97.9988309076682\\
6	98.5136310436068\\
7	98.5208826752239\\
9	98.600642737947\\
11	98.5933832212646\\
12	98.658621622268\\
13	98.6586163655579\\
14	98.6078654576584\\
16	98.6731222571472\\
17	98.6151276026958\\
18	98.6441315008091\\
19	98.6006243394615\\
20	98.5353649116176\\
21	98.6078812277887\\
22	98.6223818626679\\
23	98.6586321356883\\
25	98.5933832212646\\
26	98.6731432839877\\
27	98.6731406556327\\
29	98.7021445537461\\
30	98.7384053401867\\
31	98.7166399319151\\
32	98.6586373923984\\
33	98.6586347640434\\
34	98.6731327705675\\
35	98.6658890240155\\
36	98.6441341291642\\
37	98.6803975439599\\
38	98.6586426491086\\
39	98.7093988137182\\
43	98.6876281487365\\
46	98.6948876654188\\
47	98.7383895700563\\
48	98.6731353989225\\
49	98.71663730356\\
50	98.7021366686809\\
};
\addplot [color=black, forget plot]
  table[row sep=crcr]{%
1	77.7010097206438\\
2	87.8775482111069\\
3	92.2103397107741\\
4	93.5996636347119\\
5	95.5723711595146\\
6	96.1522864049922\\
7	96.3364735764658\\
8	96.4026757885762\\
9	96.5916045548038\\
10	96.5495654597671\\
11	96.4438520034418\\
12	96.1768040709561\\
13	96.142553128961\\
14	96.0336365449824\\
15	96.0365542977309\\
16	96.1400672447157\\
17	96.1667957283196\\
18	96.3031107022384\\
19	96.0503737102083\\
20	96.0862726689533\\
21	95.9830101979916\\
22	95.8210409960337\\
23	95.9309479190219\\
24	95.9042609678968\\
25	95.9128817148687\\
26	96.0925043863021\\
27	96.0568679837545\\
28	95.9411127176965\\
29	95.9618906313519\\
31	95.8957948136432\\
32	95.8479983214007\\
33	95.7712311690668\\
34	95.7263252244064\\
35	95.8202595205501\\
36	95.901069674323\\
37	95.8008395111219\\
38	95.7207125757569\\
39	95.6742748972227\\
40	95.6963219583253\\
41	95.6892030741858\\
42	95.6290713975559\\
43	95.5221362562465\\
44	95.5033374139915\\
45	95.6114703130698\\
46	95.6067106267384\\
47	95.5610121759469\\
48	95.4448792019834\\
49	95.4337424037886\\
50	95.3869774272053\\
nan	nan\\
50	96.3762729648795\\
48	96.3763789863283\\
47	96.3472655917702\\
46	96.4320544564056\\
45	96.4997900594049\\
43	96.5456090698154\\
42	96.5981835405318\\
41	96.5670689037907\\
40	96.5744532828855\\
39	96.6254911003261\\
38	96.6225579547844\\
36	96.68874581778\\
35	96.653529865679\\
34	96.7764444047406\\
33	96.7750377363626\\
32	96.8142966899025\\
31	96.8970190533478\\
29	96.8309284923493\\
27	96.7214899303935\\
26	96.7728520804108\\
25	96.7784461064536\\
24	96.8160576097636\\
23	96.818382441817\\
22	96.9137545613103\\
21	96.9548205312114\\
20	96.7645910479455\\
19	96.7569697035676\\
18	96.7362744098649\\
17	96.8435460603443\\
16	96.9282613133346\\
15	96.9302514176798\\
14	96.9766947302612\\
13	97.0418067916731\\
12	97.167065461704\\
11	97.0740461746087\\
10	97.1713468633019\\
9	97.1728070445477\\
8	97.2456834213507\\
7	96.9639545037232\\
6	97.0031826367962\\
5	96.2199593529805\\
4	95.0167413485442\\
3	93.302950435181\\
2	89.906517176059\\
1	80.5933928555647\\
};
\addplot [color=black!80!green, forget plot]
  table[row sep=crcr]{%
1	86.369919588866\\
2	90.7576444751129\\
3	91.5219293884986\\
4	92.5977180024671\\
5	92.4120933850905\\
6	92.4329083484411\\
7	92.7057126208932\\
8	92.0867533610352\\
9	92.5575718755273\\
10	92.304189817332\\
11	91.6230122710067\\
12	91.6645284167859\\
13	91.4104010118256\\
14	91.0347494561662\\
15	91.0354363421779\\
16	90.8176280670333\\
17	90.465643245502\\
18	90.4812611511186\\
19	90.2160628141167\\
20	90.2392797413932\\
21	89.4707839338981\\
22	89.0879932675199\\
23	88.7697637035184\\
24	88.5410719051033\\
25	88.1681648447666\\
26	87.8278326006238\\
27	87.3832479253271\\
28	87.1036580992142\\
29	86.9046544143713\\
30	86.5835036326836\\
32	86.2255159327648\\
33	85.9904261592187\\
34	85.6956838485922\\
35	85.5062952278591\\
36	85.1751373253508\\
37	84.9328127380998\\
38	84.384654814503\\
39	84.2254280660142\\
40	83.8419178880646\\
41	83.597317442393\\
42	83.2187765244889\\
43	82.7541979214219\\
44	82.3547974319281\\
45	81.9749346939399\\
46	81.7144740959679\\
47	81.4814752434394\\
48	81.3250440531357\\
49	80.8562912093323\\
50	80.4769781014053\\
nan	nan\\
50	81.9071081548998\\
49	82.1658808054674\\
48	82.3061178296182\\
47	82.5847267125298\\
46	82.9317847954287\\
45	83.120893817457\\
43	83.7627611453648\\
42	84.1537541687845\\
41	84.4857627584448\\
40	84.7632061952637\\
39	85.0612442551203\\
38	85.5110967386577\\
37	86.0360278498001\\
36	86.1417184996492\\
35	86.4340905253\\
34	86.6797419777824\\
32	86.9909467166014\\
31	87.4063971462129\\
30	87.5175634531882\\
29	87.8054340797153\\
28	88.128484790783\\
27	88.3709598740012\\
26	88.6803976990008\\
25	88.9490342959715\\
24	89.3591983160807\\
23	89.6089827866419\\
22	90.1608281123612\\
21	90.4306580079727\\
20	91.0687632090825\\
19	91.0339460565817\\
18	91.4212053235555\\
17	91.625323597536\\
16	91.6069401139426\\
15	91.6356715436495\\
14	91.8973540873558\\
13	92.2757408021232\\
12	92.5001369765303\\
11	92.8026067263232\\
10	93.2525260109638\\
9	93.2312434749073\\
8	92.8900369495864\\
7	93.5906223219572\\
6	93.1673961202093\\
5	93.2751991227046\\
4	93.2346702181917\\
3	92.019185049818\\
2	92.0004619632794\\
1	87.3829955854425\\
};
\addplot [color=blue, forget plot]
  table[row sep=crcr]{%
1	85.7021183832874\\
2	85.9471129456235\\
3	86.7044005971061\\
4	89.4192684478947\\
5	89.7166995010979\\
6	86.7565401361972\\
7	88.2740283376715\\
8	89.0100350034701\\
9	91.0305268862402\\
10	91.7729989314609\\
11	91.8404913605125\\
12	92.0829598872456\\
13	92.421971523726\\
14	92.589490490537\\
15	92.5500335019131\\
16	92.9588474041311\\
18	92.9544927510677\\
19	93.0112418787567\\
20	93.1547775570303\\
21	93.117025002663\\
22	93.2004594824271\\
23	93.1729053312008\\
24	93.1737731927983\\
25	93.2570483875708\\
26	93.2377544655535\\
27	93.2841316842504\\
29	93.2603768664902\\
31	93.2936048621562\\
32	93.2313144250924\\
33	93.2297001701083\\
34	93.2125077707941\\
35	93.2405597053028\\
36	93.2995049233876\\
37	93.1895688939404\\
38	93.2680527112941\\
42	93.3305197700892\\
44	93.2811516778258\\
45	93.4716221097034\\
46	93.4963921220288\\
47	93.5508565622345\\
48	93.5669893479133\\
49	93.4784728895994\\
50	93.5478740326192\\
nan	nan\\
50	93.8361478289869\\
49	93.8040313860772\\
48	93.8025239937484\\
47	93.7606489832005\\
46	93.7425813371046\\
45	93.7093435532032\\
44	93.6533583875913\\
43	93.6878648592372\\
42	93.5604542220746\\
41	93.6031684638887\\
40	93.6333108659388\\
39	93.62308517423\\
37	93.6724353687258\\
36	93.5624520288875\\
34	93.5334493591579\\
33	93.5887732760147\\
32	93.7031956403248\\
31	93.6843939661231\\
30	93.5868619485441\\
29	93.5726103562873\\
28	93.648724635359\\
26	93.6242392836925\\
25	93.6484288678273\\
23	93.5875498052752\\
22	93.6035001870415\\
21	93.643430133813\\
19	93.5751846123288\\
18	93.3128882324155\\
17	93.2194364076747\\
16	93.3085598629027\\
15	93.2823757455862\\
14	93.1994089672597\\
13	92.9463438136983\\
11	92.788231145908\\
10	92.551249667823\\
9	91.4956167044765\\
8	89.6877937912635\\
7	89.1622268240893\\
6	87.692473849543\\
5	90.329727762724\\
4	90.4821997775267\\
3	87.9911110398045\\
2	87.6608694738352\\
1	86.9197052167653\\
};
\addplot [color=red, forget plot]
  table[row sep=crcr]{%
1	74.4447705891167\\
2	86.9411249441942\\
3	93.9842414927358\\
4	95.8049103891504\\
5	97.6939455115177\\
6	98.4325812469221\\
7	98.4022972565978\\
8	98.4480622762415\\
9	98.5172890927368\\
10	98.4616861332457\\
11	98.496027360265\\
12	98.5089165921868\\
13	98.482632884094\\
14	98.4217495740064\\
15	98.373242643204\\
16	98.442830142854\\
17	98.4561393724132\\
18	98.5142904263309\\
19	98.4633500890582\\
20	98.3887756949583\\
21	98.5132032238429\\
22	98.5298057508957\\
23	98.5625305227293\\
24	98.5392782021753\\
25	98.4775396038177\\
26	98.5686810192877\\
27	98.5138211470985\\
28	98.5583859600022\\
29	98.5729227731785\\
30	98.6378950496178\\
31	98.5843252944403\\
32	98.5439060327274\\
34	98.5685137457548\\
35	98.5558693911641\\
36	98.4954645368527\\
37	98.555919482958\\
38	98.4965042397365\\
39	98.544092736638\\
40	98.5631448886904\\
42	98.5028665297963\\
43	98.5306807080476\\
44	98.510992783209\\
45	98.546173798151\\
46	98.5431466695506\\
47	98.6196845965957\\
48	98.4999622230123\\
49	98.5492255897569\\
50	98.5538812942216\\
nan	nan\\
50	98.8503920431401\\
49	98.8840490173631\\
48	98.8463085748327\\
47	98.857094543517\\
45	98.8290930127422\\
44	98.864263514264\\
43	98.8445755894255\\
42	98.8868825174907\\
41	98.8571390241775\\
40	98.8411442188016\\
39	98.8747048907985\\
38	98.8207810584806\\
35	98.775908656867\\
34	98.7777517953802\\
33	98.7613116935729\\
32	98.7733687520695\\
31	98.8489545693899\\
28	98.8168966210215\\
27	98.8324601641668\\
26	98.7776055486877\\
25	98.7092268387116\\
24	98.7199940431049\\
23	98.7547337486473\\
22	98.7149579744401\\
20	98.681954128277\\
19	98.7378985898648\\
18	98.7739725752874\\
17	98.7741158329783\\
16	98.9034143714404\\
15	98.9004802982092\\
14	98.7939813413103\\
13	98.8345998470218\\
12	98.8083266523493\\
11	98.6907390822643\\
10	98.7250908227038\\
9	98.6839963831571\\
8	98.6807121401915\\
6	98.5946808402915\\
5	98.3037163038188\\
4	96.2629243009836\\
3	94.9802786051126\\
2	89.2044659507634\\
1	76.9471931101458\\
};
\end{axis}
\end{tikzpicture}%

%% file: Figures/balancing1_norm0_FFTtransformed_50rSVDmodes_recall_fingers_class.tex
\definecolor{mycolor1}{rgb}{0.64, 0.76, 0.68}%
\begin{tikzpicture}

\begin{axis}[%
width=7cm,
height=5cm,
scale only axis,
unbounded coords=jump,
xmin=0,
xmax=50,
xlabel style={font=\color{white!15!black}},
xlabel={$r$},
ymin=0,
ymax=100,
ylabel style={font=\color{white!15!black}},
axis background/.style={fill=white},
title style={font=\bfseries},
legend columns = 2,
legend pos = south east,
legend style={legend cell align=left, align=left, draw=white!15!black, font=\footnotesize},
legend entries={Tree,
                NB,
                LD,
                kNN,
                97.9~\%},
]
\addlegendimage{legend image code/.code={
\draw [draw=none, fill=white!80!black] (0cm,-0.15cm) rectangle (0.6cm,0.15cm);
\draw[dashdotted] (0cm,0cm) -- (0.6cm,0cm);
\draw[-] (0cm,0.15cm) -- (0.6cm,0.15cm);
\draw[-] (0cm,-0.15cm) -- (0.6cm,-0.15cm);
}
} 
\addlegendimage{legend image code/.code={
\draw [draw=none, fill=mycolor1] (0cm,-0.15cm) rectangle (0.6cm,0.15cm);
\draw[dotted, black!80!green] (0cm,0cm) -- (0.6cm,0cm);
\draw[-, black!80!green] (0cm,0.15cm) -- (0.6cm,0.15cm);
\draw[-, black!80!green] (0cm,-0.15cm) -- (0.6cm,-0.15cm);
}
}
\addlegendimage{legend image code/.code={
\draw [draw=none, fill=white!80!blue] (0cm,-0.15cm) rectangle (0.6cm,0.15cm);
\draw[dashed, blue] (0cm,0cm) -- (0.6cm,0cm);
\draw[-, blue] (0cm,0.15cm) -- (0.6cm,0.15cm);
\draw[-, blue] (0cm,-0.15cm) -- (0.6cm,-0.15cm);
}
}
\addlegendimage{legend image code/.code={
\draw [draw=none, fill=white!80!red] (0cm,-0.15cm) rectangle (0.6cm,0.15cm);
\draw[-, red] (0cm,0cm) -- (0.6cm,0cm);
\draw[-, red] (0cm,0.15cm) -- (0.6cm,0.15cm);
\draw[-, red] (0cm,-0.15cm) -- (0.6cm,-0.15cm);
}
}
\addlegendimage{mark=none, black, very thick, dotted}

\addplot[area legend, draw=black, fill=white!80!black, dashdotted]
table[row sep=crcr] {%
x	y\\
1	60.6394901114679\\
2	73.6843145996363\\
3	82.5376647269747\\
4	84.501078287243\\
5	88.331593085875\\
6	91.3593328832876\\
7	91.679361272792\\
8	90.8991681270127\\
9	91.2371041888655\\
10	91.5085551011823\\
11	91.4665932373601\\
12	91.3098205974916\\
13	91.2776943500656\\
14	90.6660174602327\\
15	90.6266206431551\\
16	90.4115668797142\\
17	90.7771869272822\\
18	90.1784339779537\\
19	89.8211580779307\\
20	89.7927513876222\\
21	89.7183150901476\\
22	90.193740412626\\
23	90.4027875951781\\
24	90.5965669466127\\
25	90.7119703403505\\
26	90.6595383193188\\
27	90.6611772123206\\
28	90.6566272603544\\
29	90.2735297668438\\
30	90.3703528007401\\
31	90.2601088559047\\
32	90.3007833481865\\
33	90.388017635051\\
34	90.1648451733456\\
35	90.1615538941955\\
36	90.1714746820352\\
37	89.785889745309\\
38	89.9471800678897\\
39	90.0168334545488\\
40	90.0478162268027\\
41	90.0336180758169\\
42	89.9118132456715\\
43	89.9118132456715\\
44	89.8389375049728\\
45	89.8857666310331\\
46	89.8997700904282\\
47	89.8997700904282\\
48	89.8997700904282\\
49	89.8949810241911\\
50	90.0205346936556\\
50	93.0439814353766\\
49	93.0620082231208\\
48	93.1109825977439\\
47	93.1109825977439\\
46	93.1109825977439\\
45	93.1249860571389\\
44	93.0105248606186\\
43	92.9376491199199\\
42	92.9376491199199\\
41	92.8696077306347\\
40	92.9091730205091\\
39	93.1552095562039\\
38	93.0098091794221\\
37	92.8485188568415\\
36	92.8930414469971\\
35	92.9029622348368\\
34	93.0071978374071\\
33	92.9453156982823\\
32	92.9250231034264\\
31	92.9656975957082\\
30	92.9092170917331\\
29	92.9522766847692\\
28	92.94552327728\\
27	92.8334464435934\\
26	92.781321895735\\
25	92.9977070790044\\
24	92.8980567093013\\
23	92.876782297295\\
22	92.9783025981267\\
21	93.184910716304\\
20	93.1642378596896\\
19	93.2971214919618\\
18	93.6387703231216\\
17	94.1152861909973\\
16	93.7819815073825\\
15	93.9970352708234\\
14	93.8501115720254\\
13	93.7223056499344\\
12	93.4213621982073\\
11	93.6946970852205\\
10	93.4376814579575\\
9	93.2252614025323\\
8	93.6707243461056\\
7	92.8905312003263\\
6	92.4578714177877\\
5	89.7866864840174\\
4	87.756986228886\\
3	83.9677116171113\\
2	75.3479434648799\\
1	62.7476066627257\\
}--cycle;
\addlegendentry{Tree}

\addplot[area legend, draw=black!80!green, fill=mycolor1, dotted]
table[row sep=crcr] {%
x	y\\
1	74.9106464903133\\
2	77.9618018630925\\
3	80.6189999607151\\
4	85.0218648054251\\
5	85.1906209059084\\
6	85.6703310288509\\
7	83.9962009976001\\
8	83.6871546066834\\
9	82.9482134475675\\
10	83.559328823007\\
11	83.3306794656177\\
12	79.7871213560969\\
13	79.9788594268729\\
14	79.1242997034261\\
15	75.0771523529513\\
16	71.7895457638147\\
17	69.4985166596695\\
18	67.8885377877198\\
19	66.4457981458301\\
20	65.7204675045685\\
21	65.71867074286\\
22	64.2965815680708\\
23	64.2675339677986\\
24	63.7199487852759\\
25	63.0866913259286\\
26	62.455209612548\\
27	61.8546649102939\\
28	60.917215701111\\
29	60.6610400394442\\
30	60.600232702857\\
31	60.3053053131916\\
32	60.1192455751207\\
33	60.0600190007207\\
34	60.0683431388895\\
35	60.2681452172846\\
36	60.1528654703758\\
37	60.0676581168101\\
38	60.4543361534074\\
39	60.7489892926786\\
40	61.3044377135012\\
41	61.4256795669793\\
42	61.7209532852664\\
43	62.3492178419719\\
44	62.0547493913804\\
45	62.3375337307556\\
46	62.8913313070445\\
47	62.789838462322\\
48	62.9411875859359\\
49	63.3323533450385\\
50	63.7947949516875\\
50	76.4740222526136\\
49	76.2375391280797\\
48	75.8760167151393\\
47	75.8660755161726\\
46	74.6893138542458\\
45	73.0925737961261\\
44	72.1387989957164\\
43	71.0378789322216\\
42	70.6984015534433\\
41	70.1872236588272\\
40	69.6095407811224\\
39	69.2510107073214\\
38	68.90050255627\\
37	68.4269655391038\\
36	68.3955216263984\\
35	68.3340053203498\\
34	68.1574633127234\\
33	67.7356799240105\\
32	67.5689264678901\\
31	67.3291032889589\\
30	67.6255737487559\\
29	67.9948739390504\\
28	68.1688058042653\\
27	68.0915716488459\\
26	68.8888764089573\\
25	69.3864269536413\\
24	69.9359651932187\\
23	70.9475197956423\\
22	71.7786872491336\\
21	71.7006840958497\\
20	72.2902851836035\\
19	72.4789330369656\\
18	72.3802794165812\\
17	75.6090102220509\\
16	80.1997015480132\\
15	81.8045680771562\\
14	83.4563454578642\\
13	83.6232911107615\\
12	85.2666420847634\\
11	85.6478151580382\\
10	84.935294832907\\
9	84.686195154583\\
8	85.5063937804134\\
7	85.3048742712171\\
6	88.4156904765255\\
5	86.3685188790378\\
4	86.1071674526394\\
3	82.1229355231559\\
2	79.8338970616387\\
1	77.6162352301168\\
}--cycle;
\addlegendentry{NB}

\addplot[area legend, draw=blue, fill=white!80!blue, dashed]
table[row sep=crcr] {%
x	y\\
1	73.1108958663089\\
2	73.7123709893554\\
3	73.5657483827872\\
4	74.9383970592015\\
5	78.2894385068773\\
6	80.0714898111462\\
7	79.9555742650288\\
8	82.5498703493335\\
9	83.4046979298167\\
10	85.4089733744097\\
11	85.2496475829188\\
12	85.659213351396\\
13	85.5888813690386\\
14	85.3908267082529\\
15	85.86055974787\\
16	85.8224675349012\\
17	86.0215053763441\\
18	86.0856055745177\\
19	86.4532367672856\\
20	86.4918288170955\\
21	86.4400798575029\\
22	86.3911528638702\\
23	86.9382183920572\\
24	86.8460629099331\\
25	86.5100393194453\\
26	86.8016203178978\\
27	86.6855542251139\\
28	86.7427769052005\\
29	87.0084873320575\\
30	86.8895994205638\\
31	86.8234220664909\\
32	86.8687966790908\\
33	86.8687966790908\\
34	86.9853542203012\\
35	86.8772514004421\\
36	87.073842040524\\
37	87.0068587086743\\
38	86.9873291442607\\
39	86.8364815225045\\
40	86.9080093636449\\
41	86.8772514004421\\
42	86.8565204611467\\
43	86.9108454138897\\
44	86.8868354682318\\
45	86.7479229866752\\
46	86.7479229866752\\
47	86.9278736309491\\
48	86.9759166300012\\
49	86.9004045245969\\
50	86.932967390544\\
50	88.6584304589183\\
49	88.5834664431451\\
48	88.6154812194611\\
47	88.5022338959326\\
46	88.5208942176259\\
45	88.5208942176259\\
44	88.4357451769295\\
43	88.519262112992\\
42	88.573587065735\\
41	88.6603830081601\\
40	88.7371519266776\\
39	88.8624432086783\\
38	88.8728859095027\\
37	88.6920660225085\\
36	88.6250826906588\\
35	88.6603830081601\\
34	88.6598070700214\\
33	88.6150742886512\\
32	88.6150742886512\\
31	88.6604489012511\\
30	88.4867446654577\\
29	88.5291470765447\\
28	88.5798037399607\\
27	88.6370264200474\\
26	88.7897775315646\\
25	88.5974875622751\\
24	88.5840446169486\\
23	88.6531794574052\\
22	88.2862664909685\\
21	88.2373394973359\\
20	88.1855905377432\\
19	88.0091288241123\\
18	87.8928890491382\\
17	87.9032258064516\\
16	87.994736766174\\
15	87.9566445532053\\
14	87.2973453347579\\
13	87.314344437413\\
12	87.1364855733352\\
11	87.1159438149306\\
10	86.8490911417193\\
9	86.9178827153446\\
8	84.2243231990536\\
7	81.7648558424981\\
6	82.1328112641226\\
5	80.3664754716173\\
4	76.190635198863\\
3	75.3589828000085\\
2	75.0510698708596\\
1	75.2224374670245\\
}--cycle;
\addlegendentry{LD}

\addplot[area legend, draw=red, fill=white!80!red]
table[row sep=crcr] {%
x	y\\
1	59.4313236700477\\
2	75.0109417262328\\
3	86.8253031215317\\
4	89.8440621623035\\
5	93.6369948829465\\
6	95.8709980603081\\
7	96.2851521230868\\
8	96.4960958317243\\
9	96.5225039609428\\
10	96.5618333549468\\
11	96.6717308398882\\
12	96.6243501177149\\
13	96.6196806128024\\
14	96.5678045825359\\
15	96.588503483839\\
16	96.6004234086943\\
17	96.5678045825359\\
18	96.5277983299843\\
19	96.5625105668185\\
20	96.681944174055\\
21	96.8324624469304\\
22	96.6835706163744\\
23	96.8481868625291\\
24	96.802629593937\\
25	96.7445529199869\\
26	96.7741935483871\\
27	96.7022131846865\\
28	96.6440472128651\\
29	96.715970372995\\
30	96.7229120859895\\
31	96.6622069321348\\
32	96.5710320640966\\
33	96.7793619270438\\
34	96.6550318080187\\
35	96.6468698560933\\
36	96.7793619270438\\
37	96.7022131846865\\
38	96.7678875664298\\
39	96.9557002504845\\
40	96.8557999988234\\
41	96.8452842141453\\
42	96.750307961828\\
43	96.8059535031934\\
44	96.7826607611551\\
45	96.696919168969\\
46	96.696919168969\\
47	96.5763258125001\\
48	96.6255978319318\\
49	96.5781820054381\\
50	96.7054406662471\\
50	97.7569249251508\\
49	97.8304201450996\\
48	97.8367677594661\\
47	97.7785128971774\\
46	97.7654464224288\\
45	97.7654464224288\\
44	97.8409951528234\\
43	97.7639389699249\\
42	97.7120576295699\\
41	97.724608258973\\
40	97.6065655925744\\
39	97.7754825452144\\
38	97.7482414658282\\
37	97.652625524991\\
36	97.6830036643541\\
35	97.7617322944443\\
34	97.7535703425189\\
33	97.6830036643541\\
32	97.6225163230002\\
31	97.6388683366824\\
30	97.7394535054083\\
29	97.6926317775426\\
28	97.6570280559521\\
27	97.652625524991\\
26	97.5806451612903\\
25	97.6640492305507\\
24	97.4984456748802\\
23	97.5066518471483\\
22	97.6712680933031\\
21	97.6299031444675\\
20	97.565367653902\\
19	97.6310378202783\\
18	97.5044597345318\\
17	97.5182169228404\\
16	97.5393615375423\\
15	97.6050449032578\\
14	97.5182169228404\\
13	97.4125774517138\\
12	97.6229617102421\\
11	97.7368713106494\\
10	97.900532236451\\
9	97.9398616304551\\
8	97.7512159962327\\
7	97.4782887371283\\
6	97.3010449504447\\
5	95.1264459772686\\
4	91.6613141817826\\
3	89.1424388139521\\
2	76.3869077361328\\
1	60.998783856834\\
}--cycle;
\addlegendentry{kNN}

\addplot[mark=none, black, very thick, dotted, domain=0:50] {97.94};
\addlegendentry{97.9~\%}

\addplot [color=black, forget plot, dashdotted]
  table[row sep=crcr]{%
1	61.6935483870968\\
2	74.5161290322581\\
3	83.252688172043\\
4	86.1290322580645\\
5	89.0591397849462\\
6	91.9086021505376\\
7	92.2849462365591\\
8	92.2849462365591\\
9	92.2311827956989\\
10	92.4731182795699\\
11	92.5806451612903\\
12	92.3655913978494\\
13	92.5\\
14	92.258064516129\\
15	92.3118279569892\\
16	92.0967741935484\\
17	92.4462365591398\\
18	91.9086021505376\\
19	91.5591397849462\\
20	91.4784946236559\\
21	91.4516129032258\\
22	91.5860215053763\\
23	91.6397849462366\\
25	91.8548387096774\\
26	91.7204301075269\\
27	91.747311827957\\
28	91.8010752688172\\
29	91.6129032258065\\
30	91.6397849462366\\
31	91.6129032258064\\
32	91.6129032258064\\
33	91.6666666666667\\
34	91.5860215053763\\
35	91.5322580645161\\
36	91.5322580645161\\
37	91.3172043010753\\
38	91.4784946236559\\
39	91.5860215053763\\
40	91.4784946236559\\
42	91.4247311827957\\
44	91.4247311827957\\
45	91.505376344086\\
48	91.505376344086\\
49	91.4784946236559\\
50	91.5322580645161\\
};
\addplot [color=black!80!green, forget plot, dotted]
  table[row sep=crcr]{%
1	76.263440860215\\
2	78.8978494623656\\
3	81.3709677419355\\
4	85.5645161290323\\
5	85.7795698924731\\
6	87.0430107526882\\
7	84.6505376344086\\
8	84.5967741935484\\
9	83.8172043010753\\
10	84.247311827957\\
11	84.489247311828\\
12	82.5268817204301\\
13	81.8010752688172\\
14	81.2903225806452\\
15	78.4408602150538\\
16	75.994623655914\\
17	72.5537634408602\\
18	70.1344086021505\\
19	69.4623655913978\\
20	69.005376344086\\
21	68.7096774193548\\
22	68.0376344086022\\
23	67.6075268817204\\
24	66.8279569892473\\
25	66.2365591397849\\
26	65.6720430107527\\
27	64.9731182795699\\
28	64.5430107526882\\
30	64.1129032258064\\
31	63.8172043010753\\
32	63.8440860215054\\
33	63.8978494623656\\
34	64.1129032258064\\
35	64.3010752688172\\
37	64.247311827957\\
38	64.6774193548387\\
39	65\\
40	65.4569892473118\\
41	65.8064516129032\\
42	66.2096774193548\\
43	66.6935483870968\\
44	67.0967741935484\\
45	67.7150537634409\\
46	68.7903225806452\\
47	69.3279569892473\\
48	69.4086021505376\\
49	69.7849462365591\\
50	70.1344086021505\\
};
\addplot [color=blue, forget plot, dashed]
  table[row sep=crcr]{%
1	74.1666666666667\\
2	74.3817204301075\\
3	74.4623655913979\\
4	75.5645161290323\\
5	79.3279569892473\\
6	81.1021505376344\\
7	80.8602150537634\\
8	83.3870967741936\\
9	85.1612903225806\\
10	86.1290322580645\\
11	86.1827956989247\\
12	86.3978494623656\\
13	86.4516129032258\\
14	86.3440860215054\\
15	86.9086021505376\\
16	86.9086021505376\\
17	86.9623655913978\\
18	86.989247311828\\
19	87.2311827956989\\
20	87.3387096774194\\
22	87.3387096774194\\
23	87.7956989247312\\
24	87.7150537634409\\
25	87.5537634408602\\
26	87.7956989247312\\
27	87.6612903225806\\
28	87.6612903225806\\
29	87.7688172043011\\
30	87.6881720430108\\
31	87.741935483871\\
33	87.741935483871\\
34	87.8225806451613\\
35	87.7688172043011\\
36	87.8494623655914\\
37	87.8494623655914\\
38	87.9301075268817\\
39	87.8494623655914\\
40	87.8225806451613\\
42	87.7150537634409\\
43	87.7150537634409\\
44	87.6612903225806\\
45	87.6344086021505\\
46	87.6344086021505\\
48	87.7956989247312\\
49	87.741935483871\\
50	87.7956989247312\\
};
\addplot [color=red, forget plot]
  table[row sep=crcr]{%
1	60.2150537634409\\
2	75.6989247311828\\
3	87.9838709677419\\
4	90.752688172043\\
5	94.3817204301075\\
6	96.5860215053764\\
7	96.8817204301075\\
8	97.1236559139785\\
9	97.2311827956989\\
10	97.2311827956989\\
11	97.2043010752688\\
12	97.1236559139785\\
13	97.0161290322581\\
14	97.0430107526882\\
15	97.0967741935484\\
18	97.0161290322581\\
19	97.0967741935484\\
20	97.1236559139785\\
21	97.2311827956989\\
22	97.1774193548387\\
23	97.1774193548387\\
24	97.1505376344086\\
25	97.2043010752688\\
26	97.1774193548387\\
27	97.1774193548387\\
28	97.1505376344086\\
29	97.2043010752688\\
30	97.2311827956989\\
31	97.1505376344086\\
32	97.0967741935484\\
33	97.2311827956989\\
34	97.2043010752688\\
35	97.2043010752688\\
36	97.2311827956989\\
37	97.1774193548387\\
38	97.258064516129\\
39	97.3655913978495\\
40	97.2311827956989\\
41	97.2849462365592\\
42	97.2311827956989\\
43	97.2849462365592\\
44	97.3118279569893\\
45	97.2311827956989\\
46	97.2311827956989\\
47	97.1774193548387\\
48	97.2311827956989\\
49	97.2043010752688\\
50	97.2311827956989\\
};
\addplot [color=black, forget plot]
  table[row sep=crcr]{%
1	60.6394901114679\\
2	73.6843145996363\\
3	82.5376647269747\\
4	84.501078287243\\
5	88.331593085875\\
6	91.3593328832876\\
7	91.679361272792\\
8	90.8991681270127\\
9	91.2371041888655\\
10	91.5085551011823\\
11	91.4665932373601\\
12	91.3098205974916\\
13	91.2776943500656\\
14	90.6660174602327\\
15	90.6266206431551\\
16	90.4115668797142\\
17	90.7771869272822\\
18	90.1784339779537\\
19	89.8211580779307\\
20	89.7927513876222\\
21	89.7183150901476\\
22	90.193740412626\\
23	90.4027875951781\\
24	90.5965669466127\\
25	90.7119703403505\\
26	90.6595383193188\\
28	90.6566272603544\\
29	90.2735297668438\\
30	90.3703528007401\\
31	90.2601088559047\\
32	90.3007833481865\\
33	90.388017635051\\
34	90.1648451733456\\
35	90.1615538941955\\
36	90.1714746820352\\
37	89.785889745309\\
38	89.9471800678897\\
39	90.0168334545488\\
40	90.0478162268027\\
41	90.0336180758169\\
42	89.9118132456715\\
43	89.9118132456715\\
44	89.8389375049728\\
45	89.8857666310331\\
46	89.8997700904282\\
49	89.8949810241911\\
50	90.0205346936556\\
nan	nan\\
50	93.0439814353766\\
49	93.0620082231208\\
48	93.1109825977439\\
46	93.1109825977439\\
45	93.1249860571389\\
44	93.0105248606186\\
43	92.9376491199199\\
42	92.9376491199199\\
41	92.8696077306347\\
40	92.9091730205091\\
39	93.1552095562039\\
38	93.0098091794221\\
37	92.8485188568415\\
36	92.8930414469971\\
35	92.9029622348368\\
34	93.0071978374071\\
33	92.9453156982823\\
32	92.9250231034264\\
31	92.9656975957082\\
30	92.9092170917331\\
29	92.9522766847692\\
28	92.94552327728\\
27	92.8334464435934\\
26	92.781321895735\\
25	92.9977070790044\\
24	92.8980567093013\\
23	92.876782297295\\
22	92.9783025981267\\
21	93.184910716304\\
20	93.1642378596896\\
19	93.2971214919618\\
18	93.6387703231216\\
17	94.1152861909973\\
16	93.7819815073825\\
15	93.9970352708234\\
14	93.8501115720254\\
13	93.7223056499344\\
12	93.4213621982073\\
11	93.6946970852205\\
10	93.4376814579575\\
9	93.2252614025323\\
8	93.6707243461056\\
7	92.8905312003263\\
6	92.4578714177877\\
5	89.7866864840174\\
4	87.756986228886\\
3	83.9677116171113\\
2	75.3479434648799\\
1	62.7476066627257\\
};
\addplot [color=black!80!green, forget plot]
  table[row sep=crcr]{%
1	74.9106464903133\\
2	77.9618018630925\\
3	80.6189999607151\\
4	85.0218648054251\\
5	85.1906209059084\\
6	85.6703310288509\\
7	83.9962009976001\\
8	83.6871546066834\\
9	82.9482134475675\\
10	83.559328823007\\
11	83.3306794656177\\
12	79.7871213560969\\
13	79.9788594268729\\
14	79.1242997034261\\
15	75.0771523529513\\
16	71.7895457638147\\
17	69.4985166596695\\
18	67.8885377877198\\
19	66.4457981458301\\
20	65.7204675045685\\
21	65.71867074286\\
22	64.2965815680708\\
23	64.2675339677986\\
24	63.7199487852759\\
26	62.455209612548\\
27	61.8546649102939\\
28	60.917215701111\\
29	60.6610400394442\\
30	60.600232702857\\
31	60.3053053131916\\
32	60.1192455751207\\
33	60.0600190007207\\
34	60.0683431388895\\
35	60.2681452172846\\
36	60.1528654703758\\
37	60.0676581168102\\
38	60.4543361534074\\
39	60.7489892926786\\
40	61.3044377135012\\
41	61.4256795669793\\
42	61.7209532852664\\
43	62.3492178419719\\
44	62.0547493913804\\
45	62.3375337307556\\
46	62.8913313070445\\
47	62.789838462322\\
48	62.9411875859359\\
49	63.3323533450385\\
50	63.7947949516875\\
nan	nan\\
50	76.4740222526136\\
49	76.2375391280797\\
48	75.8760167151393\\
47	75.8660755161726\\
46	74.6893138542458\\
45	73.0925737961261\\
44	72.1387989957164\\
43	71.0378789322216\\
42	70.6984015534433\\
41	70.1872236588272\\
40	69.6095407811224\\
38	68.90050255627\\
37	68.4269655391038\\
36	68.3955216263984\\
35	68.3340053203498\\
34	68.1574633127234\\
33	67.7356799240105\\
32	67.5689264678901\\
31	67.3291032889589\\
30	67.6255737487559\\
29	67.9948739390504\\
28	68.1688058042653\\
27	68.0915716488459\\
26	68.8888764089573\\
25	69.3864269536413\\
24	69.9359651932187\\
23	70.9475197956423\\
22	71.7786872491336\\
21	71.7006840958497\\
20	72.2902851836035\\
19	72.4789330369656\\
18	72.3802794165812\\
17	75.6090102220509\\
16	80.1997015480132\\
15	81.8045680771562\\
14	83.4563454578642\\
13	83.6232911107615\\
12	85.2666420847634\\
11	85.6478151580382\\
10	84.935294832907\\
9	84.686195154583\\
8	85.5063937804134\\
7	85.3048742712171\\
6	88.4156904765255\\
5	86.3685188790378\\
4	86.1071674526394\\
3	82.1229355231559\\
2	79.8338970616387\\
1	77.6162352301168\\
};
\addplot [color=blue, forget plot]
  table[row sep=crcr]{%
1	73.1108958663089\\
2	73.7123709893554\\
3	73.5657483827872\\
4	74.9383970592015\\
5	78.2894385068773\\
6	80.0714898111462\\
7	79.9555742650288\\
8	82.5498703493335\\
9	83.4046979298167\\
10	85.4089733744097\\
11	85.2496475829188\\
12	85.659213351396\\
13	85.5888813690386\\
14	85.3908267082529\\
15	85.86055974787\\
16	85.8224675349012\\
17	86.0215053763441\\
18	86.0856055745177\\
19	86.4532367672856\\
20	86.4918288170955\\
22	86.3911528638702\\
23	86.9382183920572\\
24	86.8460629099331\\
25	86.5100393194453\\
26	86.8016203178978\\
27	86.6855542251139\\
28	86.7427769052005\\
29	87.0084873320575\\
30	86.8895994205638\\
31	86.8234220664909\\
32	86.8687966790908\\
33	86.8687966790908\\
34	86.9853542203012\\
35	86.8772514004421\\
36	87.073842040524\\
37	87.0068587086743\\
38	86.9873291442607\\
39	86.8364815225045\\
40	86.9080093636449\\
42	86.8565204611467\\
43	86.9108454138897\\
44	86.8868354682318\\
45	86.7479229866752\\
46	86.7479229866752\\
47	86.9278736309491\\
48	86.9759166300012\\
49	86.9004045245969\\
50	86.932967390544\\
nan	nan\\
50	88.6584304589183\\
49	88.5834664431451\\
48	88.6154812194611\\
47	88.5022338959326\\
46	88.5208942176259\\
45	88.5208942176259\\
44	88.4357451769295\\
43	88.519262112992\\
42	88.573587065735\\
40	88.7371519266776\\
39	88.8624432086783\\
38	88.8728859095027\\
37	88.6920660225085\\
36	88.6250826906588\\
35	88.6603830081601\\
34	88.6598070700214\\
33	88.6150742886512\\
32	88.6150742886512\\
31	88.6604489012511\\
30	88.4867446654577\\
28	88.5798037399607\\
27	88.6370264200474\\
26	88.7897775315646\\
25	88.5974875622751\\
24	88.5840446169486\\
23	88.6531794574052\\
22	88.2862664909685\\
20	88.1855905377432\\
19	88.0091288241123\\
18	87.8928890491382\\
17	87.9032258064516\\
16	87.994736766174\\
15	87.9566445532053\\
14	87.2973453347579\\
13	87.314344437413\\
12	87.1364855733352\\
11	87.1159438149306\\
10	86.8490911417193\\
9	86.9178827153446\\
8	84.2243231990536\\
7	81.7648558424981\\
6	82.1328112641226\\
5	80.3664754716173\\
4	76.190635198863\\
3	75.3589828000085\\
2	75.0510698708596\\
1	75.2224374670245\\
};
\addplot [color=red, forget plot]
  table[row sep=crcr]{%
1	59.4313236700477\\
2	75.0109417262328\\
3	86.8253031215317\\
4	89.8440621623035\\
5	93.6369948829465\\
6	95.8709980603081\\
7	96.2851521230868\\
8	96.4960958317243\\
9	96.5225039609428\\
10	96.5618333549468\\
11	96.6717308398882\\
12	96.6243501177149\\
13	96.6196806128024\\
14	96.5678045825359\\
16	96.6004234086943\\
18	96.5277983299843\\
19	96.5625105668185\\
20	96.681944174055\\
21	96.8324624469304\\
22	96.6835706163744\\
23	96.8481868625291\\
24	96.802629593937\\
25	96.7445529199869\\
26	96.7741935483871\\
27	96.7022131846865\\
28	96.6440472128651\\
29	96.715970372995\\
30	96.7229120859895\\
31	96.6622069321348\\
32	96.5710320640966\\
33	96.7793619270438\\
34	96.6550318080187\\
35	96.6468698560933\\
36	96.7793619270438\\
37	96.7022131846865\\
38	96.7678875664298\\
39	96.9557002504845\\
40	96.8557999988234\\
41	96.8452842141453\\
42	96.750307961828\\
43	96.8059535031934\\
44	96.7826607611551\\
45	96.696919168969\\
46	96.696919168969\\
47	96.5763258125001\\
48	96.6255978319318\\
49	96.5781820054381\\
50	96.7054406662471\\
nan	nan\\
50	97.7569249251508\\
49	97.8304201450996\\
48	97.8367677594661\\
47	97.7785128971774\\
46	97.7654464224288\\
45	97.7654464224288\\
44	97.8409951528234\\
43	97.7639389699249\\
42	97.7120576295699\\
41	97.724608258973\\
40	97.6065655925744\\
39	97.7754825452144\\
38	97.7482414658282\\
37	97.652625524991\\
36	97.6830036643541\\
35	97.7617322944443\\
34	97.7535703425189\\
32	97.6225163230002\\
31	97.6388683366824\\
30	97.7394535054083\\
29	97.6926317775426\\
28	97.6570280559521\\
27	97.652625524991\\
26	97.5806451612903\\
25	97.6640492305507\\
24	97.4984456748802\\
23	97.5066518471483\\
22	97.6712680933031\\
21	97.6299031444675\\
20	97.565367653902\\
19	97.6310378202783\\
18	97.5044597345318\\
16	97.5393615375423\\
15	97.6050449032578\\
14	97.5182169228404\\
13	97.4125774517138\\
12	97.6229617102421\\
11	97.7368713106494\\
10	97.900532236451\\
9	97.9398616304551\\
8	97.7512159962327\\
7	97.4782887371283\\
6	97.3010449504447\\
5	95.1264459772686\\
4	91.6613141817826\\
3	89.1424388139521\\
2	76.3869077361328\\
1	60.9987838568341\\
};
\end{axis}
\end{tikzpicture}%